\documentclass[a4paper,fleqn]{cas-dc}

\usepackage[authoryear]{natbib}
\usepackage{hyperref}
\usepackage{url}
\usepackage{soul}
\usepackage[utf8]{inputenc}
\usepackage{caption}
\usepackage{url}
\usepackage[list=true]{subcaption}
\usepackage{graphicx}
\usepackage{amsmath}
\usepackage{amssymb}
\usepackage{booktabs}
\usepackage{tabu}
\usepackage{wrapfig}
\usepackage{arydshln}
\usepackage[normalem]{ulem}
\usepackage{multirow}
\usepackage{siunitx}

\def\tsc#1{\csdef{#1}{\textsc{\lowercase{#1}}\xspace}}
\tsc{WGM}
\tsc{QE}

\newdefinition{rmk}{Remark}
\newproof{pf}{Proof}
\newproof{pot}{Proof of Theorem \ref{thm}}

\begin{document}
\let\WriteBookmarks\relax
\def\floatpagepagefraction{1}
\def\textpagefraction{.001}

\shorttitle{Fusing SfM and Simulation-Augmented Pose Regression from Optical Flow for Challenging Indoor Environments}
\shortauthors{F. Ott, L. Heublein, D. Rügamer, B. Bischl, C. Mutschler}
\title[mode=title]{Fusing Structure from Motion and Simulation-Augmented Pose Regression from Optical Flow for Challenging Indoor Environments}

\tnotemark[1]
\tnotetext[1]{Evaluating the performance of absolute pose regression methods, i.e., SfM and PoseNet, in challenging scenarios such as repetitive patterns, motion blur, structure-less surfaces, and difficult lighting conditions. Fusing absolute and relative poses using either PGO or recurrent networks.}

\author[1,2,4]{Felix Ott}[style=german, linkedin=felix-ott-494b06146, orcid=0000-0002-4392-0830]
\cormark[1]
\cortext[1]{Corresponding author}
\ead{felix.ott@iis.fraunhofer.de}
\ead[url]{https://www.slds.stat.uni-muenchen.de/people/ott/}

\credit{conceptualization of this study, methodology, software, validation, formal analysis, investigation, data curation, writing -- original draft, writing -- review \& editing, visualization, data recording, project administration}

\affiliation[1]{organization={Fraunhofer IIS, Fraunhofer Institute for Integrated Circuits IIS},
            addressline={Nordostpark 84}, 
            city={Nuremberg},
            citysep={},
            postcode={90411},
            country={Germany}}

\affiliation[2]{organization={LMU Munich},
            addressline={Ludwigstraße 33}, 
            city={Munich},
            citysep={},
            postcode={80539},
            country={Germany}}

\affiliation[3]{organization={Technical University of Dortmund},
            addressline={August-Schmidt-Straße 1}, 
            city={Dortmund},
            citysep={},
            postcode={44227},
            country={Germany}}

\affiliation[4]{organization={Munich Center for Machine Learning (MCML)},
            addressline={Ludwigstraße 33},
            city={Munich},
            citysep={},
            postcode={80539},
            country={Germany}}

\author[1]{Lucas Heublein}[style=german, orcid=0000-0001-6670-3698]
\ead{heublels@iis.fraunhofer.de}
\credit{conceptualization of this study, methodology, software, investigation, data curation}

\author[2,3,4]{David Rügamer}[style=german, linkedin=david-ruegamer, orcid=0000-0002-8772-9202]
\ead{david.ruegamer@stat.uni-muenchen.de}
\ead[url]{https://www.slds.stat.uni-muenchen.de/people/ruegamer/}
\credit{formal analysis, writing -- review \& editing, visualization, supervision}

\author[2,4]{Bernd Bischl}[style=german, orcid=0000-0001-6002-6980]
\ead{bernd.bischl@stat.uni-muenchen.de}
\ead[url]{https://www.statistik.uni-muenchen.de/personen/professoren/bischl/index.html}
\credit{supervision}

\author[1]{Christopher Mutschler}[style=german, linkedin=christopher-mutschler-28431576, orcid=0000-0001-8108-0230]
\ead{christopher.mutschler@iis.fraunhofer.de}
\ead[url]{https://cmutschler.de/}
\credit{resources, writing -- review \& editing, supervision, funding acquisition}

\begin{abstract}
The localization of objects is essential in many applications, such as robotics, virtual and augmented reality, and warehouse logistics. Recent advancements in deep learning have enabled localization using monocular cameras. Traditionally, structure from motion (SfM) techniques predict an object's absolute position from a point cloud, while absolute pose regression (APR) methods use neural networks to understand the environment semantically. However, both approaches face challenges from environmental factors like motion blur, lighting changes, repetitive patterns, and featureless areas. This study addresses these challenges by incorporating additional information and refining absolute pose estimates with relative pose regression (RPR) methods. RPR also struggles with issues like motion blur. To overcome this, we compute the optical flow between consecutive images using the Lucas-Kanade algorithm and use a small recurrent convolutional network to predict relative poses. Combining absolute and relative poses is difficult due to differences between global and local coordinate systems. Current methods use pose graph optimization (PGO) to align these poses. In this work, we propose recurrent fusion networks to better integrate absolute and relative pose predictions, enhancing the accuracy of absolute pose estimates. We evaluate eight different recurrent units and create a simulation environment to pre-train the APR and RPR networks for improved generalization. Additionally, we record a large dataset of various scenarios in a challenging indoor environment resembling a warehouse with transportation robots. Through hyperparameter searches and experiments, we demonstrate that our recurrent fusion method outperforms PGO in effectiveness.
\end{abstract}

\begin{keywords}
visual self-localization \sep structure from motion \sep pose regression \sep optical flow \sep pose graph optimization \sep recurrent pose fusion \sep synthetic transfer learning \sep challenging environment \sep \textit{Datasets and Source Code}: \\ \url{https://gitlab.cc-asp.fraunhofer.de/ottf/industry_datasets}
\end{keywords}

\maketitle
\section{Introduction}
\label{sec_introduction}

Accurate localization of objects is essential for various path planning applications, such as robotic systems operating in vast warehouses \citep{radwan_valada_burgard,loeffler_riechel}. Achieving this requires highly precise pose recognition, including both the robot's position and orientation. While some localization systems rely on LiDAR, radio, or radar-based technologies \citep{stahlke_sensors} or inertial sensors \citep{silva_uchiyama}, visual self-localization using monocular cameras has become increasingly popular due to advancements in deep learning techniques \citep{kendall_grimes_cipolla}. The effectiveness of pose estimation methods heavily depends on the invariance properties of the features used, such as scale and rotation invariance \citep{ott_tro}.

The 3D structure of a scene can be estimated from a sequence of 2D images using structure from motion (SfM) methods \citep{sfm_github,resch_lensch_wang,jiang_jiang_jiang,brachmann_cavallari,yu_feng_ye_jiang,li_cao_liu_li_zhu}, which involve camera motion estimation by detecting and matching feature points between pairs of images, and 3D structure estimation by triangulating the feature points. Absolute pose regression (APR) techniques \citep{kendall_grimes_cipolla,loeffler_riechel,radwan_valada_burgard} use neural networks to extract features from images and estimate the 6DoF global pose of an object with respect to a known coordinate system. Classical visual odometry (VO) methods \citep{mansur_habib_pratama} estimate camera motion by analyzing consecutive images to compute the relative pose of an object. Relative pose regression (RPR) methods \citep{wang_clark_wen,iyer_murthy_gupta,kreuzig_ochs_mester,idan_shavit_keller} have emerged, which predict the relative pose by using convolutional neural networks (CNNs) and recurrent neural networks (RNNs). RPR methods based on optical flow, which captures the motion of pixels between two frames, are particularly promising due to their robustness to environmental changes, as evidenced by recent studies \citep{muller_savakis,muller_savakis_ptucha,zhou_luo_shen,ott_cvprw}.

Each method for absolute pose prediction, including SfM, APR, and RPR, has its own drawbacks, and each pose is affected by different sources of noise, as highlighted by \cite{sattler_zhou_pollefeys}. APR algorithms such as PoseNet~\citep{kendall_grimes_cipolla} learn a set of base poses such that the poses of all training images can be expressed as a linear combination of these base poses. The base translations estimated by PoseNet are significantly more noisy than those by SfM and do not all align on a straight line. Hence, PoseNet is still not able to generalize well across different scenarios \cite{sattler_zhou_pollefeys}. To improve the performance of these methods, it is desirable to combine absolute poses (from SfM or APR) and relative poses (through RPR from optical flow). The motivation for fusing absolute and relative pose methods lies in addressing the limitations and enhancing the robustness of pose estimation systems: (1) Utilizing complementary information, i.e., APR methods provide global positioning but may suffer from long-term drift over time, while RPR methods excel in short-term accuracy but lack a global reference. By combining both approaches, the strengths of each can be leveraged, compensating for the weaknesses and creating a more reliable and accurate overall pose estimation \citep{huang_bachrach}. (2) Integrating relative pose information allows for periodic correction, mitigating the effects of drift and enhancing long-term accuracy \citep{pacheco_ascencio}. (3) APR methods may struggle in dynamic or complex environments where changes occur rapidly. By incorporating relative pose information, the system's adaptability to local changes is enhanced \citep{marros_michel}. (4) A fusion of absolute and relative pose methods creates a hybrid system that can be adaptable to various applications and scenarios. (5) Methods that utilize multiple sensors with different strengths and weaknesses can be improved by fusing absolute and relative pose information \citep{kazerouni_fitzgerald}. While some approaches combine both fields or different modalities by a common representation between networks \citep{ott_tro,brieger_ion_gnss}, others directly combine the absolute and relative poses \citep{mitsuki_ryogo}. However, fusing absolute and relative poses is a challenging task since RPR requires knowledge of the global pose, which may contain errors. Pose graph optimization (PGO) \citep{brahmbhatt_gu_kim,mirowski_grimes} is a technique that estimates smooth and globally consistent pose predictions during inference. Another approach is to use a fusion module based on recurrent cells \citep{ott_cvprw}, which can learn an improved (i.e., smoothed), long-term trajectory, but requires time-distributions (i.e., applies a layer to every temporal slice of an input) to learn the orientation of the relative to the global pose. Despite recent advances, optimizing the absolute pose with neural networks remains an open research topic.

Another research goal of visual self-localization is to develop methods that can adapt to new and unknown scenes or remain robust to environmental changes \citep{idan_shavit_keller,winkelbauer_denninger,wang_qi,acharya_tatli}, such as adding or removing racks in warehouses. To achieve this, some approaches use multi-scene APR techniques (e.g., employing Transformers) to learn multiple scenes simultaneously \citep{shavit_ferens_keller}, use auto-encoders \citep{shavit_keller}, or apply transfer learning between different scenes \citep{chidlovskii_sadek}. To improve the performance of the initial weights, we pre-train the APR and RPR models using synthetic data.

\textbf{Contributions.} The primary objective of this work is to optimize the accuracy of absolute poses by fusing absolute and relative poses. To achieve this goal, the following steps are taken: (1) To enhance the performance of SfM in retrieving absolute poses, we conduct a large hyperparameter search. PoseNet~\citep{kendall_grimes_cipolla} is alternatively evaluated as an APR technique. The relative poses are regressed from optical flow computed with the Lucas-Kanade~\citep{baker_matthews} algorithm. The objective is to use these techniques as baseline black box models and focus on the optimization of fusion methods. (2) We evaluate PGO~\citep{mirowski_grimes} to optimize the pose and conduct a benchmark to assess the performance of eight recurrent and 17 convolutional, recurrent, and Transformer networks in absolute and relative pose fusion. (3) As there is currently no publicly available dataset to evaluate large-scale challenging environments, we record and publish\footnote{Datasets and source code: \url{https://gitlab.cc-asp.fraunhofer.de/ottf/industry_datasets}} a large database containing various scenes with changes between scenes. This dataset is used to conduct experiments assessing the robustness of methods against volatile environments. (4) We develop a simulation framework to generate and pre-train the APR and RPR models on synthetic data. Advanced techniques may be used as black box models in place of the baseline models (SfM, APR, RPR) to further enhance the results. However, the focus of this study is on assessing the performance of the fusion techniques with respect to environmental changes and motion dynamics on the localization task.

The remainder of this article is organized as follows. We discuss related work in Section~\ref{sec_related_work}. Section~\ref{sec_method} introduces our method, including SfM, APR, RPR, and fusion modules. We present novel datasets and our simulation for data generation in Section~\ref{sec_experiments}, and discuss experimental results in Section~\ref{sec_evaluation}.
\section{Related Work}
\label{sec_related_work}

The fields of SfM and APR have experienced significant growth in recent years, resulting in a broad and diverse body of research. This work, however, primarily focuses on fusing absolute and relative poses. For a comprehensive overview of SfM approaches, please refer to \cite{jiang_jiang_jiang,zheng_yang_tian,piasco_sidibe,brachmann_cavallari,radanovic_khoshelham}. For a detailed review of APR techniques, see \cite{sattler_zhou_pollefeys,ott_tro,xu_wang_xu_zhang,hunter_blanton,qiao_xiang_fan}. An overview of APR loss functions can be found in \cite{boittiaux_marxer_dune,pepe_lasenby}. In Section~\ref{sec_rw_rpr}, we introduce techniques for RPR estimation using optical flow. State-of-the-art techniques for fusing absolute and relative poses are discussed in Section~\ref{sec_rw_apr_rpr_fusion}, while Section~\ref{sec_rw_simulation} presents related work on scene generalization through simulation.

\subsection{RPR from Optical Flow}
\label{sec_rw_rpr}

Numerous techniques exist for estimating the relative pose from successive image pairs by utilizing optical flow, including Flowdometry~\citep{muller_savakis,muller_savakis_ptucha}, ViPR~\citep{ott_cvprw}, DeepVIO~\citep{han_lim_du_lian}, KFNet~\citep{zhou_luo_shen}, and the model by \cite{ott_tro}. These methods employ either FlowNet or FlowNetSimple by \cite{dosovitskiy_fischer_ilg} or FlowNet2~\citep{ilg_mayer_saikia} for estimating optical flow. While LS-VO~\citep{constante_ciarfuglia} uses an autoencoder to predict optical flow and estimate a trajectory from a neural network, we compute the optical flow using the Lucas-Kanade~\citep{baker_matthews} algorithm. In contrast, \cite{zhi_yu_po_heng} utilizes the relative pose as auxiliary information for optical flow prediction of FlowNet2. Our method uses the Lucas-Kanade~\citep{baker_matthews} algorithm in combination with a small relative model. DeepVO~\citep{wang_clark_wen}, CTCNet~\citep{iyer_murthy_gupta}, DistanceNet~\citep{kreuzig_ochs_mester}, and MotionNet~\citep{ding_wang_tang} do not rely on optical flow but combine CNNs with RNNs, such as stacked LSTMs, bidirectional LSTMs, or fully connected layers, to model sequential dynamics and relationships to predict relative poses. Similarly, our relative model consists of convolutional layer in combination with LSTMs. \cite{muller_smith} use pretrained semantic segmentation and optical flow to extract ground plane correspondences between  aligned images and RANSAC to find the best fitting homography. In contract, DiffPoseNet~\citep{parameshwara_hari} uses normal flow to estimate relative camera pose based on the cheirality (depth positivity) constraint by formulating a optimization problem as a differentiable layer. While DiffPoseNet uses flow to improve the coarse absolute predictions from PoseNet, our model directly predicts the relative poses to refine the absolute poses.

\subsection{APR \& RPR Fusion}
\label{sec_rw_apr_rpr_fusion}

LM-Reloc~\citep{stumberg_wenzel} formulates its loss based on the Levenberg-Marquardt algorithm to improve direct image alignment through learned features. It also incorporates an RPR model to bootstrap the direct image alignment. ViPR~\citep{ott_cvprw} enhances absolute poses by concatenating relative poses predicted from optical flow and absolute poses, and then refining them using an LSTM network. MapNet+PGO~\citep{brahmbhatt_gu_kim} refines predicted poses from APR and RPR using PGO. VLocNet~\citep{valada_radwan_burgard} estimates global poses and combines it with VO, while VLocNet++~\citep{radwan_valada_burgard} adds features from semantic segmentation. RCNN \citep{lin_liu_huang} fuses relative and global sub-networks to smooth VO and prevent drift. \cite{mitsuki_ryogo} propose a graph neural network that uses image-to-nodes and image-to-edges to create similarity-preserving mappings, with nodes representing absolute features and edges representing relative features. Kalman filters are frequently employed for pose fusion, such as in the work by \cite{emter_schirg_woock}. This approach combines absolute and relative measurements, leading to a correlation between a past state and present state that creates correlations. Our objective is to mitigate these correlations by utilizing an optimal recurrent cell. In general, LSTM cells are employed for representing temporal dependencies in localization tasks. Recently, Transformer networks have been applied in visual self-localization, as demonstrated by the methods proposed by \cite{shavit_ferens_keller,li_ling,kim_kim,li_cao_zhu}. Nevertheless, Transformers are not well-suited for integrating absolute and relative poses due to their low-dimensional information. Additionally, the bidirectional properties \citep{graves_liwicki_fernandez} of LSTMs are deemed unfeasible for pose modeling. \cite{ruan_he_guan_zhang} fuse APR with scene coordinate regression. While \cite{lu_lu} combine a depth map from a single-view depth network with relative poses from a pose network, \cite{yang_stumberg} proposed D3VO that combines the depth map with relative poses from PoseNet. In contrast, \cite{zhan_weerasekera} proposed the combination of a depth network with optical flow. However, this method is designed solely for visual odometry. \cite{li_zhao_song} independently utilize a depth network, optical flow network, and camera motion network, which are jointly optimized during the training phase. The objective is to predict various masks by combining four types of loss functions specifically for relative prediction. 

\cite{das_dubbelman} fuse absolute poses derived from GNSS data with relative poses obtained from vehicle odometry. Two subsequent works integrate relative poses from inertial data with absolute poses from visual data. VINet~\citep{clark_wang_wen} uses a concatenation approach to merge relative features, extracted by an inertial LSTM encoder, with absolute features, extracted by a visual encoder. In contrast, VI-DSO~\citep{stumberg_usenko} employs a combined energy functional to jointly estimate camera poses and sparse scene geometry by minimizing both visual and inertial measurement errors.

A related yet distinct field is depth estimation, which is also employed in APR and RPR techniques. Research in this area often emphasizes self-supervised learning \citep{godard_aodha}, diffusion models \citep{saxena_kar} due to recent advancements, and Transformer architectures \citep{zhang_nex}. Many additional depth estimation techniques are available and could be integrated into fusion methods, albeit not within the context of SfM and RPR methods.

\subsection{Simulation for Scene Generalization}
\label{sec_rw_simulation}

Todays regression-based techniques for APR are scene-specific with respect to their training and evaluation. They rely on the coordinate system of the training dataset, leading to poor generalization across different scenes \citep{chidlovskii_sadek}. The Transformer network, proposed by \cite{shavit_ferens_keller}, learns multi-scene APR. The model employs encoders to aggregate activation maps via self-attention and decoders to transform latent features and scenes encoding into predicted pose. This approach allows the model to learn informative features while embedding multiple scenes in parallel. To address the issue of dataset shift, \cite{chidlovskii_sadek} developed a deep adaptation network that can learn scene-invariant image representations for transfer learning. Furthermore, \cite{chidlovskii_sadek} use adversarial learning to generate such representations for the model transfer. \cite{wang_qi} report that single-input images can cause confusion in relocalization when dealing with scenes that share similar views but differ in position, which motivates the use of a time-distributed network. Furthermore, they highlight the difficulty of relocalization in variable or dynamic scenes. \cite{wang_qi} created a variable scene dataset comprising three scenes: an office, a bedroom, and a sitting room. They use semi-automatic processing to develop their MMLNet method, which can regress both camera pose and scene point cloud. Similarly, our approach involves the fusion of point clouds and relative poses. \cite{acharya_tatli} uses a generative network to model fake synthetic images for APR.

\begin{figure*}[!t]
    \centering
    \includegraphics[width=1.0\linewidth]{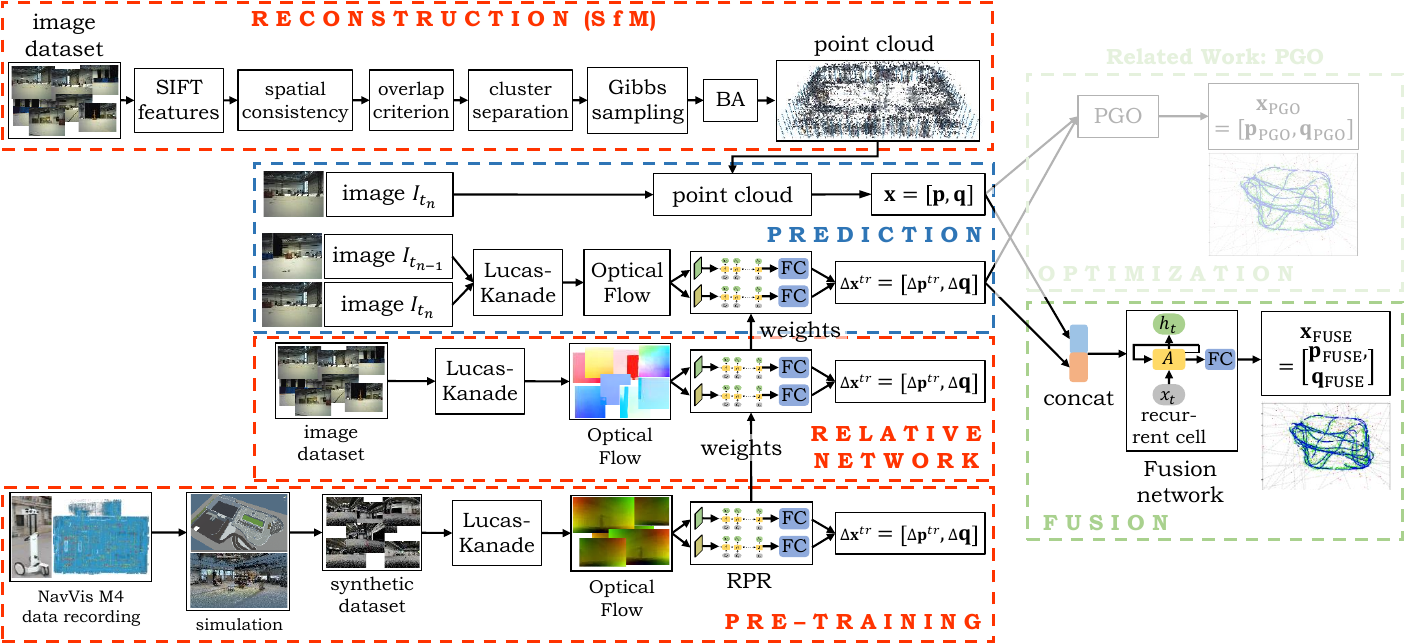}
    \caption{\textbf{Method overview}. First, a point cloud is constructed from an image dataset using SfM to extract features, spatial consistency, overlap criterion, cluster sampling, and bundle adjustment (BA). Second, we train a small convolutional recurrent neural network to predict the relative pose $\Delta \mathbf{x}$ between two consecutive images. For the training of the fusion model and the evaluation step, the absolute pose $\mathbf{x}$ from the point cloud and the relative pose $\Delta \mathbf{x}^{tr}$ from the RPR model is retrieved for a query image. Last the absolute pose is optimized with either PGO or a recurrent network. To compare with state-of-the-art methods, we replace the reconstruction and the point cloud in the prediction steps with the absolute pose prediction from the APR model.}
    \label{figure_method_overview}
\end{figure*}

While APR techniques learn absolute scene parameters, RPR methods can localize in unseen environments by learning the residual pose between image pairs. However, RPR methods often perform poorly in unfamiliar scenes. To improve the generalization of RPR methods, \cite{idan_shavit_keller} enhanced their performance by aggregating paired feature maps into latent codes. Additionally, \cite{winkelbauer_denninger} proposed an RPR approach based on ResNet that includes changes to the model architecture, such as an extended regression part, hierarchical correlation layers, and the incorporation of uncertainty and scale information, to better generalize to new scenes.

In summary, prior works have focused on incorporating auxiliary information into models to learn both general features and scene-specific features that can generalize across various scenes. However, no existing approach learns scene information from simulated environments (absolute) or motion from simulated dynamics (relative) \citep{idan_shavit_keller,winkelbauer_denninger,shavit_ferens_keller}.
\section{Methodology}
\label{sec_method}

Firstly, we present the reconstruction of a point cloud from an image dataset using SfM (see Section~\ref{sec_method_sfm}). Section~\ref{sec_method_apr} describes an alternative approach based on APR. The Lucas-Kanade method for computing optical flow and our RPR network are presented in Section~\ref{sec_method_rpr}. Description of absolute and relative pose fusion is provided based on these APR and RPR methods. Furthermore, we evaluate PGO in Section~\ref{sec_method_pgo} and propose a framework of models that utilize various RNNs in Section~\ref{sec_method_fusion}. Finally, in Section~\ref{sec_method_simulation}, we discuss the rationale for pre-training APR and RPR to improve generalizability to various scenarios. Figure~\ref{figure_method_overview} presents an overview of the method, illustrating all steps. 

\subsection{Structure from Motion}
\label{sec_method_sfm}

\begin{figure}[!t]
    \centering
    \includegraphics[trim=1 1 1 1, clip, width=1.0\linewidth]{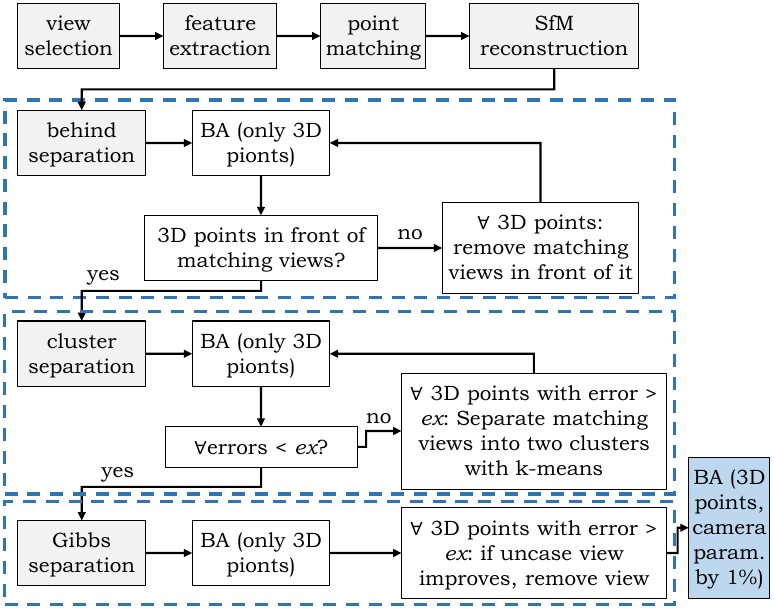}
    \caption{SfM pipeline using bundle adjustment (BA) to reconstruct a point cloud from input images.}
    \label{figure_sfm_pipeline}
\end{figure}

This section outlines the steps involved in using SfM to create a point cloud. The SfM pipeline is proposed in Figure~\ref{figure_sfm_pipeline}, while Table~\ref{table_hyperparameters} provides a summary of the hyperparameters used in each step.

\paragraph{SfM.} We adopt the source code provided by \cite{sfm_github} as the baseline for our SfM pipeline, which is used to recover 3D structure of a given environment. SfM is employed to determine the relative pose of each image with respect to the first image. Inspired by the findings of \cite{resch_lensch_wang}, which suggest using every fifth image for reconstruction, we manually pre-select around 600 images from each dataset (comprising between 7,000 to 138,000 images) for point cloud reconstruction. Additionally, we obtain the intrinsic matrix, essential for recalibration, by capturing images of a checkerboard using our camera setup. For the sake of reproducibility, we report the following calibration matrix:
\begin{equation}
K =
\begin{bmatrix}
548.44934818 & 0.0 & 317.73762648\\
0.0 & 540.17600512 & 249.00614224\\
0.0 & 0.0 & 1.0
\end{bmatrix} 
\end{equation}
We ensure to avoid matching two points from the same image with a single point from the point cloud.

\paragraph{Feature Extraction.} We utilize the opencv-python cv2 library \citep{cv2_sift} to extract image features, which yields descriptors of size 128 for each image. In a pre-defined study, we assess the impact of three feature extraction techniques, namely, scale-invariant feature transform (SIFT) proposed by \cite{david_lowe}, speeded up robust features (SURF) introduced by \cite{bay_tuytelaars}, and oriented fast and rotated brief (ORB) proposed by \cite{rublee_rabaud}, on the reconstruction performance. Based on our evaluation, we select SIFT as the feature extraction technique for our pipeline.

\paragraph{Spatial Consistency \& Overlap Criterion.} In addition, we employ spatial consistency to enhance the discriminative capability of the raw feature points by evaluating the matching quality of feature points in a larger spatial neighborhood. We follow the approach proposed in \cite{jiang_jiang_jiang}, which involves computing the ratios of features that fall into two corresponding regions. We introduce a hyperparameter $sc$ that specifies the spatial consistency of the point neighborhood in terms of the number of pixels. We also apply an overlap criterion to filter out unnecessary image pairs. We divide the ground floor into a grid and require each image to contain a specific percentage (controlled by the parameter $oc$) from these grid points.

\begin{table}[t!]
\begin{center}
\setlength{\tabcolsep}{3.3pt}
    \caption{Overview of SfM hyperparameters.}
    \label{table_hyperparameters}
    \small \begin{tabular}{ p{0.5cm} | p{0.5cm}}
    \multicolumn{1}{c|}{\textbf{Parameter}} & \multicolumn{1}{c}{\textbf{Description}} \\ \hline
    \multicolumn{1}{l|}{$sc$} & \multicolumn{1}{l}{Spatial consistency of the point neighborhood} \\
    \multicolumn{1}{l|}{} & \multicolumn{1}{l}{in pixel} \\
    \multicolumn{1}{l|}{$oc$} & \multicolumn{1}{l}{Overlap criterion of floor pixels between} \\
    \multicolumn{1}{l|}{} & \multicolumn{1}{l}{images in \%} \\
    \multicolumn{1}{l|}{$mm$} & \multicolumn{1}{l}{Minimal number of matches for each point} \\
    \multicolumn{1}{l|}{} & \multicolumn{1}{l}{over all images} \\
    \multicolumn{1}{l|}{$ex$} & \multicolumn{1}{l}{Exclude points with reprojection error larger} \\
    \multicolumn{1}{l|}{} & \multicolumn{1}{l}{than separation limit in pixel} \\
    \multicolumn{1}{l|}{$gibbs$} & \multicolumn{1}{l}{Boolean of \textit{gibbs} factor of pixel improvement} \\
    \multicolumn{1}{l|}{} & \multicolumn{1}{l}{by excluding single points} \\
    \multicolumn{1}{l|}{$std$} & \multicolumn{1}{l}{Standard deviation for point exclusion for BA} \\
    \end{tabular}
\end{center}
\end{table}

\paragraph{Cluster Separation.} Points from multiple images can correspond to the same point in the point cloud. To handle such cases, we partition these image points into clusters and explore the possibility of improving the match separation, as proposed in \cite{jiang_jiang_jiang}. This results in the creation of additional clusters in the point cloud. We use the hyperparameter $mm$ to determine the number of matches. For neighborhood clustering, we use k-means with two clusters. We apply this strategy iteratively with bundle adjustment (BA) until the hyperparameter $ex$ falls below a predefined threshold in terms of pixels.

\paragraph{Gibbs Sampling.} We first apply BA. Next, we iterate through all image points for each point in the point cloud. We employ Gibbs sampling to iteratively remove one point from the selected image points and evaluate whether the remaining points improve the performance of the BA. We adjust the hyperparameter $gibbs$ for this process.

\paragraph{Bundle Adjustment.} We adopt the sparse BA method proposed in \cite{ba_github}. Additionally, we define parameters to regularize the point cloud to remain within the environment. In all BA steps, we keep the rotation matrix fixed, and only change the rotation matrix of the last BA step. We tune the hyperparameter $std$, which represents the standard deviation for point exclusions.

\subsection{Absolute Pose Regression}
\label{sec_method_apr}

In order to reduce the computational requirements associated with hyperparameter tuning for SfM (provided in Section~\ref{sec_experiments_search}), an alternative approach involves using APR methods, as discussed in Section~\ref{sec_rw_apr_rpr_fusion}. Rather than tuning hyperparameters, a CNN-based APR model can learn to directly regress the camera pose from a single input image in conjunction with their corresponding ground truth poses. In our case, we utilize a time-distributed network that takes a set of three consecutive images as input (at timesteps $t_{n-2}$, $t_{n-1}$, and $t_n$) to predict absolute positions $\mathbf{p} \in \mathbb{R}^3$ in Euclidean coordinates and absolute orientations $\mathbf{q} \in \mathbb{R}^4$ as quaternions of the absolute pose $\mathbf{x} = [\mathbf{p}, \mathbf{q}]$. We use the PoseNet architecture \citep{kendall_grimes_cipolla} based on GoogLeNet~\citep{szegedy_liu_jia} with time-distribution~\citep{ott_cvprw}. Our network includes a fully connected (FC) layer of 2,048 units, and two parallel FC layers, each with three and four units. We minimize the root mean squared error (RMSE) loss function
\begin{equation}
    \mathcal{L}_{\text{APR}} = ||\hat{\mathbf{p}} - \mathbf{p}||_2^2 + \beta_1 \big|\big|\hat{\mathbf{q}} - \frac{\mathbf{q}}{||\mathbf{q}||_2}\big|\big|_2^2,
\end{equation}
between the predicted pose $\mathbf{x} = [\mathbf{p}, \mathbf{q}]$ and ground truth pose $\hat{\mathbf{x}} = [\hat{\mathbf{p}}, \hat{\mathbf{q}}]$, weighted by the hyperparameter $\beta_1 = 50$. We use a batch size of 50, the Adam optimizer without decay, and a learning rate of $10^{-4}$.

\subsection{Relative Pose Regression from Optical Flow}
\label{sec_method_rpr}

\begin{figure}[!t]
    \centering
	\begin{minipage}[t]{1.0\linewidth}
	\begin{minipage}[t]{0.24\linewidth}
        \centering
    	\includegraphics[width=1.0\linewidth]{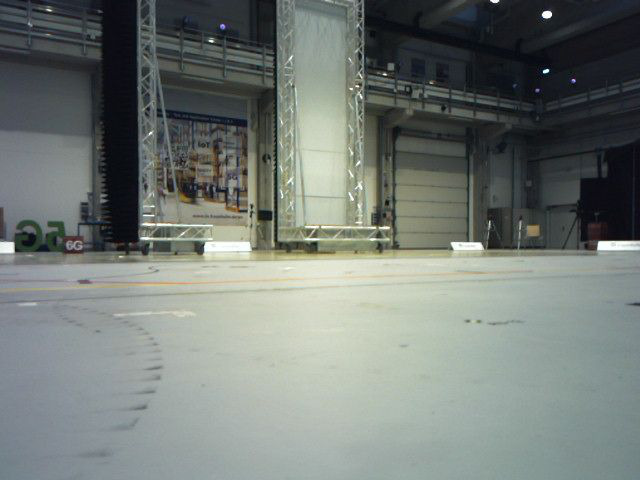}
    \end{minipage}
    \hfill
	\begin{minipage}[t]{0.24\linewidth}
        \centering
    	\includegraphics[width=1.0\linewidth]{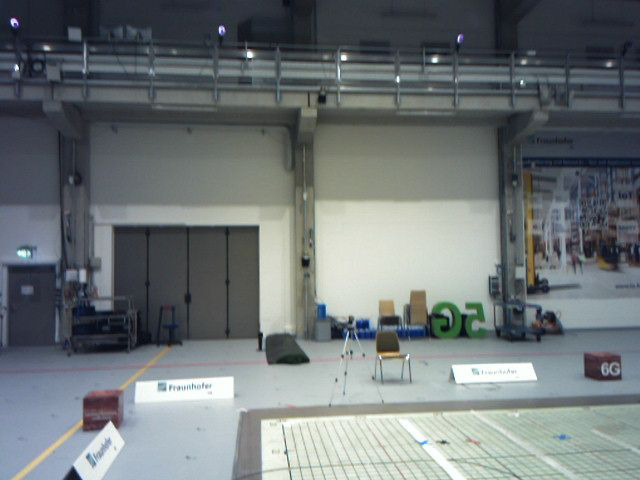}
    \end{minipage}
    \hfill
	\begin{minipage}[t]{0.24\linewidth}
        \centering
    	\includegraphics[width=1.0\linewidth]{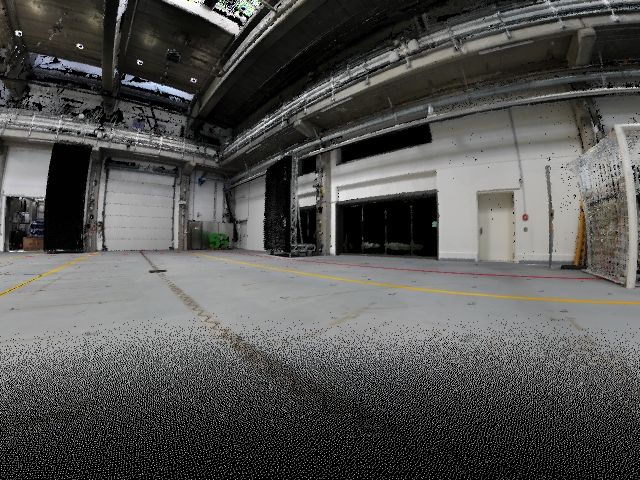}
    \end{minipage}
    \hfill
	\begin{minipage}[t]{0.24\linewidth}
        \centering
    	\includegraphics[width=1.0\linewidth]{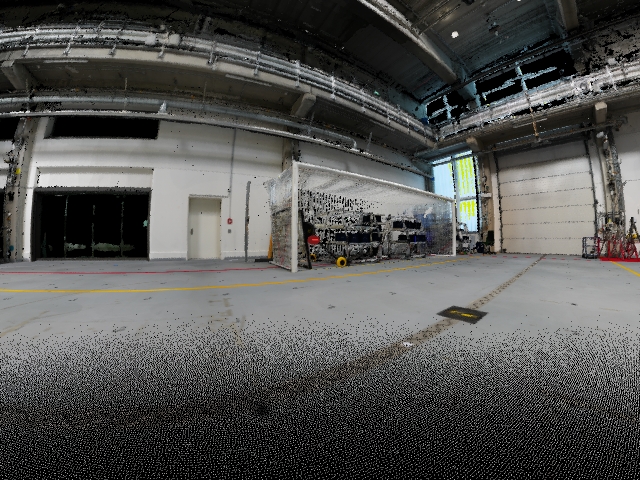}
    \end{minipage}
    \subcaption{Images of timestep $t_{n-1}$.}
    \label{figure_example_of1}
    \end{minipage}
	\begin{minipage}[t]{1.0\linewidth}
	\begin{minipage}[t]{0.24\linewidth}
        \centering
    	\includegraphics[width=1.0\linewidth]{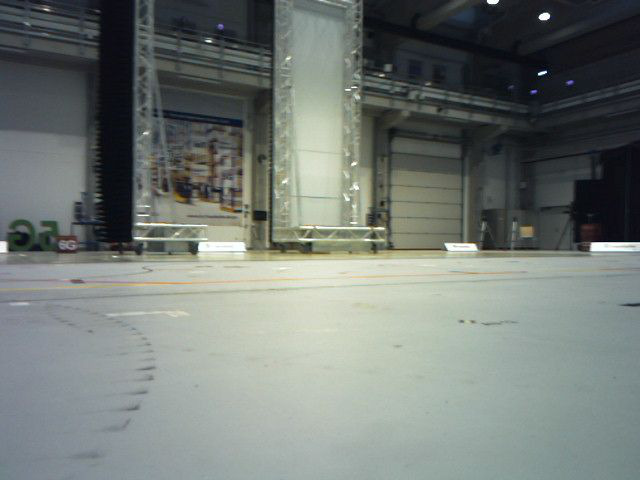}
    \end{minipage}
    \hfill
	\begin{minipage}[t]{0.24\linewidth}
        \centering
    	\includegraphics[width=1.0\linewidth]{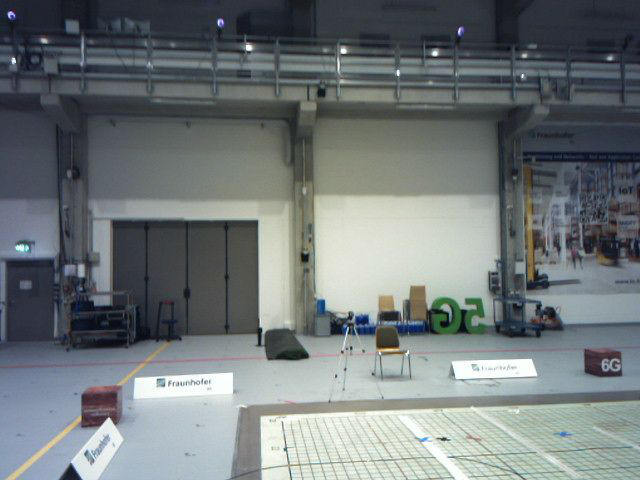}
    \end{minipage}
    \hfill
	\begin{minipage}[t]{0.24\linewidth}
        \centering
    	\includegraphics[width=1.0\linewidth]{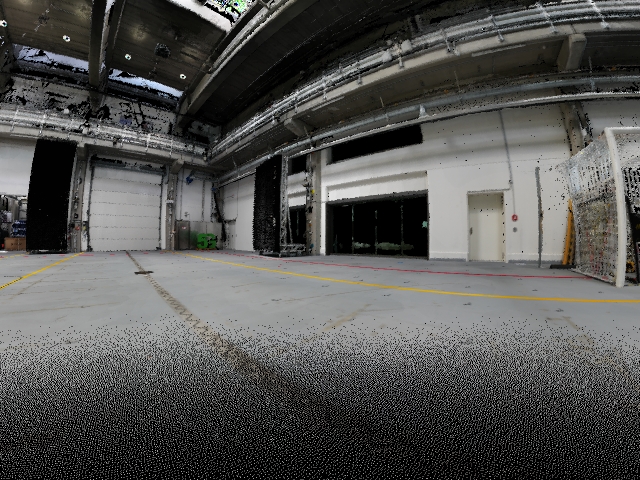}
    \end{minipage}
    \hfill
	\begin{minipage}[t]{0.24\linewidth}
        \centering
    	\includegraphics[width=1.0\linewidth]{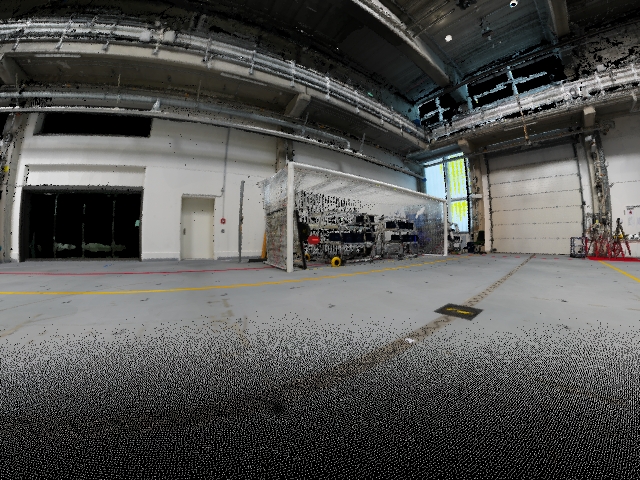}
    \end{minipage}
    \subcaption{Images of timestep $t_{n}$.}
    \label{figure_example_of2}
    \end{minipage}
	\begin{minipage}[t]{1.0\linewidth}
	\begin{minipage}[c]{0.24\linewidth}
        \centering
    	\includegraphics[trim=60 40 50 40, clip, width=1.0\linewidth]{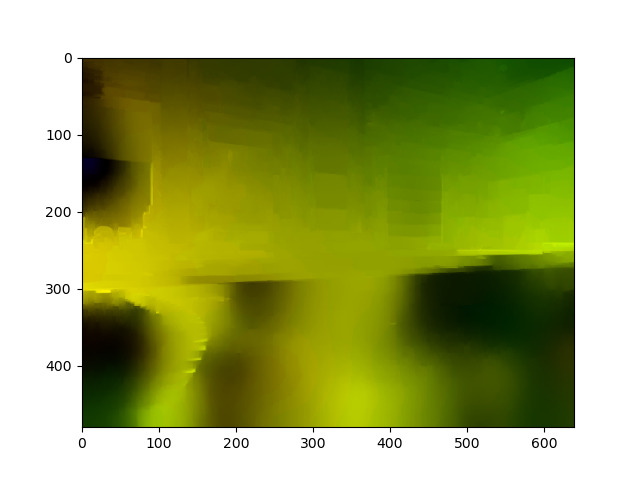}
    \end{minipage}
    \hfill
	\begin{minipage}[c]{0.24\linewidth}
        \centering
    	\includegraphics[trim=60 40 50 40, clip, width=1.0\linewidth]{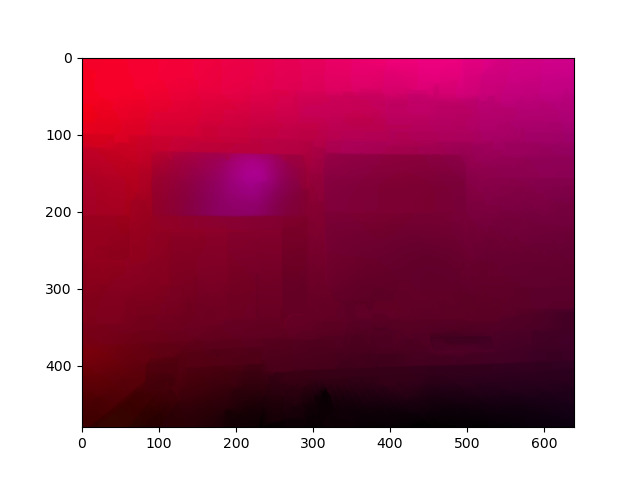}
    \end{minipage}
    \hfill
	\begin{minipage}[c]{0.24\linewidth}
        \centering
    	\includegraphics[trim=70 71 70 71, clip, width=1.0\linewidth]{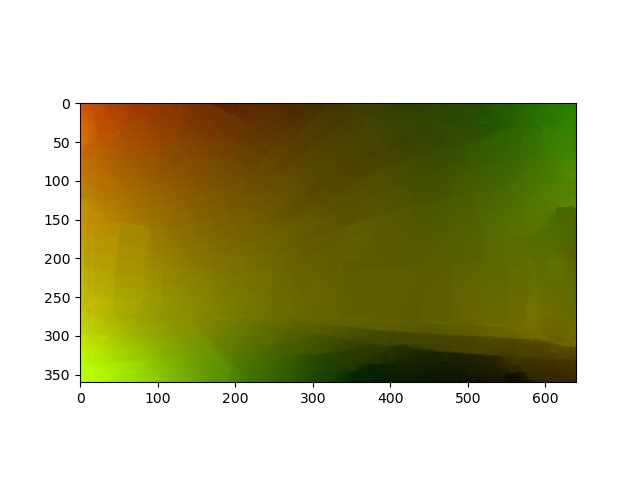}
    \end{minipage}
    \hfill
	\begin{minipage}[c]{0.24\linewidth}
        \centering
    	\includegraphics[trim=70 71 70 71, clip, width=1.0\linewidth]{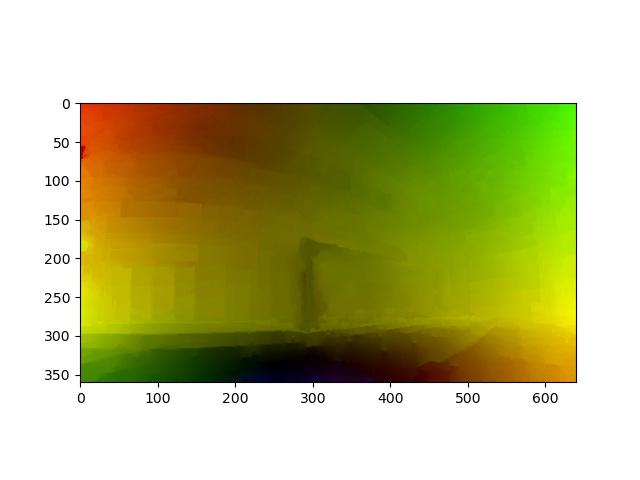}
    \end{minipage}
    \subcaption{Optical flow visualization.}
    \label{figure_example_of3}
    \end{minipage}
    \caption{Exemplary optical flow visualizations (c) between two consecutive images (a and b) of real-world scenarios (image 1 and 2) and simulated scenarios (image 3 and 4). Note the small movement between timestep $t_{n-1}$ (a) and timestep $t_{n}$ (b).}
    \label{figure_example_of}
\end{figure}

We utilize the Lucas-Kanade~\citep{baker_matthews} algorithm to compute the optical flow between two images captured at timesteps $t_{n-1}$ and $t_n$ for learning the relative movements and rotations of an object. Optical flow is a vector field representation of the movement of objects or surfaces in a sequence of images or video, as captured by a camera. Each vector in the optical flow field corresponds to the displacement of a small portion of the image from frame $t_{n-1}$ to frame $t_n$. Hence, we receive the pixel movements in $u$ and $v$ direction. The Lucas-Kanade algorithm is a classic differential method that assumes uniform motion of objects in a small local neighborhood and solves a system of linear equations relating the spatial and temporal derivatives of the image intensity to the local motion parameters. We present an exemplary visualization of optical flow in Figure~\ref{figure_example_of}.

We propose the following recurrent network for the purpose of regressing the relative pose from optical flow. To achieve this, we initially subject the vector input, which has a size of $480 \times 640 \times 2$ pixels, to $4 \times 4$ mean average pooling with stride 1, resulting in a tensor size of $120 \times 160 \times 2$. In this case, $u$ and $v$ represent the third dimension of the optical flow. We then proceed to search for an optimal network configuration, which may involve batch normalization, adding one or two stacked LSTM layer, and ReLU, softmax, or no activation. Our chosen network configuration involves the absence of batch normalization and activation, and the use of one LSTM layer with 50 units for each direction of $u$ and $v$ by slicing the third dimension of the optical flow. The output of the LSTM layers is then concatenated, and two FC layers of size 3 and 4 are added to predict the relative pose $\Delta\mathbf{x} = [\Delta\mathbf{p}^{tr}, \Delta\mathbf{q}]$, i.e., the relative position (translation) $\Delta \mathbf{p}^{tr} \in \mathbb{R}^3$ and the relative orientation (rotation) $\Delta \mathbf{q} \in \mathbb{R}^4$. To minimize error, we apply a mean squared error (MSE) loss function
\begin{equation}
    \mathcal{L}_{\text{RPR}} = ||\Delta\hat{\mathbf{p}}^{tr} - \Delta\mathbf{p}^{tr}||_2 + \beta_2 \big|\big|\Delta\hat{\mathbf{q}} - \frac{\Delta\mathbf{q}}{||\Delta\mathbf{q}||_2}\big|\big|_2,
\end{equation}
with the ground truth relative pose $\Delta\hat{\mathbf{x}} = [\Delta\hat{\mathbf{p}}^{tr}, \Delta\hat{\mathbf{q}}]$. Therefore, we transform the global coordinate systems of consecutive cameras to a local coordinate system $\overline{\mathbf{p}} = \mathbf{R} \mathbf{p}$ with the rotation matrix $\mathbf{R}$, such that $\mathbf{R}^T = \mathbf{R}^{-1}$ and $\mathbf{R}^T\mathbf{R} = \mathbf{R}\mathbf{R}^T = \mathbf{I}$. The relative poses $\Delta\mathbf{p}^{tr}$ represent the differences between the rotated poses. We weight the orientational loss function with $\beta_2 = 50$. Additionally, we use a batch size of 50, the Adam optimizer without decay, and a learning rate of $10^{-4}$. Finally, the relative ground truth labels are transformed based on the current orientation of the absolute pose.

\subsection{Pose Graph Optimization}
\label{sec_method_pgo}

We employ the PGO algorithm \citep{mirowski_grimes} as the state-of-the-art method for fusing absolute and relative poses, which provides smooth and globally consistent pose estimates. PGO can be formulated as a non-convex minimization problem represented by a graph with vertices corresponding to the estimated global poses and edges representing the relative poses. The goal is to ensure that the refined relative poses align with the input relative poses \citep{brahmbhatt_gu_kim,ott_tro}. To reduce runtime complexity, we split the absolute and relative poses into 100-sample chunks with a 20-sample overlap. We conducted a search for the optimal parameters that results in the lowest positioning error.

\subsection{APR-RPR Fusion based on Recurrent Cells}
\label{sec_method_fusion}

Our contribution is a recurrent model for fusing absolute and relative poses to achieve a smoother absolute pose estimation and optimize localization error. To accomplish this, we concatenate the output pose of the absolute method (refer to Section~\ref{sec_method_sfm} and Section~\ref{sec_method_apr}) of size ($N_t \times BS \times 7$) and the output pose of the relative method (refer to Section~\ref{sec_method_rpr}) of size ($N_t \times BS \times 7$), resulting in a network input of size ($N_t \times BS \times 14$), where $N_t$ is the number of timesteps and $BS$ is the batch size. The objective is to predict an improved absolute pose at timestep $t_{n}$ from the concatenated absolute and relative poses of timesteps $\{t_{n-N_t-1}, \ldots, t_{n}\}$. For pose refinement, we implement the following recurrent network, which includes one or two stacked RNN cells. The output size of the first cell is ($N_t \times BS \times 14$), while the output size of the second cell is ($BS \times r_u$), where $r_u$ represents the number of units in the RNN cell. Finally, we include two FC layers, similar to the APR model (refer to Section~\ref{sec_method_apr}), with the RMSE loss function
\begin{equation}
    \mathcal{L}_{\text{APR-RPR}} = ||\hat{\mathbf{p}} - \mathbf{p}||_2^2 + \beta_3 \big|\big|\hat{\mathbf{q}} - \frac{\mathbf{q}}{||\mathbf{q}||_2}\big|\big|_2^2,
\end{equation}
and weighting $\beta_3 = 50$, a batch size of 100, the Adam optimizer without decay, and a learning rate of $10^{-4}$. In Section~\ref{sec_evaluation_results_pgo}, we present a hyperparamter search for $N_t$, $r_u$, and the number of stacked recurrent cells (one or two).

The vanishing gradient problem is a major issue with RNNs, which occurs when gradients become too small as they propagate backwards through time during weight updates. This can be problematic in our fusion model due to small orientation entries (i.e., quaternions with magnitudes below 1), hindering the learning of long-term dependencies. To address this issue, we explore various recurrent cells that can allow the model to learn the dynamics of an object (i.e., slow moving robot versus fast moving human) on the low-level pose representation. Object movements typically follow predictable, non-chaotic patterns that follow physical behavior \citep{ott_wacv}. One solution is to use gating mechanisms such as those found in long short-term memory (LSTM) \citep{hochreiter_schmidhuber} and gated recurrent unit (GRU) \citep{chung_gulcehre_cho}. Another option is the minimal gated unit (MGU) \citep{zhou_wu_zhang,heck_salem}, which is a simplified version of LSTM and GRU that uses only an update and reset gate. The recurrent additive network (RAN) \citep{lee_levy_zettlemoyer} is another type of gates RNN that uses purely additive latent state updates. For greater parallelism, we evaluate the simple recurrent unit (SRU) \citep{lei_zhang_artzi}, which simplifies computations as the majority of computations for each step are independent of recurrence. Additionally, quasi-recurrent neural networks (QRNN) \citep{bradbury_merity_xiong} apply minimalistic recurrent pooling functions in parallel across channels. \cite{balduzzi_ghifary} proposed strongly-typed RNN (TRNN), which learns simple semantic interpretations via dynamic average pooling. The chaos free network (CFN), proposed by \cite{laurent_brecht}, is a type of RNN that models non-chaotic dynamics.

Furthermore, we evaluate convolutional networks, such as FCN \citep{wang_yan_oates}, TCN \citep{bai_kolter_koltun}, ResNet \citep{wang_yan_oates}, ResCNN \citep{zou_wang_li}, InceptionTime \citep{fawaz_lucas_forestier}, XceptionTime~\citep{rahimian_zabihi_atashzar}, and OmniScaleCNN \citep{tang_long_liu}, a combination of recurrent with fully convolutional networks, i.e., LSTM-FCN \citep{karim_majumdar_darabi}, GRU-FCN \citep{elsaye_maida_bayoumi}, and MLSTM-FCN \citep{karim_majumdar}, Transformer models, i.e., TST \citep{zerveas_jayaraman_patel}, TSPerceiver \citep{jaegle_borgeaud_alayrac}, and TSSequencerPlus \citep{tatsunami_taki}, as well as mWDN \citep{wang_wang_li}, XCM \citep{fauvel_fromont_masson}, and gMLP \citep{liu_dai_so_le}.

\subsection{Simulation-Augmented Pre-Training}
\label{sec_method_simulation}

\begin{figure}[!t]
    \centering
	\begin{minipage}[t]{0.48\linewidth}
        \centering
    	\includegraphics[trim=0 0 0 50, clip, width=1.0\linewidth]{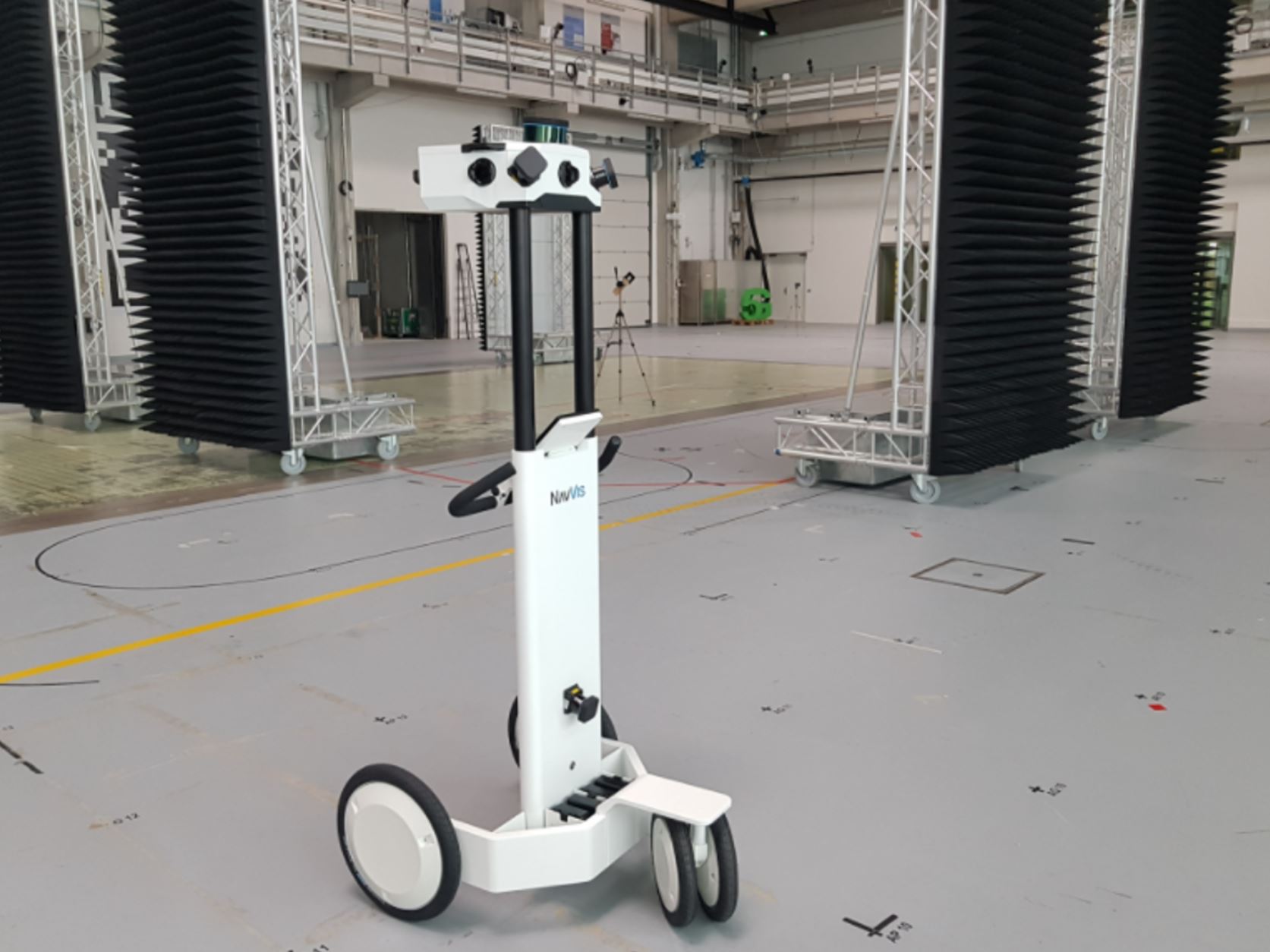}
    	\subcaption{NavVis M4.}
    	\label{figure_simulation_images1}
    \end{minipage}
    \hfill
	\begin{minipage}[t]{0.48\linewidth}
        \centering
    	\includegraphics[width=1.0\linewidth]{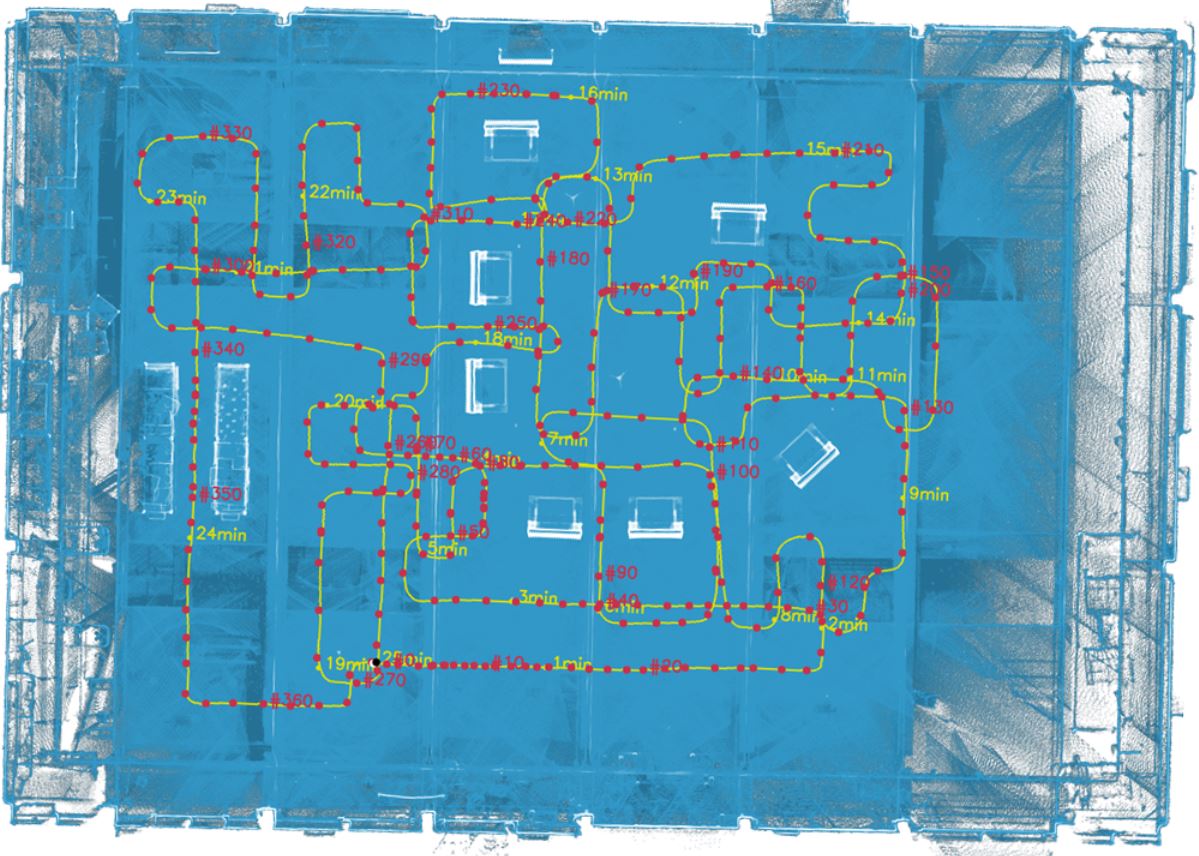}
    	\subcaption{Recording trajectory.}
    	\label{figure_simulation_images2}
    \end{minipage}
	\begin{minipage}[t]{0.48\linewidth}
        \centering
    	\includegraphics[width=1.0\linewidth]{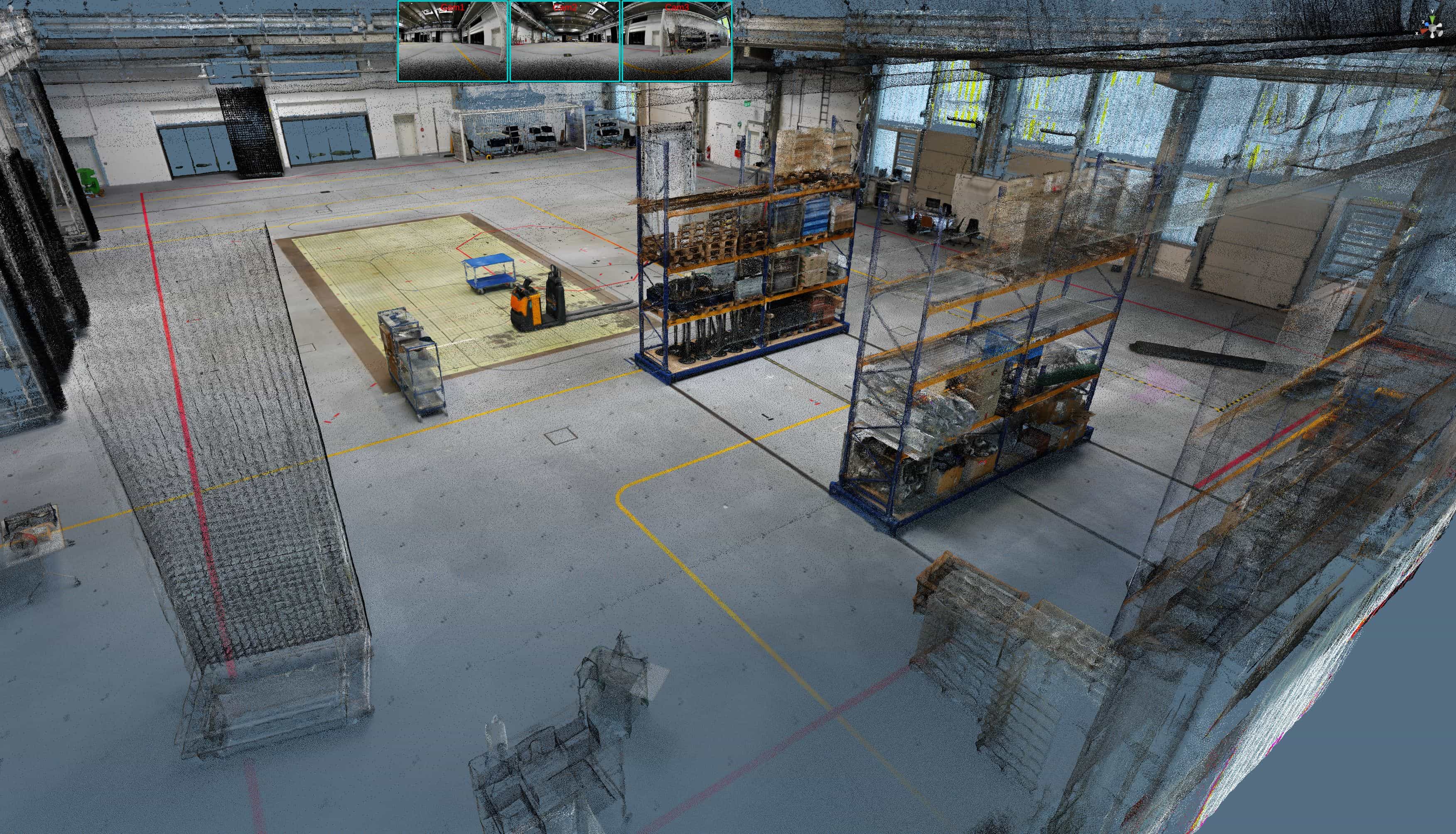}
    	\subcaption{Simulated point cloud.}
    	\label{figure_simulation_images3}
    \end{minipage}
    \hfill
	\begin{minipage}[t]{0.48\linewidth}
        \centering
    	\includegraphics[width=1.0\linewidth]{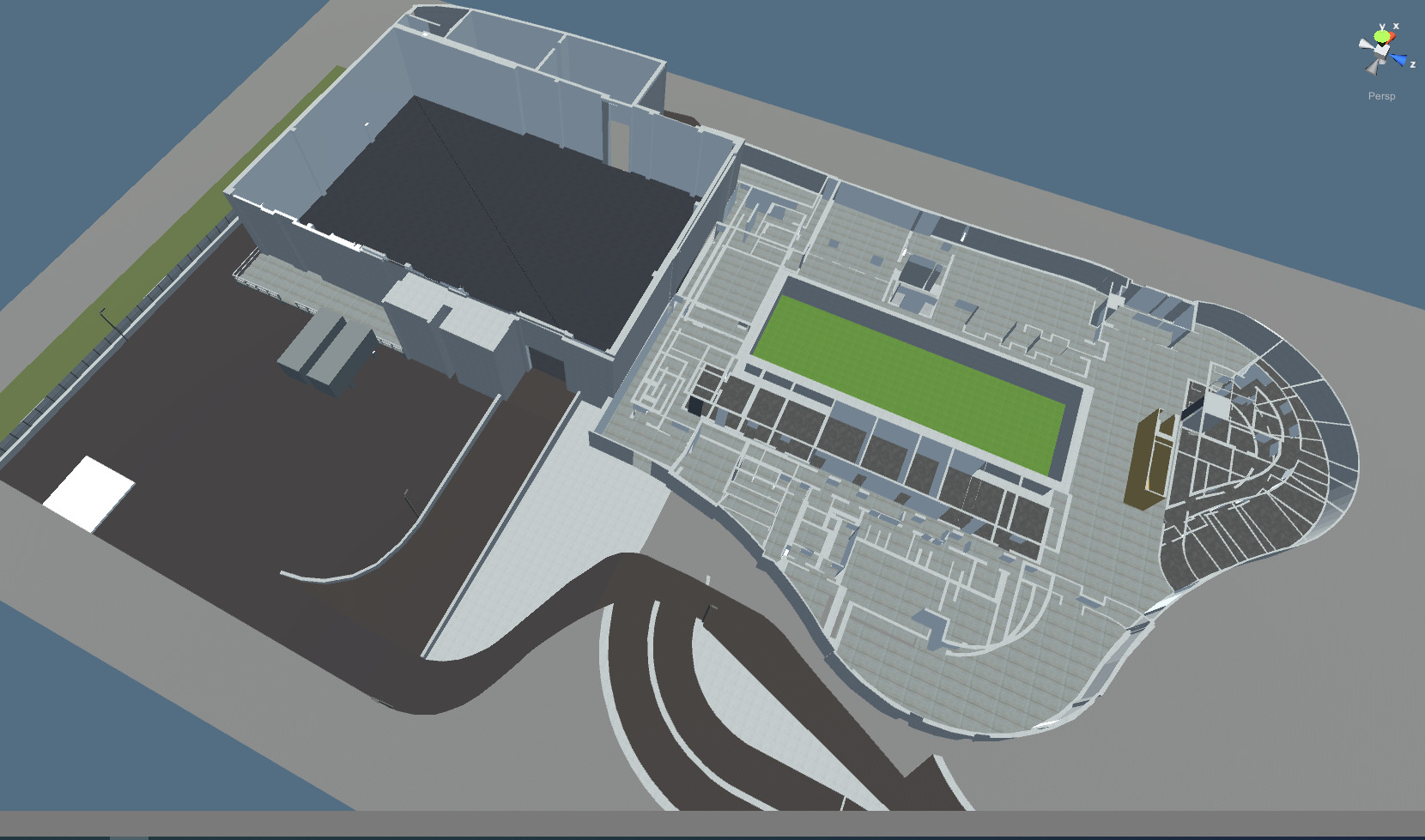}
    	\subcaption{Simulation framework.}
    	\label{figure_simulation_images4}
    \end{minipage}
    \caption{Creation of simulated images for pre-training.}
    \label{figure_simulation_images}
\end{figure}

Pre-training neural networks using simulated data can enhance their performance in various tasks, especially in the case of data scarcity, domain adaptation, data augmentation, and providing a good initialization for transfer learning to real data \citep{zhang_bengio_hardt,shrivastava_pfister_tuzel}. Previous studies in the context of visual self-localization \citep{winkelbauer_denninger,idan_shavit_keller,wang_qi} have demonstrated the potential benefits of pre-training on synthetic data for improved performance. To this end, we develop a simulation environment with a loaded point cloud of the real-world environment to generate a large number of samples and augment the APR and RPR models for better initial weights, improved adaptability to unseen scenarios, and enhanced transfer learning to changes in the environment. Figure~\ref{figure_simulation_images} illustrates the data generation process. Initially, we record a large dataset using a NavVis M4 system (see Figure~\ref{figure_simulation_images1}) while moving in an environment with seven large black absorber walls and two warehouse racks, as shown in Figure~\ref{figure_simulation_images2}. Next, we create a detailed point cloud (see Figure~\ref{figure_simulation_images3}) from this dataset, which was then loaded into the simulation framework implemented by Shhuna GmbH (see Figure~\ref{figure_simulation_images4}). We only use the large hall of the simulation framework (left part) for data generation. Using a fish-eye camera (in the simulation), we generated 319,955 images mimicking a slow moving robot, see Figure~\ref{figure_example_of} (third and fourth column) for exemplary images. We concurrently capture images from three cameras, including one front-facing camera and two side-facing cameras. We utilized this dataset to pre-train the APR and RPR models on a large dataset.
\section{Experiments}
\label{sec_experiments}

In the following section, we provide details regarding the data collection and experiments. Specifically, Section~\ref{sec_experiments_datasets} offers an overview of our datasets and the challenges associated with localization. Section~\ref{sec_evaluation_metrics} provides a summary of our experiments. Finally, in Section~\ref{sec_experiments_hardware}, we describe the hardware setup and evaluation metrics used in our study.

\begin{table*}[t!]
\begin{center}
    \caption{Overview of the visual Industry scenario \#4 datasets recorded in a large-scale indoor environment with a robot or handheld.}
    \label{table_dataset_overview}
    \small \begin{tabular}{ p{0.5cm} | p{0.5cm} | p{0.5cm}}
    \multicolumn{1}{c|}{\textbf{Dataset}} & \multicolumn{1}{c|}{\textbf{Setup}} & \multicolumn{1}{c}{\textbf{\# Images}} \\ \hline
    \multicolumn{1}{l|}{Train 1} & \multicolumn{1}{l|}{Robot, clear environment} & \multicolumn{1}{r}{92,668} \\
    \multicolumn{1}{l|}{Train 2} & \multicolumn{1}{l|}{Robot, four absorber walls with objects} & \multicolumn{1}{r}{19,872} \\
    \multicolumn{1}{l|}{Train 3} & \multicolumn{1}{l|}{Robot, four absorber walls} & \multicolumn{1}{r}{138,233} \\
    \multicolumn{1}{l|}{Train 4} & \multicolumn{1}{l|}{Handheld, person 1, four absorber walls} & \multicolumn{1}{r}{27,856} \\
    \multicolumn{1}{l|}{Train 5} & \multicolumn{1}{l|}{Handheld, person 2, four absorber walls} & \multicolumn{1}{r}{28,113} \\
    \multicolumn{1}{l|}{Train 6} & \multicolumn{1}{l|}{Robot, open environment with objects} & \multicolumn{1}{r}{114,998} \\
    \multicolumn{1}{l|}{Train 7} & \multicolumn{1}{l|}{Robot, open environment with objects and labyrinth} & \multicolumn{1}{r}{29,240} \\
    \multicolumn{1}{l|}{Train 8} & \multicolumn{1}{l|}{Robot, open environment with object, labyrinth, and absorber walls} & \multicolumn{1}{r}{110,923} \\
    \multicolumn{1}{l|}{Test 1} & \multicolumn{1}{l|}{Robot, clear environment} & \multicolumn{1}{r}{23,168} \\
    \multicolumn{1}{l|}{Test 2} & \multicolumn{1}{l|}{Robot, one absorber wall} & \multicolumn{1}{r}{30,379} \\
    \multicolumn{1}{l|}{Test 3} & \multicolumn{1}{l|}{Robot, two absorber walls} & \multicolumn{1}{r}{24,752} \\
    \multicolumn{1}{l|}{Test 4} & \multicolumn{1}{l|}{Robot, three absorber walls} & \multicolumn{1}{r}{26,877} \\
    \multicolumn{1}{l|}{Test 5} & \multicolumn{1}{l|}{Robot, four absorber walls with objects} & \multicolumn{1}{r}{4,968} \\
    \multicolumn{1}{l|}{Test 6} & \multicolumn{1}{l|}{Robot, four absorber walls} & \multicolumn{1}{r}{34,559} \\
    \multicolumn{1}{l|}{Test 7} & \multicolumn{1}{l|}{Handheld, person 1, four absorber walls} & \multicolumn{1}{r}{6,964} \\
    \multicolumn{1}{l|}{Test 8} & \multicolumn{1}{l|}{Handheld, person 2, four absorber walls} & \multicolumn{1}{r}{7,029} \\
    \multicolumn{1}{l|}{Test 9} & \multicolumn{1}{l|}{Robot, open environment with objects} & \multicolumn{1}{r}{28,750} \\
    \multicolumn{1}{l|}{Test 10} & \multicolumn{1}{l|}{Robot, open environment with objects and labyringth} & \multicolumn{1}{r}{7,311} \\
    \multicolumn{1}{l|}{Test 11} & \multicolumn{1}{l|}{Robot, open environment with objects, labyrinth,  and absorber walls} & \multicolumn{1}{r}{27,732} \\
    \end{tabular}
\end{center}
\end{table*}

\subsection{Datasets \& Challenges}
\label{sec_experiments_datasets}

\begin{figure*}[!t]
    \centering
	\begin{minipage}[t]{0.245\linewidth}
        \centering
    	\includegraphics[trim=0 11 0 10, clip, width=1.0\linewidth]{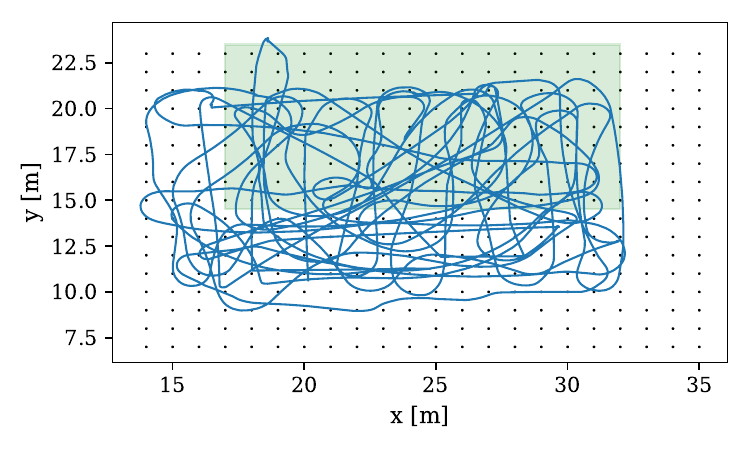}
    	\subcaption{Train 1.}
    	\label{figure_dataset_traj1}
    \end{minipage}
    \hfill
	\begin{minipage}[t]{0.245\linewidth}
        \centering
    	\includegraphics[trim=0 11 0 10, clip, width=1.0\linewidth]{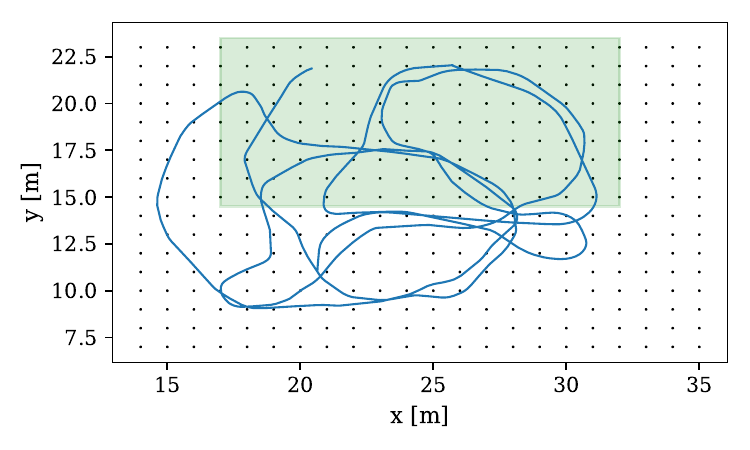}
    	\subcaption{Train 2.}
    	\label{figure_dataset_traj2}
    \end{minipage}
    \hfill
	\begin{minipage}[t]{0.245\linewidth}
        \centering
    	\includegraphics[trim=0 11 0 10, clip, width=1.0\linewidth]{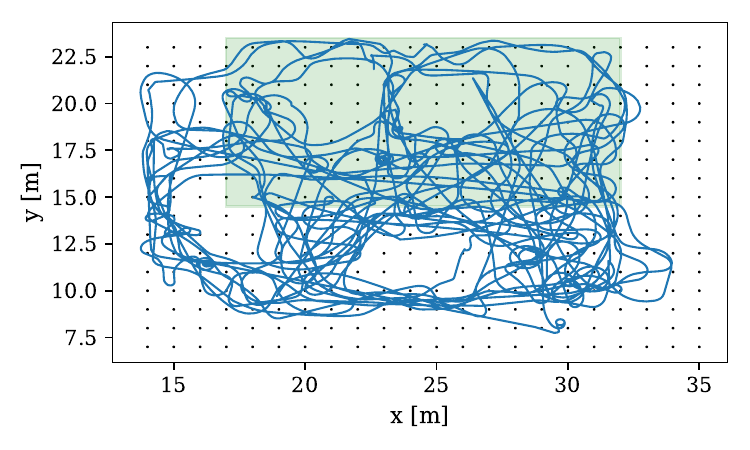}
    	\subcaption{Train 3.}
    	\label{figure_dataset_traj3}
    \end{minipage}
    \hfill
	\begin{minipage}[t]{0.245\linewidth}
        \centering
    	\includegraphics[trim=0 11 0 10, clip, width=1.0\linewidth]{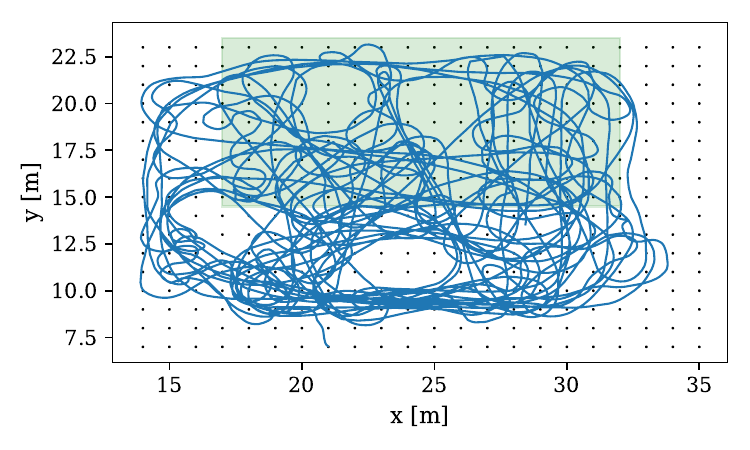}
    	\subcaption{Train 4.}
    	\label{figure_dataset_traj4}
    \end{minipage}
	\begin{minipage}[t]{0.245\linewidth}
        \centering
    	\includegraphics[trim=0 11 0 10, clip, width=1.0\linewidth]{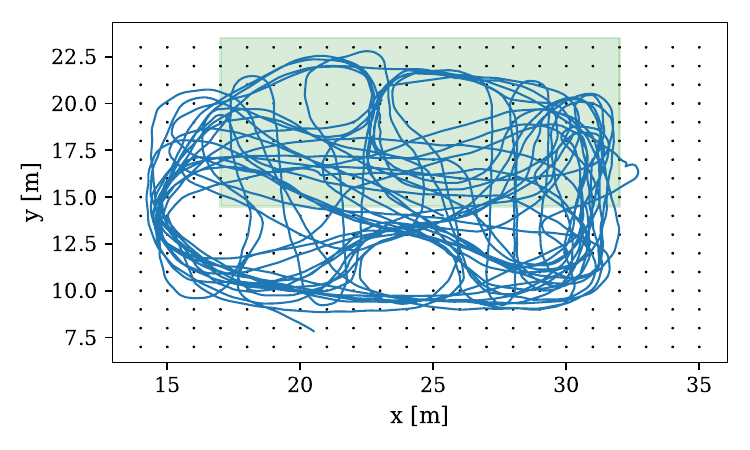}
    	\subcaption{Train 5.}
    	\label{figure_dataset_traj5}
    \end{minipage}
    \hfill
	\begin{minipage}[t]{0.245\linewidth}
        \centering
    	\includegraphics[trim=0 11 0 10, clip, width=1.0\linewidth]{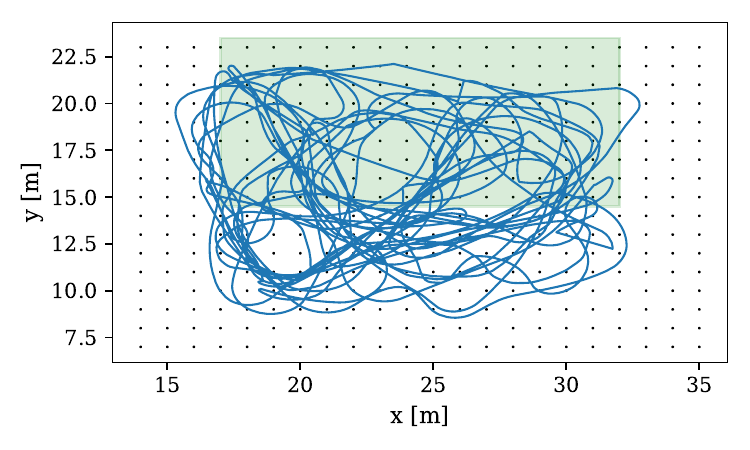}
    	\subcaption{Train 6.}
    	\label{figure_dataset_traj6}
    \end{minipage}
    \hfill
	\begin{minipage}[t]{0.245\linewidth}
        \centering
    	\includegraphics[trim=0 11 0 10, clip, width=1.0\linewidth]{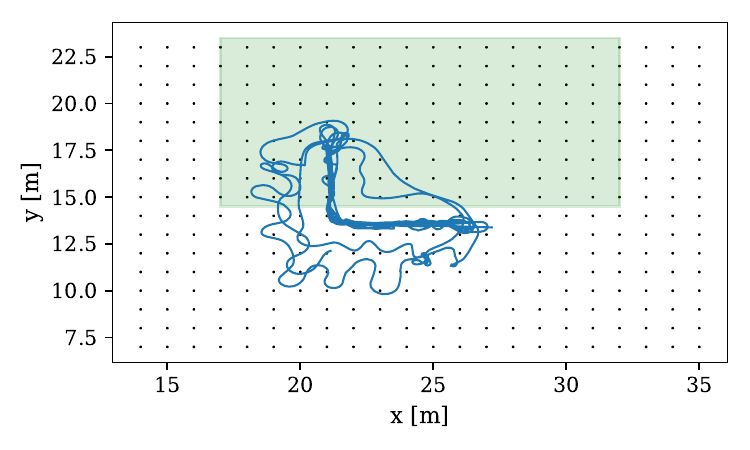}
    	\subcaption{Train 7.}
    	\label{figure_dataset_traj7}
    \end{minipage}
    \hfill
	\begin{minipage}[t]{0.245\linewidth}
        \centering
    	\includegraphics[trim=0 11 0 10, clip, width=1.0\linewidth]{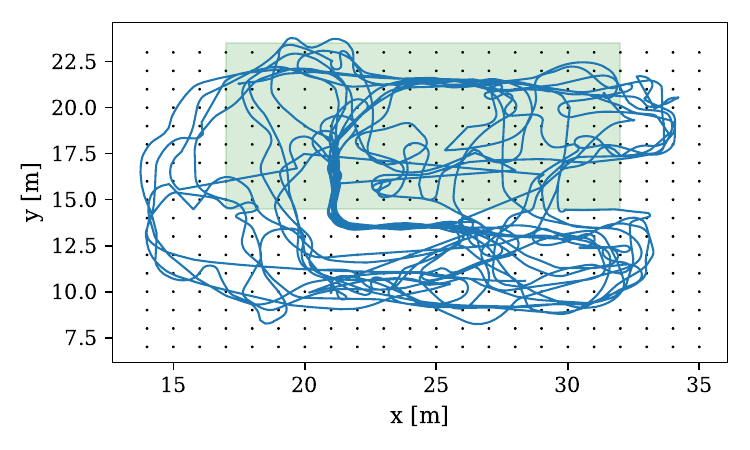}
    	\subcaption{Train 8.}
    	\label{figure_dataset_traj8}
    \end{minipage}
	\begin{minipage}[t]{0.245\linewidth}
        \centering
    	\includegraphics[trim=0 11 0 10, clip, width=1.0\linewidth]{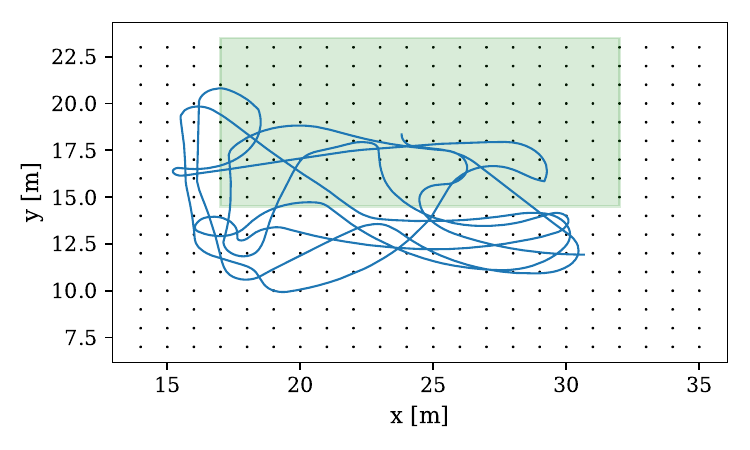}
    	\subcaption{Test 1.}
    	\label{figure_dataset_traj9}
    \end{minipage}
    \hfill
	\begin{minipage}[t]{0.245\linewidth}
        \centering
    	\includegraphics[trim=0 11 0 10, clip, width=1.0\linewidth]{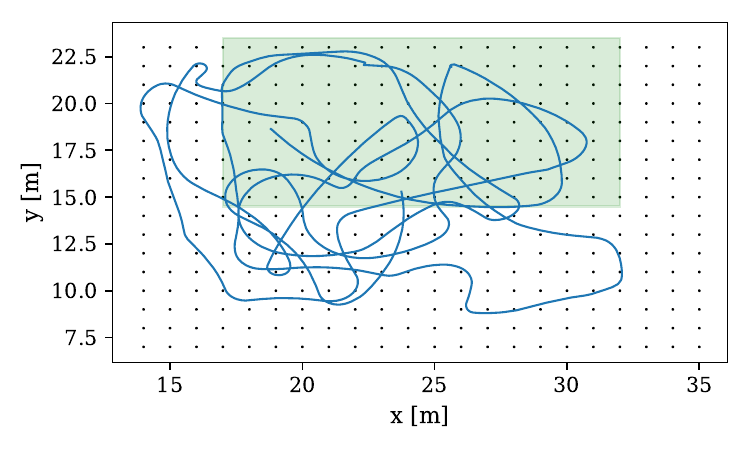}
    	\subcaption{Test 2.}
    	\label{figure_dataset_traj10}
    \end{minipage}
    \hfill
	\begin{minipage}[t]{0.245\linewidth}
        \centering
    	\includegraphics[trim=0 11 0 10, clip, width=1.0\linewidth]{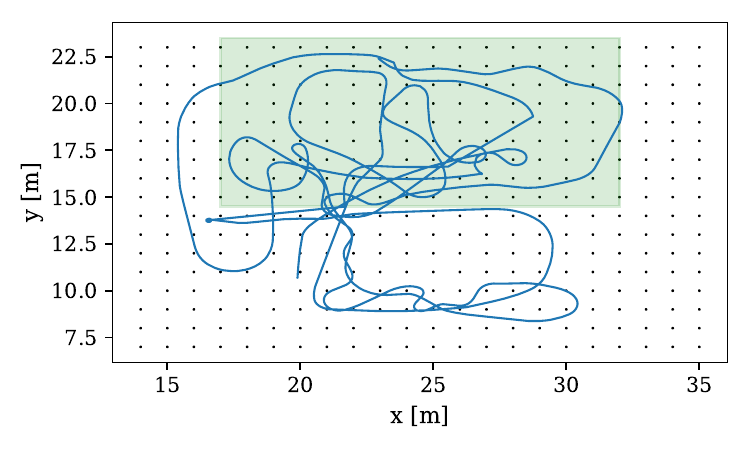}
    	\subcaption{Test 3.}
    	\label{figure_dataset_traj11}
    \end{minipage}
    \hfill
	\begin{minipage}[t]{0.245\linewidth}
        \centering
    	\includegraphics[trim=0 11 0 10, clip, width=1.0\linewidth]{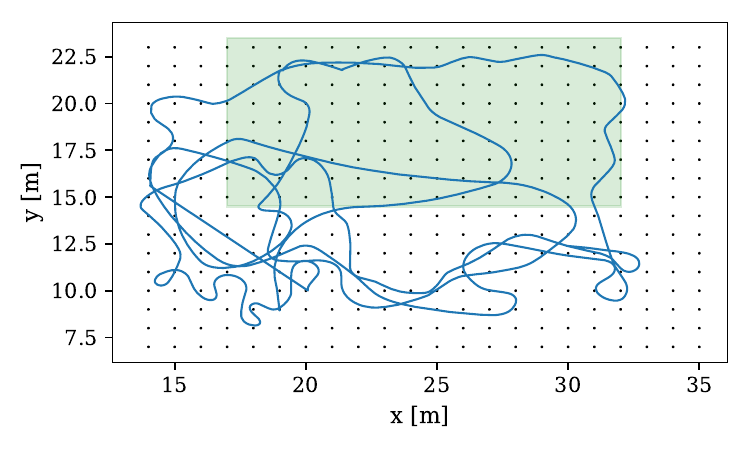}
    	\subcaption{Test 4.}
    	\label{figure_dataset_traj12}
    \end{minipage}
	\begin{minipage}[t]{0.245\linewidth}
        \centering
    	\includegraphics[trim=0 11 0 10, clip, width=1.0\linewidth]{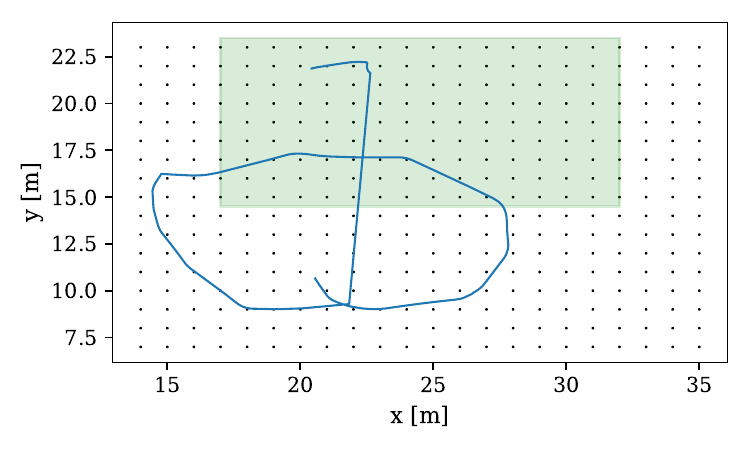}
    	\subcaption{Test 5.}
    	\label{figure_dataset_traj13}
    \end{minipage}
    \hfill
	\begin{minipage}[t]{0.245\linewidth}
        \centering
    	\includegraphics[trim=0 11 0 10, clip, width=1.0\linewidth]{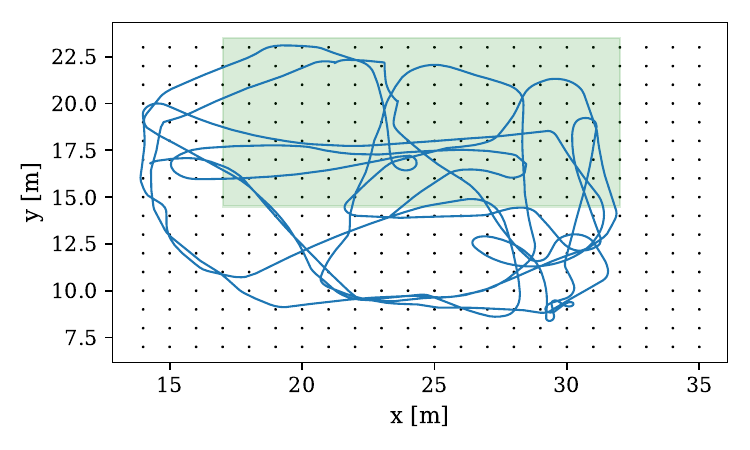}
    	\subcaption{Test 6.}
    	\label{figure_dataset_traj14}
    \end{minipage}
    \hfill
	\begin{minipage}[t]{0.245\linewidth}
        \centering
    	\includegraphics[trim=0 11 0 10, clip, width=1.0\linewidth]{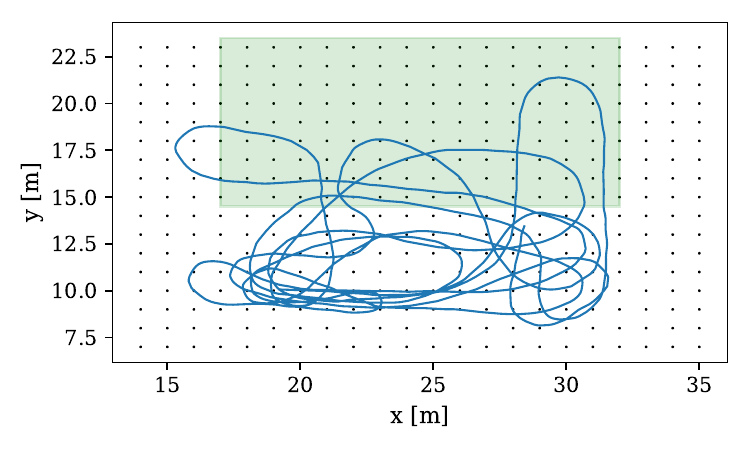}
    	\subcaption{Test 7.}
    	\label{figure_dataset_traj15}
    \end{minipage}
    \hfill
	\begin{minipage}[t]{0.245\linewidth}
        \centering
    	\includegraphics[trim=0 11 0 10, clip, width=1.0\linewidth]{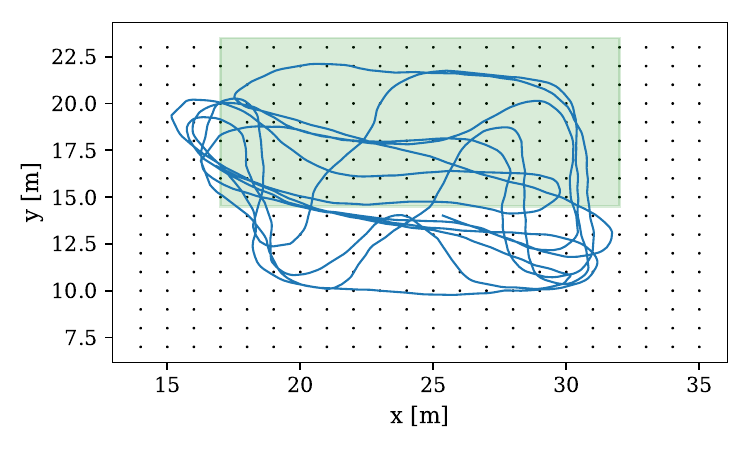}
    	\subcaption{Test 8.}
    	\label{figure_dataset_traj16}
    \end{minipage}
	\begin{minipage}[t]{0.245\linewidth}
        \centering
    	\includegraphics[trim=0 11 0 10, clip, width=1.0\linewidth]{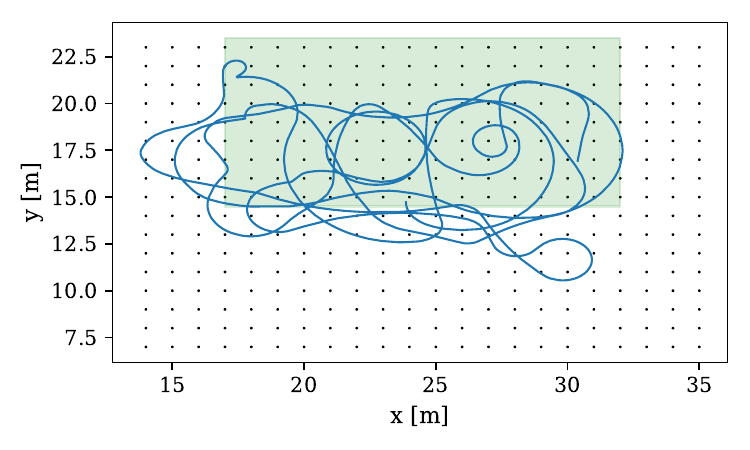}
    	\subcaption{Test 9.}
    	\label{figure_dataset_traj17}
    \end{minipage}
	\begin{minipage}[t]{0.245\linewidth}
        \centering
    	\includegraphics[trim=0 11 0 10, clip, width=1.0\linewidth]{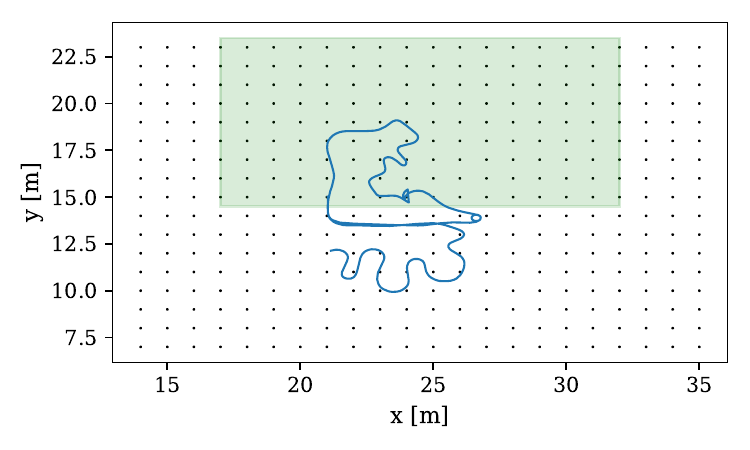}
    	\subcaption{Test 10.}
    	\label{figure_dataset_traj18}
    \end{minipage}
	\begin{minipage}[t]{0.245\linewidth}
        \centering
    	\includegraphics[trim=0 11 0 10, clip, width=1.0\linewidth]{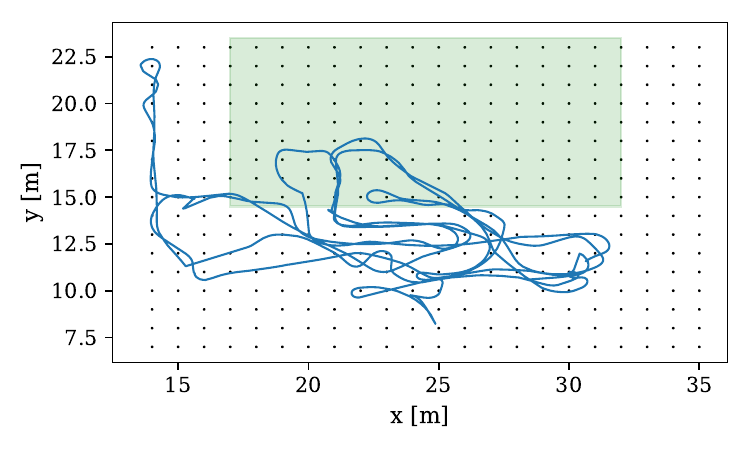}
    	\subcaption{Test 11.}
    	\label{figure_dataset_traj19}
    \end{minipage}
    \hfill
    \caption{Trajectories of the Industry datasets with $x \in [14, 34]$ and $y \in [7, 23]$.}
    \label{figure_dataset_traj}
\end{figure*}

Numerous datasets are currently accessible for assessing APR techniques. Nevertheless, the present datasets possess limitations such as being captured outside the industrial environments or with equipment such as micro aerial vehicles (MAVs) or handheld devices, or are inadequate for evaluating specific scenarios like changes in the environment. Therefore, we have captured the Industry dataset in a vast large-scale industrial environment covering an area of $1,320m^2$, similar to the environment in \cite{loeffler_riechel,ott_cvprw,ott_tro,stahlke_sensors}. A small robotic recording platform is constructed, equipped with an Orbbec3D camera featuring an RGB image resolution of $640 \times 480$ pixels and a recording frequency of 23\,Hz. To measure reference poses at a high-precision level ($< 1mm$), a motion capture system operating at 140\,Hz is utilized. Eight training and 11 testing datasets are recorded, comprising 10 distinct scenarios, as presented in Table~\ref{table_dataset_overview}. Trajectories are depicted in Figure~\ref{figure_dataset_traj}. Changes are introduced to the environment between recordings to asses the model's robustness against volatile objects, i.e., the removal or addition of absorber walls, and the ability to generalize and adapt to various scenarios and motion dynamics. Initially, a large-scale clear environment without objects was recorded. Subsequently, we introduced one, two, three, or four absorber walls, along with smaller objects. Additionally, a small ``labyrinth'' in an L-like configuration was constructed, as shown in Figure~\ref{figure_dataset_traj7} and Figure~\ref{figure_dataset_traj18}. Finally, we captured the motion dynamics of two individuals randomly walking in the environment, with four absorber walls present (see trajectories in Figure~\ref{figure_dataset_traj4}, \ref{figure_dataset_traj5}, \ref{figure_dataset_traj15}, and \ref{figure_dataset_traj16}).

\begin{figure}[!t]
    \centering
	\begin{minipage}[t]{0.325\linewidth}
        \centering
    	\includegraphics[width=1.0\linewidth]{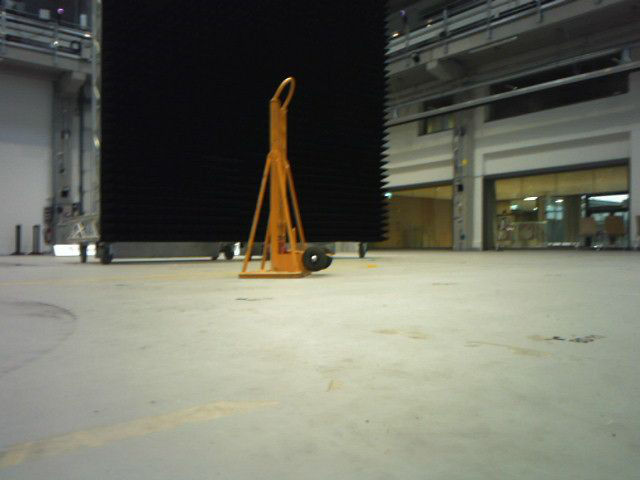}
    	\subcaption{Feature-less absorber walls.}
    	\label{figure_data_images1}
    \end{minipage}
    \hfill
	\begin{minipage}[t]{0.325\linewidth}
        \centering
    	\includegraphics[width=1.0\linewidth]{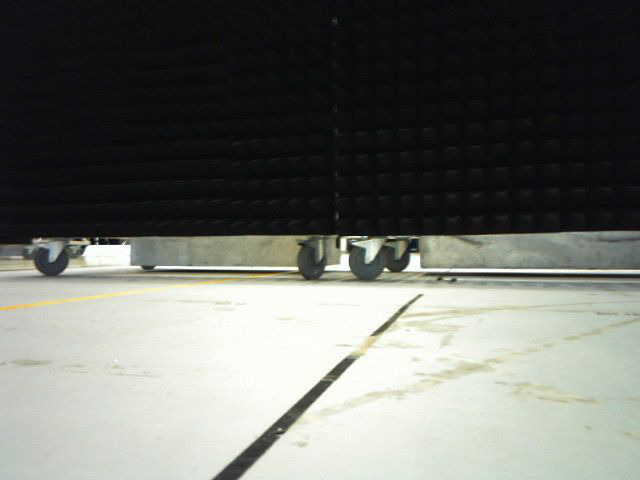}
    	\subcaption{Feature-less absorber walls.}
    	\label{figure_data_images2}
    \end{minipage}
    \hfill
	\begin{minipage}[t]{0.325\linewidth}
        \centering
    	\includegraphics[width=1.0\linewidth]{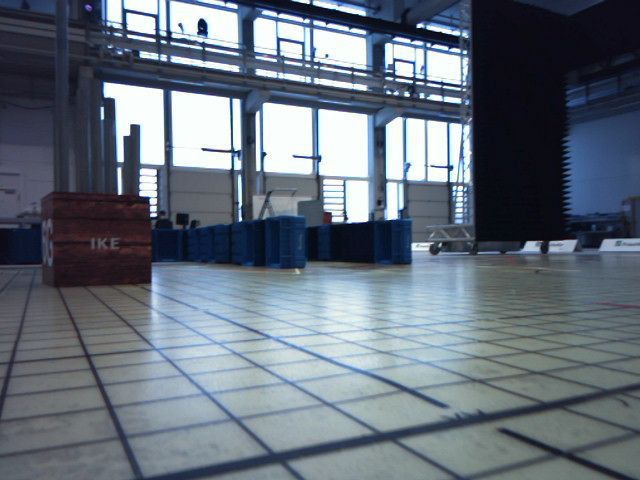}
    	\subcaption{Harsh lighting conditions.}
    	\label{figure_data_images3}
    \end{minipage}
	\begin{minipage}[t]{0.325\linewidth}
        \centering
    	\includegraphics[width=1.0\linewidth]{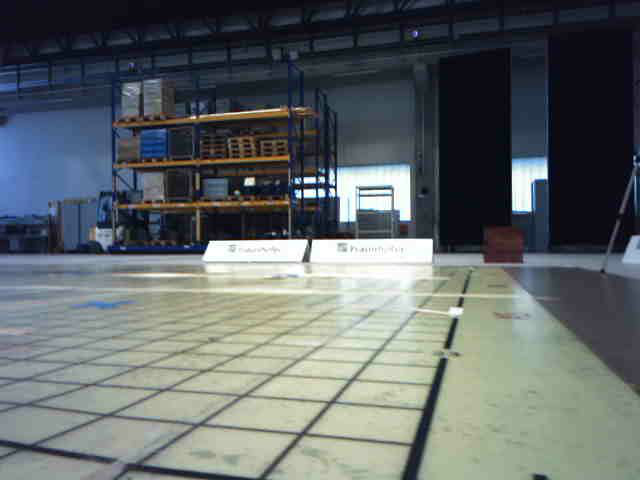}
    	\subcaption{Warehouse racks.}
    	\label{figure_data_images4}
    \end{minipage}
    \hfill
	\begin{minipage}[t]{0.325\linewidth}
        \centering
    	\includegraphics[width=1.0\linewidth]{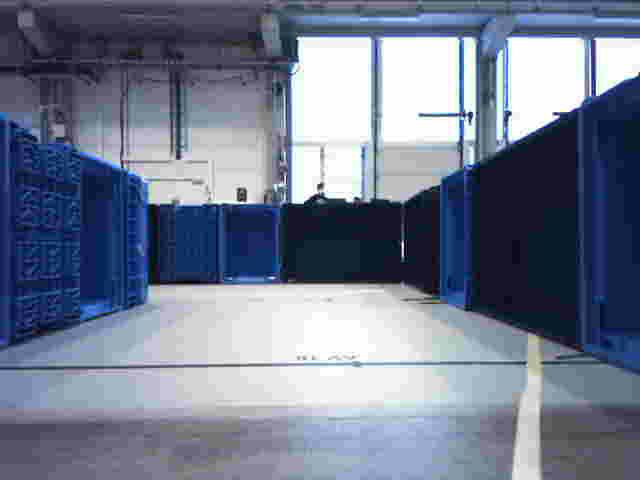}
    	\subcaption{Labyrinth.}
    	\label{figure_data_images5}
    \end{minipage}
    \hfill
	\begin{minipage}[t]{0.325\linewidth}
        \centering
    	\includegraphics[width=1.0\linewidth]{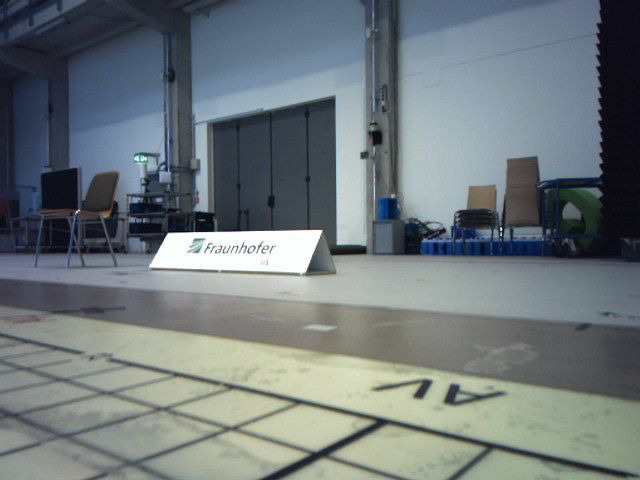}
    	\subcaption{Texture-less walls.}
    	\label{figure_data_images6}
    \end{minipage}
    \caption{Exemplary challenging images of the large-scale indoor environment.}
    \label{figure_data_images}
\end{figure}

The motion dynamics between the robot and humans are different. Cross-validation is conducted for all training and test datasets. Figure~\ref{figure_data_images} presents images that are challenging with respect to the localization task, including the feature-less structure of the absorber walls (Figure~\ref{figure_data_images1} and Figure~\ref{figure_data_images2}), which makes it particularly hard for SfM to extract features and match image points, or different scalings between distant warehouse racks (Figure~\ref{figure_data_images4}) and the small-scale labyrinth (Figure~\ref{figure_data_images5}). See \cite{zangeneh_bruns_dekel}, for challenges under ambiguous scenes.

\subsection{Overview of Experiments}
\label{sec_evaluation_metrics}

Table~\ref{table_evaluation_overview} provides a comprehensive summary of the experiments conducted in our study. Our approach to estimating absolute pose involves either applying SfM (see Section~\ref{sec_method_sfm}) or using the time-distributed APR model (see Section~\ref{sec_method_apr}). To predict relative poses, we utilize the RPR model (see Section~\ref{sec_method_rpr}). We evaluate the performance of both APR and RPR models with and without pre-training on the synthetically generated dataset (see Section~\ref{sec_method_simulation}). In terms of fusion, we compare our proposed fusion framework with eight different recurrent cells, namely LSTM, GRU, MGU, RAN, SRU, QRNN, TRNN, and CFN (see Section~\ref{sec_method_fusion}), against the state-of-the-art PGO technique (see Section~\ref{sec_method_pgo}).

\begin{table}[t!]
\begin{center}
    \caption{Overview of different fusion combinations.}
    \label{table_evaluation_overview}
    \footnotesize \begin{tabular}{ p{0.5cm} | p{0.5cm} | p{0.5cm}}
    \multicolumn{1}{c|}{\textbf{Absolute Pose}} & \multicolumn{1}{c|}{\textbf{Relative Pose}} & \multicolumn{1}{c}{\textbf{Fusion}} \\ \hline
    \multicolumn{1}{l|}{SfM} & \multicolumn{1}{l|}{RPR} & \multicolumn{1}{l}{PGO} \\
    \multicolumn{1}{l|}{SfM} & \multicolumn{1}{l|}{RPR (pre-trained)} & \multicolumn{1}{l}{PGO} \\
    \multicolumn{1}{l|}{APR} & \multicolumn{1}{l|}{RPR} & \multicolumn{1}{l}{PGO} \\
    \multicolumn{1}{l|}{APR} & \multicolumn{1}{l|}{RPR (pre-trained)} & \multicolumn{1}{l}{PGO} \\
    \multicolumn{1}{l|}{APR (pre-trained)} & \multicolumn{1}{l|}{RPR} & \multicolumn{1}{l}{PGO} \\
    \multicolumn{1}{l|}{APR (pre-trained)} & \multicolumn{1}{l|}{RPR (pre-trained)} & \multicolumn{1}{l}{PGO} \\
    \multicolumn{1}{l|}{SfM} & \multicolumn{1}{l|}{RPR} & \multicolumn{1}{l}{Recurrent model} \\
    \multicolumn{1}{l|}{SfM} & \multicolumn{1}{l|}{RPR (pre-trained)} & \multicolumn{1}{l}{Recurrent model} \\
    \multicolumn{1}{l|}{APR} & \multicolumn{1}{l|}{RPR} & \multicolumn{1}{l}{Recurrent model} \\
    \multicolumn{1}{l|}{APR} & \multicolumn{1}{l|}{RPR (pre-trained)} & \multicolumn{1}{l}{Recurrent model} \\
    \multicolumn{1}{l|}{APR (pre-trained)} & \multicolumn{1}{l|}{RPR} & \multicolumn{1}{l}{Recurrent model} \\
    \multicolumn{1}{l|}{APR (pre-trained)} & \multicolumn{1}{l|}{RPR (pre-trained)} & \multicolumn{1}{l}{Recurrent model} \\
    \end{tabular}
\end{center}
\end{table}

\subsection{Hardware Setup \& Evaluation Metrics}
\label{sec_experiments_hardware}

For all experiments, we use Nvidia Tesla V100-SXM2 GPUs with 32 GB VRAM equipped with Core Xeon CPUs and 192 GB RAM. We use the Adam optimizer with a learning rate of $10^{-4}$. We run each experiment for $\text{epochs} = (\text{iterations} \cdot \text{BS})/\text{dataset\_size}$, where we set iterations to 150,000 for APR and RPR and to 75,000 for the fusion models, the batch size BS is 50 for APR and RPR and 100 for the fusion models, and the dataset size depends on the scenario according to Table~\ref{table_dataset_overview}. We report results for the last epoch. For the evaluation of absolute predictions, we report the median absolute position in $m$ and the median absolute orientation in \textdegree.

The network framework proposed in this paper comprises multiple training stages. Initially, the dataset undergoes recording and pre-processing procedures. Subsequently, the SfM algorithm is employed to construct a point cloud, facilitating the computation of absolute poses from testing images. Additionally, the RPR network is trained on optical flow images generated using the Lucas-Kanade algorithm. The RPR network predicts relative poses between pairs of consecutive images. Finally, the third network, known as the fusion model, is trained using both absolute and relative poses from the training set. During the evaluation stage, the fusion model utilizes the absolute and relative poses predicted from testing images to estimate absolute poses.
\section{Evaluation}
\label{sec_evaluation}

\subsection{Hyperparameter Search for SfM}
\label{sec_experiments_search}

\begin{figure}[!t]
    \centering
	\begin{minipage}[t]{0.493\linewidth}
        \centering
    	\includegraphics[trim=10 10 10 10, clip, width=1.0\linewidth]{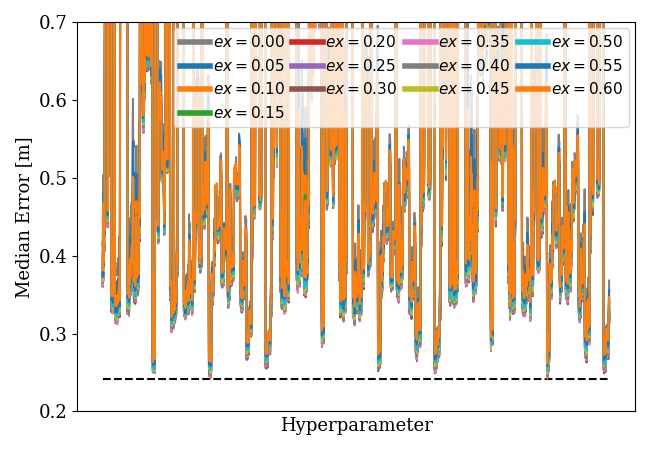}
    	\subcaption{Median position error in $m$.}
    	\label{figure_sfm_robot1}
    \end{minipage}
    \hfill
	\begin{minipage}[t]{0.493\linewidth}
        \centering
    	\includegraphics[trim=10 10 10 10, clip, width=1.0\linewidth]{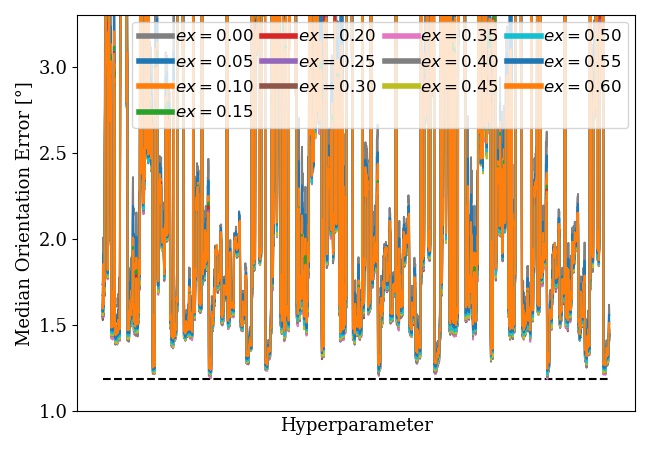}
    	\subcaption{Median orientation error in \textdegree.}
    	\label{figure_sfm_robot2}
    \end{minipage}
    \caption{Evaluation of all SfM point clouds (for different hyperparameters, see x-axis) for the robot train 3 and test 6 datasets. The dashed lines indicate the lowest error.}
    \label{figure_sfm_robot}
\end{figure}

\begin{figure}[!t]
    \centering
	\begin{minipage}[t]{0.493\linewidth}
        \centering
    	\includegraphics[trim=10 10 10 10, clip, width=1.0\linewidth]{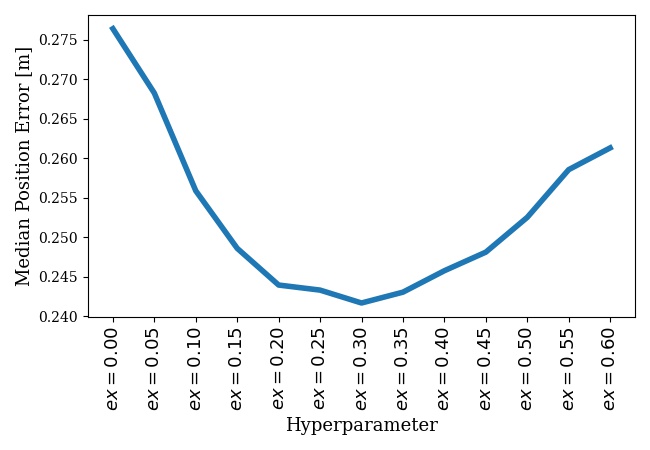}
    	\subcaption{Median position error in $m$.}
    	\label{figure_best_robot1}
    \end{minipage}
    \hfill
	\begin{minipage}[t]{0.493\linewidth}
        \centering
    	\includegraphics[trim=10 10 10 10, clip, width=1.0\linewidth]{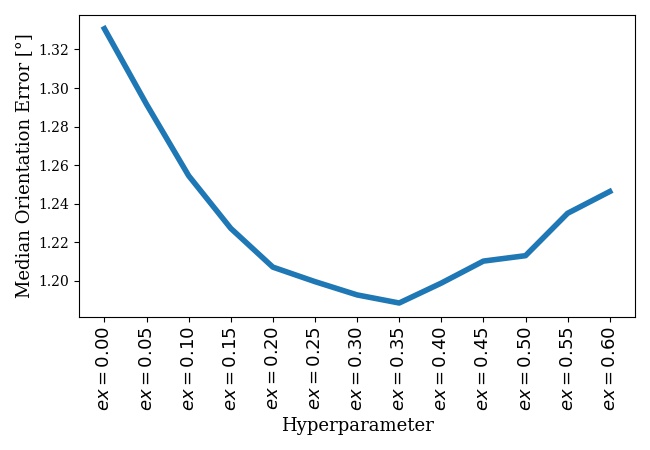}
    	\subcaption{Median orientation error in \textdegree.}
    	\label{figure_best_robot2}
    \end{minipage}
    \caption{Evaluation of the best SfM point cloud (from Figure~\ref{figure_sfm_robot}) for different hyperparameters $exclude$ for the robot train 3 and test 6 datasets.}
    \label{figure_best_robot}
\end{figure}

\begin{figure}[!t]
    \centering
	\begin{minipage}[t]{0.493\linewidth}
        \centering
    	\includegraphics[trim=10 10 10 10, clip, width=1.0\linewidth]{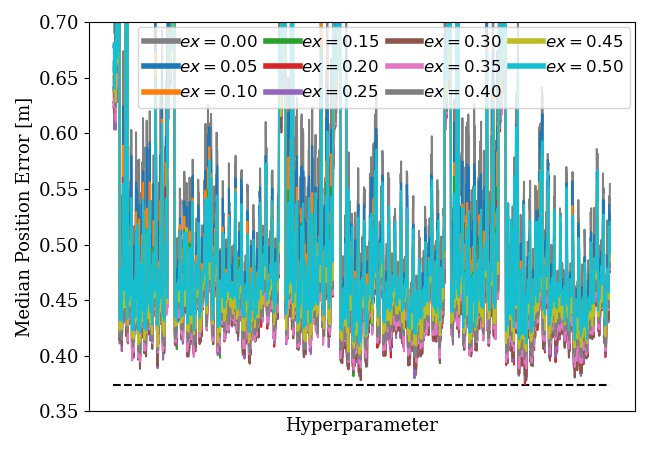}
    	\subcaption{Median position error in $m$.}
    	\label{figure_sfm_handheld1}
    \end{minipage}
    \hfill
	\begin{minipage}[t]{0.493\linewidth}
        \centering
    	\includegraphics[trim=10 10 10 10, clip, width=1.0\linewidth]{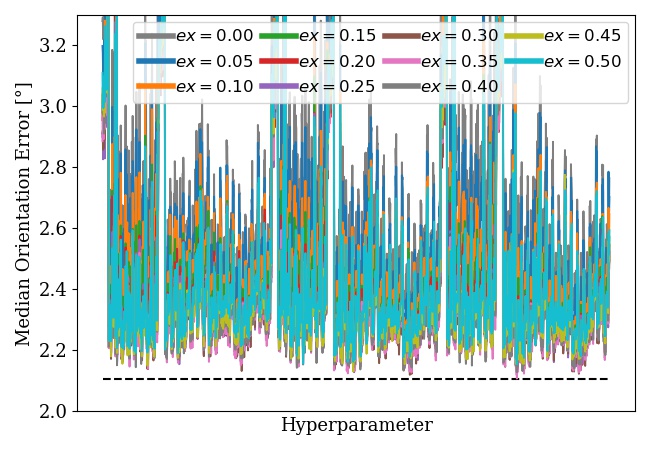}
    	\subcaption{Median orientation error in \textdegree.}
    	\label{figure_sfm_handheld2}
    \end{minipage}
    \caption{Evaluation of all SfM point clouds (for different hyperparameters, see x-axis) for the handheld train 4 and test 7 datasets. The dashed lines indicate the lowest error.}
    \label{figure_sfm_handheld}
\end{figure}

\begin{figure}[!t]
    \centering
	\begin{minipage}[t]{0.493\linewidth}
        \centering
    	\includegraphics[trim=10 10 10 10, clip, width=1.0\linewidth]{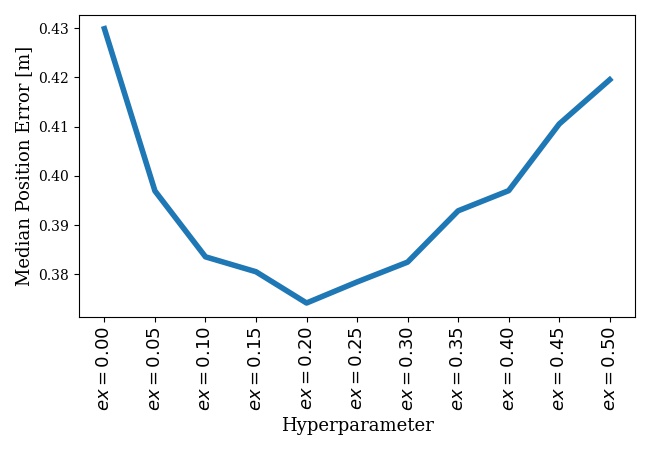}
    	\subcaption{Median position error in $m$.}
    	\label{figure_best_handheld1}
    \end{minipage}
    \hfill
	\begin{minipage}[t]{0.493\linewidth}
        \centering
    	\includegraphics[trim=10 10 10 10, clip, width=1.0\linewidth]{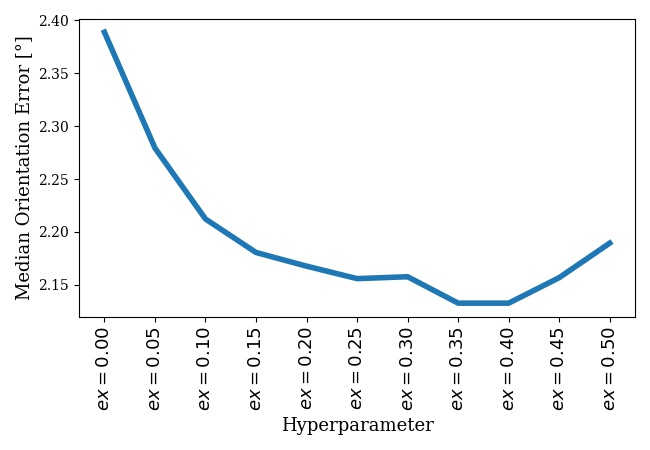}
    	\subcaption{Median orientation error in \textdegree.}
    	\label{figure_best_handheld2}
    \end{minipage}
    \caption{Evaluation of the best SfM point cloud (from Figure~\ref{figure_sfm_handheld}) for different hyperparameters $exclude$ for the handheld train 4 and test 7 datasets.}
    \label{figure_best_handheld}
\end{figure}

In the first step, we conduct an extensive hyperparameter search for the SfM parameters described in Table~\ref{table_hyperparameters} using two datasets: a dataset (train3 and test 6) and a handheld dataset (train 4 and test 7). We select the best hyperparameter combinations for point cloud reconstruction and apply them to the remaining robotic and handheld datasets. Each dataset (robotic and handheld) comprises a total of 1,752 parameters. Figure~\ref{figure_sfm_robot} illustrates all hyperparameter results for the robot dataset, while we select the best point clouds and evaluate the parameter $exclude$ with values in $\{0.0$, $0.05$, $0.1$, $0.15$, $0.2$, $0.25$, $0.3$, $0.35$, $0.4$, $0.45$, $0.5$, $0.55$, $0.6\}$ in Figure~\ref{figure_best_robot}. We observe that the parameter $exclude$ has a significant impact on the position error, with values ranging from 0.24\textit{m} to over 0.7\textit{m}. Based on Figure~\ref{figure_best_robot}, we find that the parameter $ex=0.30$ yields a low position error of 0.242\textit{m}, while the parameter $ex=0.35$ yields a low orientation error of 1.19\textdegree. The hyperparameter search results for the handheld dataset are shown in Figure~\ref{figure_sfm_handheld} and Figure~\ref{figure_best_handheld}. The error in this case increases due to the higher dynamics of the human, resulting in a position error of 0.37\textit{m} and orientation error of 2.14\textdegree. The optimal value for the parameter \textit{exclude} varies with the position (with $ex=0.20$ being optimal) and orientation (with $ex=0.35$ being optimal). To obtain highly accurate orientation predictions, a point cloud with a few and very accurate points is preferred, while a dense point cloud is better for low position error.

\begin{figure*}[!t]
    \centering
	\begin{minipage}[t]{0.245\linewidth}
        \centering
    	\includegraphics[trim=10 10 10 10, clip, width=1.0\linewidth]{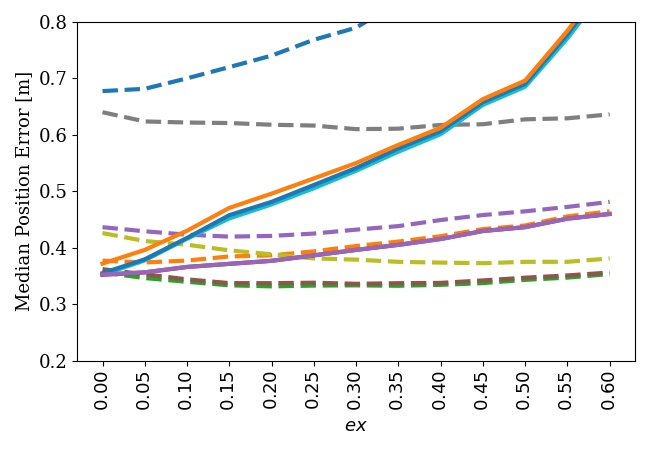}
    	\subcaption{Median position error in $m$.}
    	\label{figure_hyp_robot1}
    \end{minipage}
    \hfill
	\begin{minipage}[t]{0.245\linewidth}
        \centering
    	\includegraphics[trim=10 10 10 10, clip, width=1.0\linewidth]{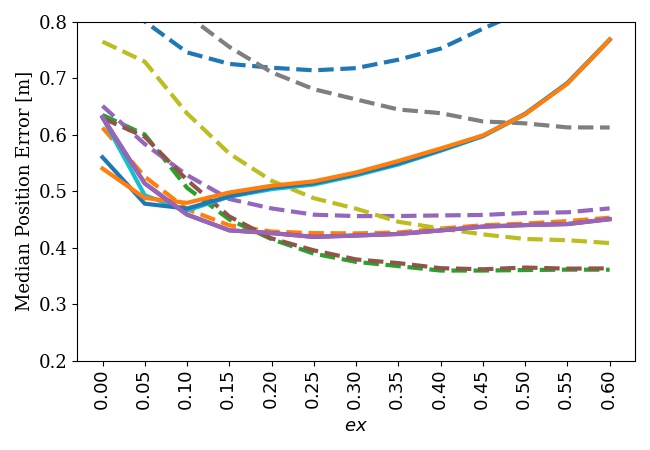}
    	\subcaption{Median position error in $m$.}
    	\label{figure_hyp_robot2}
    \end{minipage}
    \hfill
	\begin{minipage}[t]{0.245\linewidth}
        \centering
    	\includegraphics[trim=10 10 10 10, clip, width=1.0\linewidth]{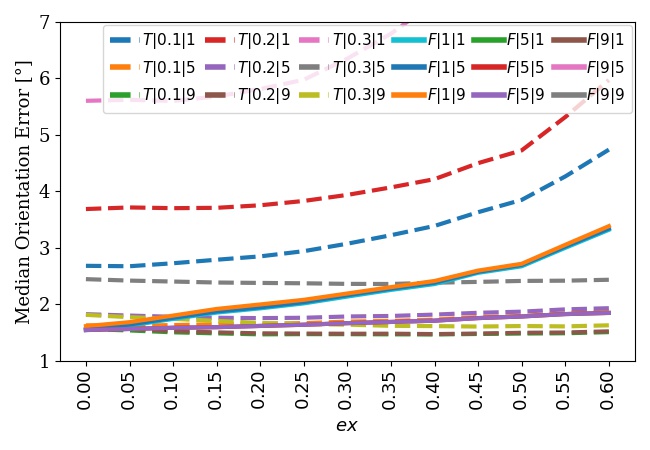}
    	\subcaption{Median orientation error in \textdegree.}
    	\label{figure_hyp_robot3}
    \end{minipage}
    \hfill
	\begin{minipage}[t]{0.245\linewidth}
        \centering
    	\includegraphics[trim=10 10 10 10, clip, width=1.0\linewidth]{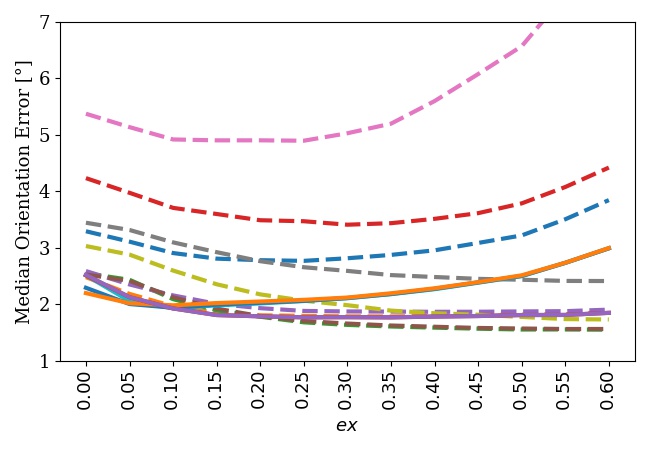}
    	\subcaption{Median orientation error in \textdegree.}
    	\label{figure_hyp_robot4}
    \end{minipage}
    \caption{SfM hyperparameter search for the robot train 3 and test 6 datasets. Figures a) and c) are with the three-point criterion, and Figures b) and d) are without the three-point criterion. Legend definition in the corresponding order: With use of the three-point-criterion, i.e., true (\textit{T}), without use of the three-point criterion, i.e., false (\textit{F}). If true: \textit{T}|\textit{oc}|\textit{sc}. If false: \textit{F}|\textit{std}|\textit{sc}. The legend is equal for all subplots.}
    \label{figure_hyp_robot}
\end{figure*}

\begin{figure*}[!t]
    \centering
	\begin{minipage}[t]{0.245\linewidth}
        \centering
    	\includegraphics[trim=10 10 10 10, clip, width=1.0\linewidth]{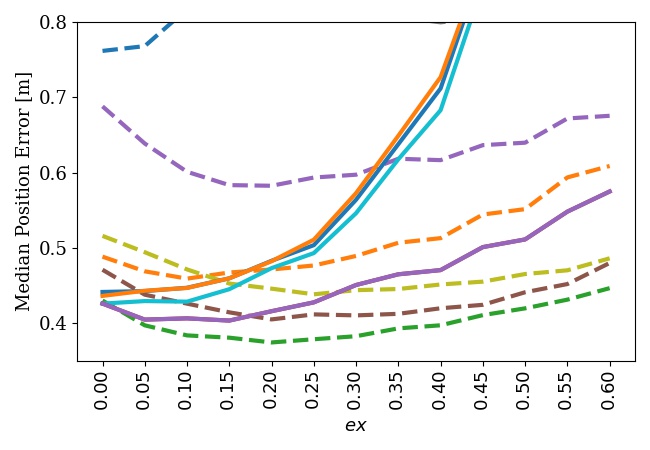}
    	\subcaption{Median position error in $m$.}
    	\label{figure_hyp_handheld1}
    \end{minipage}
    \hfill
	\begin{minipage}[t]{0.245\linewidth}
        \centering
    	\includegraphics[trim=10 10 10 10, clip, width=1.0\linewidth]{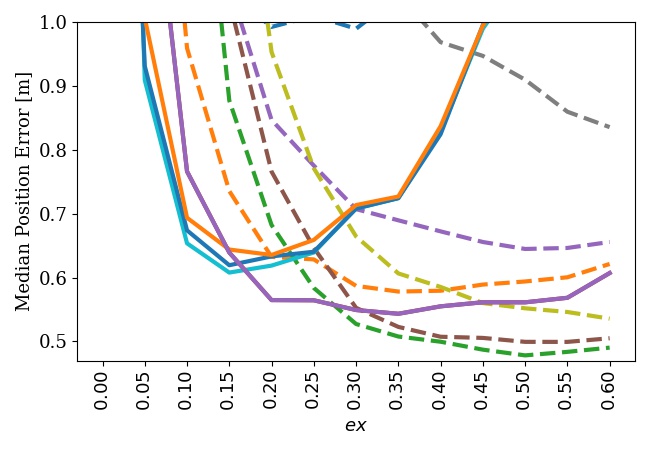}
    	\subcaption{Median position error in $m$.}
    	\label{figure_hyp_handheld2}
    \end{minipage}
    \hfill
	\begin{minipage}[t]{0.245\linewidth}
        \centering
    	\includegraphics[trim=10 10 10 10, clip, width=1.0\linewidth]{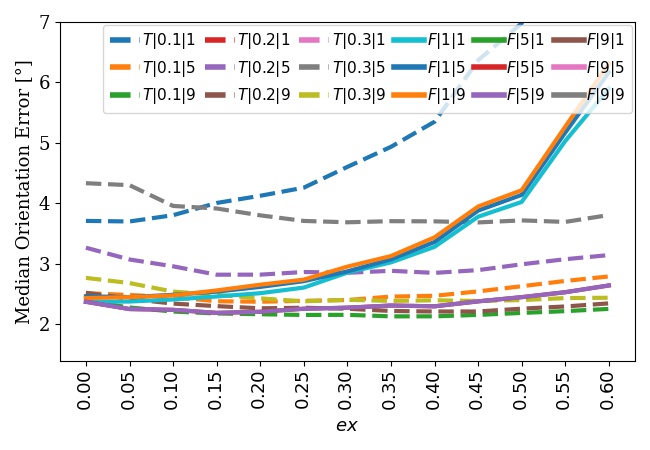}
    	\subcaption{Median orientation error in \textdegree.}
    	\label{figure_hyp_handheld3}
    \end{minipage}
    \hfill
	\begin{minipage}[t]{0.245\linewidth}
        \centering
    	\includegraphics[trim=10 10 10 10, clip, width=1.0\linewidth]{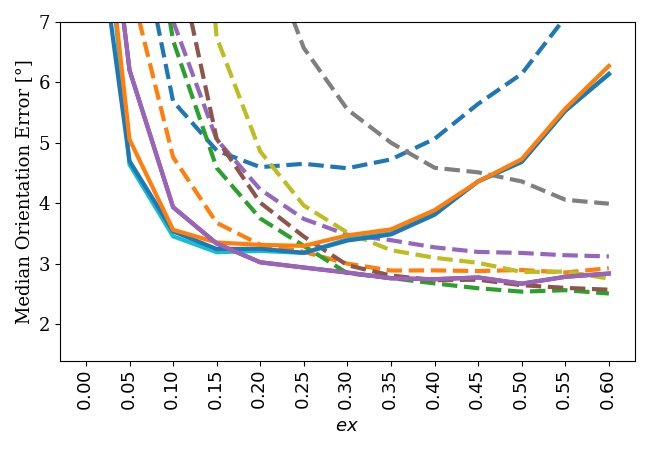}
    	\subcaption{Median orientation error in \textdegree.}
    	\label{figure_hyp_handheld4}
    \end{minipage}
    \caption{SfM hyperparameter search for the handheld train 4 and test 7 datasets. Figures a) and c) are with the three-point criterion, and Figures b) and d) are without the three-point criterion. Legend definition in the corresponding order: With use of the three-point-criterion, i.e., true (\textit{T}), without use of the three-point criterion, i.e., false (\textit{F}). If true: \textit{T}|\textit{oc}|\textit{sc}. If false: \textit{F}|\textit{std}|\textit{sc}. The legend is equal for all subplots.}
    \label{figure_hyp_handheld}
\end{figure*}

\begin{figure*}[!b]
    \centering
	\begin{minipage}[t]{0.245\linewidth}
        \centering
    	\includegraphics[trim=430 290 430 400, clip, width=1.0\linewidth]{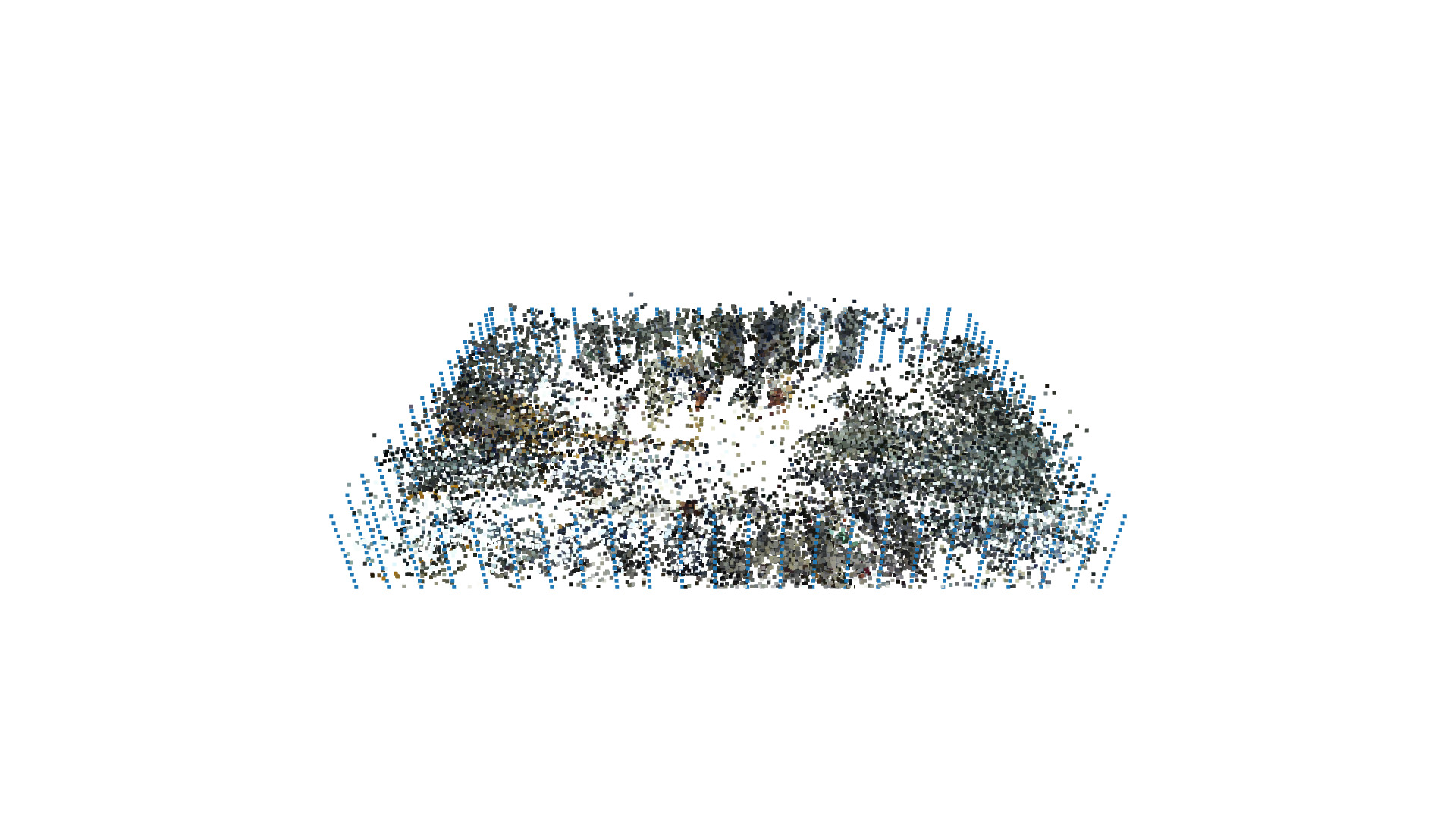}
    	\subcaption{Train 1.}
    	\label{figure_sfm_pcd1}
    \end{minipage}
    \hfill
	\begin{minipage}[t]{0.245\linewidth}
        \centering
    	\includegraphics[trim=430 290 430 400, clip, width=1.0\linewidth]{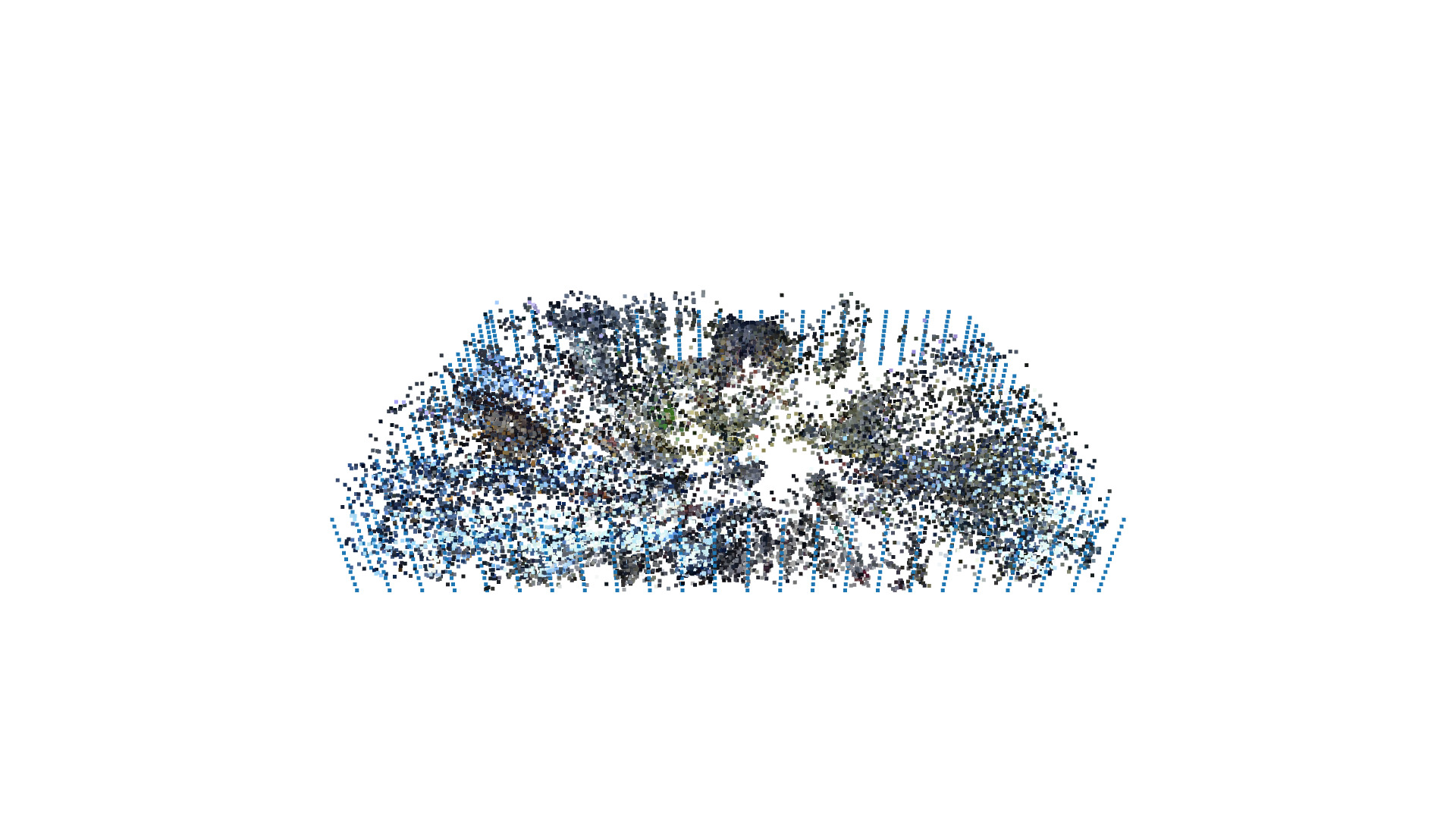}
    	\subcaption{Train 2.}
    	\label{figure_sfm_pcd2}
    \end{minipage}
    \hfill
	\begin{minipage}[t]{0.245\linewidth}
        \centering
    	\includegraphics[trim=430 290 430 400, clip, width=1.0\linewidth]{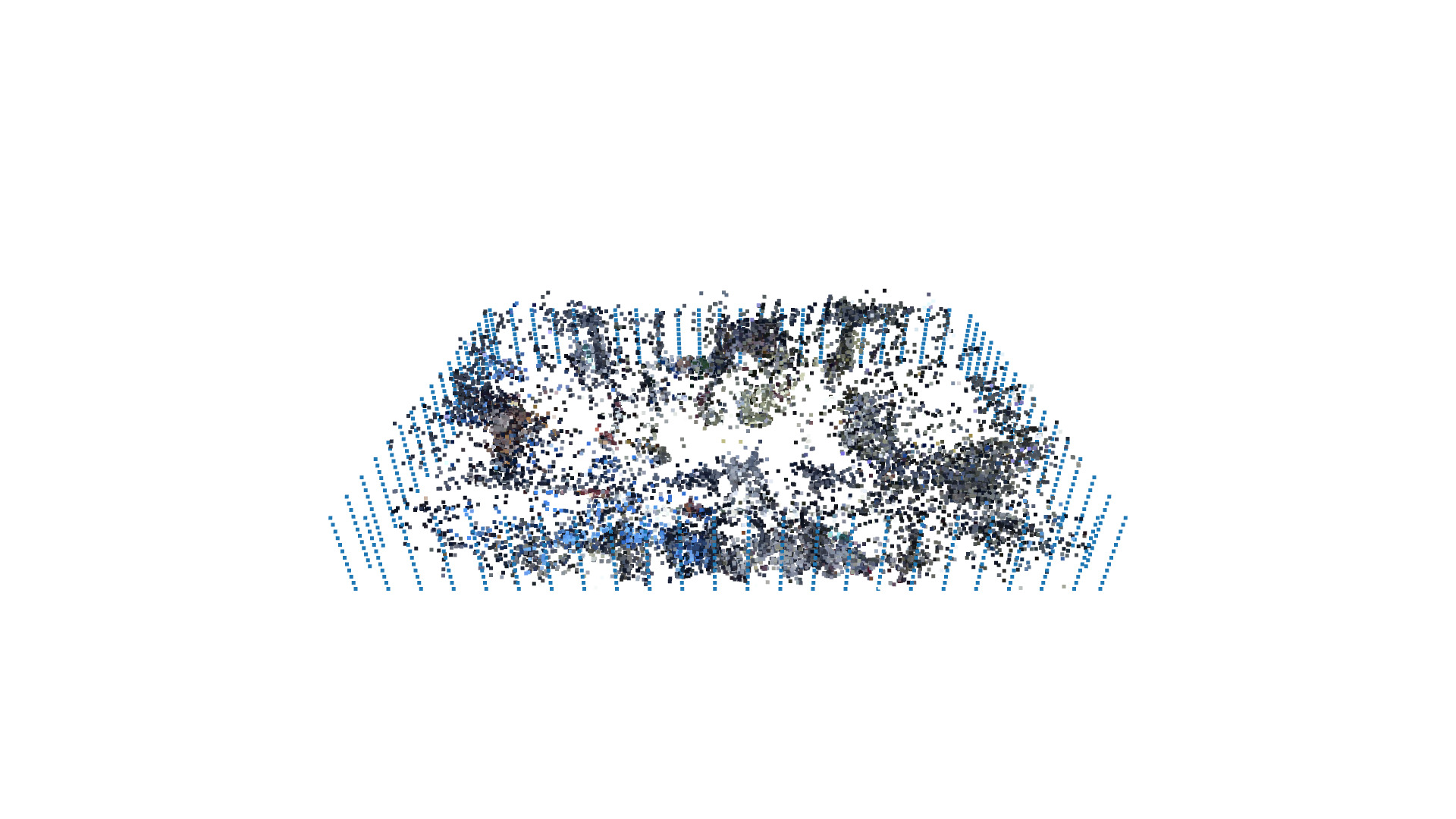}
    	\subcaption{Train 3.}
    	\label{figure_sfm_pcd3}
    \end{minipage}
    \hfill
	\begin{minipage}[t]{0.245\linewidth}
        \centering
    	\includegraphics[trim=430 290 430 400, clip, width=1.0\linewidth]{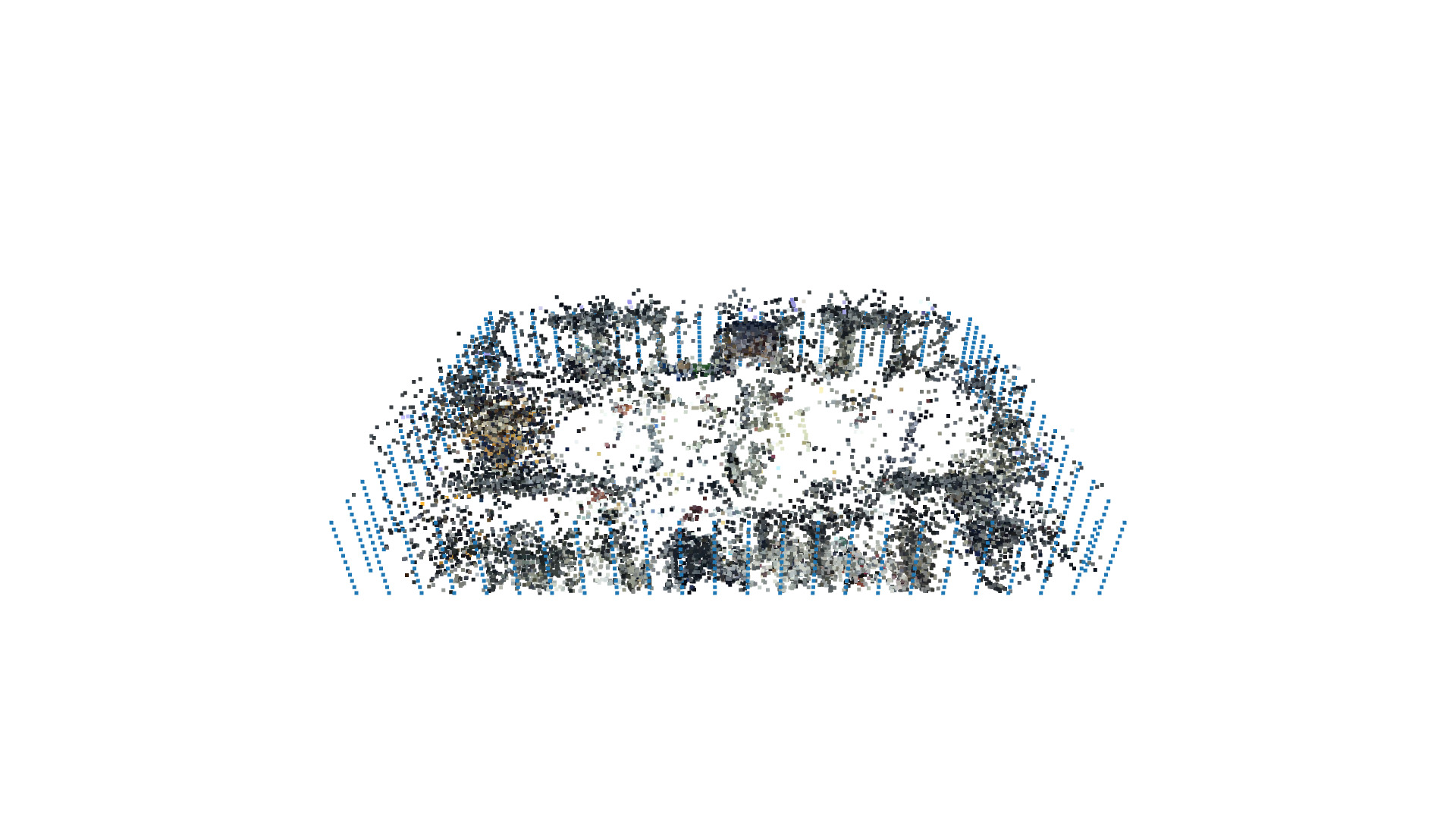}
    	\subcaption{Train 4.}
    	\label{figure_sfm_pcd4}
    \end{minipage}
	\begin{minipage}[t]{0.245\linewidth}
        \centering
    	\includegraphics[trim=430 290 430 400, clip, width=1.0\linewidth]{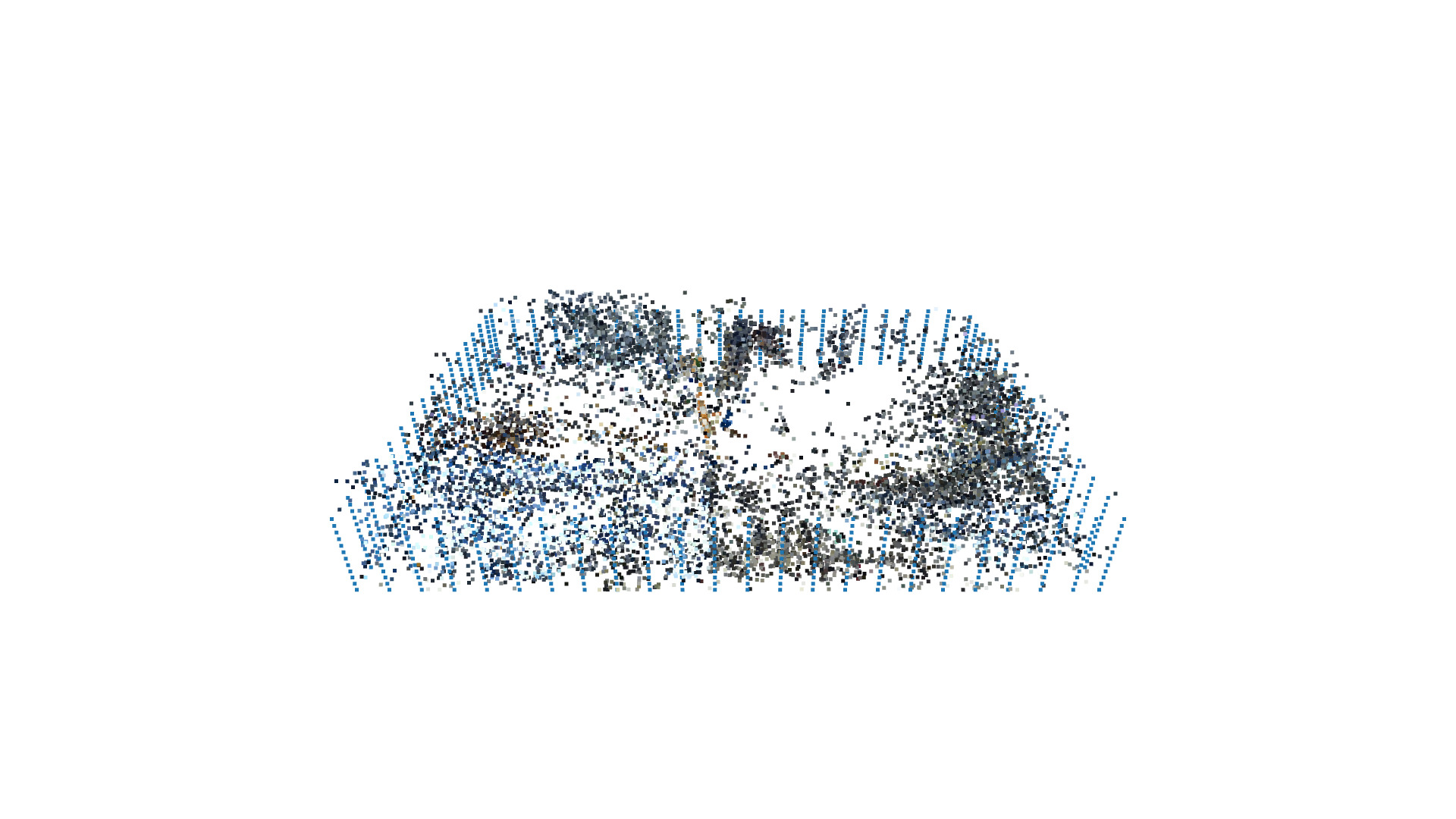}
    	\subcaption{Train 5.}
    	\label{figure_sfm_pcd5}
    \end{minipage}
    \hfill
	\begin{minipage}[t]{0.245\linewidth}
        \centering
    	\includegraphics[trim=430 290 430 400, clip, width=1.0\linewidth]{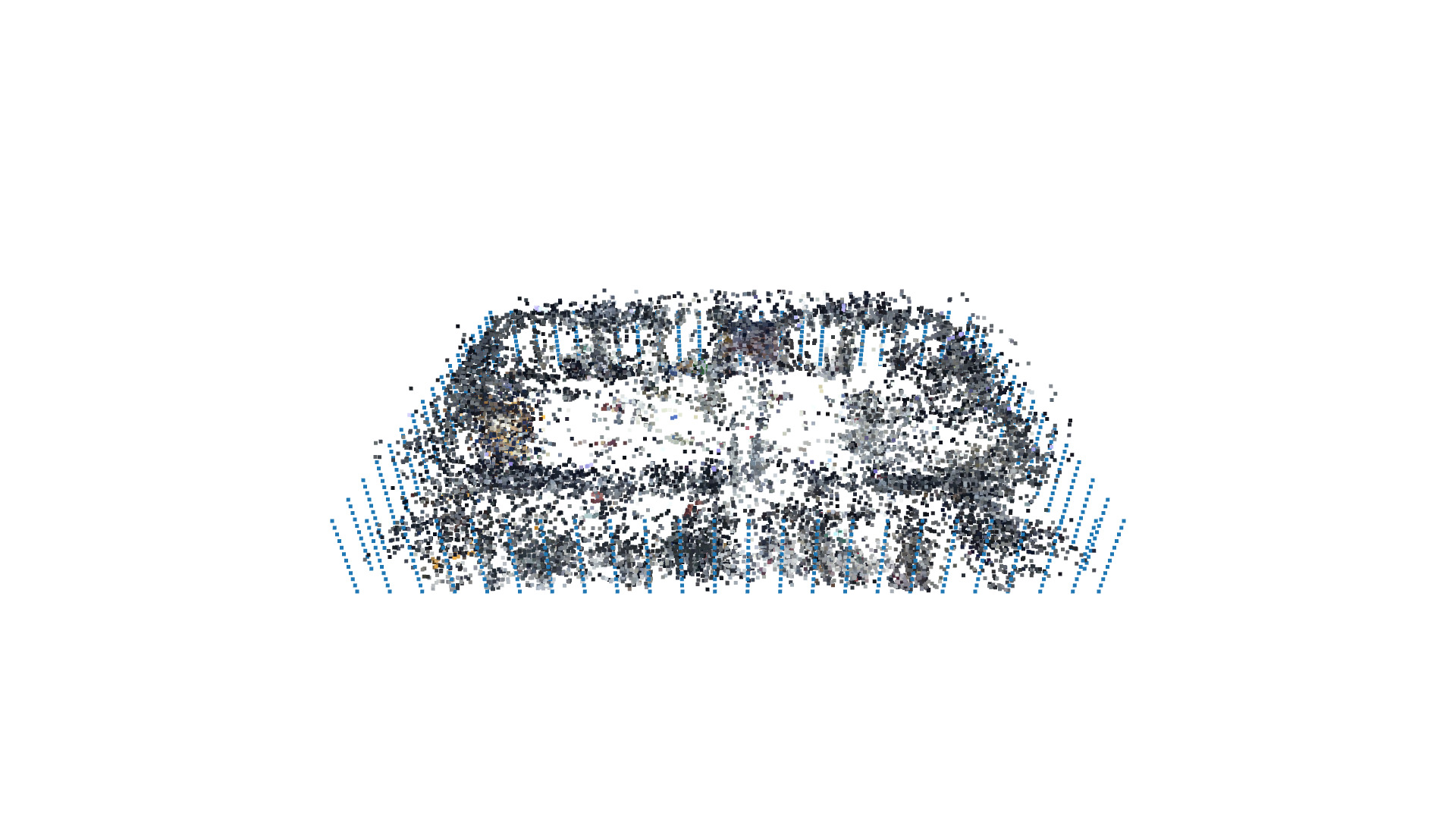}
    	\subcaption{Train 6.}
    	\label{figure_sfm_pcd6}
    \end{minipage}
    \hfill
	\begin{minipage}[t]{0.245\linewidth}
        \centering
    	\includegraphics[trim=430 290 430 400, clip, width=1.0\linewidth]{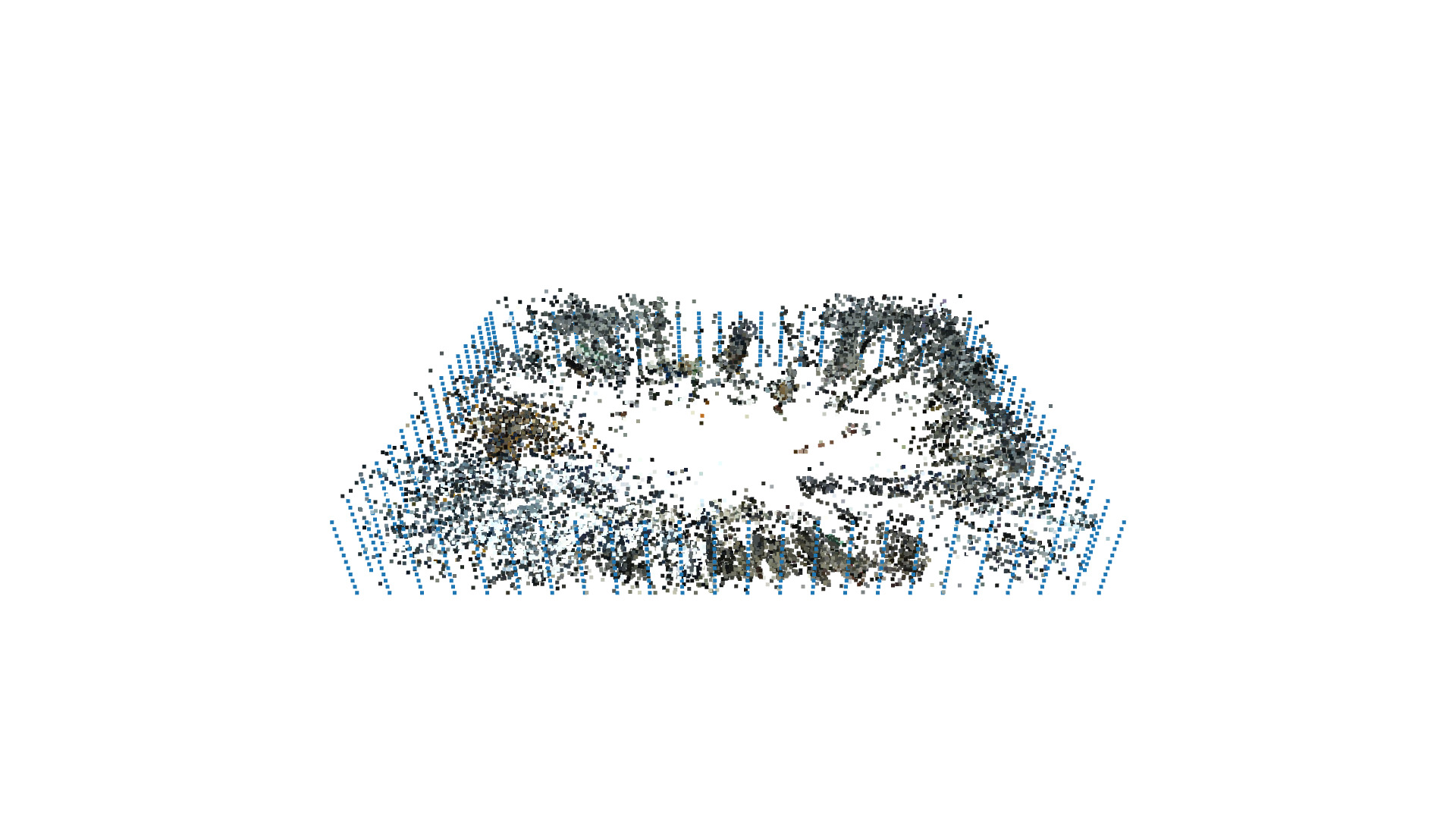}
    	\subcaption{Train 7.}
    	\label{figure_sfm_pcd7}
    \end{minipage}
    \hfill
	\begin{minipage}[t]{0.245\linewidth}
        \centering
    	\includegraphics[trim=430 290 430 400, clip, width=1.0\linewidth]{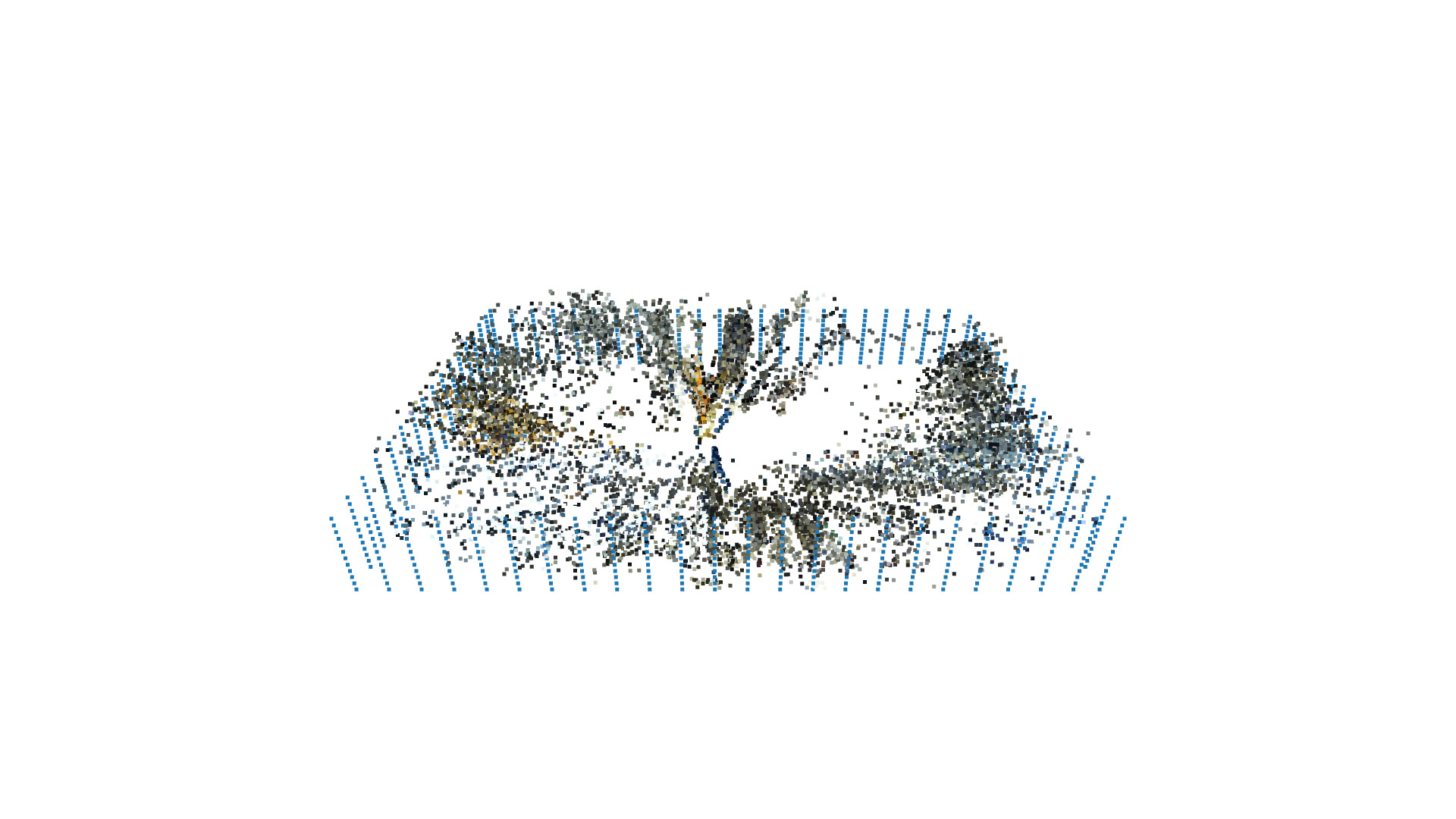}
    	\subcaption{Train 8.}
    	\label{figure_sfm_pcd8}
    \end{minipage}
    \caption{Overview of SfM point clouds for all eight training datasets. Blue dotted lines indicate the borders of the environment.}
    \label{figure_sfm_pcd}
\end{figure*}

Figure~\ref{figure_hyp_robot} and Figure~\ref{figure_hyp_handheld} present the selective search for the parameters of the three-point criterion for the robot dataset and the handheld dataset, respectively. We observe that the three-point criterion results in decreased performance (Figures a and c represent the results with the criterion, while Figures b and d show the results without it). Additionally, we explore three other parameters (refer to the legend in Figure c). The first parameter is a hard limit to exclude points, which can be set to either \textit{True} or \textit{False}. We observe an improvement in results when using the limit. If the limit is used, the second parameter specifies the percentage of lower values to remove from $[0.1, 0.2, 0.3]$, with a preference for lower values. If the limit is set to \textit{False}, we search for the minimum number of neighbors required to not exclude a point, with values in the range $[1, 5, 9]$. The best results are obtained when the number of neighbors is set to 5. The third parameter is the radius, which determines when a point counts as a neighbor for another point, with values in the range $[1, 5, 9$]. Here, a high value is preferred.

The evaluation of additional hyperparameters is presented in Appendix~\ref{app_appendix_search}. Figure~\ref{figure_robot_full_t} and Figure~\ref{figure_robot_full_r} show the results for the robot dataset, while Figure~\ref{figure_handheld_full_t} and Figure~\ref{figure_handheld_full_r} present the results for the handheld dataset. We select the best parameters for fusing with the RPR model.

\subsection{Evaluation Results: SfM}
\label{sec_evaluation_results_sfm}

\begin{figure*}[!t]
    \centering
	\begin{minipage}[t]{0.325\linewidth}
        \centering
    	\includegraphics[trim=10 10 9 11, clip, width=1.0\linewidth]{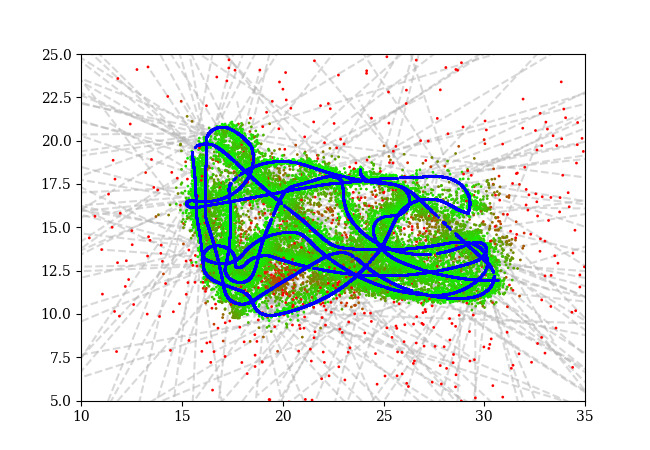}
    	\subcaption{Train 1, test 1.}
    	\label{figure_eval_traj1}
    \end{minipage}
    \hfill
	\begin{minipage}[t]{0.325\linewidth}
        \centering
    	\includegraphics[trim=10 10 9 11, clip, width=1.0\linewidth]{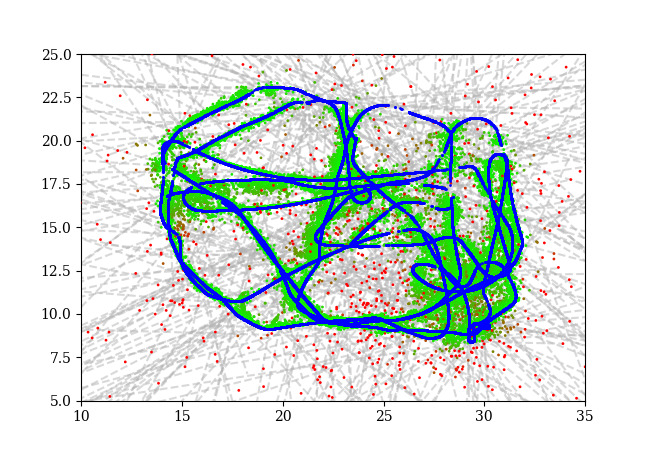}
    	\subcaption{Train 3, test 6.}
    	\label{figure_eval_traj2}
    \end{minipage}
    \hfill
	\begin{minipage}[t]{0.325\linewidth}
        \centering
    	\includegraphics[trim=10 10 9 11, clip, width=1.0\linewidth]{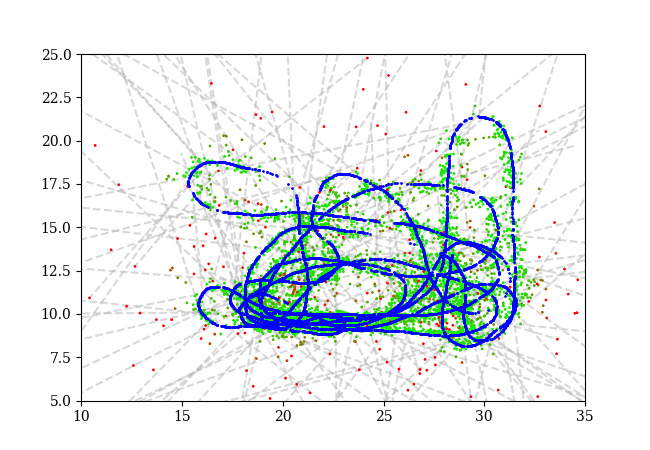}
    	\subcaption{Train 4, test 7.}
    	\label{figure_eval_traj3}
    \end{minipage}
    \caption{Evaluation of the predicted positions (green $\widehat{=}$ low position error, red $\widehat{=}$ large position error) against the ground truth trajectories (blue) for SfM.}
    \label{figure_eval_traj}
\end{figure*}

\begin{table}[t!]
\begin{center}
    \caption{Number of matches per point for the eight training datasets (average and standard deviation).}
    \label{table_matches_per_point}
    \small \begin{tabular}{ p{0.5cm} | p{0.5cm} | p{0.5cm} | p{0.5cm} }
    \multicolumn{1}{c|}{\textbf{Dataset}} & \multicolumn{1}{c||}{\textbf{\# Matches}} & \multicolumn{1}{c|}{\textbf{Dataset}} & \multicolumn{1}{c}{\textbf{\# Matches}} \\ \hline
    \multicolumn{1}{c|}{Train 1} & \multicolumn{1}{c||}{9.23\,$\pm$\,8.41} & \multicolumn{1}{c|}{Train 5} & \multicolumn{1}{c}{6.86\,$\pm$\,4.50} \\
    \multicolumn{1}{c|}{Train 2} & \multicolumn{1}{c||}{8.57\,$\pm$\,6.93} & \multicolumn{1}{c|}{Train 6} & \multicolumn{1}{c}{7.83\,$\pm$\,5.73} \\
    \multicolumn{1}{c|}{Train 3} & \multicolumn{1}{c||}{7.72\,$\pm$\,6.51} & \multicolumn{1}{c|}{Train 7} & \multicolumn{1}{c}{7.12\,$\pm$\,4.72} \\
    \multicolumn{1}{c|}{Train 4} & \multicolumn{1}{c||}{7.84\,$\pm$\,6.29} & \multicolumn{1}{c|}{Train 8} & \multicolumn{1}{c}{7.89\,$\pm$\,6.42} \\
    \end{tabular}
\end{center}
\end{table}

\begin{table*}[t!]
\begin{center}
\setlength{\tabcolsep}{2.1pt}
    \caption{Evaluation results for SfM. Results are shown as median position error in $m$ and median orientation error in \textdegree. We select the hyperparameter $ex=0.3$ for the robotic datasets and the hyperparameter $ex=0.2$ for the handheld datasets. The cell color indicates the relatedness between training and test environments and dynamics (light gray $\widehat{=}$ similar environments, dark gray $\widehat{=}$ strong environmental differences).}
    \label{table_results_sfm}
    \small \begin{tabular}{ p{0.5cm} | p{0.5cm} | p{0.5cm} | p{0.5cm} | p{0.5cm} | p{0.5cm} | p{0.5cm} | p{0.5cm} | p{0.5cm} | p{0.5cm} | p{0.5cm} | p{0.5cm} | p{0.5cm} | p{0.5cm} | p{0.5cm} | p{0.5cm} | p{0.5cm} }
    \multicolumn{1}{c|}{} & \multicolumn{2}{c|}{\textbf{Train 1}} & \multicolumn{2}{c|}{\textbf{Train 2}} & \multicolumn{2}{c|}{\textbf{Train 3}} & \multicolumn{2}{c|}{\textbf{Train 4}} & \multicolumn{2}{c|}{\textbf{Train 5}} & \multicolumn{2}{c|}{\textbf{Train 6}} & \multicolumn{2}{c|}{\textbf{Train 7}} & \multicolumn{2}{c}{\textbf{Train 8}} \\ \hline
    \multicolumn{1}{l|}{\textbf{Test 1}} & \multicolumn{1}{r}{0.4406} & \multicolumn{1}{r|}{1.47} & \multicolumn{1}{r}{\cellcolor{gray!10}0.6720} & \multicolumn{1}{r|}{\cellcolor{gray!10}2.18} & \multicolumn{1}{r}{\cellcolor{gray!30}0.6820} & \multicolumn{1}{r|}{\cellcolor{gray!30}2.40} & \multicolumn{1}{r}{\cellcolor{gray!45}1.1758} & \multicolumn{1}{r|}{\cellcolor{gray!45}31.83} & \multicolumn{1}{r}{\cellcolor{gray!45}1.7932} & \multicolumn{1}{r|}{\cellcolor{gray!45}6.55} & \multicolumn{1}{r}{\cellcolor{gray!10}0.6600} & \multicolumn{1}{r|}{\cellcolor{gray!10}2.30} & \multicolumn{1}{r}{\cellcolor{gray!30}2.0009} & \multicolumn{1}{r|}{\cellcolor{gray!30}6.06} & \multicolumn{1}{r}{\cellcolor{gray!45}6.6378} & \multicolumn{1}{r}{\cellcolor{gray!45}20.30} \\

    \multicolumn{1}{l|}{\textbf{Test 2}} & \multicolumn{1}{r}{\cellcolor{gray!10}0.5736} & \multicolumn{1}{r|}{\cellcolor{gray!10}1.94} & \multicolumn{1}{r}{\cellcolor{gray!10}0.5984} & \multicolumn{1}{r|}{\cellcolor{gray!10}2.14} & \multicolumn{1}{r}{\cellcolor{gray!20}0.4296} & \multicolumn{1}{r|}{\cellcolor{gray!20}1.71} & \multicolumn{1}{r}{\cellcolor{gray!45}1.1756} & \multicolumn{1}{r|}{\cellcolor{gray!45}4.02} & \multicolumn{1}{r}{\cellcolor{gray!45}1.0385} & \multicolumn{1}{r|}{\cellcolor{gray!45}3.64} & \multicolumn{1}{r}{\cellcolor{gray!10}0.7775} & \multicolumn{1}{r|}{\cellcolor{gray!10}2.69} & \multicolumn{1}{r}{\cellcolor{gray!45}2.6908} & \multicolumn{1}{r|}{\cellcolor{gray!45}8.47} & \multicolumn{1}{r}{\cellcolor{gray!45}5.9536} & \multicolumn{1}{r}{\cellcolor{gray!45}19.22} \\
    
    \multicolumn{1}{l|}{\textbf{Test 3}} & \multicolumn{1}{r}{\cellcolor{gray!20}0.8155} & \multicolumn{1}{r|}{\cellcolor{gray!20}2.90} & \multicolumn{1}{r}{\cellcolor{gray!10}0.7514} & \multicolumn{1}{r|}{\cellcolor{gray!10}2.78} & \multicolumn{1}{r}{\cellcolor{gray!10}0.4529} & \multicolumn{1}{r|}{\cellcolor{gray!10}1.89} & \multicolumn{1}{r}{\cellcolor{gray!45}1.7175} & \multicolumn{1}{r|}{\cellcolor{gray!45}6.40} & \multicolumn{1}{r}{\cellcolor{gray!45}1.5590} & \multicolumn{1}{r|}{\cellcolor{gray!45}5.62} & \multicolumn{1}{r}{\cellcolor{gray!20}1.1602} & \multicolumn{1}{r|}{\cellcolor{gray!20}3.91} & \multicolumn{1}{r}{\cellcolor{gray!45}3.4231} & \multicolumn{1}{r|}{\cellcolor{gray!45}11.23} & \multicolumn{1}{r}{\cellcolor{gray!45}6.5891} & \multicolumn{1}{r}{\cellcolor{gray!45}21.41} \\
    
    \multicolumn{1}{l|}{\textbf{Test 4}} & \multicolumn{1}{r}{\cellcolor{gray!30}1.6407} & \multicolumn{1}{r|}{\cellcolor{gray!30}5.50} & \multicolumn{1}{r}{\cellcolor{gray!20}0.6393} & \multicolumn{1}{r|}{\cellcolor{gray!20}2.48} & \multicolumn{1}{r}{\cellcolor{gray!10}0.3407} & \multicolumn{1}{r|}{\cellcolor{gray!10}1.54} & \multicolumn{1}{r}{\cellcolor{gray!45}1.8450} & \multicolumn{1}{r|}{\cellcolor{gray!45}6.79} & \multicolumn{1}{r}{\cellcolor{gray!45}1.4746} & \multicolumn{1}{r|}{\cellcolor{gray!45}5.40} & \multicolumn{1}{r}{\cellcolor{gray!45}1.1674} & \multicolumn{1}{r|}{\cellcolor{gray!45}4.03} & \multicolumn{1}{r}{\cellcolor{gray!45}7.2445} & \multicolumn{1}{r|}{\cellcolor{gray!45}26.19} & \multicolumn{1}{r}{\cellcolor{gray!45}7.0458} & \multicolumn{1}{r}{\cellcolor{gray!45}22.88} \\
    
    \multicolumn{1}{l|}{\textbf{Test 5}} & \multicolumn{1}{r}{\cellcolor{gray!45}6.5806} & \multicolumn{1}{r|}{\cellcolor{gray!45}20.27} & \multicolumn{1}{r}{0.1312} & \multicolumn{1}{r|}{0.82} & \multicolumn{1}{r}{\cellcolor{gray!10}0.2076} & \multicolumn{1}{r|}{\cellcolor{gray!10}1.05} & \multicolumn{1}{r}{\cellcolor{gray!45}1.4789} & \multicolumn{1}{r|}{\cellcolor{gray!45}4.79} & \multicolumn{1}{r}{\cellcolor{gray!45}0.5336} & \multicolumn{1}{r|}{\cellcolor{gray!45}1.85} & \multicolumn{1}{r}{\cellcolor{gray!45}4.0145} & \multicolumn{1}{r|}{\cellcolor{gray!45}12.14} & \multicolumn{1}{r}{\cellcolor{gray!45}5.7685} & \multicolumn{1}{r|}{\cellcolor{gray!45}19.70} & \multicolumn{1}{r}{\cellcolor{gray!45}9.4618} & \multicolumn{1}{r}{\cellcolor{gray!45}29.55} \\
    
    \multicolumn{1}{l|}{\textbf{Test 6}} & \multicolumn{1}{r}{\cellcolor{gray!45}3.8720} & \multicolumn{1}{r|}{\cellcolor{gray!45}13.03} & \multicolumn{1}{r}{\cellcolor{gray!30}0.7598} & \multicolumn{1}{r|}{\cellcolor{gray!30}2.93} & \multicolumn{1}{r}{0.2417} & \multicolumn{1}{r|}{1.19} & \multicolumn{1}{r}{\cellcolor{gray!20}0.5971} & \multicolumn{1}{r|}{\cellcolor{gray!20}2.25} & \multicolumn{1}{r}{\cellcolor{gray!20}0.5417} & \multicolumn{1}{r|}{\cellcolor{gray!20}1.99} & \multicolumn{1}{r}{\cellcolor{gray!45}4.3004} & \multicolumn{1}{r|}{\cellcolor{gray!45}13.06} & \multicolumn{1}{r}{\cellcolor{gray!45}6.7546} & \multicolumn{1}{r|}{\cellcolor{gray!45}23.65} & \multicolumn{1}{r}{\cellcolor{gray!45}10.3992} & \multicolumn{1}{r}{\cellcolor{gray!45}35.50} \\
    
    \multicolumn{1}{l|}{\textbf{Test 7}} & \multicolumn{1}{r}{\cellcolor{gray!45}24.5902} & \multicolumn{1}{r|}{\cellcolor{gray!45}94.50} & \multicolumn{1}{r}{\cellcolor{gray!45}15.2681} & \multicolumn{1}{r|}{\cellcolor{gray!45}64.65} & \multicolumn{1}{r}{\cellcolor{gray!20}1.5336} & \multicolumn{1}{r|}{\cellcolor{gray!20}6.05} & \multicolumn{1}{r}{0.3742} & \multicolumn{1}{r|}{2.17} & \multicolumn{1}{r}{\cellcolor{gray!10}0.3458} & \multicolumn{1}{r|}{\cellcolor{gray!10}1.99} & \multicolumn{1}{r}{\cellcolor{gray!45}21.3927} & \multicolumn{1}{r|}{\cellcolor{gray!45}80.61} & \multicolumn{1}{r}{\cellcolor{gray!45}18.3089} & \multicolumn{1}{r|}{\cellcolor{gray!45}75.64} & \multicolumn{1}{r}{\cellcolor{gray!45}20.9043} & \multicolumn{1}{r}{\cellcolor{gray!45}79.34} \\
    
    \multicolumn{1}{l|}{\textbf{Test 8}} & \multicolumn{1}{r}{\cellcolor{gray!45}21.6899} & \multicolumn{1}{r|}{\cellcolor{gray!45}81.23} & \multicolumn{1}{r}{\cellcolor{gray!45}12.6190} & \multicolumn{1}{r|}{\cellcolor{gray!45}43.84} & \multicolumn{1}{r}{\cellcolor{gray!30}1.0662} & \multicolumn{1}{r|}{\cellcolor{gray!30}4.41} & \multicolumn{1}{r}{\cellcolor{gray!10}0.4205} & \multicolumn{1}{r|}{\cellcolor{gray!10}2.22} & \multicolumn{1}{r}{0.3049} & \multicolumn{1}{r|}{1.82} & \multicolumn{1}{r}{\cellcolor{gray!45}10.3005} & \multicolumn{1}{r|}{\cellcolor{gray!45}35.54} & \multicolumn{1}{r}{\cellcolor{gray!45}11.2760} & \multicolumn{1}{r|}{\cellcolor{gray!45}42.33} & \multicolumn{1}{r}{\cellcolor{gray!45}14.2804} & \multicolumn{1}{r}{\cellcolor{gray!45}52.24} \\
    
    \multicolumn{1}{l|}{\textbf{Test 9}} & \multicolumn{1}{r}{\cellcolor{gray!20}1.0352} & \multicolumn{1}{r|}{\cellcolor{gray!20}3.99} & \multicolumn{1}{r}{\cellcolor{gray!30}2.0459} & \multicolumn{1}{r|}{\cellcolor{gray!30}6.84} & \multicolumn{1}{r}{\cellcolor{gray!30}0.8040} & \multicolumn{1}{r|}{\cellcolor{gray!30}3.81} & \multicolumn{1}{r}{\cellcolor{gray!45}5.9806} & \multicolumn{1}{r|}{\cellcolor{gray!45}18.24} & \multicolumn{1}{r}{\cellcolor{gray!45}1.0407} & \multicolumn{1}{r|}{\cellcolor{gray!45}4.30} & \multicolumn{1}{r}{0.5175} & \multicolumn{1}{r|}{2.52} & \multicolumn{1}{r}{\cellcolor{gray!20}1.5983} & \multicolumn{1}{r|}{\cellcolor{gray!20}5.55} & \multicolumn{1}{r}{\cellcolor{gray!10}3.8926} & \multicolumn{1}{r}{\cellcolor{gray!10}12.44} \\
    
    \multicolumn{1}{l|}{\textbf{Test 10}} & \multicolumn{1}{r}{\cellcolor{gray!30}1.1925} & \multicolumn{1}{r|}{\cellcolor{gray!30}5.56} & \multicolumn{1}{r}{\cellcolor{gray!45}5.6020} & \multicolumn{1}{r|}{\cellcolor{gray!45}17.17} & \multicolumn{1}{r}{\cellcolor{gray!45}1.1586} & \multicolumn{1}{r|}{\cellcolor{gray!45}5.47} & \multicolumn{1}{r}{\cellcolor{gray!45}2.8885} & \multicolumn{1}{r|}{\cellcolor{gray!45}10.17} & \multicolumn{1}{r}{\cellcolor{gray!45}1.6000} & \multicolumn{1}{r|}{\cellcolor{gray!45}6.32} & \multicolumn{1}{r}{\cellcolor{gray!10}0.5515} & \multicolumn{1}{r|}{\cellcolor{gray!10}3.40} & \multicolumn{1}{r}{0.7954} & \multicolumn{1}{r|}{4.14} & \multicolumn{1}{r}{\cellcolor{gray!20}4.9886} & \multicolumn{1}{r}{\cellcolor{gray!20}16.21} \\
    
    \multicolumn{1}{l|}{\textbf{Test 11}} & \multicolumn{1}{r}{\cellcolor{gray!45}12.1940} & \multicolumn{1}{r|}{\cellcolor{gray!45}39.80} & \multicolumn{1}{r}{\cellcolor{gray!45}4.1512} & \multicolumn{1}{r|}{\cellcolor{gray!45}14.22} & \multicolumn{1}{r}{\cellcolor{gray!20}0.4923} & \multicolumn{1}{r|}{\cellcolor{gray!20}3.27} & \multicolumn{1}{r}{\cellcolor{gray!45}1.9017} & \multicolumn{1}{r|}{\cellcolor{gray!45}7.29} & \multicolumn{1}{r}{\cellcolor{gray!45}0.9849} & \multicolumn{1}{r|}{\cellcolor{gray!45}4.58} & \multicolumn{1}{r}{\cellcolor{gray!20}0.9164} & \multicolumn{1}{r|}{\cellcolor{gray!20}4.34} & \multicolumn{1}{r}{\cellcolor{gray!10}1.8022} & \multicolumn{1}{r|}{\cellcolor{gray!10}6.10} & \multicolumn{1}{r}{2.1391} & \multicolumn{1}{r}{7.25} \\
    \end{tabular}
\end{center}
\end{table*}

We visualize the resulting SfM point clouds for all training datasets in Figure~\ref{figure_sfm_pcd} with the dashed blue lines as the borders of the environment. The open environment of the train 1 dataset with no objects allows for a dense reconstruction, while train 2 also yields a high-density point cloud with wall features. However, on the top right and bottom left sections of the hall, points are missing due to the underrepresentation of images in these areas. In contrast, datasets train 3, train 7, and train 8 contain many object features, while train 6 has more ceiling points. Dataset train 4, recorded with slow motions from person 1, has many feature-rich object points, while train 5, with fast motions from person 2, has a higher noise of points due to motion blur. The average number of matches per point cloud, reported in Table~\ref{table_matches_per_point}, supports these findings, with train 1 having the most matches at an average of 9.23 per point, resulting in a dense point cloud. In contrast, train 5 has a lower density with an average of 6.86 matches per point, and the remaining datasets have an average number of matches ranging from 7.12 to 8.57.

Table~\ref{table_results_sfm} provides a summary of the SfM results, with the grey cells indicating the difference between the training and testing datasets. The evaluation shows that SfM is highly effective when tested on the same scenario as the training scenario, as evidenced by the low errors on the train 2 and test 5 dataset (0.1312\textit{m} and 0.82\textdegree) and the train 3 and test 6 dataset (0.2417\textit{m} and 1.19\textdegree). For instance, with the point cloud from the train 3 dataset, the error increases consistent with increasing changes in the environment. Therefore, the error correlates with the changes in the environment, i.e., the color of the table cells. On the other hand, the prediction of orientation from the reconstruction is relatively robust against changes, with errors ranging from 1.19\textdegree\,on the test 6 dataset to 2.40\textdegree\,on the test 1 dataset. SfM also performs well on the handheld datasets (train 4 and train 5) with high motion dynamics, when evaluated on the same scenarios (test 7 and test 8). However, SfM fails when the reconstructions are applied to the robot evaluation datasets (and vice versa). In conclusion, while SfM is robust against environmental changes, it is sensitive to dynamics, such as motion blur.

In Figure~\ref{figure_eval_traj}, we present three representative trajectory predictions from SfM. The trajectory predictions for all datasets can be found in Appendix~\ref{app_appendix_trajectories}, spanning from Figure~\ref{figure_2dtraj1} to Figure~\ref{figure_2dtraj8}. Overall, SfM produces a small number of outliers (specifically, fewer than 100), particularly in scenarios where the test images lack distinctive features. Moreover, SfM can accurately localize the robot in both open environments (see Figure~\ref{figure_eval_traj1}) and those with absorber walls (see Figure~\ref{figure_eval_traj1}). However, the predicted trajectory may not be smooth in some cases (see Figure~\ref{figure_eval_traj3}).

\subsection{Evaluation Results: APR and Augmented APR}
\label{sec_evaluation_results_posenet_augmented}

\begin{table*}[t!]
\begin{center}
\setlength{\tabcolsep}{2.1pt}
    \caption{Evaluation results for the APR model. Results are shown as median position error in $m$ and median orientation error in \textdegree. We select the hyperparameter $ex=0.3$ for the robotic datasets and the hyperparameter $ex=0.2$ for the handheld datasets. The cell color indicates the relatedness between training and test environments and dynamics (light gray $\widehat{=}$ similar environments, dark gray $\widehat{=}$ strong environmental differences).}
    \label{table_results_posenet}
    \small \begin{tabular}{ p{0.5cm} | p{0.5cm} | p{0.5cm} | p{0.5cm} | p{0.5cm} | p{0.5cm} | p{0.5cm} | p{0.5cm} | p{0.5cm} | p{0.5cm} | p{0.5cm} | p{0.5cm} | p{0.5cm} | p{0.5cm} | p{0.5cm} | p{0.5cm} | p{0.5cm} }
    \multicolumn{1}{c|}{} & \multicolumn{2}{c|}{\textbf{Train 1}} & \multicolumn{2}{c|}{\textbf{Train 2}} & \multicolumn{2}{c|}{\textbf{Train 3}} & \multicolumn{2}{c|}{\textbf{Train 4}} & \multicolumn{2}{c|}{\textbf{Train 5}} & \multicolumn{2}{c|}{\textbf{Train 6}} & \multicolumn{2}{c|}{\textbf{Train 7}} & \multicolumn{2}{c}{\textbf{Train 8}} \\ \hline
    \multicolumn{1}{l|}{\textbf{Test 1}} & \multicolumn{1}{r}{0.8076} & \multicolumn{1}{r|}{2.12} & \multicolumn{1}{r}{\cellcolor{gray!10}3.0364} & \multicolumn{1}{r|}{\cellcolor{gray!10}9.56} & \multicolumn{1}{r}{\cellcolor{gray!30}2.0245} & \multicolumn{1}{r|}{\cellcolor{gray!30}10.92} & \multicolumn{1}{r}{\cellcolor{gray!45}4.4608} & \multicolumn{1}{r|}{\cellcolor{gray!45}33.05} & \multicolumn{1}{r}{\cellcolor{gray!45}5.4224} & \multicolumn{1}{r|}{\cellcolor{gray!45}36.99} & \multicolumn{1}{r}{\cellcolor{gray!10}1.1224} & \multicolumn{1}{r|}{\cellcolor{gray!10}3.03} & \multicolumn{1}{r}{\cellcolor{gray!30}3.1848} & \multicolumn{1}{r|}{\cellcolor{gray!30}9.73} & \multicolumn{1}{r}{\cellcolor{gray!45}1.8310} & \multicolumn{1}{r}{\cellcolor{gray!45}5.64} \\
    \multicolumn{1}{l|}{\textbf{Test 2}} & \multicolumn{1}{r}{\cellcolor{gray!10}1.0096} & \multicolumn{1}{r|}{\cellcolor{gray!10}2.98} & \multicolumn{1}{r}{\cellcolor{gray!10}3.4245} & \multicolumn{1}{r|}{\cellcolor{gray!10}10.36} & \multicolumn{1}{r}{\cellcolor{gray!20}2.0611} & \multicolumn{1}{r|}{\cellcolor{gray!20}9.47} & \multicolumn{1}{r}{\cellcolor{gray!45}5.6366} & \multicolumn{1}{r|}{\cellcolor{gray!45}34.33} & \multicolumn{1}{r}{\cellcolor{gray!45}5.6771} & \multicolumn{1}{r|}{\cellcolor{gray!45}46.69} & \multicolumn{1}{r}{\cellcolor{gray!10}1.6957} & \multicolumn{1}{r|}{\cellcolor{gray!10}4.83} & \multicolumn{1}{r}{\cellcolor{gray!45}3.9469} & \multicolumn{1}{r|}{\cellcolor{gray!45}13.93} & \multicolumn{1}{r}{\cellcolor{gray!45}2.2105} & \multicolumn{1}{r}{\cellcolor{gray!45}6.09} \\
    \multicolumn{1}{l|}{\textbf{Test 3}} & \multicolumn{1}{r}{\cellcolor{gray!20}1.3636} & \multicolumn{1}{r|}{\cellcolor{gray!20}4.72} & \multicolumn{1}{r}{\cellcolor{gray!10}3.0526} & \multicolumn{1}{r|}{\cellcolor{gray!10}12.44} & \multicolumn{1}{r}{\cellcolor{gray!10}1.9051} & \multicolumn{1}{r|}{\cellcolor{gray!10}9.32} & \multicolumn{1}{r}{\cellcolor{gray!45}4.5158} & \multicolumn{1}{r|}{\cellcolor{gray!45}35.70} & \multicolumn{1}{r}{\cellcolor{gray!45}5.9039} & \multicolumn{1}{r|}{\cellcolor{gray!45}42.06} & \multicolumn{1}{r}{\cellcolor{gray!20}2.3052} & \multicolumn{1}{r|}{\cellcolor{gray!20}8.79} & \multicolumn{1}{r}{\cellcolor{gray!45}3.8144} & \multicolumn{1}{r|}{\cellcolor{gray!45}18.73} & \multicolumn{1}{r}{\cellcolor{gray!45}2.5504} & \multicolumn{1}{r}{\cellcolor{gray!45}8.30} \\
    \multicolumn{1}{l|}{\textbf{Test 4}} & \multicolumn{1}{r}{\cellcolor{gray!30}2.2635} & \multicolumn{1}{r|}{\cellcolor{gray!30}5.85} & \multicolumn{1}{r}{\cellcolor{gray!20}3.5413} & \multicolumn{1}{r|}{\cellcolor{gray!20}9.95} & \multicolumn{1}{r}{\cellcolor{gray!10}1.9869} & \multicolumn{1}{r|}{\cellcolor{gray!10}8.66} & \multicolumn{1}{r}{\cellcolor{gray!45}5.8072} & \multicolumn{1}{r|}{\cellcolor{gray!45}29.51} & \multicolumn{1}{r}{\cellcolor{gray!45}7.2380} & \multicolumn{1}{r|}{\cellcolor{gray!45}52.85} & \multicolumn{1}{r}{\cellcolor{gray!45}2.4261} & \multicolumn{1}{r|}{\cellcolor{gray!45}7.85} & \multicolumn{1}{r}{\cellcolor{gray!45}4.6354} & \multicolumn{1}{r|}{\cellcolor{gray!45}19.50} & \multicolumn{1}{r}{\cellcolor{gray!45}2.3831} & \multicolumn{1}{r}{\cellcolor{gray!45}7.14} \\
    \multicolumn{1}{l|}{\textbf{Test 5}} & \multicolumn{1}{r}{\cellcolor{gray!45}2.1091} & \multicolumn{1}{r|}{\cellcolor{gray!45}6.93} & \multicolumn{1}{r}{0.7340} & \multicolumn{1}{r|}{1.99} & \multicolumn{1}{r}{\cellcolor{gray!10}0.6417} & \multicolumn{1}{r|}{\cellcolor{gray!10}2.32} & \multicolumn{1}{r}{\cellcolor{gray!45}3.9771} & \multicolumn{1}{r|}{\cellcolor{gray!45}24.13} & \multicolumn{1}{r}{\cellcolor{gray!45}7.1118} & \multicolumn{1}{r|}{\cellcolor{gray!45}71.50} & \multicolumn{1}{r}{\cellcolor{gray!45}2.6327} & \multicolumn{1}{r|}{\cellcolor{gray!45}9.32} & \multicolumn{1}{r}{\cellcolor{gray!45}3.9958} & \multicolumn{1}{r|}{\cellcolor{gray!45}18.60} & \multicolumn{1}{r}{\cellcolor{gray!45}1.8708} & \multicolumn{1}{r}{\cellcolor{gray!45}6.64} \\
    \multicolumn{1}{l|}{\textbf{Test 6}} & \multicolumn{1}{r}{\cellcolor{gray!45}3.2216} & \multicolumn{1}{r|}{\cellcolor{gray!45}23.71} & \multicolumn{1}{r}{\cellcolor{gray!30}3.7292} & \multicolumn{1}{r|}{\cellcolor{gray!30}19.39} & \multicolumn{1}{r}{0.4232} & \multicolumn{1}{r|}{1.74} & \multicolumn{1}{r}{\cellcolor{gray!20}3.9909} & \multicolumn{1}{r|}{\cellcolor{gray!20}16.07} & \multicolumn{1}{r}{\cellcolor{gray!20}5.2587} & \multicolumn{1}{r|}{\cellcolor{gray!20}43.72} & \multicolumn{1}{r}{\cellcolor{gray!45}3.3173} & \multicolumn{1}{r|}{\cellcolor{gray!45}30.38} & \multicolumn{1}{r}{\cellcolor{gray!45}5.2294} & \multicolumn{1}{r|}{\cellcolor{gray!45}54.94} & \multicolumn{1}{r}{\cellcolor{gray!45}2.8280} & \multicolumn{1}{r}{\cellcolor{gray!45}23.32} \\
    \multicolumn{1}{l|}{\textbf{Test 7}} & \multicolumn{1}{r}{\cellcolor{gray!45}8.4233} & \multicolumn{1}{r|}{\cellcolor{gray!45}79.54} & \multicolumn{1}{r}{\cellcolor{gray!45}7.1442} & \multicolumn{1}{r|}{\cellcolor{gray!45}87.50} & \multicolumn{1}{r}{\cellcolor{gray!20}6.5239} & \multicolumn{1}{r|}{\cellcolor{gray!20}39.58} & \multicolumn{1}{r}{0.6277} & \multicolumn{1}{r|}{2.92} & \multicolumn{1}{r}{\cellcolor{gray!10}0.9070} & \multicolumn{1}{r|}{\cellcolor{gray!10}4.10} & \multicolumn{1}{r}{\cellcolor{gray!45}7.6766} & \multicolumn{1}{r|}{\cellcolor{gray!45}76.02} & \multicolumn{1}{r}{\cellcolor{gray!45}5.8731} & \multicolumn{1}{r|}{\cellcolor{gray!45}82.03} & \multicolumn{1}{r}{\cellcolor{gray!45}6.6052} & \multicolumn{1}{r}{\cellcolor{gray!45}68.35} \\
    \multicolumn{1}{l|}{\textbf{Test 8}} & \multicolumn{1}{r}{\cellcolor{gray!45}6.2944} & \multicolumn{1}{r|}{\cellcolor{gray!45}83.48} & \multicolumn{1}{r}{\cellcolor{gray!45}5.9621} & \multicolumn{1}{r|}{\cellcolor{gray!45}82.11} & \multicolumn{1}{r}{\cellcolor{gray!30}4.6108} & \multicolumn{1}{r|}{\cellcolor{gray!30}18.77} & \multicolumn{1}{r}{\cellcolor{gray!10}0.8599} & \multicolumn{1}{r|}{\cellcolor{gray!10}3.88} & \multicolumn{1}{r}{0.8892} & \multicolumn{1}{r|}{3.85} & \multicolumn{1}{r}{\cellcolor{gray!45}5.7339} & \multicolumn{1}{r|}{\cellcolor{gray!45}60.65} & \multicolumn{1}{r}{\cellcolor{gray!45}6.0106} & \multicolumn{1}{r|}{\cellcolor{gray!45}69.61} & \multicolumn{1}{r}{\cellcolor{gray!45}5.3620} & \multicolumn{1}{r}{\cellcolor{gray!45}54.91} \\
    \multicolumn{1}{l|}{\textbf{Test 9}} & \multicolumn{1}{r}{\cellcolor{gray!20}1.7769} & \multicolumn{1}{r|}{\cellcolor{gray!20}5.63} & \multicolumn{1}{r}{\cellcolor{gray!30}3.9475} & \multicolumn{1}{r|}{\cellcolor{gray!30}18.32} & \multicolumn{1}{r}{\cellcolor{gray!30}3.0726} & \multicolumn{1}{r|}{\cellcolor{gray!30}21.53} & \multicolumn{1}{r}{\cellcolor{gray!45}4.6508} & \multicolumn{1}{r|}{\cellcolor{gray!45}34.17} & \multicolumn{1}{r}{\cellcolor{gray!45}4.8097} & \multicolumn{1}{r|}{\cellcolor{gray!45}41.53} & \multicolumn{1}{r}{0.8045} & \multicolumn{1}{r|}{2.59} & \multicolumn{1}{r}{\cellcolor{gray!20}3.9290} & \multicolumn{1}{r|}{\cellcolor{gray!20}8.69} & \multicolumn{1}{r}{\cellcolor{gray!10}1.6533} & \multicolumn{1}{r}{\cellcolor{gray!10}4.95} \\
    \multicolumn{1}{l|}{\textbf{Test 10}} & \multicolumn{1}{r}{\cellcolor{gray!30}4.1898} & \multicolumn{1}{r|}{\cellcolor{gray!30}12.43} & \multicolumn{1}{r}{\cellcolor{gray!45}2.8045} & \multicolumn{1}{r|}{\cellcolor{gray!45}37.11} & \multicolumn{1}{r}{\cellcolor{gray!45}4.2577} & \multicolumn{1}{r|}{\cellcolor{gray!45}48.94} & \multicolumn{1}{r}{\cellcolor{gray!45}3.9608} & \multicolumn{1}{r|}{\cellcolor{gray!45}41.43} & \multicolumn{1}{r}{\cellcolor{gray!45}4.5296} & \multicolumn{1}{r|}{\cellcolor{gray!45}46.22} & \multicolumn{1}{r}{\cellcolor{gray!10}2.1423} & \multicolumn{1}{r|}{\cellcolor{gray!10}8.28} & \multicolumn{1}{r}{0.8444} & \multicolumn{1}{r|}{3.66} & \multicolumn{1}{r}{\cellcolor{gray!20}0.8224} & \multicolumn{1}{r}{\cellcolor{gray!20}4.01} \\
    \multicolumn{1}{l|}{\textbf{Test 11}} & \multicolumn{1}{r}{\cellcolor{gray!45}3.3845} & \multicolumn{1}{r|}{\cellcolor{gray!45}16.54} & \multicolumn{1}{r}{\cellcolor{gray!45}4.3289} & \multicolumn{1}{r|}{\cellcolor{gray!45}26.20} & \multicolumn{1}{r}{\cellcolor{gray!20}2.1233} & \multicolumn{1}{r|}{\cellcolor{gray!20}8.81} & \multicolumn{1}{r}{\cellcolor{gray!45}3.9488} & \multicolumn{1}{r|}{\cellcolor{gray!45}30.08} & \multicolumn{1}{r}{\cellcolor{gray!45}4.9100} & \multicolumn{1}{r|}{\cellcolor{gray!45}54.42} & \multicolumn{1}{r}{\cellcolor{gray!20}2.3541} & \multicolumn{1}{r|}{\cellcolor{gray!20}8.53} & \multicolumn{1}{r}{\cellcolor{gray!10}2.7667} & \multicolumn{1}{r|}{\cellcolor{gray!10}13.89} & \multicolumn{1}{r}{0.8000} & \multicolumn{1}{r}{3.82} \\
    \end{tabular}
\end{center}
\end{table*}

In this section, an evaluation of the APR model is presented and the results are compared to SfM. The summary of the results is provided in Table~\ref{table_results_posenet}. Similar to SfM, the APR model achieves the lowest position and orientation errors when evaluated in the same training and testing scenario. However, the errors produced by the APR model are higher compared to SfM (e.g., 0.4232\textit{m} for APR and 0.2417\textit{m} for SfM on the train 3 and test 6 dataset). The error increases significantly when there are variations in the environment (e.g., 2.0245\textit{m} for the test 1 dataset), indicating that the APR model is less robust to changes compared to SfM. On the other hand, the APR model produces fewer outliers due to the restricted area in the learning process.

\begin{figure*}[!t]
    \centering
	\begin{minipage}[t]{0.493\linewidth}
        \centering
    	\includegraphics[trim=10 10 10 11, clip, width=1.0\linewidth]{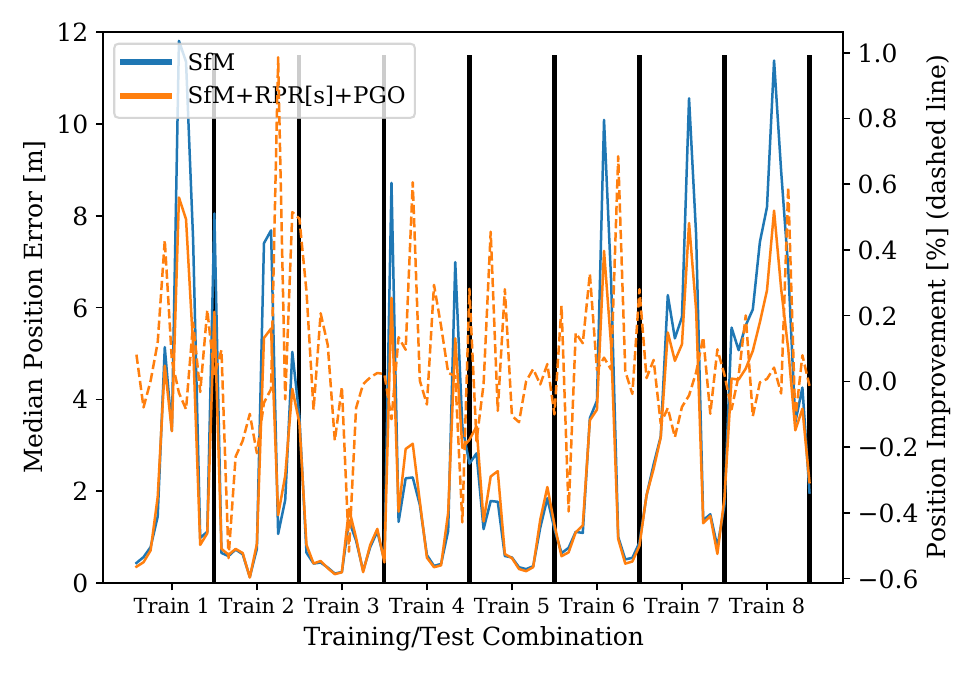}
    	\subcaption{Fusion of SfM and RPR with PGO.}
    	\label{figure_sfm_apr_of_pgo1}
    \end{minipage}
    \hfill
	\begin{minipage}[t]{0.493\linewidth}
        \centering
    	\includegraphics[trim=10 10 10 11, clip, width=1.0\linewidth]{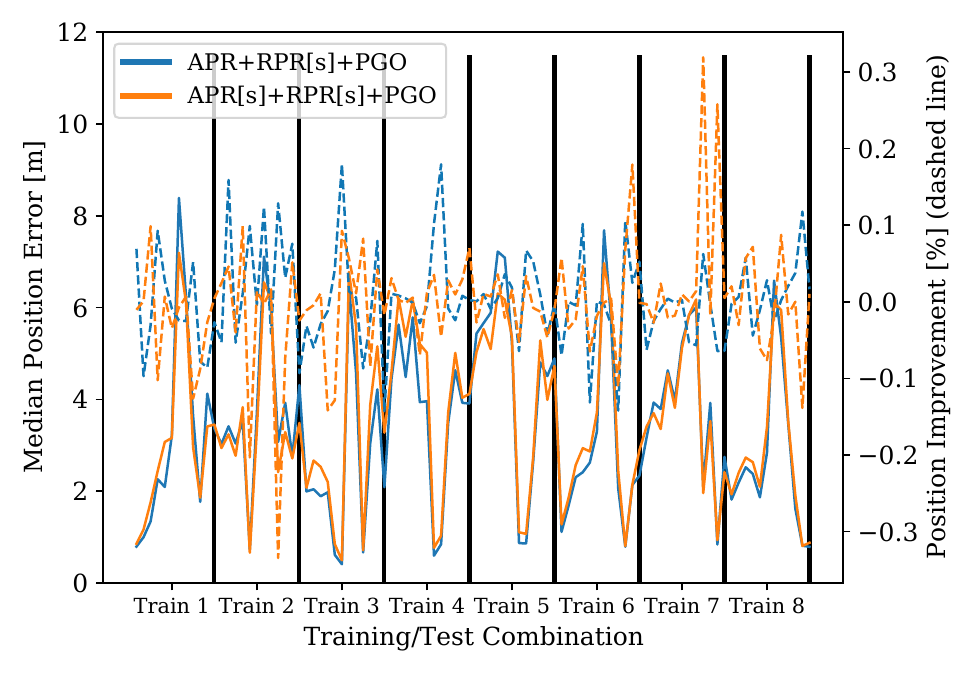}
    	\subcaption{Fusion of APR and RPR with PGO.}
    	\label{figure_sfm_apr_of_pgo2}
    \end{minipage}
    \caption{Comparison of SfM, respectively for APR, with the optimization of SfM, respectively of APR, with RPR and PGO. We evaluate APR and RPR pre-trained and non-pre-trained. Dashed lines show the position improvements in \%. [s] in APR[s] and RPR[s] defines the pre-trained model with synthetically generated data.}
    \label{figure_sfm_apr_of_pgo}
\end{figure*}

\begin{figure}[!t]
    \centering
	\begin{minipage}[t]{0.493\linewidth}
        \centering
    	\includegraphics[trim=10 0 34 34, clip, width=1.0\linewidth]{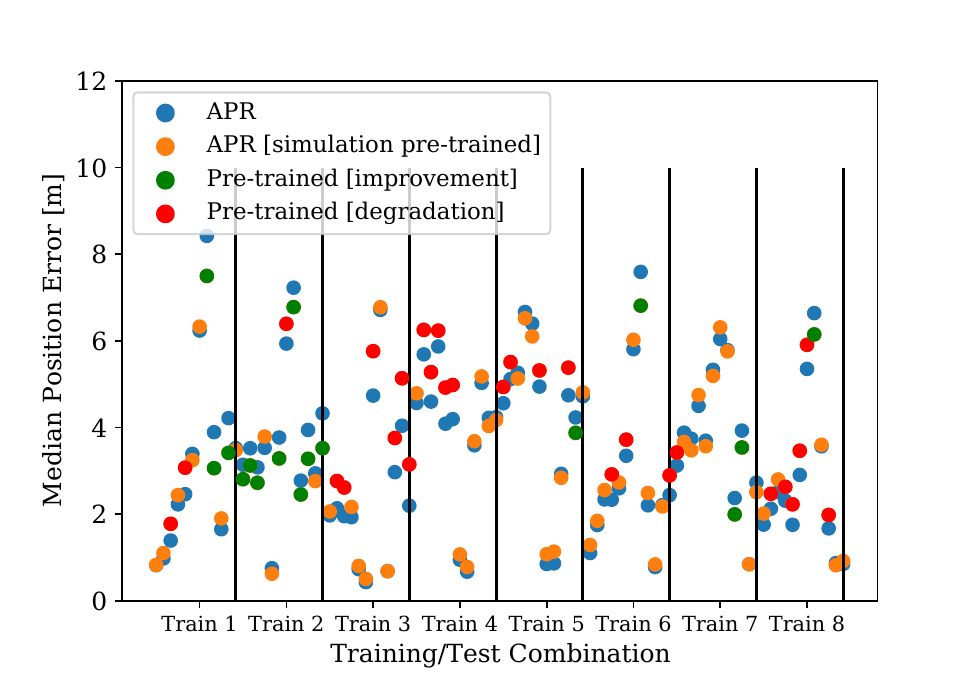}
    	\subcaption{Median position error in $m$.}
    	\label{figure_posenet_synthetic1}
    \end{minipage}
    \hfill
	\begin{minipage}[t]{0.493\linewidth}
        \centering
    	\includegraphics[trim=10 0 34 34, clip, width=1.0\linewidth]{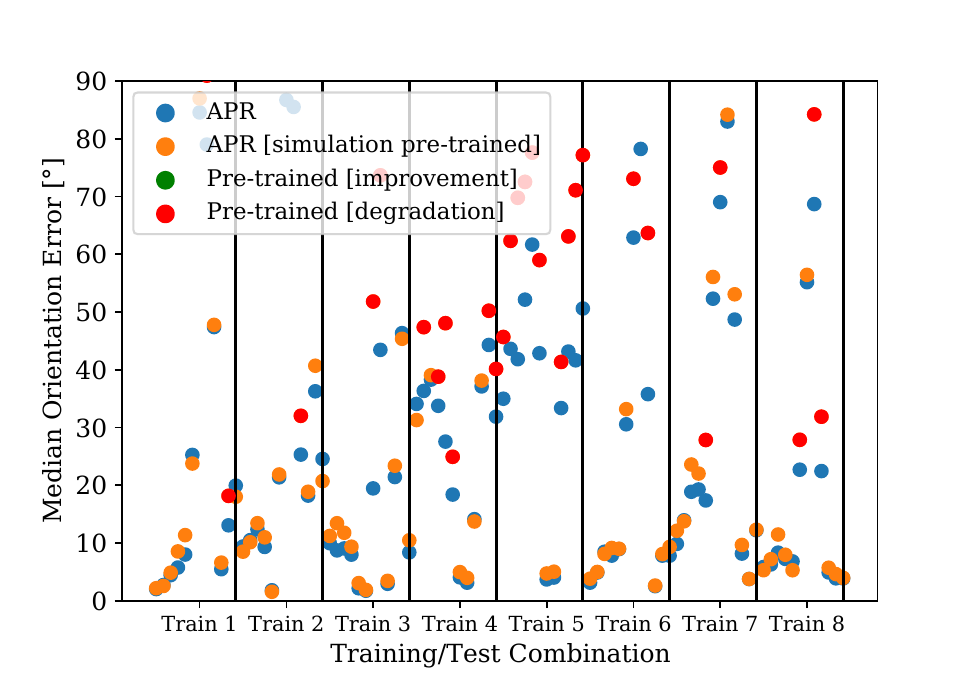}
    	\subcaption{Median orientation error in \textdegree.}
    	\label{figure_posenet_synthetic2}
    \end{minipage}
    \caption{Comparison of APR with APR pre-trained on data generated from simulation.}
    \label{figure_posenet_synthetic}
\end{figure}

The motivation behind pre-training the APR model on a synthetically generated dataset is to enhance its generalization capabilities when objects are removed or added. We propose a comparison between the results of the APR model and the pre-trained APR model in Figure~\ref{figure_posenet_synthetic}. The x-axis represents all possible combinations of training and testing scenarios. The green dots indicate an improvement in the error, while the red dots indicate a degradation in the error. The simulation environment contains many absorber walls that resemble the environment of dataset train 2, leading to a significant improvement in results for this scenario. However, the results decrease for the train 4 and train 5 datasets since the synthetic dataset lacks human dynamics and motion blur. When the predicted position error by APR is high, augmenting the data can further improve the results. However, the changes are marginal when the error is already low.

\subsection{Fusion Results}
\label{sec_evaluation_results_pgo}

In the next step, we conduct an evaluation of the fusion of SfM and RPR, as well as the fusion of APR and RPR. The evaluation initially involves pose refinement using PGO, followed by recurrent fusion cells. For an overview of the evaluation, please refer to Table~\ref{table_evaluation_overview}.

\paragraph{PGO \& Augmentation.} Initially, we refine the absolute poses using relative poses with the state-of-the-art PGO algorithm. Figure~\ref{figure_sfm_apr_of_pgo1} presents a comparison between the position error of SfM with the refined poses obtained from PGO (SfM+RPR+PGO). PGO has a significant positive impact on results in challenging scenarios but does not affect good localization results (e.g., train 2, train 3, and train 6 datasets). However, PGO marginally decreases the results for handheld datasets (train 4 and train 5). We also present the percentage improvement achieved when pre-training RPR with simulated data (dashed lines). While pre-training has a negative impact (-0.6\%) on the train 2 and train 3 datasets, results can be slightly improved for the remaining datasets, up to 1.0\%. Figure~\ref{figure_sfm_apr_of_pgo2} presents results for refining APR with RPR. In contrast to SfM, this combination has no effect on results. Additionally, pre-training only leads to marginal improvements or decreases in results.

\begin{figure}[!t]
    \centering
    \includegraphics[trim=10 10 9 11, clip, width=1.0\linewidth]{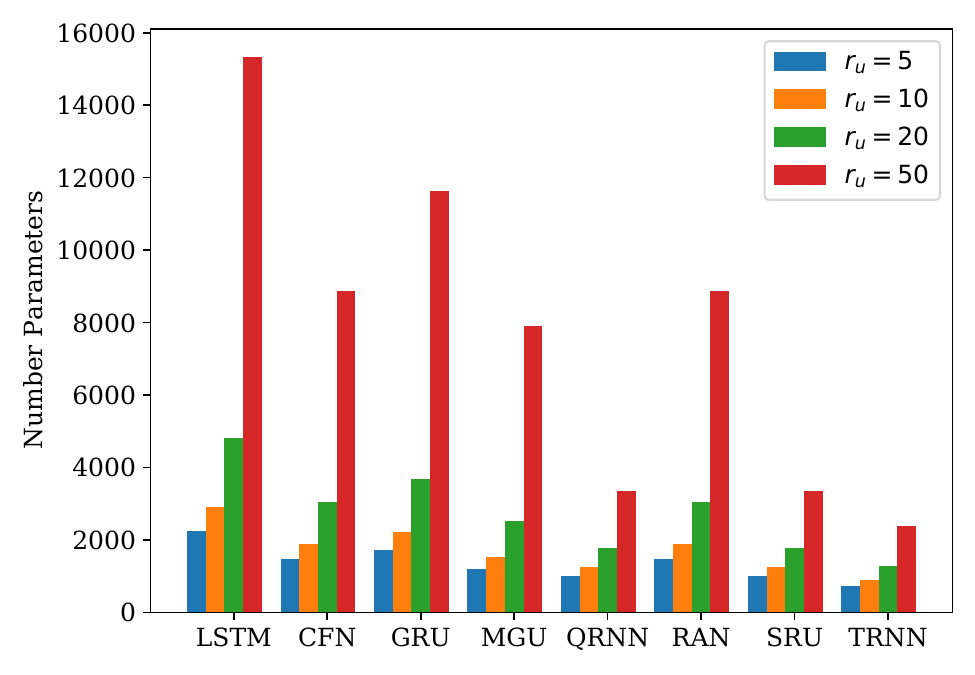}
    \caption{Comparison of trainable model parameters of all eight stacked recurrent networks for APR and RPR fusion for different number of units $r_u \in [5, 10, 20, 50]$.}
    \label{figure_fusion_parameters}
\end{figure}

\begin{figure*}[!t]
    \centering
	\begin{minipage}[t]{0.495\linewidth}
        \centering
    	\includegraphics[trim=10 10 0 11, clip, width=1.0\linewidth]{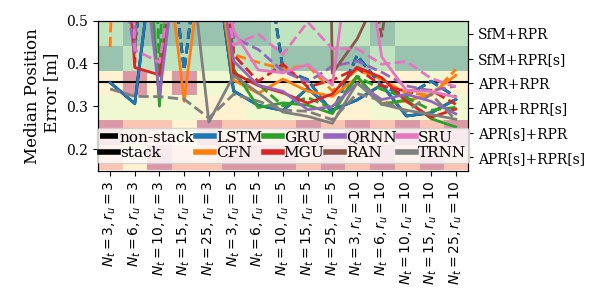}
    	\subcaption{Train 1.}
    	\label{figure_recurrent_fusion1}
    \end{minipage}
    \hfill
	\begin{minipage}[t]{0.495\linewidth}
        \centering
    	\includegraphics[trim=10 10 0 11, clip, width=1.0\linewidth]{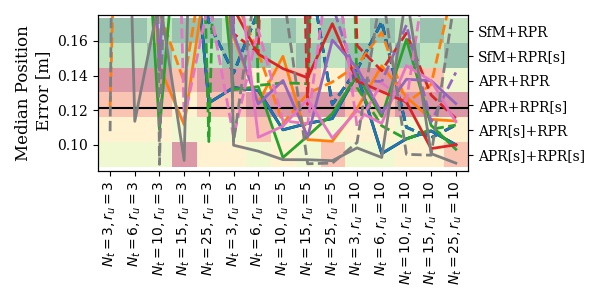}
    	\subcaption{Train 2.}
    	\label{figure_recurrent_fusion2}
    \end{minipage}
	\begin{minipage}[t]{0.495\linewidth}
        \centering
    	\includegraphics[trim=10 10 0 11, clip, width=1.0\linewidth]{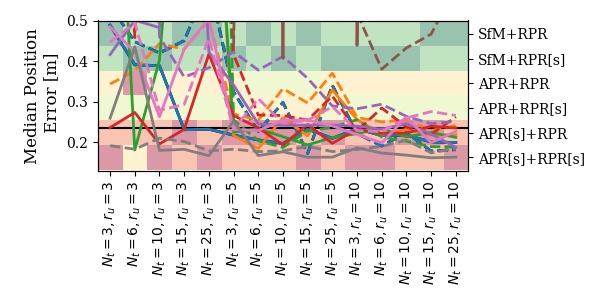}
    	\subcaption{Train 3.}
    	\label{figure_recurrent_fusion3}
    \end{minipage}
    \hfill
	\begin{minipage}[t]{0.495\linewidth}
        \centering
    	\includegraphics[trim=10 10 0 11, clip, width=1.0\linewidth]{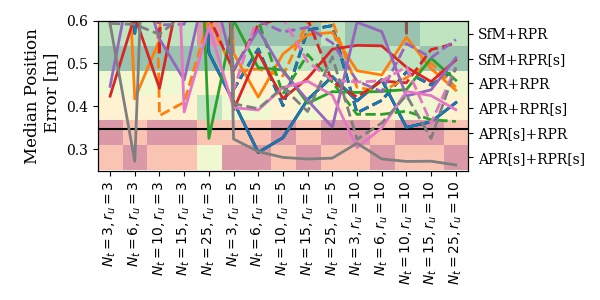}
    	\subcaption{Train 4.}
    	\label{figure_recurrent_fusion4}
    \end{minipage}
	\begin{minipage}[t]{0.495\linewidth}
        \centering
    	\includegraphics[trim=10 10 0 11, clip, width=1.0\linewidth]{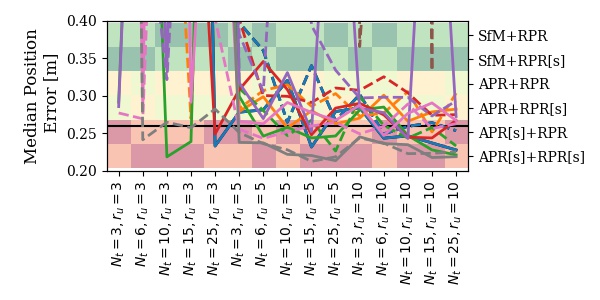}
    	\subcaption{Train 5.}
    	\label{figure_recurrent_fusion5}
    \end{minipage}
    \hfill
	\begin{minipage}[t]{0.495\linewidth}
        \centering
    	\includegraphics[trim=10 10 0 11, clip, width=1.0\linewidth]{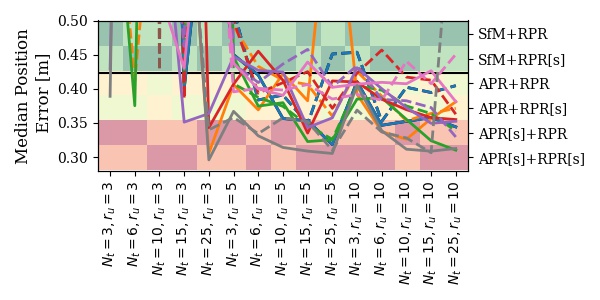}
    	\subcaption{Train 6.}
    	\label{figure_recurrent_fusion6}
    \end{minipage}
	\begin{minipage}[t]{0.495\linewidth}
        \centering
    	\includegraphics[trim=10 10 0 11, clip, width=1.0\linewidth]{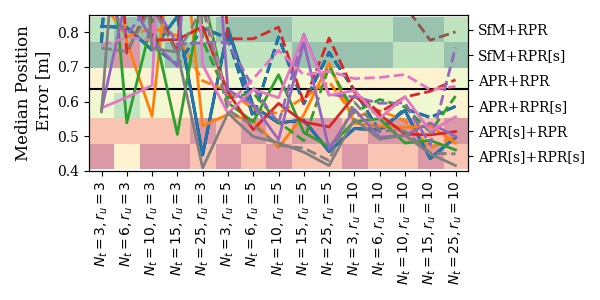}
    	\subcaption{Train 7.}
    	\label{figure_recurrent_fusion7}
    \end{minipage}
    \hfill
	\begin{minipage}[t]{0.495\linewidth}
        \centering
    	\includegraphics[trim=10 10 0 11, clip, width=1.0\linewidth]{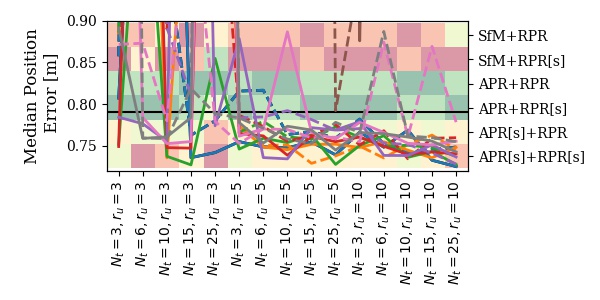}
    	\subcaption{Train 8.}
    	\label{figure_recurrent_fusion8}
    \end{minipage}
    \caption{Evaluation of recurrent absolute (i.e., SfM or APR) and relative (i.e., RPR) pose fusion for eight training datasets evaluated on the corresponding testing dataset. Lines evaluate for recurrent cells and stacked (solid) or non-stacked (dashed) cells. The heatmap in the background ranks the best fusion method (green indicates best ranked method and red indicates the last ranked method). [s] indicates the simulation-augmented pre-training of APR or RPR. Black horizontal line indicates the optimization of SfM and RPR with PGO. For readability, we set $y$ limits. The legend is defined in subfigure a) and is consistent.}
    \label{figure_recurrent_fusion}
\end{figure*}

\begin{table*}[t!]
\begin{center}
\setlength{\tabcolsep}{1.9pt}
    \caption{Evaluation results for the best recurrent cell for fusing SfM or APR with pre-trained RPR utilizing the stacked TRNN cell with $N_t$ timesteps and $r_u$ recurrent units. Results are shown as median position error in $m$ and position improvement in \%. The cell color indicates the relatedness between training and test environments and dynamics for the position error (light gray $\widehat{=}$ similar environments, dark gray $\widehat{=}$ strong environmental differences), and the degree of improvement (green $\widehat{=}$ improvement, red $\widehat{=}$ degradation) against SfM-only (see Table~\ref{table_results_sfm}).}
    \label{table_results_fusion}
    \small \begin{tabular}{ p{0.5cm} | p{0.5cm} | p{0.5cm} | p{0.5cm} | p{0.5cm} | p{0.5cm} | p{0.5cm} | p{0.5cm} | p{0.5cm} | p{0.5cm} | p{0.5cm} | p{0.5cm} | p{0.5cm} | p{0.5cm} | p{0.5cm} | p{0.5cm} | p{0.5cm} }
    \multicolumn{1}{c|}{} & \multicolumn{2}{c|}{\textbf{Train 1}} & \multicolumn{2}{c|}{\textbf{Train 2}} & \multicolumn{2}{c|}{\textbf{Train 3}} & \multicolumn{2}{c|}{\textbf{Train 4}} & \multicolumn{2}{c|}{\textbf{Train 5}} & \multicolumn{2}{c|}{\textbf{Train 6}} & \multicolumn{2}{c|}{\textbf{Train 7}} & \multicolumn{2}{c}{\textbf{Train 8}} \\ \hline
    \multicolumn{1}{l|}{\textbf{Test 1}} & \multicolumn{1}{r}{0.2606} & \multicolumn{1}{r|}{\cellcolor{green!20}+40.9} & \multicolumn{1}{r}{\cellcolor{gray!10}0.5909} & \multicolumn{1}{r|}{\cellcolor{green!10}+12.1} & \multicolumn{1}{r}{\cellcolor{gray!30}0.5249} & \multicolumn{1}{r|}{\cellcolor{green!10}+23.0} & \multicolumn{1}{r}{\cellcolor{gray!45}2.5297} & \multicolumn{1}{r|}{\cellcolor{red!30}--115.1} & \multicolumn{1}{r}{\cellcolor{gray!45}2.0405} & \multicolumn{1}{r|}{\cellcolor{red!10}--13.8} & \multicolumn{1}{r}{\cellcolor{gray!10}0.3877} & \multicolumn{1}{r|}{\cellcolor{green!20}+41.3} & \multicolumn{1}{r}{\cellcolor{gray!30}1.4115} & \multicolumn{1}{r|}{\cellcolor{green!10}+29.5} & \multicolumn{1}{r}{\cellcolor{gray!45}1.8464} & \multicolumn{1}{r}{\cellcolor{green!50}+72.2} \\
    \multicolumn{1}{l|}{\textbf{Test 2}} & \multicolumn{1}{r}{\cellcolor{gray!10}0.3297} & \multicolumn{1}{r|}{\cellcolor{green!20}+42.5} & \multicolumn{1}{r}{\cellcolor{gray!10}0.4970} & \multicolumn{1}{r|}{\cellcolor{green!10}+16.9} & \multicolumn{1}{r}{\cellcolor{gray!20}0.3091} & \multicolumn{1}{r|}{\cellcolor{green!10}+28.0} & \multicolumn{1}{r}{\cellcolor{gray!45}1.7065} & \multicolumn{1}{r|}{\cellcolor{red!10}--45.1} & \multicolumn{1}{r}{\cellcolor{gray!45}1.8001} & \multicolumn{1}{r|}{\cellcolor{red!20}--73.3} & \multicolumn{1}{r}{\cellcolor{gray!10}0.4709} & \multicolumn{1}{r|}{\cellcolor{green!20}+39.4} & \multicolumn{1}{r}{\cellcolor{gray!45}1.7347} & \multicolumn{1}{r|}{\cellcolor{green!20}+35.5} & \multicolumn{1}{r}{\cellcolor{gray!45}2.1859} & \multicolumn{1}{r}{\cellcolor{green!30}+63.3} \\
    \multicolumn{1}{l|}{\textbf{Test 3}} & \multicolumn{1}{r}{\cellcolor{gray!20}0.4724} & \multicolumn{1}{r|}{\cellcolor{green!30}+42.1} & \multicolumn{1}{r}{\cellcolor{gray!10}0.5831} & \multicolumn{1}{r|}{\cellcolor{green!10}+22.4} & \multicolumn{1}{r}{\cellcolor{gray!10}0.3335} & \multicolumn{1}{r|}{\cellcolor{green!10}+26.4} & \multicolumn{1}{r}{\cellcolor{gray!45}1.8748} & \multicolumn{1}{r|}{\cellcolor{red!10}--9.2} & \multicolumn{1}{r}{\cellcolor{gray!45}1.9356} & \multicolumn{1}{r|}{\cellcolor{red!10}--24.2} & \multicolumn{1}{r}{\cellcolor{gray!20}0.8070} & \multicolumn{1}{r|}{\cellcolor{green!10}+30.4} & \multicolumn{1}{r}{\cellcolor{gray!45}1.8795} & \multicolumn{1}{r|}{\cellcolor{green!20}+45.1} & \multicolumn{1}{r}{\cellcolor{gray!45}2.5305} & \multicolumn{1}{r}{\cellcolor{green!30}+61.6} \\
    \multicolumn{1}{l|}{\textbf{Test 4}} & \multicolumn{1}{r}{\cellcolor{gray!30}0.9710} & \multicolumn{1}{r|}{\cellcolor{green!20}+40.8} & \multicolumn{1}{r}{\cellcolor{gray!20}0.4942} & \multicolumn{1}{r|}{\cellcolor{green!10}+22.7} & \multicolumn{1}{r}{\cellcolor{gray!10}0.2394} & \multicolumn{1}{r|}{\cellcolor{green!10}+29.7} & \multicolumn{1}{r}{\cellcolor{gray!45}1.8955} & \multicolumn{1}{r|}{\cellcolor{red!10}--2.7} & \multicolumn{1}{r}{\cellcolor{gray!45}1.9427} & \multicolumn{1}{r|}{\cellcolor{red!10}--31.7} & \multicolumn{1}{r}{\cellcolor{gray!45}0.7822} & \multicolumn{1}{r|}{\cellcolor{green!20}+33.0} & \multicolumn{1}{r}{\cellcolor{gray!45}3.5102} & \multicolumn{1}{r|}{\cellcolor{green!20}+51.5} & \multicolumn{1}{r}{\cellcolor{gray!45}2.3277} & \multicolumn{1}{r}{\cellcolor{green!30}+66.9} \\
    \multicolumn{1}{l|}{\textbf{Test 5}} & \multicolumn{1}{r}{\cellcolor{gray!45}2.4365} & \multicolumn{1}{r|}{\cellcolor{green!30}+63.0} & \multicolumn{1}{r}{0.0888} & \multicolumn{1}{r|}{\cellcolor{green!10}+32.3} & \multicolumn{1}{r}{\cellcolor{gray!10}0.1399} & \multicolumn{1}{r|}{\cellcolor{green!20}+32.6} & \multicolumn{1}{r}{\cellcolor{gray!45}1.7744} & \multicolumn{1}{r|}{\cellcolor{red!10}--22.0} & \multicolumn{1}{r}{\cellcolor{gray!45}1.6923} & \multicolumn{1}{r|}{\cellcolor{red!50}--217.4} & \multicolumn{1}{r}{\cellcolor{gray!45}2.2221} & \multicolumn{1}{r|}{\cellcolor{green!20}+44.6} & \multicolumn{1}{r}{\cellcolor{gray!45}3.3965} & \multicolumn{1}{r|}{\cellcolor{green!20}+41.1} & \multicolumn{1}{r}{\cellcolor{gray!45}1.8474} & \multicolumn{1}{r}{\cellcolor{green!50}+80.5} \\
    \multicolumn{1}{l|}{\textbf{Test 6}} & \multicolumn{1}{r}{\cellcolor{gray!45}1.9714} & \multicolumn{1}{r|}{\cellcolor{green!30}+49.1} & \multicolumn{1}{r}{\cellcolor{gray!30}0.6641} & \multicolumn{1}{r|}{\cellcolor{green!10}+12.6} & \multicolumn{1}{r}{0.1620} & \multicolumn{1}{r|}{\cellcolor{green!20}+33.0} & \multicolumn{1}{r}{\cellcolor{gray!20}1.5941} & \multicolumn{1}{r|}{\cellcolor{red!40}--184.1} & \multicolumn{1}{r}{\cellcolor{gray!20}1.6885} & \multicolumn{1}{r|}{\cellcolor{red!50}--211.7} & \multicolumn{1}{r}{\cellcolor{gray!45}2.5828} & \multicolumn{1}{r|}{\cellcolor{green!20}+39.9} & \multicolumn{1}{r}{\cellcolor{gray!45}3.6296} & \multicolumn{1}{r|}{\cellcolor{green!20}+46.3} & \multicolumn{1}{r}{\cellcolor{gray!45}2.7798} & \multicolumn{1}{r}{\cellcolor{green!50}+73.3} \\
    \multicolumn{1}{l|}{\textbf{Test 7}} & \multicolumn{1}{r}{\cellcolor{gray!45}5.7052} & \multicolumn{1}{r|}{\cellcolor{green!50}+76.8} & \multicolumn{1}{r}{\cellcolor{gray!45}4.2081} & \multicolumn{1}{r|}{\cellcolor{green!40}+72.4} & \multicolumn{1}{r}{\cellcolor{gray!20}1.7238} & \multicolumn{1}{r|}{\cellcolor{red!10}--12.4} & \multicolumn{1}{r}{0.2631} & \multicolumn{1}{r|}{\cellcolor{green!10}+29.7} & \multicolumn{1}{r}{\cellcolor{gray!10}0.2612} & \multicolumn{1}{r|}{\cellcolor{green!10}+24.5} & \multicolumn{1}{r}{\cellcolor{gray!45}5.6267} & \multicolumn{1}{r|}{\cellcolor{green!50}+73.7} & \multicolumn{1}{r}{\cellcolor{gray!45}4.6363} & \multicolumn{1}{r|}{\cellcolor{green!50}+74.7} & \multicolumn{1}{r}{\cellcolor{gray!45}6.5534} & \multicolumn{1}{r}{\cellcolor{green!50}+68.7} \\
    \multicolumn{1}{l|}{\textbf{Test 8}} & \multicolumn{1}{r}{\cellcolor{gray!45}4.2785} & \multicolumn{1}{r|}{\cellcolor{green!50}+80.3} & \multicolumn{1}{r}{\cellcolor{gray!45}4.8111} & \multicolumn{1}{r|}{\cellcolor{green!40}+61.9} & \multicolumn{1}{r}{\cellcolor{gray!30}1.8361} & \multicolumn{1}{r|}{\cellcolor{red!20}--61.7} & \multicolumn{1}{r}{\cellcolor{gray!10}0.3215} & \multicolumn{1}{r|}{\cellcolor{green!10}+23.5} & \multicolumn{1}{r}{0.2125} & \multicolumn{1}{r|}{\cellcolor{green!20}+30.3} & \multicolumn{1}{r}{\cellcolor{gray!45}3.5936} & \multicolumn{1}{r|}{\cellcolor{green!40}+65.1} & \multicolumn{1}{r}{\cellcolor{gray!45}4.0741} & \multicolumn{1}{r|}{\cellcolor{green!40}+63.9} & \multicolumn{1}{r}{\cellcolor{gray!45}5.3554} & \multicolumn{1}{r}{\cellcolor{green!40}+62.5} \\
    \multicolumn{1}{l|}{\textbf{Test 9}} & \multicolumn{1}{r}{\cellcolor{gray!20}0.5803} & \multicolumn{1}{r|}{\cellcolor{green!20}+43.9} & \multicolumn{1}{r}{\cellcolor{gray!30}1.4832} & \multicolumn{1}{r|}{\cellcolor{green!10}+27.5} & \multicolumn{1}{r}{\cellcolor{gray!30}0.5536} & \multicolumn{1}{r|}{\cellcolor{green!20}+31.1} & \multicolumn{1}{r}{\cellcolor{gray!45}2.3406} & \multicolumn{1}{r|}{\cellcolor{green!40}+60.9} & \multicolumn{1}{r}{\cellcolor{gray!45}1.7902} & \multicolumn{1}{r|}{\cellcolor{red!20}--72.0} & \multicolumn{1}{r}{0.2952} & \multicolumn{1}{r|}{\cellcolor{green!20}+43.0} & \multicolumn{1}{r}{\cellcolor{gray!20}1.0633} & \multicolumn{1}{r|}{\cellcolor{green!20}+33.5} & \multicolumn{1}{r}{\cellcolor{gray!10}1.6314} & \multicolumn{1}{r}{\cellcolor{green!30}+58.1} \\
    \multicolumn{1}{l|}{\textbf{Test 10}} & \multicolumn{1}{r}{\cellcolor{gray!30}0.6552} & \multicolumn{1}{r|}{\cellcolor{green!20}+45.1} & \multicolumn{1}{r}{\cellcolor{gray!45}2.9389} & \multicolumn{1}{r|}{\cellcolor{green!20}+47.5} & \multicolumn{1}{r}{\cellcolor{gray!45}0.7821} & \multicolumn{1}{r|}{\cellcolor{green!20}+32.5} & \multicolumn{1}{r}{\cellcolor{gray!45}1.9858} & \multicolumn{1}{r|}{\cellcolor{green!10}+31.3} & \multicolumn{1}{r}{\cellcolor{gray!45}1.8844} & \multicolumn{1}{r|}{\cellcolor{red!10}--17.8} & \multicolumn{1}{r}{\cellcolor{gray!10}0.3701} & \multicolumn{1}{r|}{\cellcolor{green!20}+32.9} & \multicolumn{1}{r}{0.4084} & \multicolumn{1}{r|}{\cellcolor{green!20}+43.7} & \multicolumn{1}{r}{\cellcolor{gray!20}0.7674} & \multicolumn{1}{r}{\cellcolor{green!50}+84.6} \\
    \multicolumn{1}{l|}{\textbf{Test 11}} & \multicolumn{1}{r}{\cellcolor{gray!45}2.6793} & \multicolumn{1}{r|}{\cellcolor{green!50}+78.0} & \multicolumn{1}{r}{\cellcolor{gray!45}2.4830} & \multicolumn{1}{r|}{\cellcolor{green!20}+40.2} & \multicolumn{1}{r}{\cellcolor{gray!20}0.3220} & \multicolumn{1}{r|}{\cellcolor{green!20}+34.6} & \multicolumn{1}{r}{\cellcolor{gray!45}1.9044} & \multicolumn{1}{r|}{--0.1} & \multicolumn{1}{r}{\cellcolor{gray!45}1.7822} & \multicolumn{1}{r|}{\cellcolor{red!20}--81.0} & \multicolumn{1}{r}{\cellcolor{gray!20}0.5500} & \multicolumn{1}{r|}{\cellcolor{green!20}+40.0} & \multicolumn{1}{r}{\cellcolor{gray!10}1.2626} & \multicolumn{1}{r|}{\cellcolor{green!10}+29.9} & \multicolumn{1}{r}{0.7451} & \multicolumn{1}{r}{\cellcolor{green!30}+65.2} \\
    \end{tabular}
\end{center}
\end{table*}

\begin{figure*}[!t]
    \centering
	\begin{minipage}[t]{0.8\linewidth}
        \centering
        \includegraphics[trim=10 10 10 9, clip, width=1.0\linewidth]{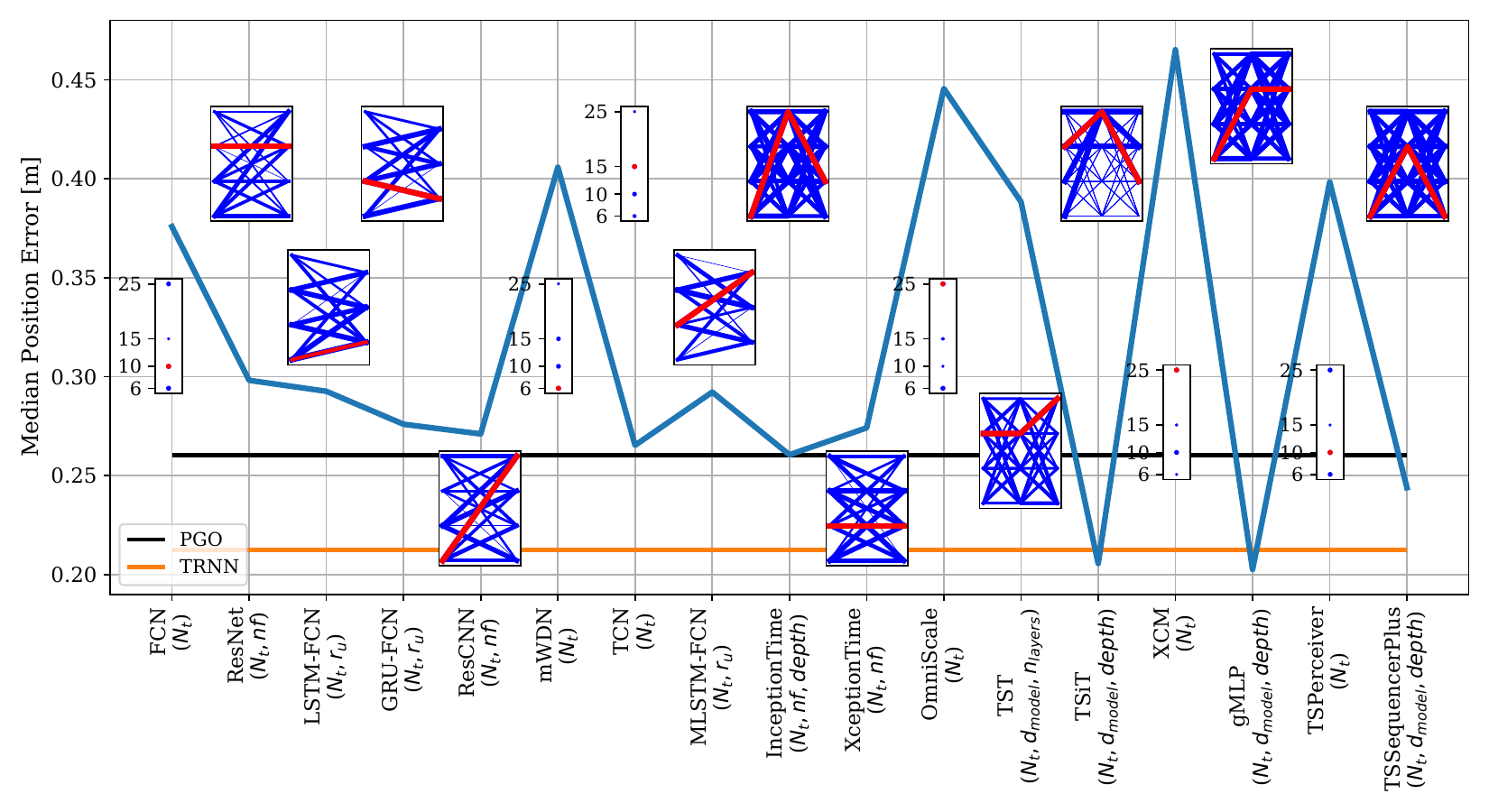}
    	\subcaption{Train 5, test 8.}
    	\label{figure_tsai1}
    \end{minipage}
	\begin{minipage}[t]{0.8\linewidth}
        \centering
        \includegraphics[trim=10 10 10 9, clip, width=1.0\linewidth]{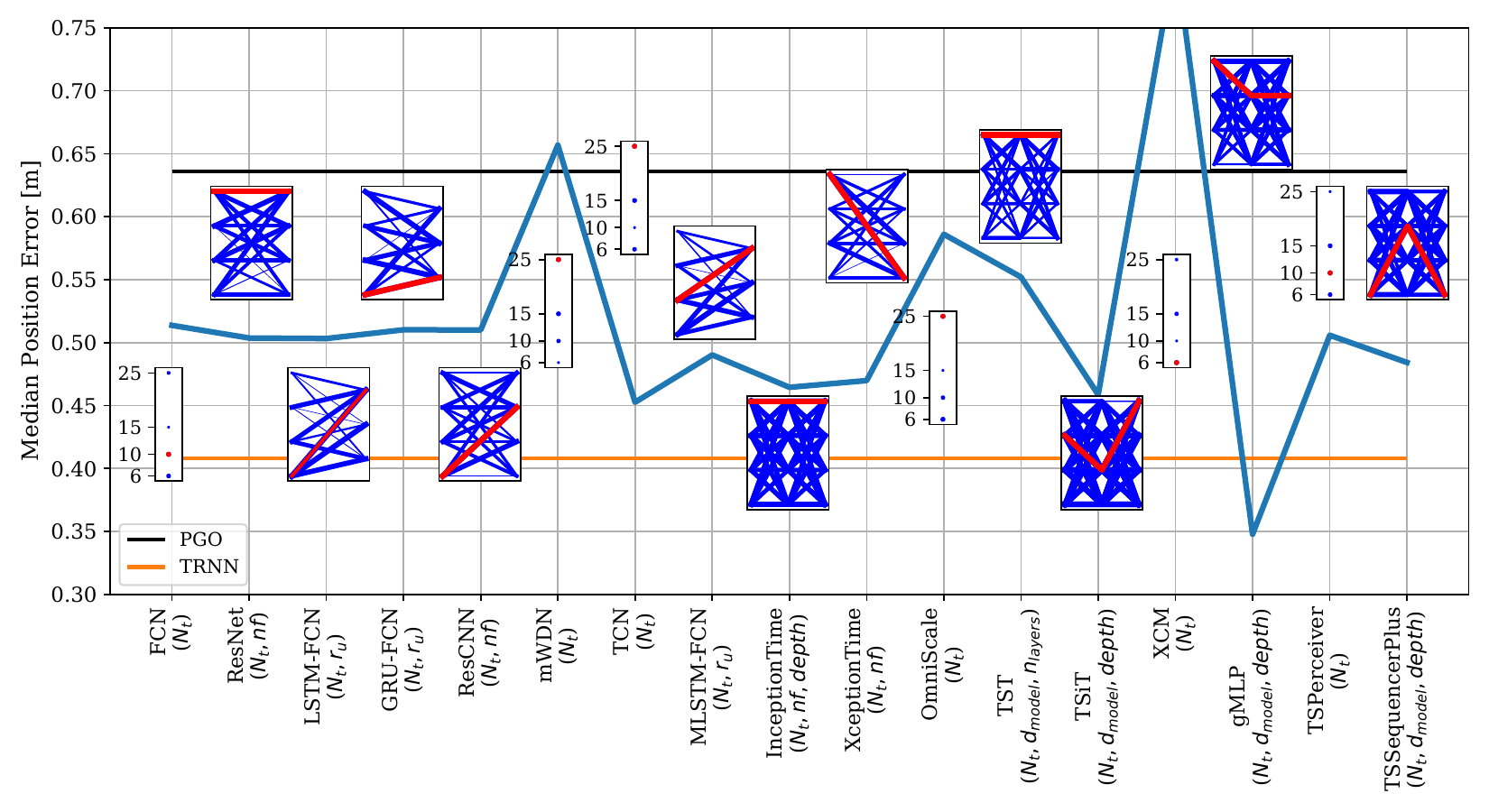}
    	\subcaption{Train 7, test 10.}
    	\label{figure_tsai2}
    \end{minipage}
    \caption{Hyperparameter search for the convolutional, recurrent convolutional, and Transformer models. We search for the hyperparameters $N_t \in [5, 10, 15, 25]$, $r_u \in [3, 5, 10]$, $nf \in [16, 32, 64, 128]$, $depth \in [3, 4, 5, 6]$, $d_{model} \in [32, 64, 128, 256]$, and $n_{layers} \in [2, 3, 4, 5]$ (ordered from bottom to top). We select the best hyperparameters (marked red) to compare with PGO (black) and TRNN (orange).}
    \label{figure_tsai}
\end{figure*}

\begin{figure*}[!t]
    \centering
	\begin{minipage}[t]{0.325\linewidth}
        \centering
    	\includegraphics[trim=10 10 9 11, clip, width=1.0\linewidth]{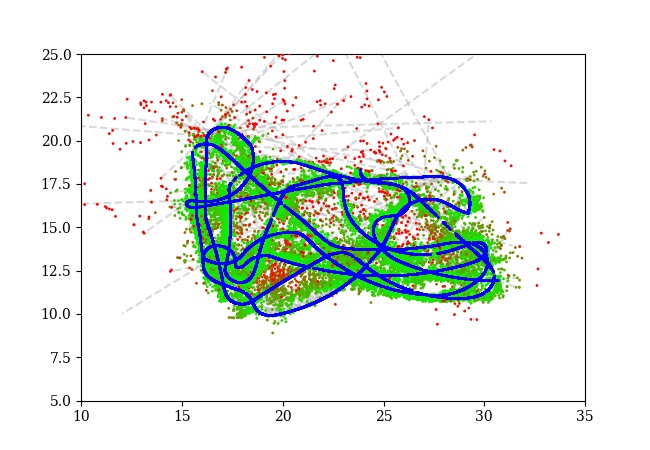}
    	\subcaption{Train 1, test 1.}
    	\label{figure_eval_traj_fusion1}
    \end{minipage}
    \hfill
	\begin{minipage}[t]{0.325\linewidth}
        \centering
    	\includegraphics[trim=10 10 9 11, clip, width=1.0\linewidth]{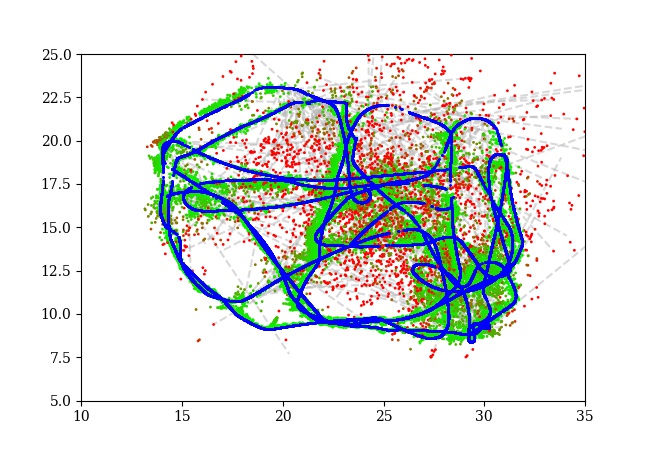}
    	\subcaption{Train 3, test 6.}
    	\label{figure_eval_traj_fusion2}
    \end{minipage}
    \hfill
	\begin{minipage}[t]{0.325\linewidth}
        \centering
    	\includegraphics[trim=10 10 9 11, clip, width=1.0\linewidth]{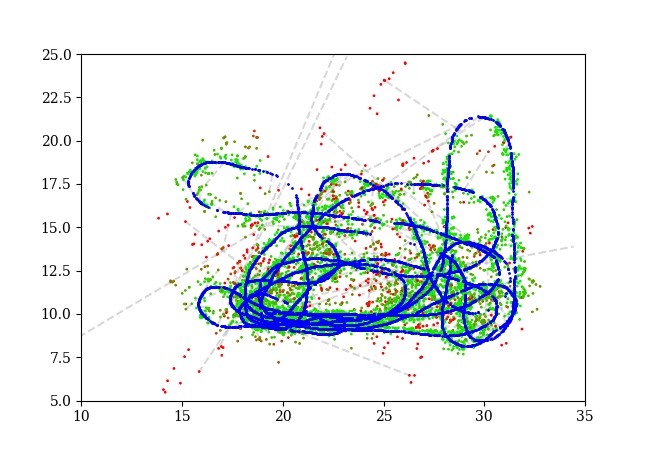}
    	\subcaption{Train 4, test 7.}
    	\label{figure_eval_traj_fusion3}
    \end{minipage}
    \caption{Evaluation of the predicted positions (green $\widehat{=}$ low position error, red $\widehat{=}$ large position error) against the ground truth trajectories (blue) for the recurrent fusion of SfM with the pre-trained RPR model utilizing the stacked TRNN cell with $N_t = 15$ timesteps and $r_u = 10$ recurrent units.}
    \label{figure_eval_traj_fusion}
\end{figure*}

\paragraph{Recurrent APR-RPR Fusion \& Augmentation.} In order to fuse APR and RPR and predict an optimized absolute pose, it is essential to consider the specific deterministic, random-walk, non-linear, or long-memory behavior (or a combination of these) of both robotic and human motion. Specifically, the motion range of objects is limited to certain velocities and orientation changes. To address these challenges, a recurrent unit is necessary to process the required pose information. However, there is currently no clear understanding of which RNN-cell structure is most suitable for each type of behavior and the characterization of these units is not clear \citep{khaldi_afia_chiheb}. Therefore, we provide a comprehensive evaluation of eight RNN units for the fusion task. According to \cite{khaldi_afia_chiheb}, an MGU cell is most suitable for a deterministic and non-linear behavior, while an LSTM cell is recommended for chaotic behavior, see also \cite{yu_si_hu_zhang}. In order to further explore the optimal RNN-cell structure for the fusion task, we train and evaluate in addition to the commonly used cells LSTM, GRU, and MGU, the cell types CFN, QRNN, RAN, SRU, and TRNN. We vary the number of input timestep values ($N_t \in [3, 6, 10, 15, 25]$) and the number of recurrent units ($r_u \in [5, 10, 20, 50]$). Both single and two stacked RNN cells are evaluated using a copy memory task, which is designed to stress test the ability of recurrent networks to propagate long-term, distant information \citep{bai_kolter_koltun}. This task assesses the model's capacity to retain information for different lengths of time ($N_t$). To compare the different models, Figure~\ref{figure_fusion_parameters} shows the number of trainable model parameters for each of the eight stacked RNN cells and the number of units $r_u$. We observe that larger cell sizes ($r_u$) result in a significant increase in trainable parameters, leading to a higher risk of overfitting. On the other hand, smaller gating mechanisms result in a decrease in the number of parameters (i.e., LSTM > GRU > MGU), with TRNN having the fewest parameters. Figure~\ref{figure_recurrent_fusion} provides an overview of the results for all eight training datasets and various fusions of absolute and relative models (with and without simulated pre-training), compared to PGO (black line). We evaluate the parameters $N_t \in [3, 6, 10, 15, 25]$ and $r_u \in [3, 5, 10]$. We provide rankings for all methods with colored backgrounds. In the following, [s] indicates pre-training. The results of the method ranking indicate that SfM+RPR and SfM+RPR[s] outperform APR on all datasets, except for the train 8 dataset. The efficacy of pre-training is dependent on the dataset. Among the fusion models, most outperform PGO on the train 1, train 2, train 3, train 5, train 6, train 7, and train 8 datasets. Using two stacked RNN cells always yields better results than using only one cell. TRNN~\citep{balduzzi_ghifary} consistently achieves the lowest position errors, which is evident in Figures~\ref{figure_recurrent_fusion1}, \ref{figure_recurrent_fusion2}, \ref{figure_recurrent_fusion3}, \ref{figure_recurrent_fusion4}, \ref{figure_recurrent_fusion5}, and \ref{figure_recurrent_fusion6}, and \ref{figure_recurrent_fusion7}. Therefore, a small model with few trainable parameters, i.e., $r_u=10$ (as shown in Figure~\ref{figure_fusion_parameters}), is adequate for fusing absolute and relative poses. A large timestep size of $N_t = 25$ is observed to be better, as it enables the model to learn long-term dependencies of the motion dynamics. The difference between the recurrent units $r_u = 5$ and $r_u = 10$ is marginal and varies depending on the dataset, as evident in Figure~\ref{figure_recurrent_fusion4} versus Figure~\ref{figure_recurrent_fusion8}. A model size of $r_u = 3$ is too small to learn usable features.

Table~\ref{table_results_fusion} displays the improvement in position compared to SfM-only (Table~\ref{table_results_sfm}) for all datasets. Depending on the dataset, the results can demonstrate an improvement ranging from +12.1\% to +84.6\% for the robotic dataset, especially when the training and testing are conducted on robotic datasets. However, there is a significant decline in results when assessing the handheld dataset on robotic datasets. This suggests that while recurrent cells can learn motion dynamics and produce a more precise predicted trajectory, they cannot adjust to new dynamics, such as transitioning from fast handheld movements to slow robotic movements. Nevertheless, the model can adapt from robotic to handheld movements. The enhancement is attributed to a significant decrease in outliers and the ability to predict smoothed trajectories, which is evident by comparing Figure~\ref{figure_eval_traj_fusion} with Figure~\ref{figure_eval_traj}.

In Figure~\ref{figure_tsai}, we present additional results involving 17 convolutional, recurrent, and Transformer models, along with a comparison to PGO and TRNN (as shown in Figure~\ref{figure_recurrent_fusion}). Hyperparameter searches are performed for the following parameters: $N_t$, $r_u$, $nf$, $depth$, $d_{model}$, and $n_{layers}$. For the handheld dataset (Figure~\ref{figure_tsai1}), only TSiT \citep{zerveas_jayaraman_patel} and gMLP \citep{liu_dai_so_le} demonstrate superior performance compared to PGO and achieve similar to those of TRNN. However, in the case of the robotic dataset (Figure~\ref{figure_tsai2}), most of the methods outperform PGO. Nevertheless, TRNN exhibits lower positioning errors compared to all other methods, except for gMLP.

\subsection{Comparison to State-of-the-art}
\label{sec_state_of_the_art}

In comparison to state-of-the-art methods, we employ the accelerated coordinate encoding (ACE) technique proposed by \cite{brachmann_cavallari}. ACE leverages a scene-agnostic feature backbone along with a scene-specific prediction head. In their work, \cite{brachmann_cavallari} utilize an MLP prediction head, enabling optimization across thousands of viewpoints simultaneously during each training iteration. These characteristics leads to a stable and rapid convergence. Notably, ACE represents the most recent advancement in this research domain and has surpassed alternative approaches such as PoseNet~\citep{kendall_grimes_cipolla}, MS-Transformer~\citep{shavit_ferens_keller}, DSAC*~\citep{brachmann_rother}, SANet~\citep{yang_bai_tang}, and SRC~\citep{dong_wang_zhuang}.

\begin{table*}[t!]
\begin{center}
\setlength{\tabcolsep}{2.2pt}
    \caption{Comparison of results for the TRNN fusion model (left column) with ACE~\citep{brachmann_cavallari} (right column). Results are shown as median position error in $m$. The cell color indicates the relatedness between training and test environments and dynamics (light gray $\widehat{=}$ similar environments, dark gray $\widehat{=}$ strong environmental differences).}
    \label{table_results_trnn_ace}
    \small \begin{tabular}{ p{0.5cm} | p{0.5cm} | p{0.5cm} | p{0.5cm} | p{0.5cm} | p{0.5cm} | p{0.5cm} | p{0.5cm} | p{0.5cm} | p{0.5cm} | p{0.5cm} | p{0.5cm} | p{0.5cm} | p{0.5cm} | p{0.5cm} | p{0.5cm} | p{0.5cm} }
    \multicolumn{1}{c|}{} & \multicolumn{2}{c|}{\textbf{Train 1}} & \multicolumn{2}{c|}{\textbf{Train 2}} & \multicolumn{2}{c|}{\textbf{Train 3}} & \multicolumn{2}{c|}{\textbf{Train 4}} & \multicolumn{2}{c|}{\textbf{Train 5}} & \multicolumn{2}{c|}{\textbf{Train 6}} & \multicolumn{2}{c|}{\textbf{Train 7}} & \multicolumn{2}{c}{\textbf{Train 8}} \\ \hline
    \multicolumn{1}{l|}{\textbf{Test 1}} & \multicolumn{1}{r}{0.2606} & \multicolumn{1}{r|}{0.271} & \multicolumn{1}{r}{\cellcolor{gray!10}0.5909} & \multicolumn{1}{r|}{\cellcolor{gray!10}0.329} & \multicolumn{1}{r}{\cellcolor{gray!30}0.5249} & \multicolumn{1}{r|}{\cellcolor{gray!30}0.492} & \multicolumn{1}{r}{\cellcolor{gray!45}2.5297} & \multicolumn{1}{r|}{\cellcolor{gray!45}2.522} & \multicolumn{1}{r}{\cellcolor{gray!45}2.0405} & \multicolumn{1}{r|}{\cellcolor{gray!45}1.701} & \multicolumn{1}{r}{\cellcolor{gray!10}0.3877} & \multicolumn{1}{r|}{\cellcolor{gray!10}0.484} & \multicolumn{1}{r}{\cellcolor{gray!30}1.4115} & \multicolumn{1}{r|}{\cellcolor{gray!30}1.619} & \multicolumn{1}{r}{\cellcolor{gray!45}1.8464} & \multicolumn{1}{r}{\cellcolor{gray!45}1.939} \\
    \multicolumn{1}{l|}{\textbf{Test 2}} & \multicolumn{1}{r}{\cellcolor{gray!10}0.3297} & \multicolumn{1}{r|}{\cellcolor{gray!10}0.135} & \multicolumn{1}{r}{\cellcolor{gray!10}0.4970} & \multicolumn{1}{r|}{\cellcolor{gray!10}0.510} & \multicolumn{1}{r}{\cellcolor{gray!20}0.3091} & \multicolumn{1}{r|}{\cellcolor{gray!20}0.414} & \multicolumn{1}{r}{\cellcolor{gray!45}1.7065} & \multicolumn{1}{r|}{\cellcolor{gray!45}1.349} & \multicolumn{1}{r}{\cellcolor{gray!45}1.8001} & \multicolumn{1}{r|}{\cellcolor{gray!45}1.557} & \multicolumn{1}{r}{\cellcolor{gray!10}0.4709} & \multicolumn{1}{r|}{\cellcolor{gray!10}0.559} & \multicolumn{1}{r}{\cellcolor{gray!45}1.7347} & \multicolumn{1}{r|}{\cellcolor{gray!45}1.829} & \multicolumn{1}{r}{\cellcolor{gray!45}2.1859} & \multicolumn{1}{r}{\cellcolor{gray!45}2.130} \\
    \multicolumn{1}{l|}{\textbf{Test 3}} & \multicolumn{1}{r}{\cellcolor{gray!20}0.4724} & \multicolumn{1}{r|}{\cellcolor{gray!20}0.166} & \multicolumn{1}{r}{\cellcolor{gray!10}0.5831} & \multicolumn{1}{r|}{\cellcolor{gray!10}0.430} & \multicolumn{1}{r}{\cellcolor{gray!10}0.3335} & \multicolumn{1}{r|}{\cellcolor{gray!10}0.369} & \multicolumn{1}{r}{\cellcolor{gray!45}1.8748} & \multicolumn{1}{r|}{\cellcolor{gray!45}1.233} & \multicolumn{1}{r}{\cellcolor{gray!45}1.9356} & \multicolumn{1}{r|}{\cellcolor{gray!45}1.369} & \multicolumn{1}{r}{\cellcolor{gray!20}0.8070} & \multicolumn{1}{r|}{\cellcolor{gray!20}0.779} & \multicolumn{1}{r}{\cellcolor{gray!45}1.8795} & \multicolumn{1}{r|}{\cellcolor{gray!45}1.714} & \multicolumn{1}{r}{\cellcolor{gray!45}2.5305} & \multicolumn{1}{r}{\cellcolor{gray!45}1.175} \\
    \multicolumn{1}{l|}{\textbf{Test 4}} & \multicolumn{1}{r}{\cellcolor{gray!30}0.9710} & \multicolumn{1}{r|}{\cellcolor{gray!30}0.263} & \multicolumn{1}{r}{\cellcolor{gray!20}0.4942} & \multicolumn{1}{r|}{\cellcolor{gray!20}0.495} & \multicolumn{1}{r}{\cellcolor{gray!10}0.2394} & \multicolumn{1}{r|}{\cellcolor{gray!10}0.273} & \multicolumn{1}{r}{\cellcolor{gray!45}1.8955} & \multicolumn{1}{r|}{\cellcolor{gray!45}1.044} & \multicolumn{1}{r}{\cellcolor{gray!45}1.9427} & \multicolumn{1}{r|}{\cellcolor{gray!45}1.267} & \multicolumn{1}{r}{\cellcolor{gray!45}0.7822} & \multicolumn{1}{r|}{\cellcolor{gray!45}0.755} & \multicolumn{1}{r}{\cellcolor{gray!45}3.5102} & \multicolumn{1}{r|}{\cellcolor{gray!45}3.687} & \multicolumn{1}{r}{\cellcolor{gray!45}2.3277} & \multicolumn{1}{r}{\cellcolor{gray!45}0.958} \\
    \multicolumn{1}{l|}{\textbf{Test 5}} & \multicolumn{1}{r}{\cellcolor{gray!45}2.4365} & \multicolumn{1}{r|}{\cellcolor{gray!45}2.715} & \multicolumn{1}{r}{0.0888} & \multicolumn{1}{r|}{0.096} & \multicolumn{1}{r}{\cellcolor{gray!10}0.1399} & \multicolumn{1}{r|}{\cellcolor{gray!10}0.114} & \multicolumn{1}{r}{\cellcolor{gray!45}1.7744} & \multicolumn{1}{r|}{\cellcolor{gray!45}0.541} & \multicolumn{1}{r}{\cellcolor{gray!45}1.6923} & \multicolumn{1}{r|}{\cellcolor{gray!45}0.618} & \multicolumn{1}{r}{\cellcolor{gray!45}2.2221} & \multicolumn{1}{r|}{\cellcolor{gray!45}2.040} & \multicolumn{1}{r}{\cellcolor{gray!45}3.3965} & \multicolumn{1}{r|}{\cellcolor{gray!45}3.583} & \multicolumn{1}{r}{\cellcolor{gray!45}1.8474} & \multicolumn{1}{r}{\cellcolor{gray!45}1.033} \\
    \multicolumn{1}{l|}{\textbf{Test 6}} & \multicolumn{1}{r}{\cellcolor{gray!45}1.9714} & \multicolumn{1}{r|}{\cellcolor{gray!45}0.660} & \multicolumn{1}{r}{\cellcolor{gray!30}0.6641} & \multicolumn{1}{r|}{\cellcolor{gray!30}0.687} & \multicolumn{1}{r}{0.1620} & \multicolumn{1}{r|}{0.175} & \multicolumn{1}{r}{\cellcolor{gray!20}1.5941} & \multicolumn{1}{r|}{\cellcolor{gray!20}0.830} & \multicolumn{1}{r}{\cellcolor{gray!20}1.6885} & \multicolumn{1}{r|}{\cellcolor{gray!20}0.788} & \multicolumn{1}{r}{\cellcolor{gray!45}2.5828} & \multicolumn{1}{r|}{\cellcolor{gray!45}2.235} & \multicolumn{1}{r}{\cellcolor{gray!45}3.6296} & \multicolumn{1}{r|}{\cellcolor{gray!45}4.430} & \multicolumn{1}{r}{\cellcolor{gray!45}2.7798} & \multicolumn{1}{r}{\cellcolor{gray!45}1.612} \\
    \multicolumn{1}{l|}{\textbf{Test 7}} & \multicolumn{1}{r}{\cellcolor{gray!45}5.7052} & \multicolumn{1}{r|}{\cellcolor{gray!45}3.550} & \multicolumn{1}{r}{\cellcolor{gray!45}4.2081} & \multicolumn{1}{r|}{\cellcolor{gray!45}3.588} & \multicolumn{1}{r}{\cellcolor{gray!20}1.7238} & \multicolumn{1}{r|}{\cellcolor{gray!20}0.418} & \multicolumn{1}{r}{0.2631} & \multicolumn{1}{r|}{0.216} & \multicolumn{1}{r}{\cellcolor{gray!10}0.2612} & \multicolumn{1}{r|}{\cellcolor{gray!10}0.227} & \multicolumn{1}{r}{\cellcolor{gray!45}5.6267} & \multicolumn{1}{r|}{\cellcolor{gray!45}5.325} & \multicolumn{1}{r}{\cellcolor{gray!45}4.6363} & \multicolumn{1}{r|}{\cellcolor{gray!45}5.317} & \multicolumn{1}{r}{\cellcolor{gray!45}6.5534} & \multicolumn{1}{r}{\cellcolor{gray!45}3.998} \\
    \multicolumn{1}{l|}{\textbf{Test 8}} & \multicolumn{1}{r}{\cellcolor{gray!45}4.2785} & \multicolumn{1}{r|}{\cellcolor{gray!45}2.281} & \multicolumn{1}{r}{\cellcolor{gray!45}4.8111} & \multicolumn{1}{r|}{\cellcolor{gray!45}1.130} & \multicolumn{1}{r}{\cellcolor{gray!30}1.8361} & \multicolumn{1}{r|}{\cellcolor{gray!30}0.402} & \multicolumn{1}{r}{\cellcolor{gray!10}0.3215} & \multicolumn{1}{r|}{\cellcolor{gray!10}0.325} & \multicolumn{1}{r}{0.2125} & \multicolumn{1}{r|}{0.225} & \multicolumn{1}{r}{\cellcolor{gray!45}3.5936} & \multicolumn{1}{r|}{\cellcolor{gray!45}3.645} & \multicolumn{1}{r}{\cellcolor{gray!45}4.0741} & \multicolumn{1}{r|}{\cellcolor{gray!45}4.441} & \multicolumn{1}{r}{\cellcolor{gray!45}5.3554} & \multicolumn{1}{r}{\cellcolor{gray!45}2.686} \\
    \multicolumn{1}{l|}{\textbf{Test 9}} & \multicolumn{1}{r}{\cellcolor{gray!30}0.5803} & \multicolumn{1}{r|}{\cellcolor{gray!30}0.319} & \multicolumn{1}{r}{\cellcolor{gray!45}1.4832} & \multicolumn{1}{r|}{\cellcolor{gray!45}0.864} & \multicolumn{1}{r}{\cellcolor{gray!45}0.5536} & \multicolumn{1}{r|}{\cellcolor{gray!45}0.977} & \multicolumn{1}{r}{\cellcolor{gray!45}2.3406} & \multicolumn{1}{r|}{\cellcolor{gray!45}1.710} & \multicolumn{1}{r}{\cellcolor{gray!45}1.7902} & \multicolumn{1}{r|}{\cellcolor{gray!45}1.894} & \multicolumn{1}{r}{\cellcolor{gray!10}0.2952} & \multicolumn{1}{r|}{\cellcolor{gray!10}0.213} & \multicolumn{1}{r}{1.0633} & \multicolumn{1}{r|}{1.014} & \multicolumn{1}{r}{\cellcolor{gray!20}1.6314} & \multicolumn{1}{r}{\cellcolor{gray!20}0.743} \\
    \multicolumn{1}{l|}{\textbf{Test 10}} & \multicolumn{1}{r}{\cellcolor{gray!45}0.6552} & \multicolumn{1}{r|}{\cellcolor{gray!45}0.423} & \multicolumn{1}{r}{\cellcolor{gray!45}2.9389} & \multicolumn{1}{r|}{\cellcolor{gray!45}1.198} & \multicolumn{1}{r}{\cellcolor{gray!20}0.7821} & \multicolumn{1}{r|}{\cellcolor{gray!20}1.233} & \multicolumn{1}{r}{\cellcolor{gray!45}1.9858} & \multicolumn{1}{r|}{\cellcolor{gray!45}1.561} & \multicolumn{1}{r}{\cellcolor{gray!45}1.8844} & \multicolumn{1}{r|}{\cellcolor{gray!45}1.733} & \multicolumn{1}{r}{\cellcolor{gray!20}0.3701} & \multicolumn{1}{r|}{\cellcolor{gray!20}0.535} & \multicolumn{1}{r}{\cellcolor{gray!10}0.4084} & \multicolumn{1}{r|}{\cellcolor{gray!10}0.365} & \multicolumn{1}{r}{0.7674} & \multicolumn{1}{r}{0.431} \\
    \multicolumn{1}{l|}{\textbf{Test 11}} & \multicolumn{1}{r}{\cellcolor{gray!45}2.6793} & \multicolumn{1}{r|}{\cellcolor{gray!45}2.676} & \multicolumn{1}{r}{\cellcolor{gray!45}2.4830} & \multicolumn{1}{r|}{\cellcolor{gray!45}1.737} & \multicolumn{1}{r}{\cellcolor{gray!20}0.3220} & \multicolumn{1}{r|}{\cellcolor{gray!20}0.426} & \multicolumn{1}{r}{\cellcolor{gray!45}1.9044} & \multicolumn{1}{r|}{\cellcolor{gray!45}0.925} & \multicolumn{1}{r}{\cellcolor{gray!45}1.7822} & \multicolumn{1}{r|}{\cellcolor{gray!45}1.526} & \multicolumn{1}{r}{\cellcolor{gray!20}0.5500} & \multicolumn{1}{r|}{\cellcolor{gray!20}0.678} & \multicolumn{1}{r}{\cellcolor{gray!10}1.2626} & \multicolumn{1}{r|}{\cellcolor{gray!10}1.239} & \multicolumn{1}{r}{0.7451} & \multicolumn{1}{r}{0.783} \\
    \end{tabular}
\end{center}
\end{table*}

\begin{figure}[!t]
    \centering
    \includegraphics[trim=10 10 10 9, clip, width=1.0\linewidth]{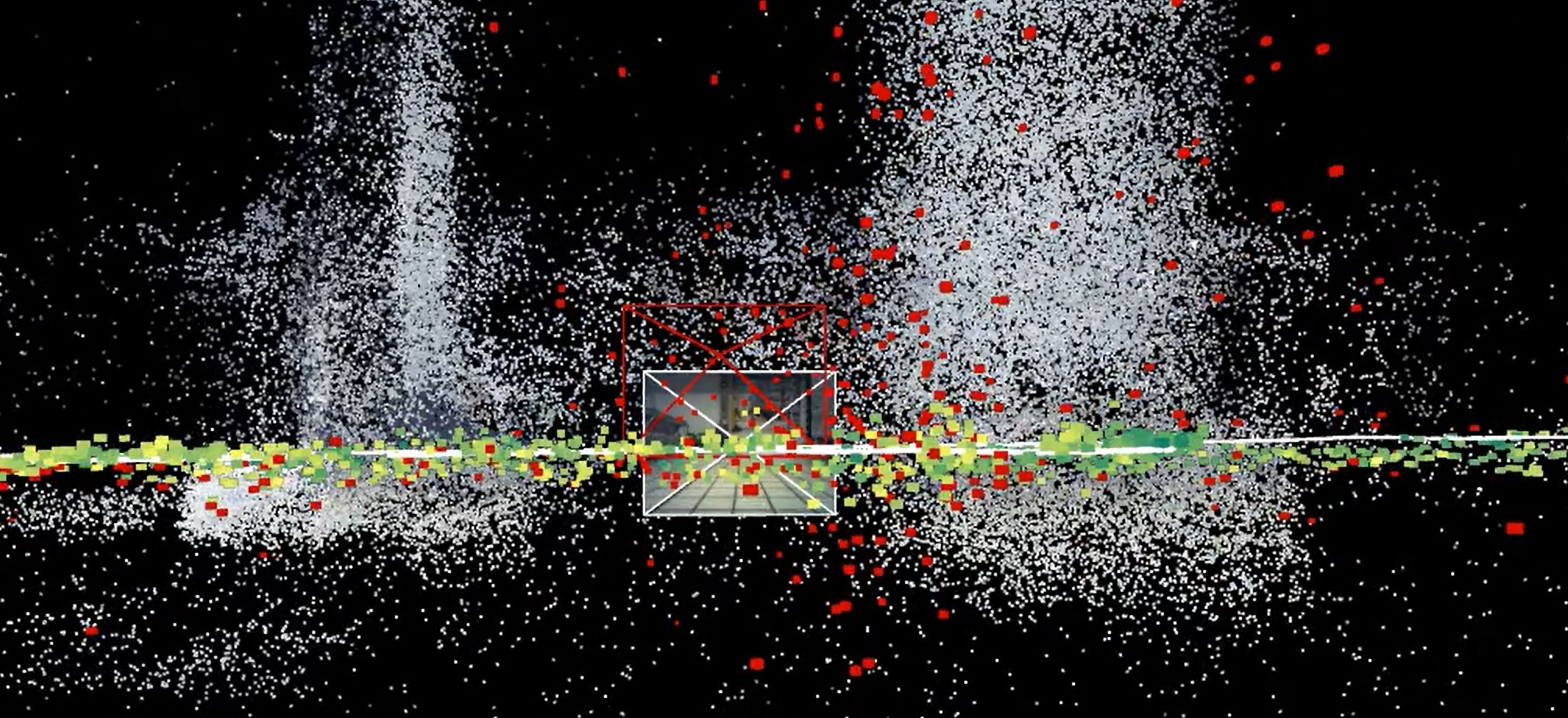}
    \caption{Retrieving absolute poses from the point cloud reconstructed with ACE~\citep{brachmann_cavallari}.}
    \label{figure_ace_point_cloud}
\end{figure}

Although ACE effectively generates a dense point cloud for nearby objects, such as the large black absorber walls, it struggles to extract features from distant white walls and other feature-rich objects at a distance (see Figure~\ref{figure_ace_point_cloud}). The performance of ACE in pose prediction is notably diminished under challenging conditions, particularly when dealing with absorber walls. In contrast, our point clouds reconstructed with SfM exhibit more evenly distributed features, even from distant walls (see Figure~\ref{figure_sfm_pcd}). Additionally, we present a comparison of SfM+RPR fusion with TRNN networks and ACE in Table~\ref{table_results_trnn_ace}. While ACE performs well for scenarios involving close objects, such as in the train 1 and test 3 datasets, our fusion model outperforms ACE in large-scale scenarios, as seen in the train 6 and test 1 dataset. Consequently, for further enhancements in localization results, ACE can be utilized as a black-box model in conjunction with our relative module to achieve a more robust localization solution.

\subsection{Robustness to Environmental Challenges}
\label{sec_evaluation_challenges}

\begin{figure*}[!t]
    \centering
	\begin{minipage}[t]{0.8\linewidth}
        \centering
    	\includegraphics[trim=10 10 10 9, clip, width=1.0\linewidth]{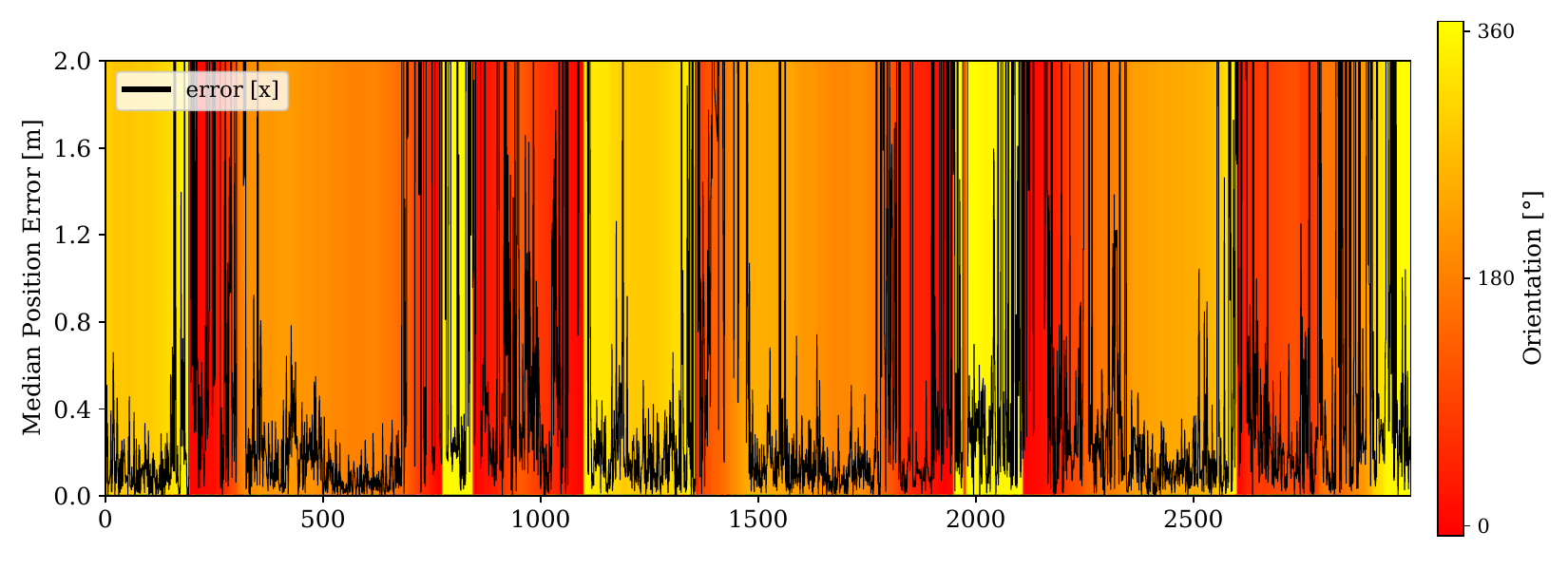}
    \end{minipage}
	\begin{minipage}[t]{0.8\linewidth}
        \centering
    	\includegraphics[trim=10 10 10 9, clip, width=1.0\linewidth]{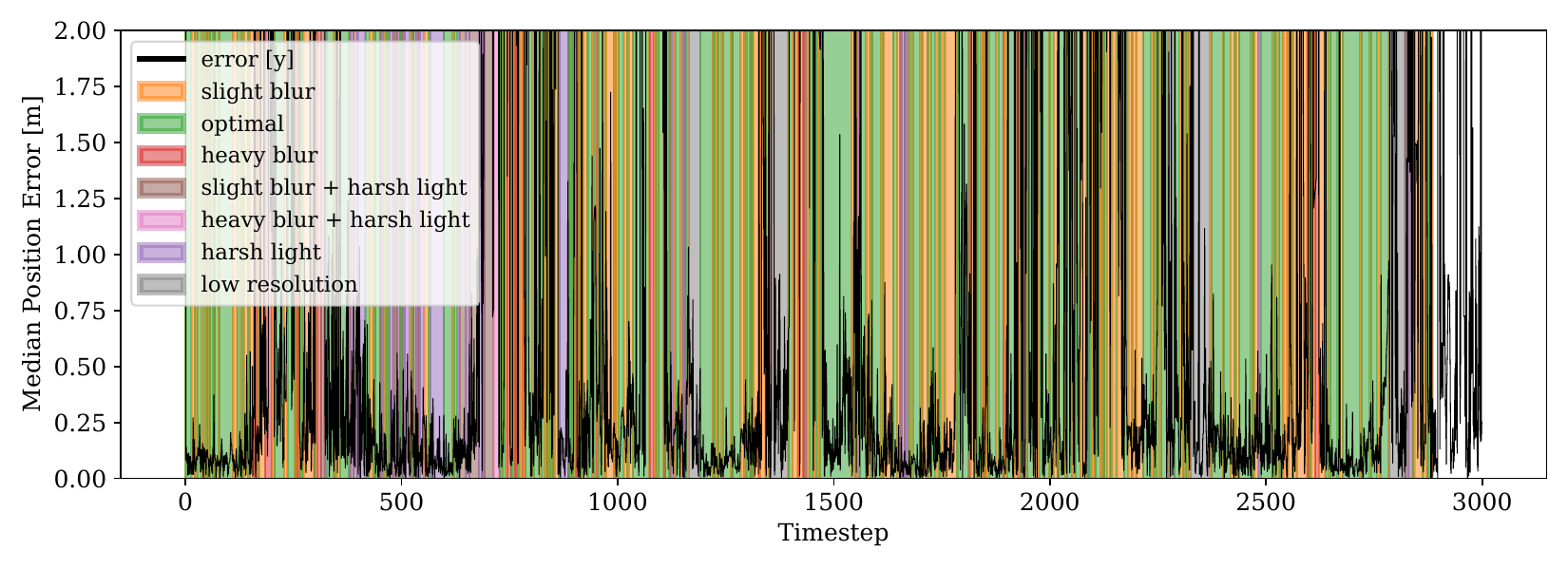}
    \end{minipage}
	\begin{minipage}[t]{0.8\linewidth}
        \centering
    	\includegraphics[trim=10 10 10 9, clip, width=1.0\linewidth]{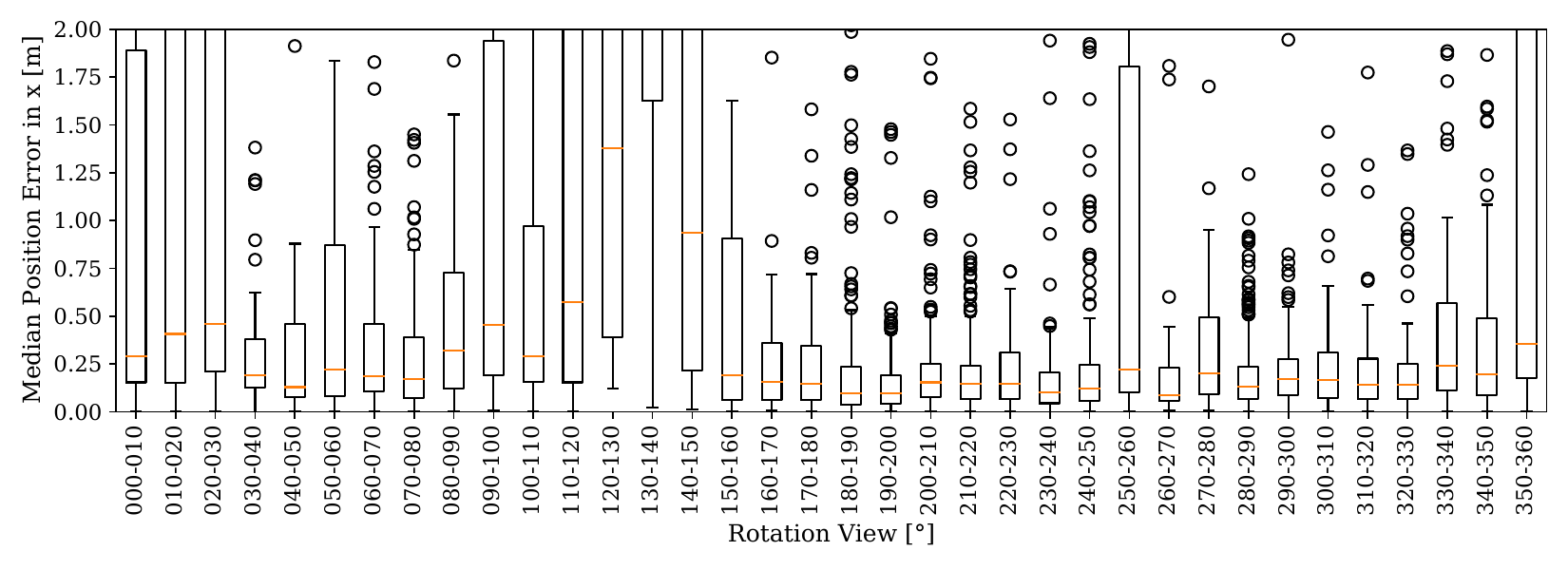}
    \end{minipage}
    \caption{Plot of the positional error with respect to the x and y-coordinates of SfM for the train 4 dataset with environmental challenges marked in color. The top plot shows the orientation between 0\textdegree and 360\textdegree. The orientation of 0\textdegree points to the east of the environment. The middle plot shows seven challenges or optimal image conditions in color. The bottom plot shows the error dependent on the orientation in the environment.}
    \label{figure_x_y_challenges}
\end{figure*}

\begin{figure}[!t]
    \centering
    \includegraphics[trim=10 10 10 9, clip, width=1.0\linewidth]{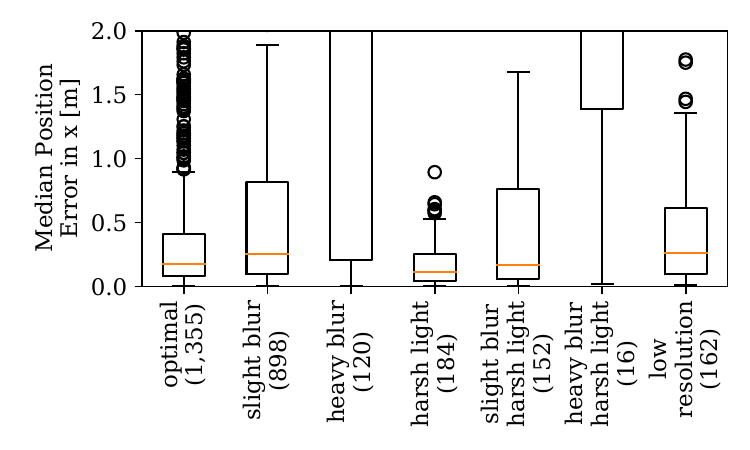}
    \caption{Position error of SfM for various environmental challenges. Number in brackets are the number of samples in the test set.}
    \label{figure_x_condition_error}
\end{figure}

\begin{figure*}[!t]
    \centering
	\begin{minipage}[t]{0.195\linewidth}
        \centering
    	\includegraphics[trim=61 38 48 46, clip, width=1.0\linewidth]{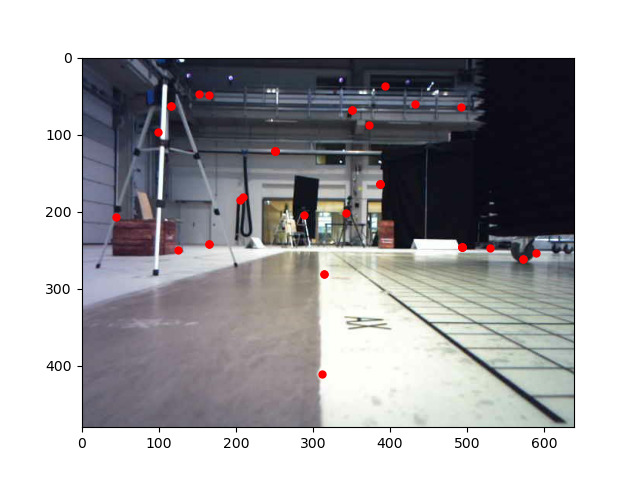}
    	\subcaption{Example 1.}
    	\label{figure_exem_images1}
    \end{minipage}
    \hfill
	\begin{minipage}[t]{0.195\linewidth}
        \centering
    	\includegraphics[trim=61 38 48 46, clip, width=1.0\linewidth]{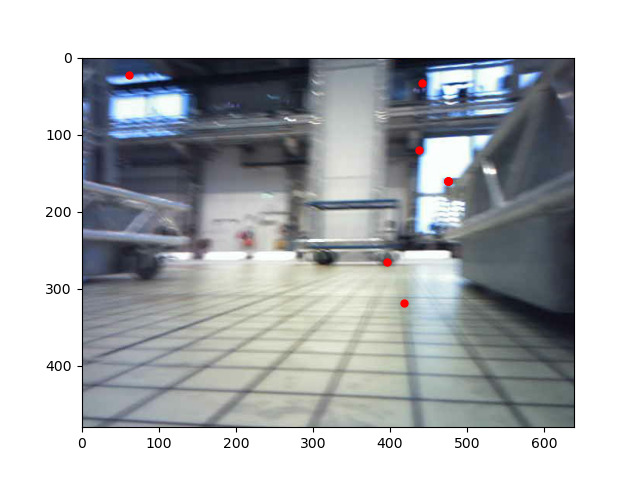}
    	\subcaption{Example 2.}
    	\label{figure_exem_images2}
    \end{minipage}
    \hfill
	\begin{minipage}[t]{0.195\linewidth}
        \centering
    	\includegraphics[trim=61 38 48 46, clip, width=1.0\linewidth]{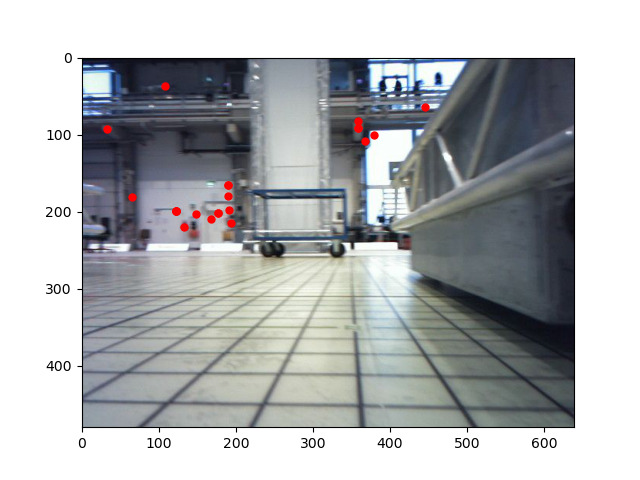}
    	\subcaption{Example 3.}
    	\label{figure_exem_images3}
    \end{minipage}
    \hfill
	\begin{minipage}[t]{0.195\linewidth}
        \centering
    	\includegraphics[trim=61 38 48 46, clip, width=1.0\linewidth]{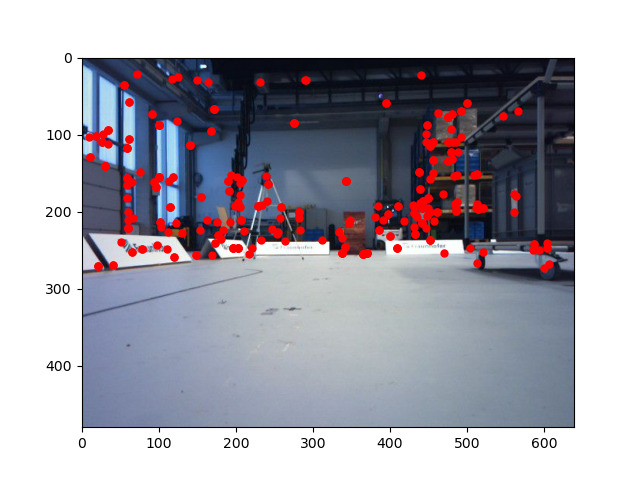}
    	\subcaption{Example 4.}
    	\label{figure_exem_images4}
    \end{minipage}
    \hfill
	\begin{minipage}[t]{0.195\linewidth}
        \centering
    	\includegraphics[trim=61 38 48 46, clip, width=1.0\linewidth]{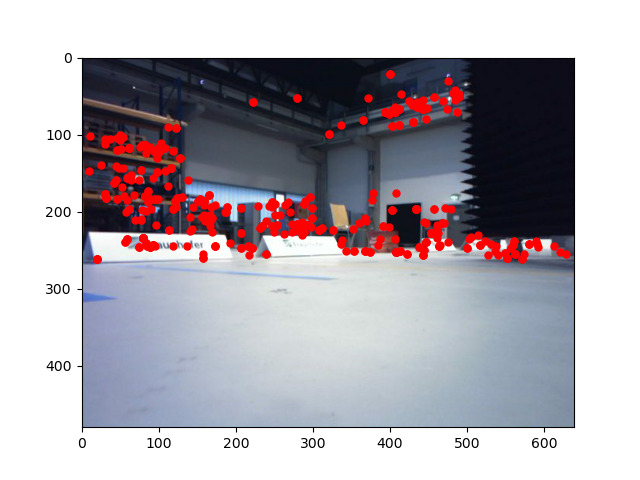}
    	\subcaption{Example 5.}
    	\label{figure_exem_images5}
    \end{minipage}
    \caption{Matches in red for five exemplary images.}
    \label{figure_exem_images}
\end{figure*}

\begin{figure*}[!t]
    \centering
	\begin{minipage}[t]{0.245\linewidth}
        \centering
    	\includegraphics[trim=10 9 10 10, clip, width=1.0\linewidth]{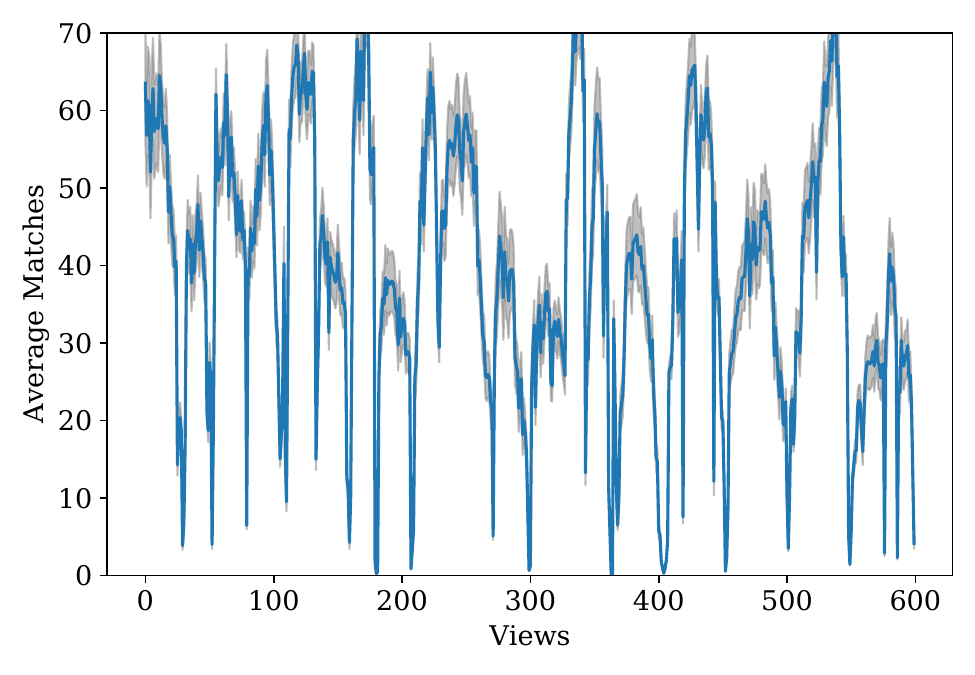}
    	\subcaption{Train 1.}
    	\label{figure_numb_matches1}
    \end{minipage}
    \hfill
	\begin{minipage}[t]{0.245\linewidth}
        \centering
    	\includegraphics[trim=10 9 10 10, clip, width=1.0\linewidth]{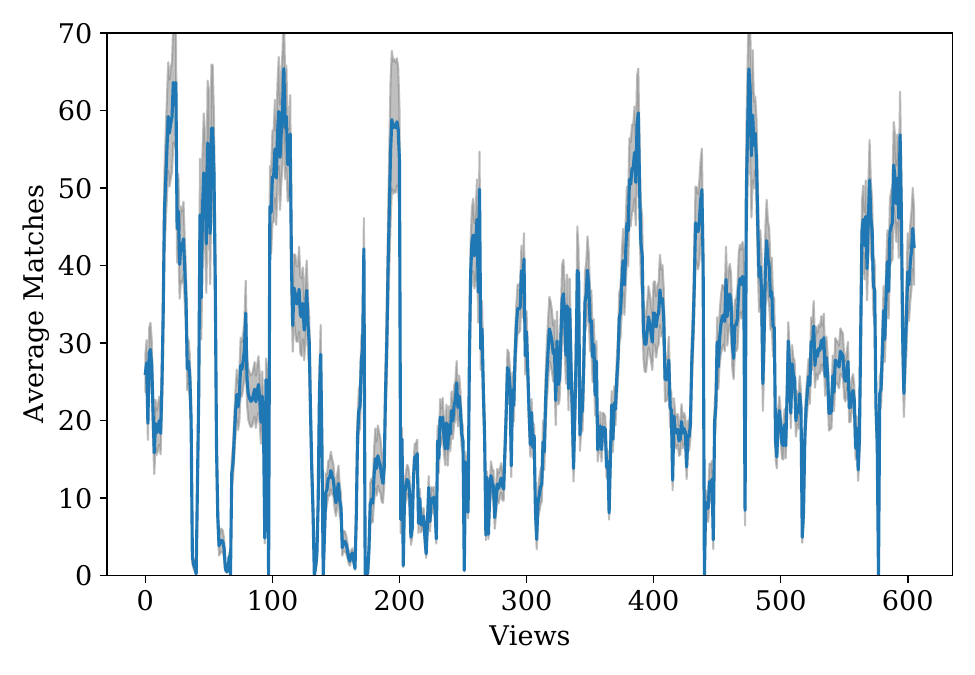}
    	\subcaption{Train 2.}
    	\label{figure_numb_matches2}
    \end{minipage}
    \hfill
	\begin{minipage}[t]{0.245\linewidth}
        \centering
    	\includegraphics[trim=10 9 10 10, clip, width=1.0\linewidth]{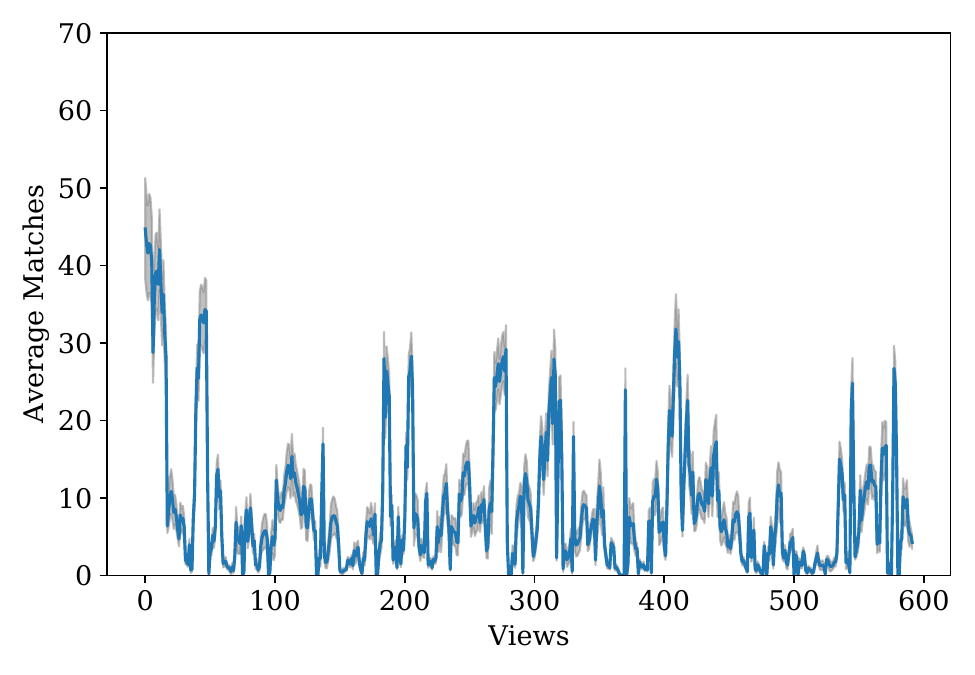}
    	\subcaption{Train 3.}
    	\label{figure_numb_matches3}
    \end{minipage}
    \hfill
	\begin{minipage}[t]{0.245\linewidth}
        \centering
    	\includegraphics[trim=10 9 10 10, clip, width=1.0\linewidth]{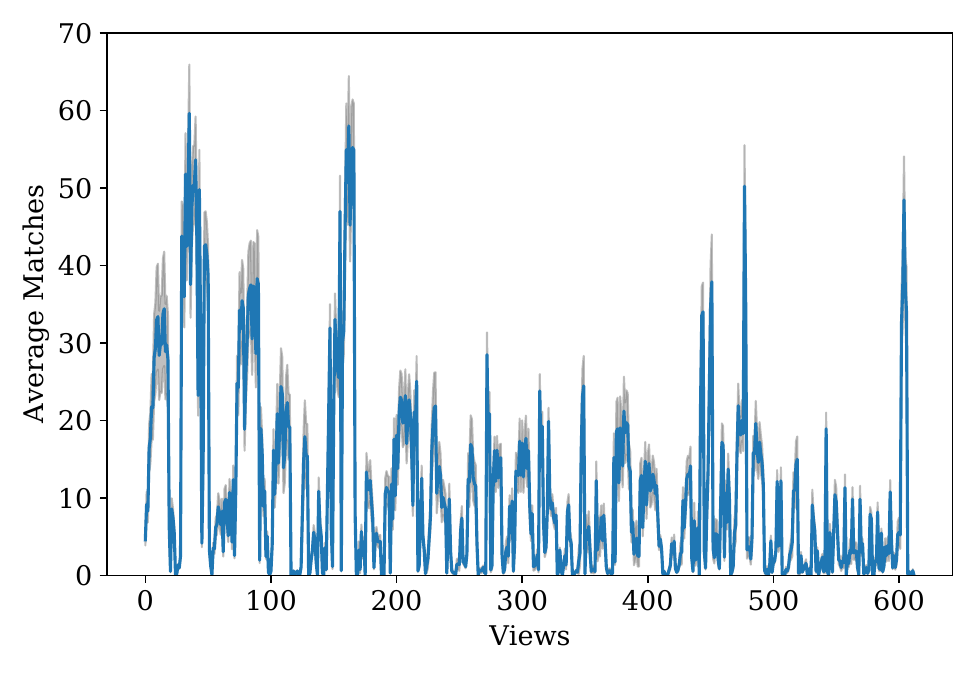}
    	\subcaption{Train 4.}
    	\label{figure_numb_matches4}
    \end{minipage}
	\begin{minipage}[t]{0.245\linewidth}
        \centering
    	\includegraphics[trim=10 9 10 10, clip, width=1.0\linewidth]{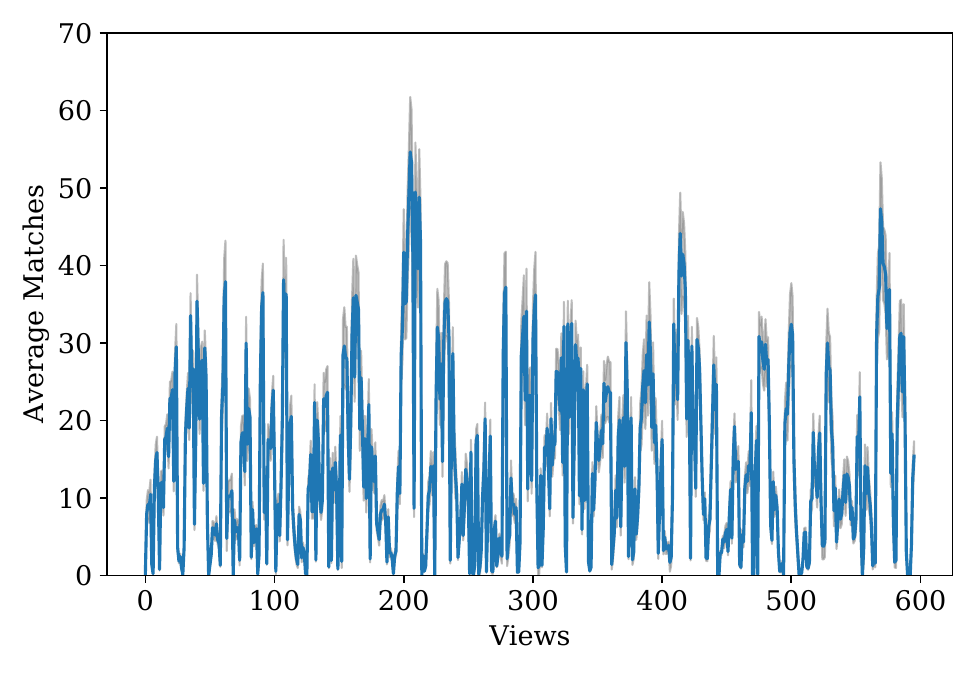}
    	\subcaption{Train 5.}
    	\label{figure_numb_matches5}
    \end{minipage}
    \hfill
	\begin{minipage}[t]{0.245\linewidth}
        \centering
    	\includegraphics[trim=10 9 10 10, clip, width=1.0\linewidth]{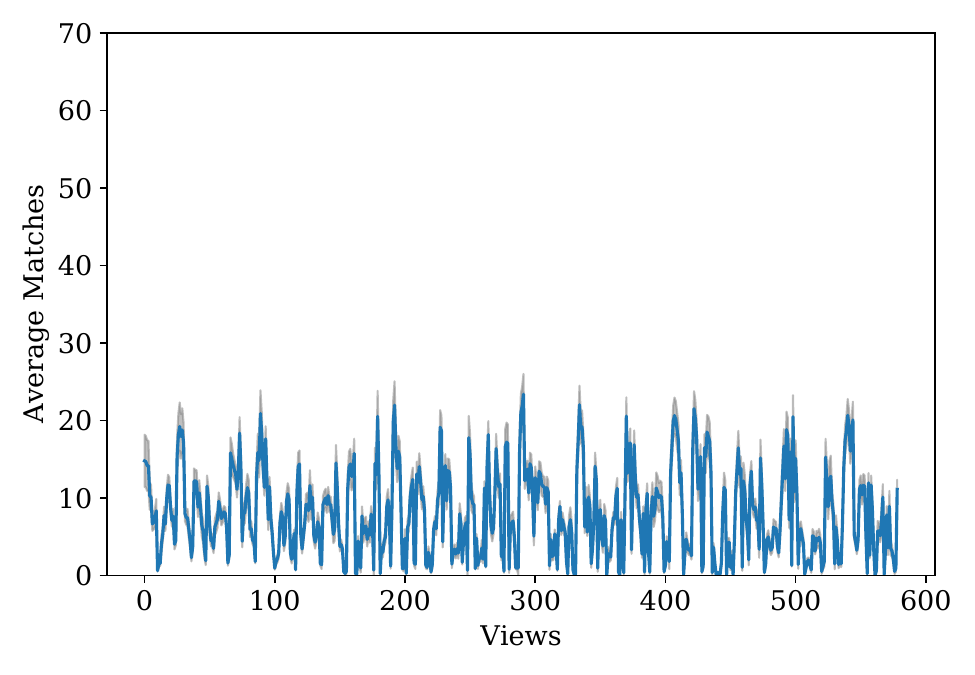}
    	\subcaption{Train 6.}
    	\label{figure_numb_matches6}
    \end{minipage}
    \hfill
	\begin{minipage}[t]{0.245\linewidth}
        \centering
    	\includegraphics[trim=10 9 10 10, clip, width=1.0\linewidth]{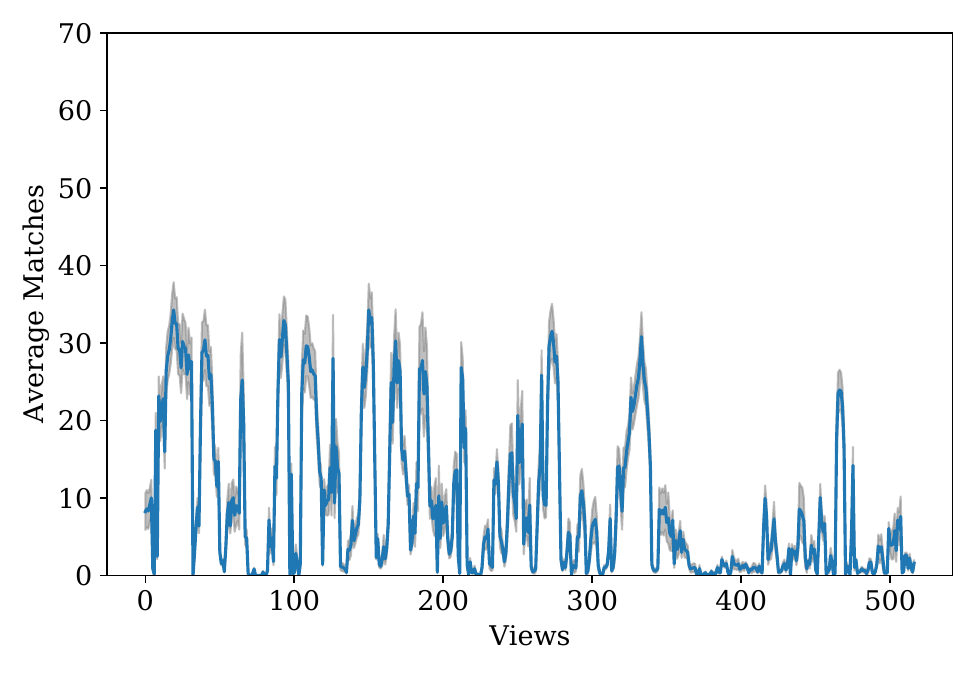}
    	\subcaption{Train 7.}
    	\label{figure_numb_matches7}
    \end{minipage}
    \hfill
	\begin{minipage}[t]{0.245\linewidth}
        \centering
    	\includegraphics[trim=10 9 10 10, clip, width=1.0\linewidth]{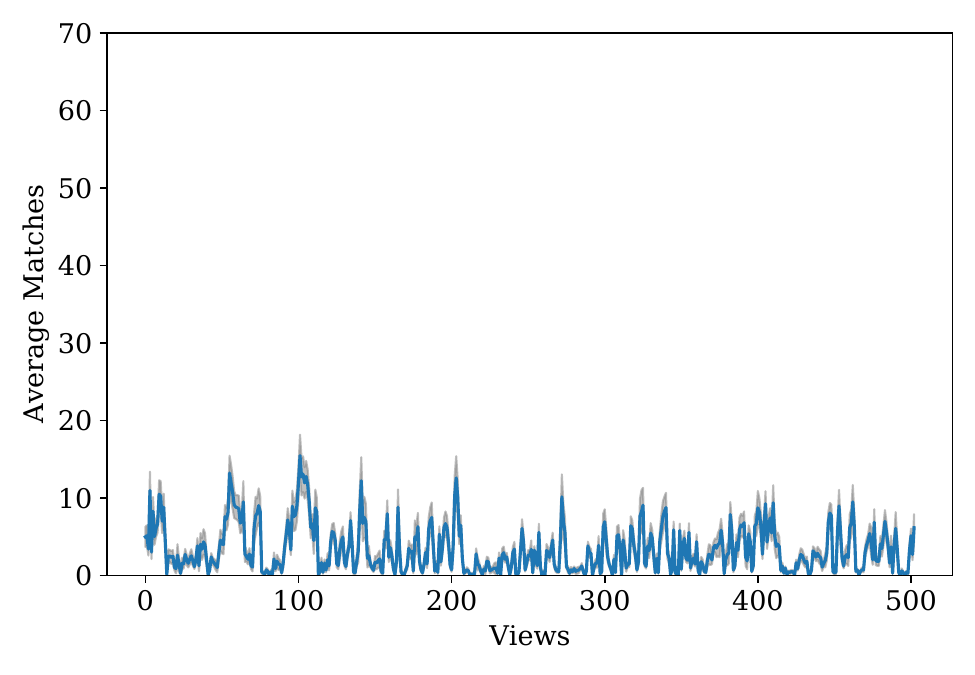}
    	\subcaption{Train 8.}
    	\label{figure_numb_matches8}
    \end{minipage}
    \caption{Average number of matches (y-axis) per view (x-axis) for all eight training datasets. Grey indicates standard deviation over all images.}
    \label{figure_numb_matches}
\end{figure*}

In this section, we aim to investigate the resilience of SfM against environmental variations and challenges. Figure~\ref{figure_x_condition_error} shows the position error for various environmental conditions. Figure~\ref{figure_exem_images} illustrates some sample images along with their corresponding matched pixels displayed as red points. One of the primary obstacles is coping with the noise in the input images caused by several factors such as lighting conditions, motion blur, and camera distortion. As depicted in Figure~\ref{figure_exem_images2}, the presence of motion blur in the image due to the rapid rotation of the robot results in only a few pixels being matched with other images. Conversely, in Figure~\ref{figure_exem_images3}, a similar background context without motion blur allows for feature extraction from the surroundings, leading to successful matching. The correspondence between the position error in the x (top) and y (bottom) coordinates with respect to the orientation of the camera in the environment, as well as challenges present in the images, is visualized in Figure~\ref{figure_x_y_challenges}. The x and y-errors exhibit similar behavior. The results indicate that the position error is low for the orientation of 180\textdegree, as the feature-rich warehouse racks are visible. However, the position error significantly increases for images that point to the right of the environment (0\textdegree\,and 360\textdegree). This is especially evident in the top figure of Figure~\ref{figure_x_y_challenges} for the timesteps 200 to 300 and 750 to 1,100, as many absorber walls and reflective surfaces are visible. Selected images recorded under optimal conditions (green) result in a low position error (e.g., at the timesteps 0 to 150 and 1,250 to 1,350). SfM is found to be robust against slight motion blur, as evident from timestep 2,600 to 2,750. The images in Figure~\ref{figure_exem_images4} and Figure~\ref{figure_exem_images5} contain rich information, leading to many pixels being matched. However, SIFT struggles to extract features from the black absorber walls present in Figure~\ref{figure_exem_images1} and Figure~\ref{figure_exem_images5}. Additionally, the images were generated using a fish-eye lens in the simulated environment, making it challenging to undistort them to construct an effective point cloud. Moreover, SfM techniques can encounter difficulties when dealing with large-scale scenes. In such scenarios, the number of features and images can become too substantial for the algorithms to handle efficiently \citep{schoenberger_frahm}. As a results, we were unable to construct point clouds from more than 1,000 images due to computation times on our hardware setup. Thus, we opted for 600 images for each dataset. One further challenge in SfM is dealing with moving objects in the scene. While our scenarios are static, there are moving objects present between the scenarios. Figure~\ref{figure_numb_matches} illustrates the average number of matches per image view, which affects the point cloud density. A larger number of matches per view, up to 70 matches (as in Figure~\ref{figure_numb_matches1}), results in a denser point cloud (as in Figure~\ref{figure_sfm_pcd1}), whereas for the train 8 dataset, the point could is sparse (see Figure~\ref{figure_sfm_pcd8}) due to numerous objects and absorber walls and a smaller number of matches per view, up to 20, as in Figure~\ref{figure_numb_matches8}. We observe that the number of matches varies significantly across different scenarios, indicating diverse environmental conditions in different areas.

\begin{figure*}[!t]
    \centering
	\begin{minipage}[t]{0.325\linewidth}
        \centering
    	\includegraphics[trim=10 10 10 9, clip, width=1.0\linewidth]{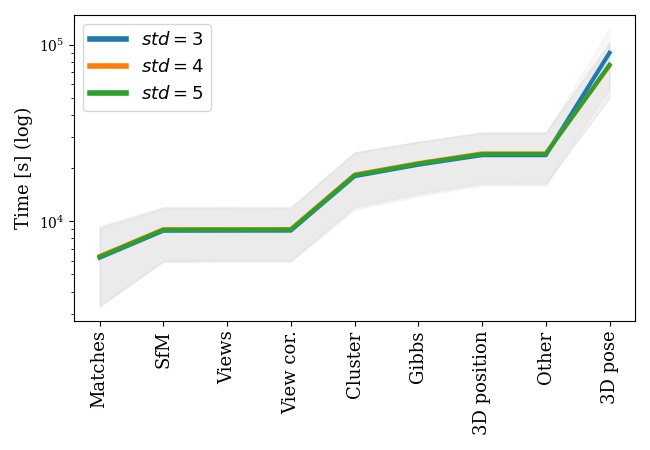}
    	\subcaption{Standard deviation limits.}
    	\label{figure_times1}
    \end{minipage}
    \hfill
	\begin{minipage}[t]{0.325\linewidth}
        \centering
    	\includegraphics[trim=10 10 10 9, clip, width=1.0\linewidth]{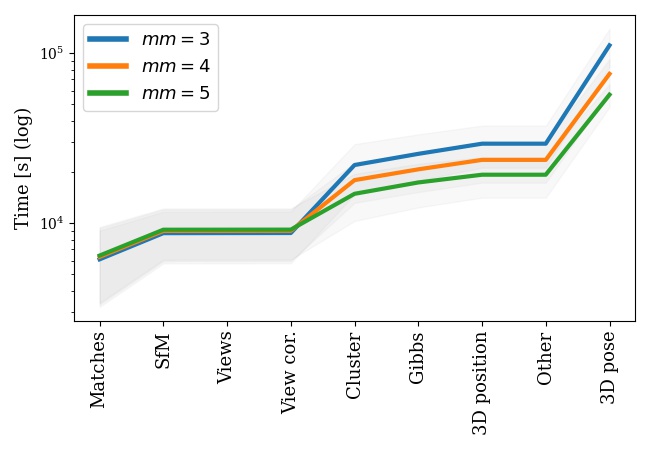}
    	\subcaption{Minimum matches.}
    	\label{figure_times2}
    \end{minipage}
    \hfill
	\begin{minipage}[t]{0.325\linewidth}
        \centering
    	\includegraphics[trim=10 10 10 9, clip, width=1.0\linewidth]{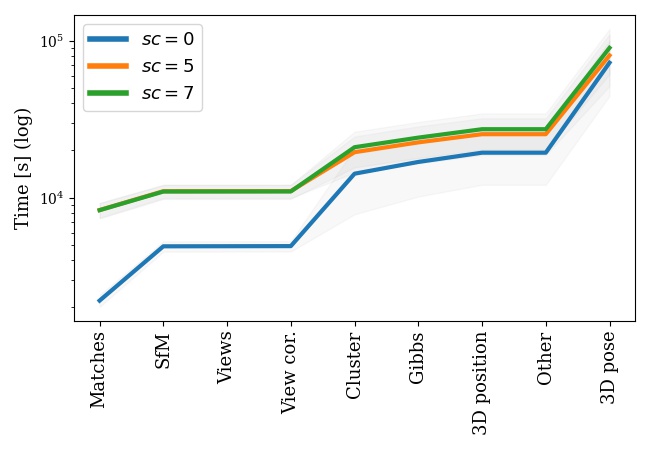}
    	\subcaption{Spatial consistency.}
    	\label{figure_times3}
    \end{minipage}
	\begin{minipage}[t]{0.325\linewidth}
        \centering
    	\includegraphics[trim=10 10 10 9, clip, width=1.0\linewidth]{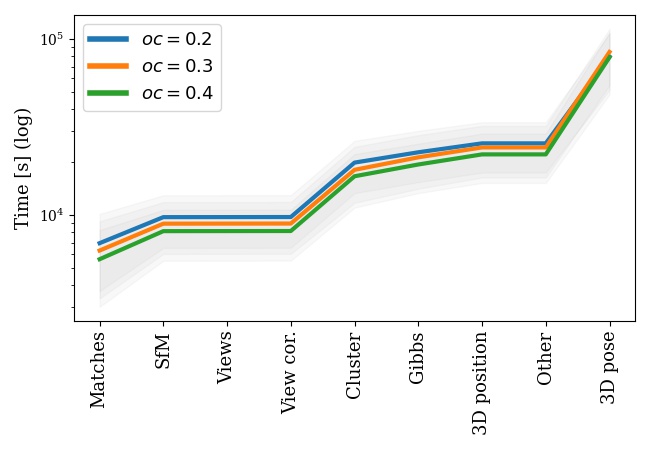}
    	\subcaption{Overlap criterion.}
    	\label{figure_times4}
    \end{minipage}
    \hfill
	\begin{minipage}[t]{0.325\linewidth}
        \centering
    	\includegraphics[trim=10 10 10 9, clip, width=1.0\linewidth]{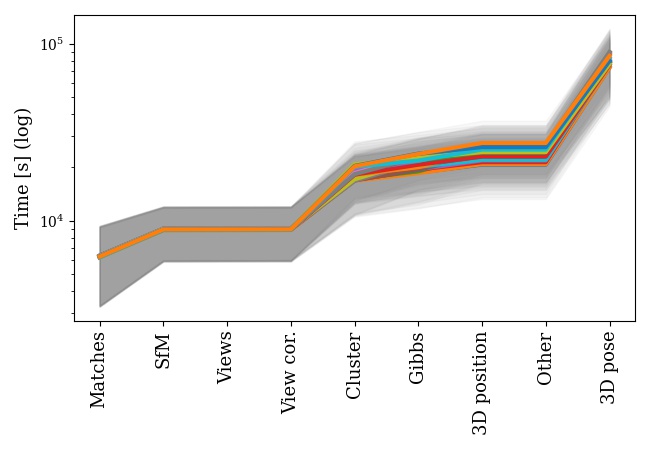}
    	\subcaption{Separation limits ($ex$).}
    	\label{figure_times5}
    \end{minipage}
    \hfill
	\begin{minipage}[t]{0.325\linewidth}
        \centering
    	\includegraphics[trim=10 10 10 9, clip, width=1.0\linewidth]{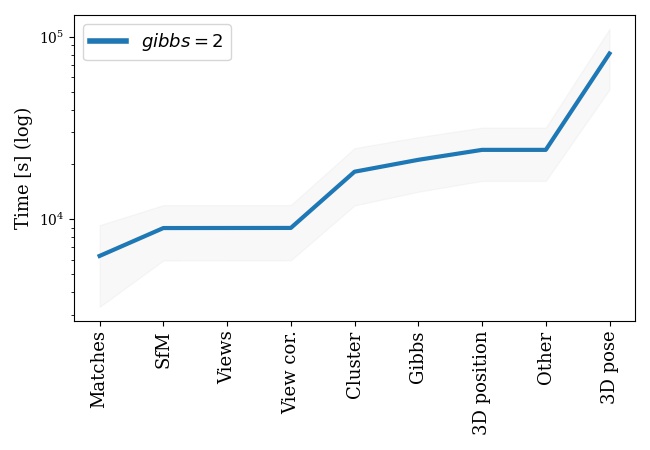}
    	\subcaption{Gibbs sampling.}
    	\label{figure_times6}
    \end{minipage}
    \caption{Overview of computation times (logarithmic in $s$) for different hyperparameter choices.}
    \label{figure_times}
\end{figure*}

In our analysis of the predicted trajectories (refer to the appendix, Figure~\ref{figure_2dtraj1} to Figure~\ref{figure_2dtraj8}), we observed an increase in error in the reconstructed position when constructing a point cloud from a clear environment (train 1) and evaluating it in environments with absorber walls. This trend is noticeable in the top region of test 2, which contains one absorber wall (see Figure~\ref{figure_2dtraj1_2}), and the lower middle region of test 6, which contains four absorber walls (see Figure~\ref{figure_2dtraj1_6}). This pattern is also evident in the other datasets we evaluated.

\textbf{Bundle Adjustment.} The verification of BA's effectiveness entails conducting experiments to evaluate its performance in 3D point reconstruction. Therefore, we assessed the performance of BA by varying the hyperparameter \textit{std} during parameter search. The \textit{std} parameter primarily influences the performance in handling outlier results. Thus, the selection of an appropriate parameter for BA is crucial, although BA exhibits satisfactory performance within the SfM framework. To further validate our findings, we compared the performance of our utilized SfM method with that of the ACE method, demonstrating that both approaches yield satisfactory reconstructions of 3D point clouds. Additionally, we conducted a reprojection error analysis by computing the reprojection error for each observed feature point before and after applying BA. This error metric quantifies the disparity between the observed feature points in the input images and their corresponding projections onto the reconstructed 3D scene. We observed a substantial reduction in reprojection error following the application of BA, indicating an enhanced alignment between the observed feature points and their respective projections.

\subsection{Computation Times}
\label{sec_computation_times}

Figure~\ref{figure_times} depicts the computation times (in $s$) of Structure from Motion (SfM) for point cloud reconstruction across various parameters: $std$, $mm$, $sc$, $oc$, $ex$, and $gibbs$. We highlight each parameter, averaging over the remaining parameters, and present the standard deviation in grey. The time is shown on a logarithmic scale. As observed in Figure~\ref{figure_times1}, the standard deviation ($std$) for point exclusion in Bundle Adjustment (BA) minimally affects computation time. The number of minimum matches ($mm \in {3, 4, 5}$) impacts the time for cluster separation, as demonstrated in Figure~\ref{figure_times2}. Higher spatial consistency ($sc$) results in increased time for match computation and view correction, as depicted in Figure~\ref{figure_times3}, due to the utilization of k-means. The overlap criterion ($oc$) in Figure~\ref{figure_times4} and separation limits ($ex$) in Figure~\ref{figure_times5} have a marginal impact on training time. Generally, cluster separation and 3D pose estimation from BA are the most time-consuming steps. SfM offers the advantage of being executable on a small CPU.

\section{Conclusion}
\label{sec_conclusion}

The application field of the proposed method extends beyond real-time self-localization of objects, such as robots and humans, within small-scale and large-scale environments. It also encompasses various other fields such as autonomous vehicles, where precise localization is crucial for navigation and obstacle avoidance. Additionally, the method can be applied in augmented reality and virtual reality systems, enabling accurate tracking of users' movements and interactions with virtual objects. Furthermore, in the context of smart cities, the method can support location-based services, urban planning, and traffic management by providing real-time localization of vehicles, pedestrians, and infrastructure elements.

The primary focus of our study was to integrate absolute poses obtained from structure from motion (SfM) or an absolute pose regression (APR) method with relative poses, aiming to optimize and smooth the trajectory of objects such as robots and humans. For estimating the relative pose, we introduced a recurrent network that learns translation and rotation by analyzing optical flow between consecutive images computed using the Lucas-Kanade algorithm. Our key contribution was a fusion framework employing eight different recurrent neural networks to combine absolute and relative poses. We compared our approach to the state-of-the-art pose graph optimization (PGO) technique for pose refinement. Additionally, we introduced a large visual database recorded in a challenging indoor environment resembling warehouse scenarios. Furthermore, we developed a simulation framework for generating synthetic images to pre-train APR and RPR models, facilitating faster training.

After conducting a thorough evaluation on the datasets, we have identified several key findings: (1) The quality of the point cloud generated by structure from motion (SfM) is heavily influenced by the chosen hyperparameters, particularly those related to point removal, impacting both position and orientation errors significantly. (2) Careful parameter selection leads to notably better outcomes compared to pose regression techniques. (3) Both SfM and absolute pose regression (APR) methods encounter challenges in extracting features from difficult images, particularly those lacking distinctive features. However, APR methods demonstrate greater resilience to environmental changes. (4) Pre-training APR and relative pose regression (RPR) models results in a slight improvement in position error, up to 1\%, particularly in environments resembling the simulated scenario. (5) Pose graph optimization (PGO) effectively refines poses, reducing high position errors while also minimizing low position errors. (6) Our framework, incorporating various fusion cells based on convolutional, recurrent, and Transformer models, significantly enhances absolute pose error, enabling smoother trajectories and greater resilience to challenges. This approach consistently surpasses PGO. Implementing a strongly-typed Recurrent Neural Network (TRNN) as a small-scale RNN model has led to an average improvement of 42.67\% in position results over SfM-only results for the robotic datasets.

\section*{Acknowledgments}
This work was supported by the Federal Ministry of Education and Research (BMBF) of Germany by Grant No. 01IS18036A (David R\"ugamer), as well as by the Bavarian Ministry for Economic Affairs, Infrastructure, Transport and Technology through the Center for Analytics-Data-Applications (ADA-Center) within the framework of ``BAYERN DIGITAL II''.

\section*{Declaration of competing interest}
The authors declare that they have no known competing financial interests or personal relationships that could have appeared to influence the work reported in this paper.

\section*{Data statement}
For reproducibility and transparency, we publish our datasets and software on the following website: \url{https://gitlab.cc-asp.fraunhofer.de/ottf/industry_datasets}. The repository includes the eight training and 11 testing datasets, the selected images for structure from motion, the checkerboard images for calibration, the synthetically generated dataset, the point clouds recorded with the NavVis M4 system, and all ground truth poses. Additionally, we publish the source code. The repository is 798 GB in size in total.

\printcredits

\bibliographystyle{JVIS2024}
\bibliography{JVIS2024}
\appendix

\section{Appendix A}
\label{app_appendix_A}

\begin{figure*}[!t]
    \centering
	\begin{minipage}[t]{0.325\linewidth}
        \centering
    	\includegraphics[trim=10 10 10 11, clip, width=1.0\linewidth]{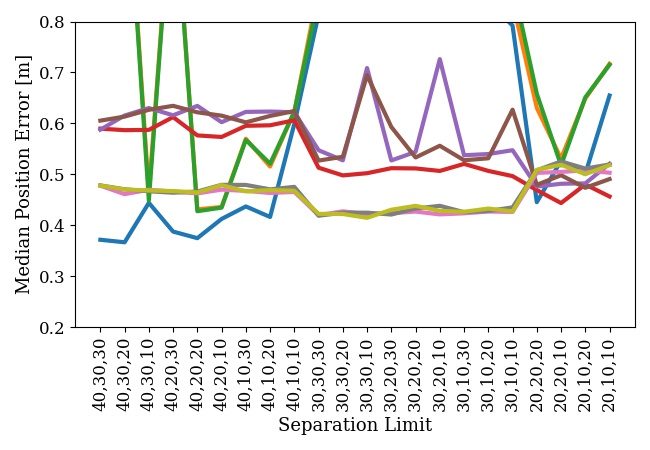}
    	\subcaption{Spatial consistency: 0, overlap criterion: 0.0.}
    	\label{figure_robot_full_t1}
    \end{minipage}
    \hfill
	\begin{minipage}[t]{0.325\linewidth}
        \centering
    	\includegraphics[trim=10 10 10 11, clip, width=1.0\linewidth]{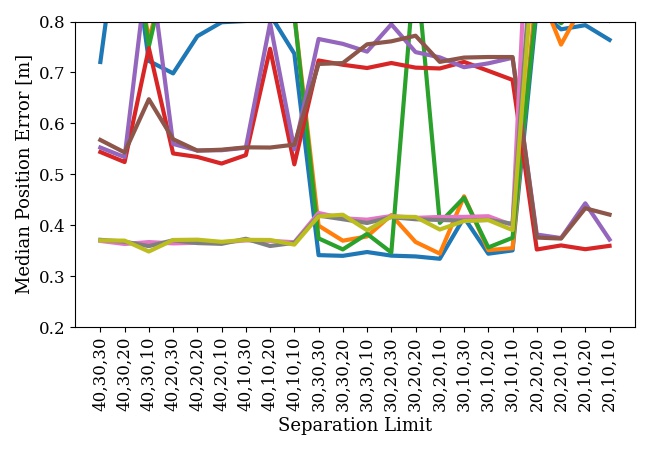}
    	\subcaption{Spatial consistency: 0, overlap criterion: 1.0.}
    	\label{figure_robot_full_t2}
    \end{minipage}
    \hfill
	\begin{minipage}[t]{0.325\linewidth}
        \centering
    	\includegraphics[trim=10 10 10 11, clip, width=1.0\linewidth]{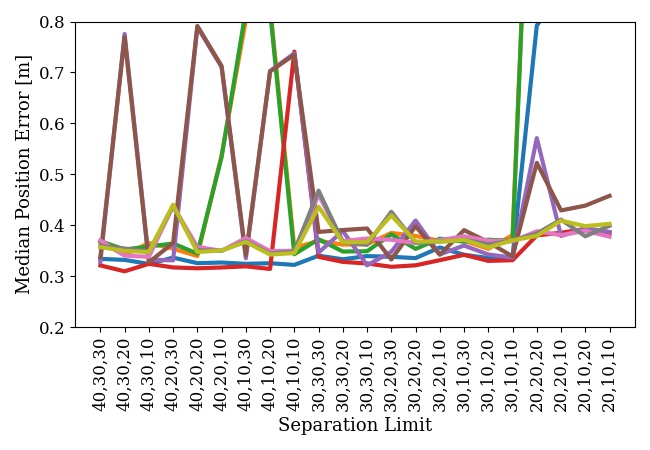}
    	\subcaption{Spatial consistency: 0, overlap criterion: 2.0.}
    	\label{figure_robot_full_t3}
    \end{minipage}
	\begin{minipage}[t]{0.325\linewidth}
        \centering
    	\includegraphics[trim=10 10 10 11, clip, width=1.0\linewidth]{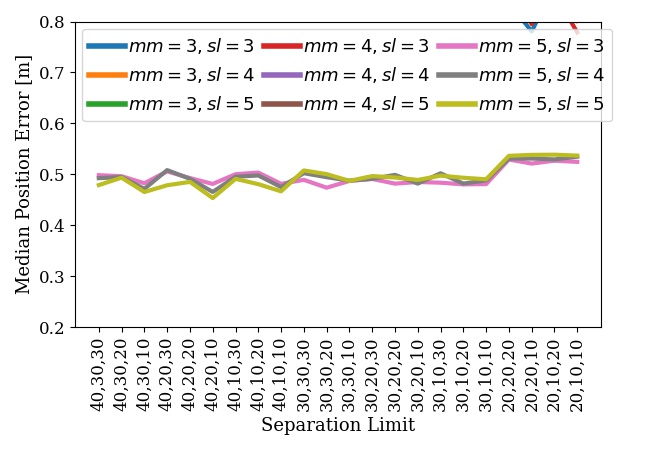}
    	\subcaption{Spatial consistency: 1, overlap criterion: 0.0.}
    	\label{figure_robot_full_t4}
    \end{minipage}
    \hfill
	\begin{minipage}[t]{0.325\linewidth}
        \centering
    	\includegraphics[trim=10 10 10 11, clip, width=1.0\linewidth]{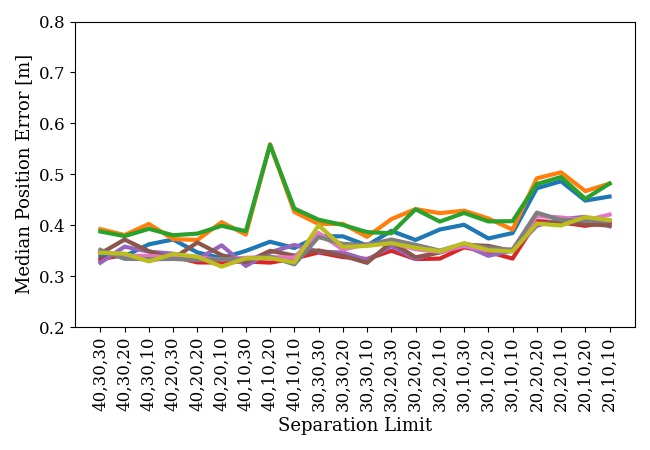}
    	\subcaption{Spatial consistency: 1, overlap criterion: 1.0.}
    	\label{figure_robot_full_t5}
    \end{minipage}
    \hfill
	\begin{minipage}[t]{0.325\linewidth}
        \centering
    	\includegraphics[trim=10 10 10 11, clip, width=1.0\linewidth]{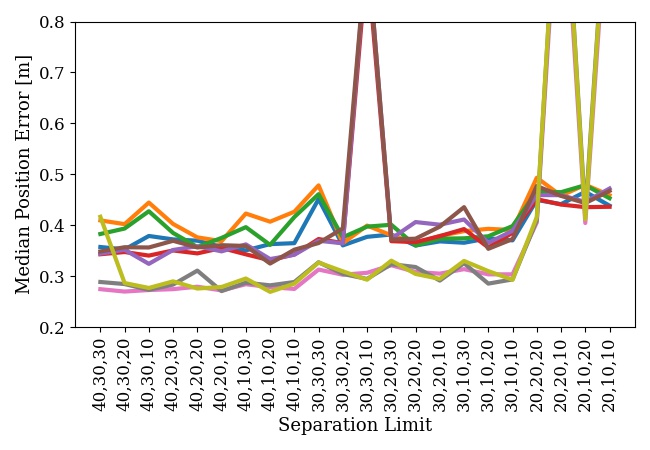}
    	\subcaption{Spatial consistency: 1, overlap criterion: 2.0.}
    	\label{figure_robot_full_t6}
    \end{minipage}
	\begin{minipage}[t]{0.325\linewidth}
        \centering
    	\includegraphics[trim=10 10 10 11, clip, width=1.0\linewidth]{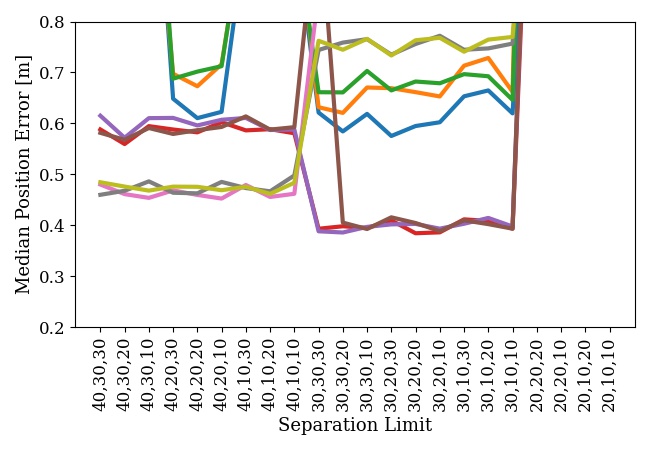}
    	\subcaption{Spatial consistency: 2, overlap criterion: 0.0.}
    	\label{figure_robot_full_t7}
    \end{minipage}
    \hfill
	\begin{minipage}[t]{0.325\linewidth}
        \centering
    	\includegraphics[trim=10 10 10 11, clip, width=1.0\linewidth]{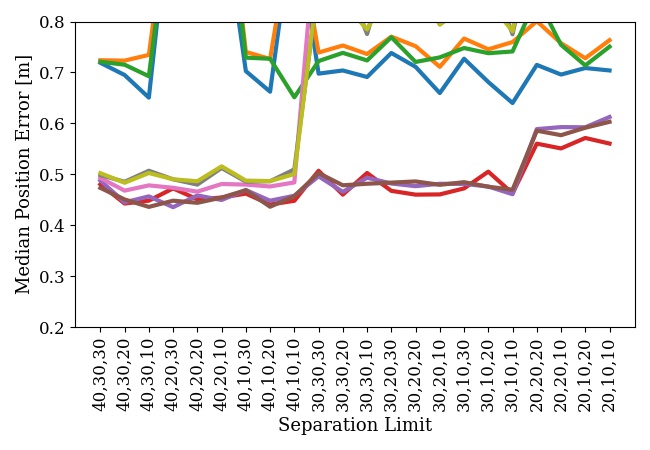}
    	\subcaption{Spatial consistency: 2, overlap criterion: 1.0.}
    	\label{figure_robot_full_t8}
    \end{minipage}
    \hfill
	\begin{minipage}[t]{0.325\linewidth}
        \centering
    	\includegraphics[trim=10 10 10 11, clip, width=1.0\linewidth]{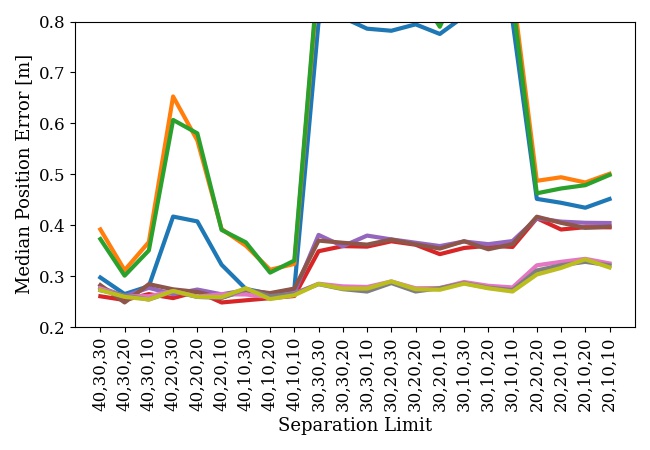}
    	\subcaption{Spatial consistency: 2, overlap criterion: 2.0.}
    	\label{figure_robot_full_t9}
    \end{minipage}
    \caption{Detailed evaluation results for the SfM hyperparameter search for the robot train 3 and test 6 datasets. Median position error in $m$. For readability, the label spacing is fixed. The legend shows the \textit{minimum matches} hyperparameter. The legend is equal for all subplots.}
    \label{figure_robot_full_t}
\end{figure*}

\begin{figure*}[!t]
    \centering
	\begin{minipage}[t]{0.325\linewidth}
        \centering
    	\includegraphics[trim=10 10 10 11, clip, width=1.0\linewidth]{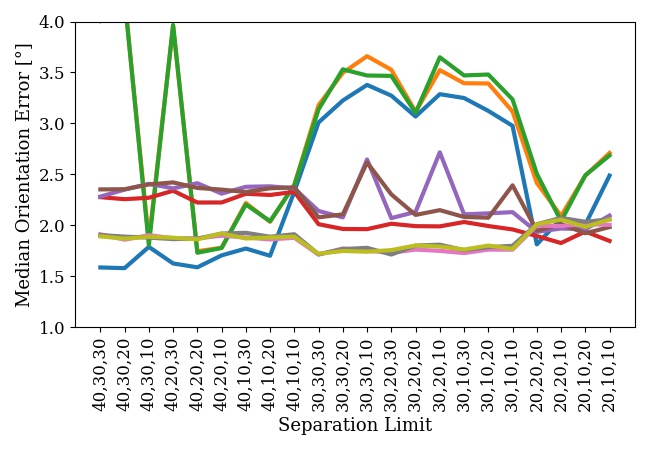}
    	\subcaption{Spatial consistency: 0, overlap criterion: 0.0.}
    	\label{figure_robot_full_r1}
    \end{minipage}
    \hfill
	\begin{minipage}[t]{0.325\linewidth}
        \centering
    	\includegraphics[trim=10 10 10 11, clip, width=1.0\linewidth]{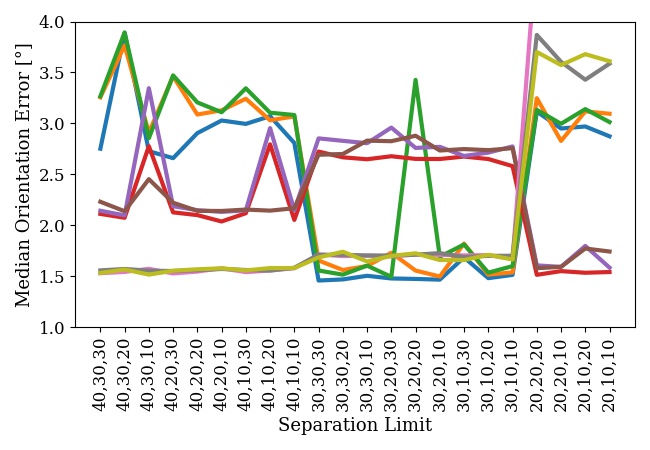}
    	\subcaption{Spatial consistency: 0, overlap criterion: 1.0.}
    	\label{figure_robot_full_r2}
    \end{minipage}
    \hfill
	\begin{minipage}[t]{0.325\linewidth}
        \centering
    	\includegraphics[trim=10 10 10 11, clip, width=1.0\linewidth]{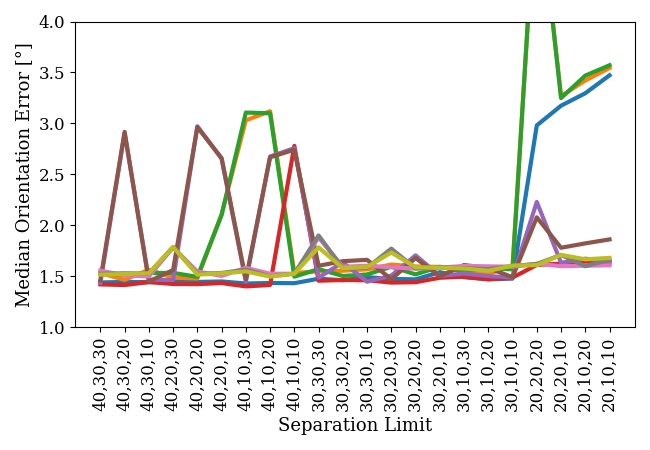}
    	\subcaption{Spatial consistency: 0, overlap criterion: 2.0.}
    	\label{figure_robot_full_r3}
    \end{minipage}
	\begin{minipage}[t]{0.325\linewidth}
        \centering
    	\includegraphics[trim=10 10 10 11, clip, width=1.0\linewidth]{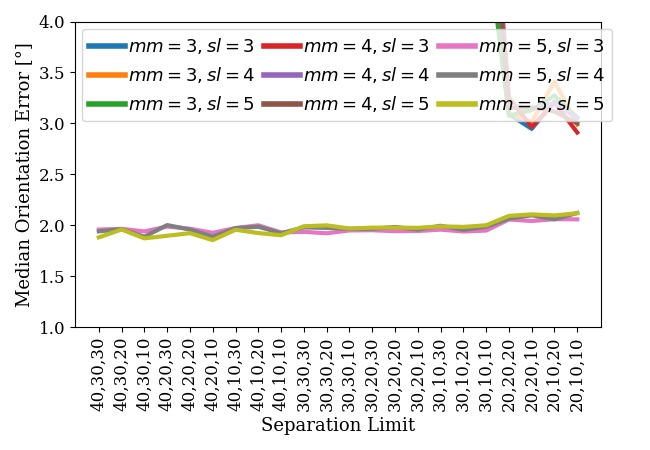}
    	\subcaption{Spatial consistency: 1, overlap criterion: 0.0.}
    	\label{figure_robot_full_r4}
    \end{minipage}
    \hfill
	\begin{minipage}[t]{0.325\linewidth}
        \centering
    	\includegraphics[trim=10 10 10 11, clip, width=1.0\linewidth]{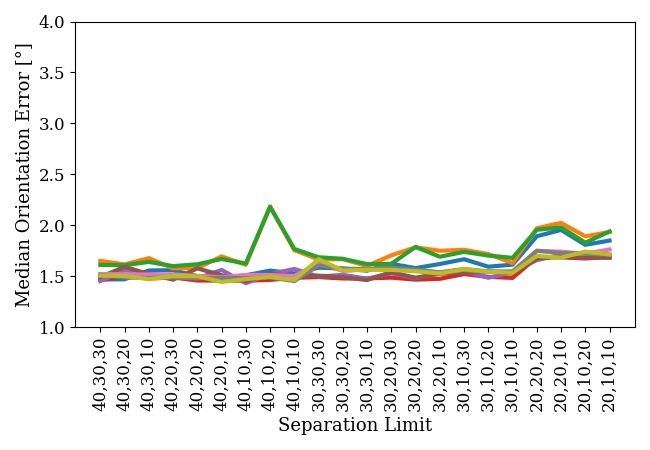}
    	\subcaption{Spatial consistency: 1, overlap criterion: 1.0.}
    	\label{figure_robot_full_r5}
    \end{minipage}
    \hfill
	\begin{minipage}[t]{0.325\linewidth}
        \centering
    	\includegraphics[trim=10 10 10 11, clip, width=1.0\linewidth]{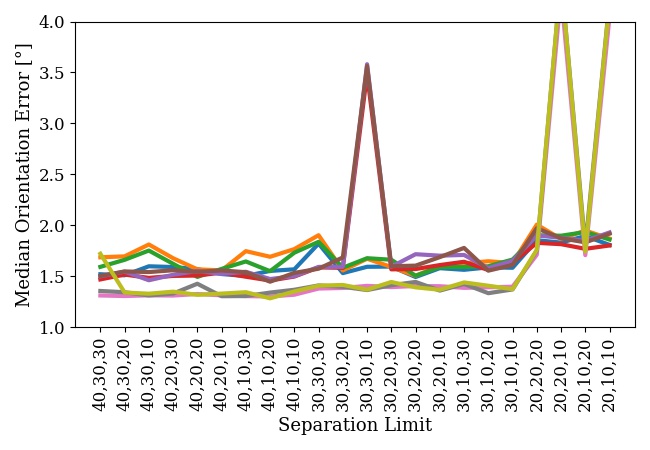}
    	\subcaption{Spatial consistency: 1, overlap criterion: 2.0.}
    	\label{figure_robot_full_r6}
    \end{minipage}
	\begin{minipage}[t]{0.325\linewidth}
        \centering
    	\includegraphics[trim=10 10 10 11, clip, width=1.0\linewidth]{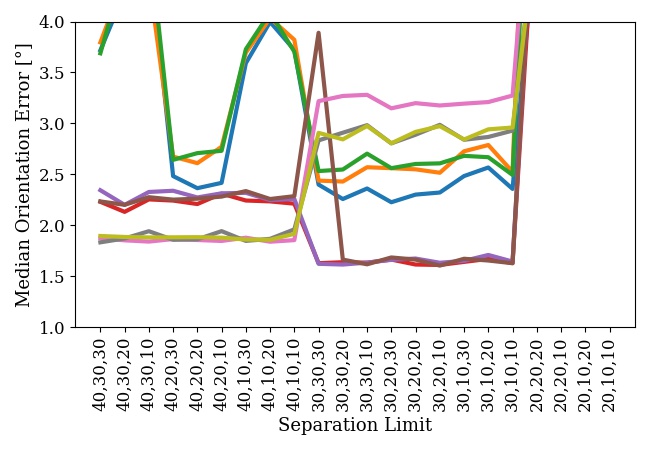}
    	\subcaption{Spatial consistency: 2, overlap criterion: 0.0.}
    	\label{figure_robot_full_r7}
    \end{minipage}
    \hfill
	\begin{minipage}[t]{0.325\linewidth}
        \centering
    	\includegraphics[trim=10 10 10 11, clip, width=1.0\linewidth]{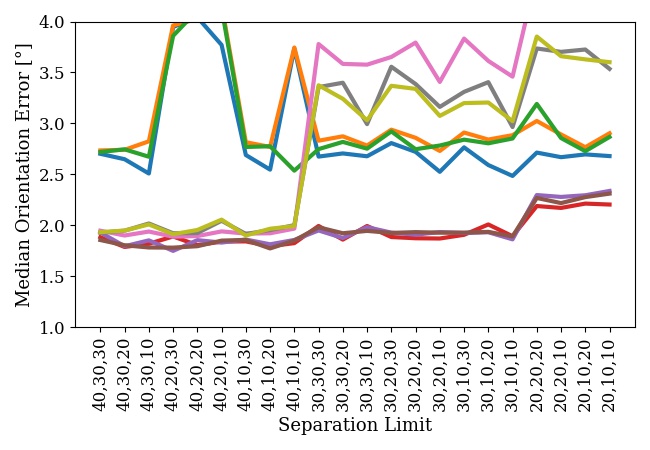}
    	\subcaption{Spatial consistency: 2, overlap criterion: 1.0.}
    	\label{figure_robot_full_r8}
    \end{minipage}
    \hfill
	\begin{minipage}[t]{0.325\linewidth}
        \centering
    	\includegraphics[trim=10 10 10 11, clip, width=1.0\linewidth]{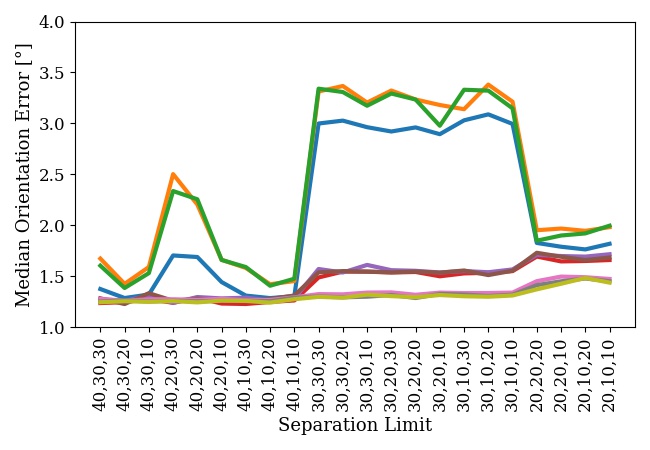}
    	\subcaption{Spatial consistency: 2, overlap criterion: 2.0.}
    	\label{figure_robot_full_r9}
    \end{minipage}
    \caption{Detailed evaluation results for the SfM hyperparameter search for the robot train 3 and test 6 datasets. Median orientation error in \textdegree. For readability, the label spacing is fixed. The legend shows the \textit{minimum matches} hyperparameter. The legend is equal for all subplots.}
    \label{figure_robot_full_r}
\end{figure*}

\subsection{Hyperparameter Search}
\label{app_appendix_search}

\begin{figure*}[!t]
    \centering
	\begin{minipage}[t]{0.325\linewidth}
        \centering
    	\includegraphics[trim=10 10 10 11, clip, width=1.0\linewidth]{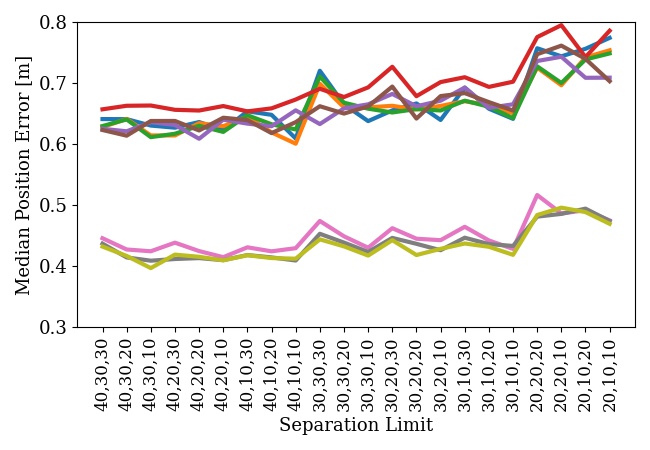}
    	\subcaption{Spatial consistency: 0, overlap criterion: 0.0.}
    	\label{figure_handheld_full_t1}
    \end{minipage}
    \hfill
	\begin{minipage}[t]{0.325\linewidth}
        \centering
    	\includegraphics[trim=10 10 10 11, clip, width=1.0\linewidth]{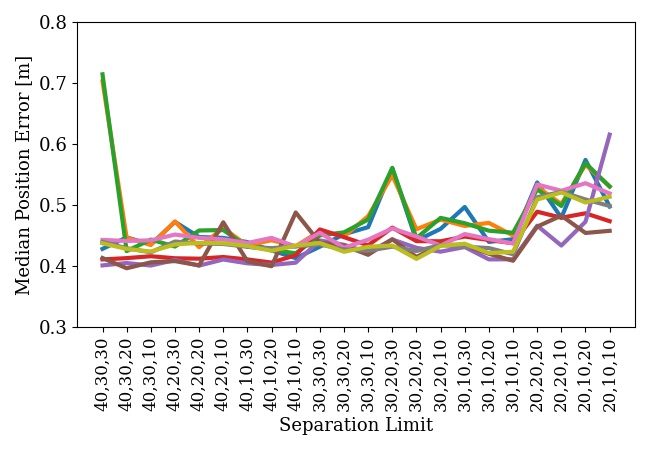}
    	\subcaption{Spatial consistency: 0, overlap criterion: 1.0.}
    	\label{figure_handheld_full_t2}
    \end{minipage}
    \hfill
	\begin{minipage}[t]{0.325\linewidth}
        \centering
    	\includegraphics[trim=10 10 10 11, clip, width=1.0\linewidth]{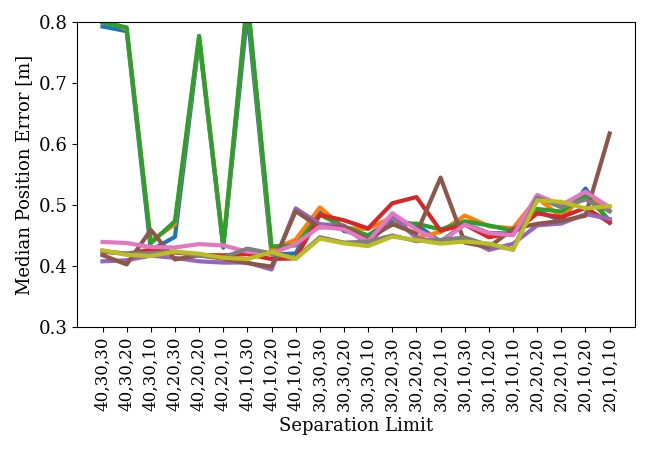}
    	\subcaption{Spatial consistency: 0, overlap criterion: 2.0.}
    	\label{figure_handheld_full_t3}
    \end{minipage}
	\begin{minipage}[t]{0.325\linewidth}
        \centering
    	\includegraphics[trim=10 10 10 11, clip, width=1.0\linewidth]{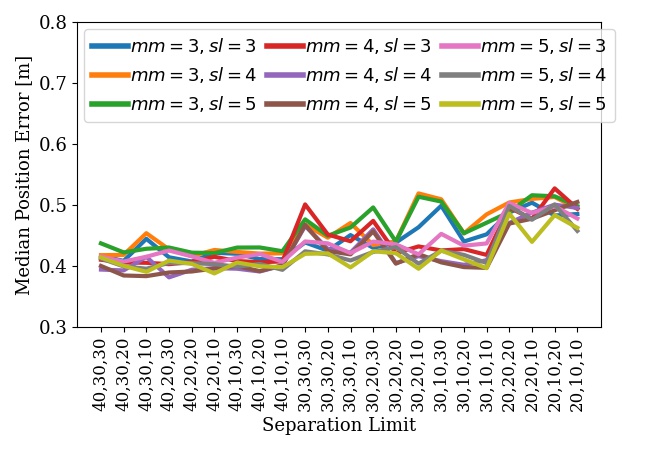}
    	\subcaption{Spatial consistency: 1, overlap criterion: 0.0.}
    	\label{figure_handheld_full_t4}
    \end{minipage}
    \hfill
	\begin{minipage}[t]{0.325\linewidth}
        \centering
    	\includegraphics[trim=10 10 10 11, clip, width=1.0\linewidth]{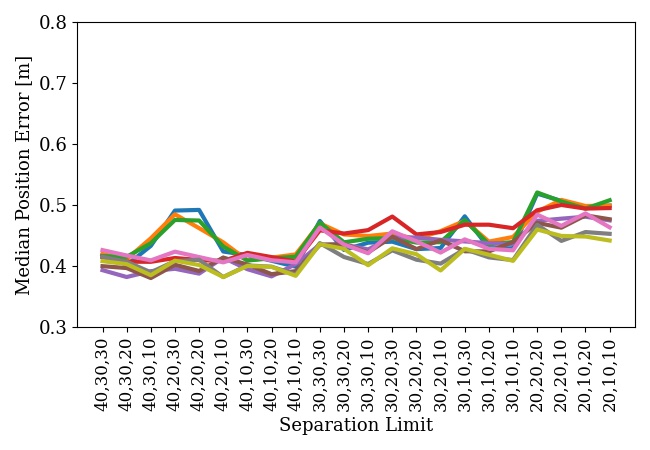}
    	\subcaption{Spatial consistency: 1, overlap criterion: 1.0.}
    	\label{figure_handheld_full_t5}
    \end{minipage}
    \hfill
	\begin{minipage}[t]{0.325\linewidth}
        \centering
    	\includegraphics[trim=10 10 10 11, clip, width=1.0\linewidth]{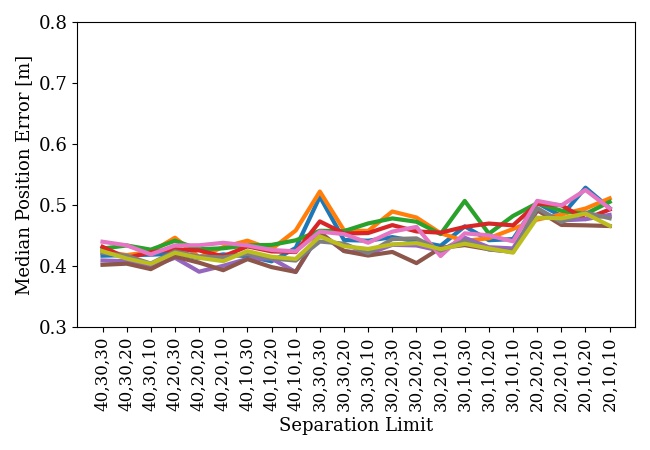}
    	\subcaption{Spatial consistency: 1, overlap criterion: 2.0.}
    	\label{figure_handheld_full_t6}
    \end{minipage}
	\begin{minipage}[t]{0.325\linewidth}
        \centering
    	\includegraphics[trim=10 10 10 11, clip, width=1.0\linewidth]{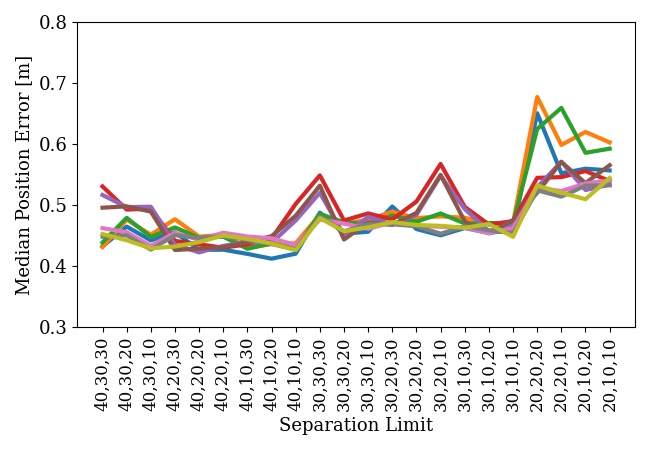}
    	\subcaption{Spatial consistency: 2, overlap criterion: 0.0.}
    	\label{figure_handheld_full_t7}
    \end{minipage}
    \hfill
	\begin{minipage}[t]{0.325\linewidth}
        \centering
    	\includegraphics[trim=10 10 10 11, clip, width=1.0\linewidth]{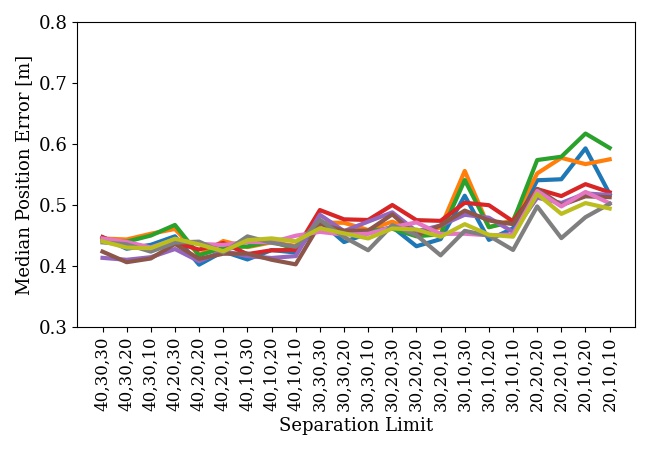}
    	\subcaption{Spatial consistency: 2, overlap criterion: 1.0.}
    	\label{figure_handheld_full_t8}
    \end{minipage}
    \hfill
	\begin{minipage}[t]{0.325\linewidth}
        \centering
    	\includegraphics[trim=10 10 10 11, clip, width=1.0\linewidth]{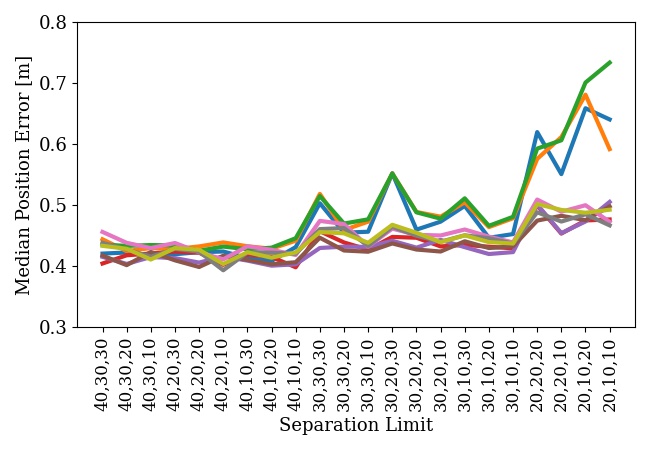}
    	\subcaption{Spatial consistency: 2, overlap criterion: 2.0.}
    	\label{figure_handheld_full_t9}
    \end{minipage}
    \caption{Detailed evaluation results for the SfM hyperparameter search for the handheld train 4 and test 7 datasets. Median position error in $m$. For readability, the label spacing is fixed. The legend shows the \textit{minimum matches} hyperparameter. The legend is equal for all subplots.}
    \label{figure_handheld_full_t}
\end{figure*}

\begin{figure*}[!t]
    \centering
	\begin{minipage}[t]{0.325\linewidth}
        \centering
    	\includegraphics[trim=10 10 10 11, clip, width=1.0\linewidth]{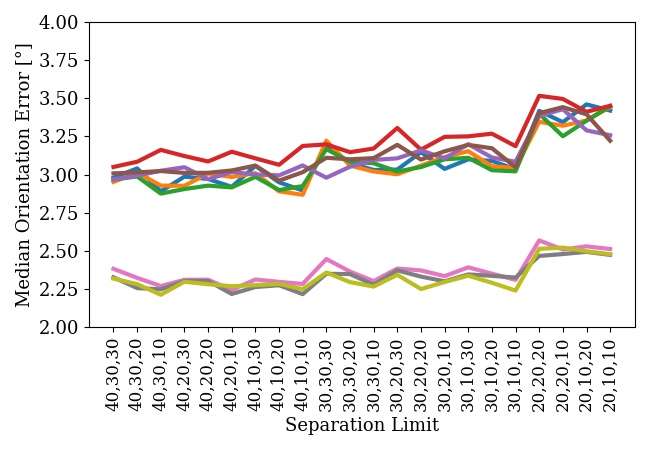}
    	\subcaption{Spatial consistency: 0, overlap criterion: 0.0.}
    	\label{figure_handheld_full_r1}
    \end{minipage}
    \hfill
	\begin{minipage}[t]{0.325\linewidth}
        \centering
    	\includegraphics[trim=10 10 10 11, clip, width=1.0\linewidth]{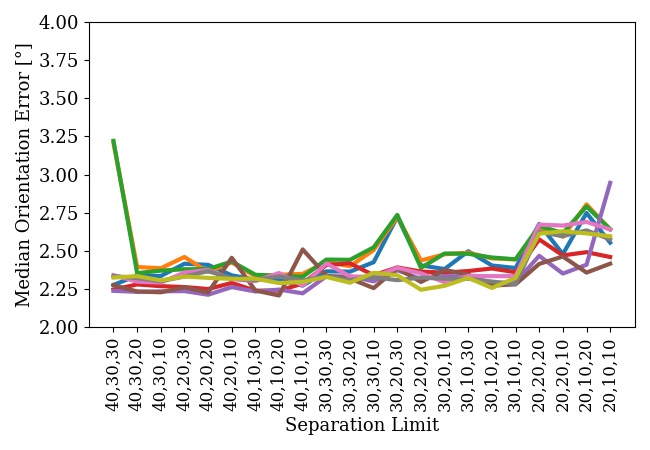}
    	\subcaption{Spatial consistency: 0, overlap criterion: 1.0.}
    	\label{figure_handheld_full_r2}
    \end{minipage}
    \hfill
	\begin{minipage}[t]{0.325\linewidth}
        \centering
    	\includegraphics[trim=10 10 10 11, clip, width=1.0\linewidth]{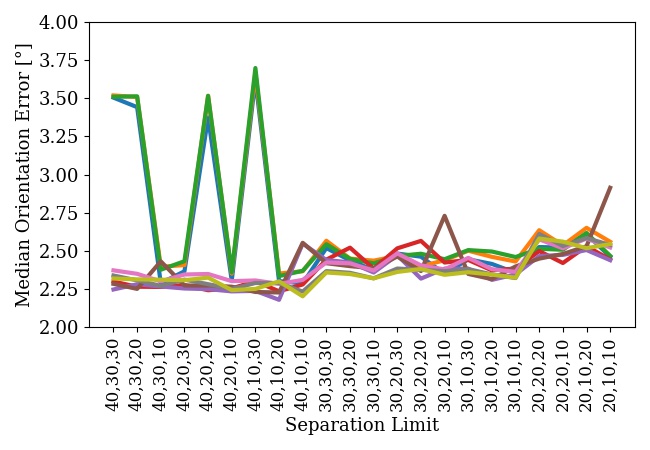}
    	\subcaption{Spatial consistency: 0, overlap criterion: 2.0.}
    	\label{figure_handheld_full_r3}
    \end{minipage}
	\begin{minipage}[t]{0.325\linewidth}
        \centering
    	\includegraphics[trim=10 10 10 11, clip, width=1.0\linewidth]{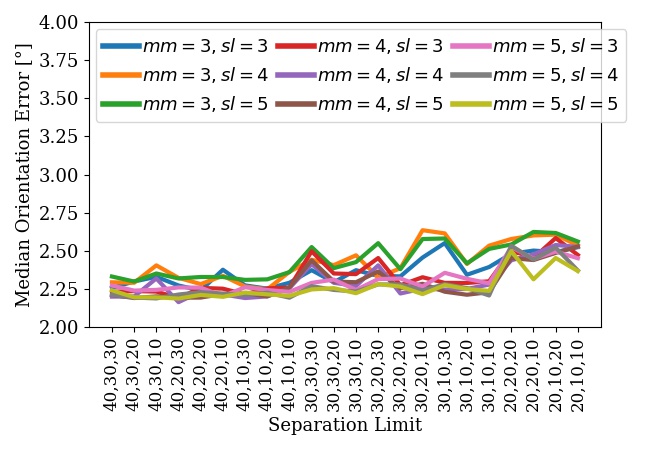}
    	\subcaption{Spatial consistency: 1, overlap criterion: 0.0.}
    	\label{figure_handheld_full_r4}
    \end{minipage}
    \hfill
	\begin{minipage}[t]{0.325\linewidth}
        \centering
    	\includegraphics[trim=10 10 10 11, clip, width=1.0\linewidth]{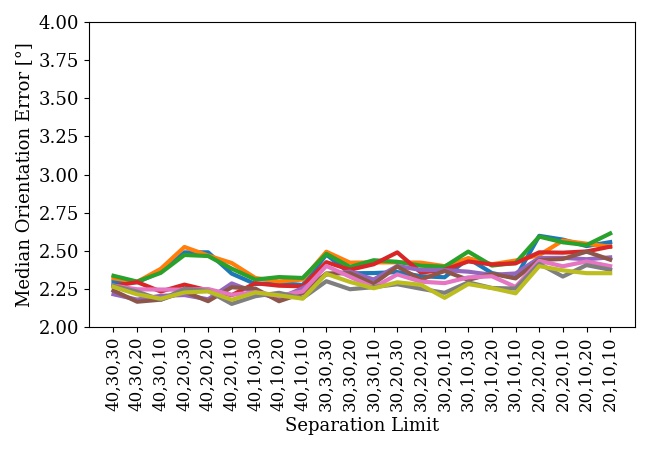}
    	\subcaption{Spatial consistency: 1, overlap criterion: 1.0.}
    	\label{figure_handheld_full_r5}
    \end{minipage}
    \hfill
	\begin{minipage}[t]{0.325\linewidth}
        \centering
    	\includegraphics[trim=10 10 10 11, clip, width=1.0\linewidth]{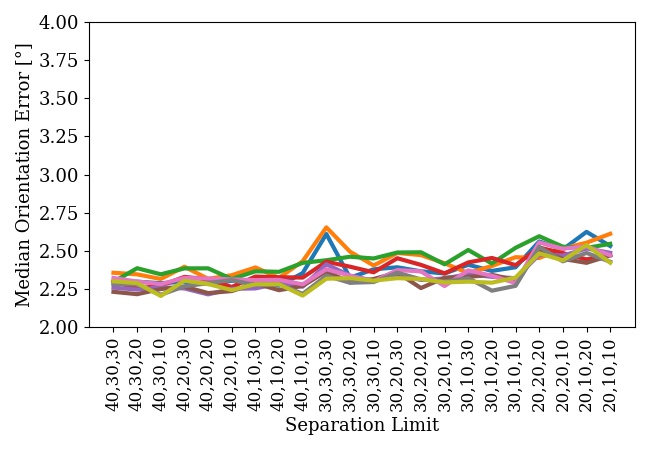}
    	\subcaption{Spatial consistency: 1, overlap criterion: 2.0.}
    	\label{figure_handheld_full_r6}
    \end{minipage}
	\begin{minipage}[t]{0.325\linewidth}
        \centering
    	\includegraphics[trim=10 10 10 11, clip, width=1.0\linewidth]{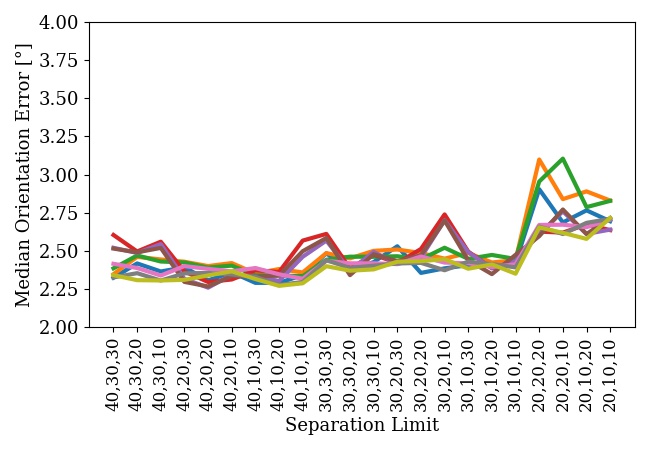}
    	\subcaption{Spatial consistency: 2, overlap criterion: 0.0.}
    	\label{figure_handheld_full_r7}
    \end{minipage}
    \hfill
	\begin{minipage}[t]{0.325\linewidth}
        \centering
    	\includegraphics[trim=10 10 10 11, clip, width=1.0\linewidth]{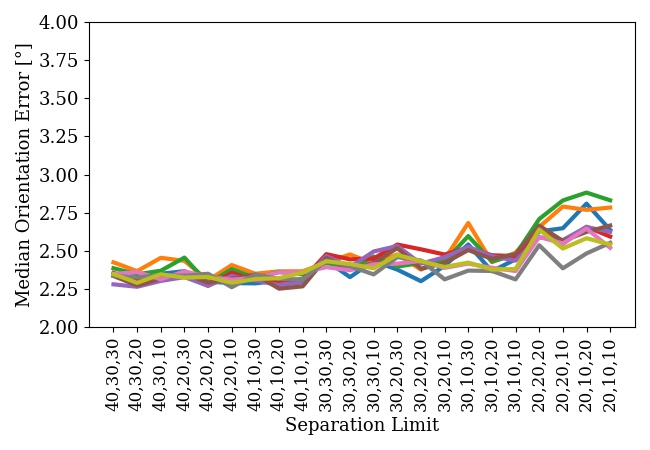}
    	\subcaption{Spatial consistency: 2, overlap criterion: 1.0.}
    	\label{figure_handheld_full_r8}
    \end{minipage}
    \hfill
	\begin{minipage}[t]{0.325\linewidth}
        \centering
    	\includegraphics[trim=10 10 10 11, clip, width=1.0\linewidth]{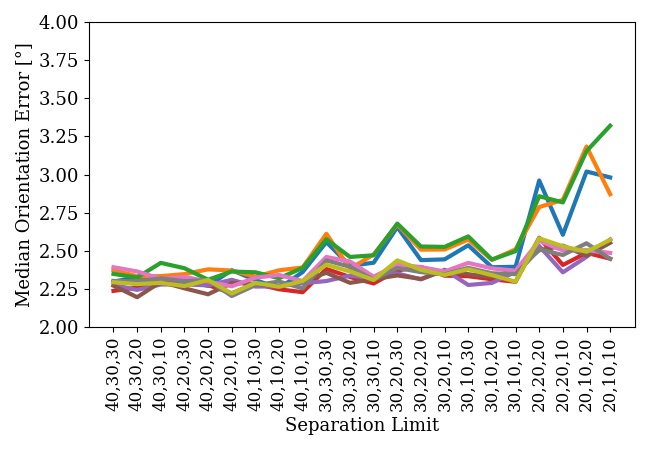}
    	\subcaption{Spatial consistency: 2, overlap criterion: 2.0.}
    	\label{figure_handheld_full_r9}
    \end{minipage}
    \caption{Detailed evaluation results for the SfM hyperparameter search for the handheld train 4 and test 7 datasets. Median orientation error in \textdegree. For readability, the label spacing is fixed. The legend shows the \textit{minimum matches} hyperparameter. The legend is equal for all subplots.}
    \label{figure_handheld_full_r}
\end{figure*}

In Figure~\ref{figure_robot_full_t} and Figure~\ref{figure_robot_full_r}, we conduct a detailed hyperparameter search for the robot dataset, while in Figure~\ref{figure_handheld_full_t} and Figure~\ref{figure_handheld_full_r}, we perform a similar search for the handheld dataset. The x-axis of each figure represents a combination of three hyperparameters ($ex_1, ex_2, ex_3$) that control the separation limit. We search for each of these parameters in the range of $[10, 20, 30, 40]$, subject to the constraints that $ex_1 \geq ex_2$, $ex_1 \geq ex_3$, and $ex_2 \geq ex_3$. Here, $ex_1$ is the cluster separation limit, $ex_2$ is the Gibbs separation parameter, and $ex_3$ is the 3D position adjustment of BA. We observe that certain combinations of these parameters result in high error rates, but higher values of the parameters generally yield lower position and orientation errors. For example, $(ex_1, ex_2, ex_3) = (40, 30, 30)$ produces the best results.

For each combination of hyperparameters, we generate a point cloud using the parameters $mm \in [3, 4, 5]$, $sl = std \in [3, 4, 5]$, the spatial consistency $sc \in [0, 1, 2]$, and the overlap criterion $oc \in [0.0, 0.1, 0.2]$. For the robot dataset, we found that the combination of $sc = 2$ and $oc = 2.0$ yields the best results (position error below 0.3\textit{m} and orientation error below 1.5\textdegree), as depicted in Figure~\ref{figure_robot_full_t9} and Figure~\ref{figure_robot_full_r9}. For the handheld dataset, we recommend the combination of $sc = 1$ and $oc = 1.0$, refer to Figure~\ref{figure_handheld_full_t5} and Figure~\ref{figure_handheld_full_r5}. We also observe that a high value for the minimum matches parameter $mm = 5$ is clearly the best choice, while $std = [3, 4, 5]$ produces similar results.

\subsection{Trajectory Plots}
\label{app_appendix_trajectories}

\begin{figure*}[!t]
    \centering
	\begin{minipage}[t]{0.195\linewidth}
        \centering
    	\includegraphics[trim=61 38 48 46, clip, width=1.0\linewidth]{test_1_1_0.3.jpg}
    	\subcaption{Test 1.}
    	\label{figure_2dtraj1_1}
    \end{minipage}
    \hfill
	\begin{minipage}[t]{0.195\linewidth}
        \centering
    	\includegraphics[trim=61 38 48 46, clip, width=1.0\linewidth]{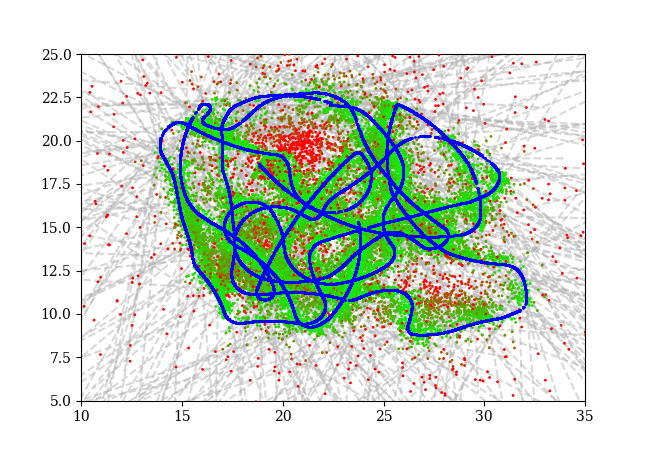}
    	\subcaption{Test 2.}
    	\label{figure_2dtraj1_2}
    \end{minipage}
    \hfill
	\begin{minipage}[t]{0.195\linewidth}
        \centering
    	\includegraphics[trim=61 38 48 46, clip, width=1.0\linewidth]{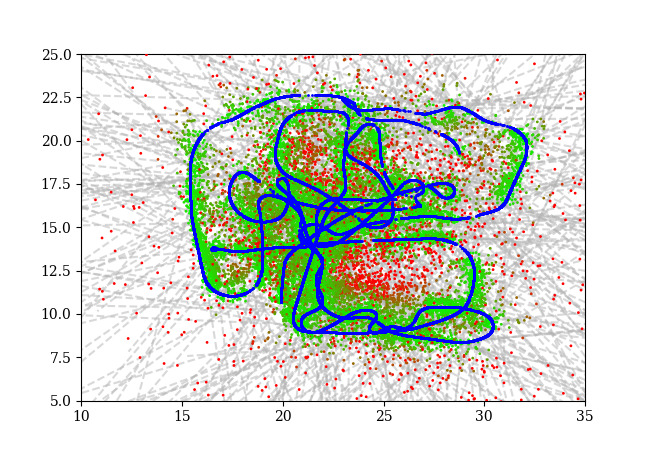}
    	\subcaption{Test 3.}
    	\label{figure_2dtraj1_3}
    \end{minipage}
    \hfill
	\begin{minipage}[t]{0.195\linewidth}
        \centering
    	\includegraphics[trim=61 38 48 46, clip, width=1.0\linewidth]{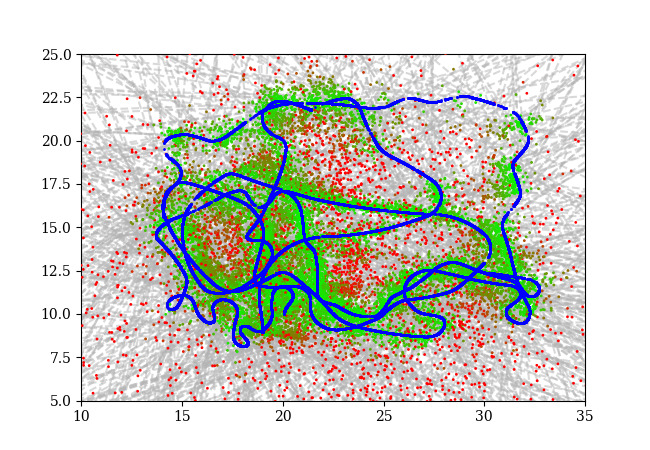}
    	\subcaption{Test 4.}
    	\label{figure_2dtraj1_4}
    \end{minipage}
    \hfill
	\begin{minipage}[t]{0.195\linewidth}
        \centering
    	\includegraphics[trim=61 38 48 46, clip, width=1.0\linewidth]{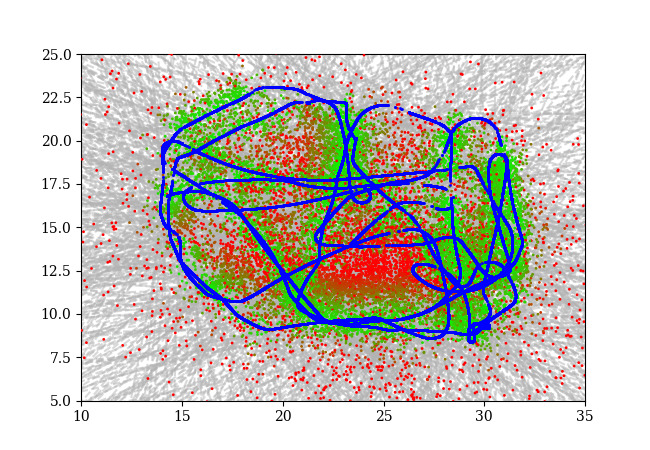}
    	\subcaption{Test 6.}
    	\label{figure_2dtraj1_6}
    \end{minipage}
	\begin{minipage}[t]{0.195\linewidth}
        \centering
    	\includegraphics[trim=61 38 48 46, clip, width=1.0\linewidth]{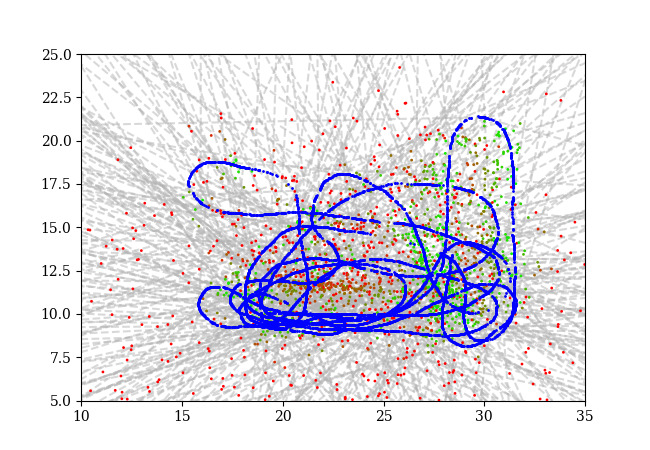}
    	\subcaption{Test 7.}
    	\label{figure_2dtraj1_7}
    \end{minipage}
    \hfill
	\begin{minipage}[t]{0.195\linewidth}
        \centering
    	\includegraphics[trim=61 38 48 46, clip, width=1.0\linewidth]{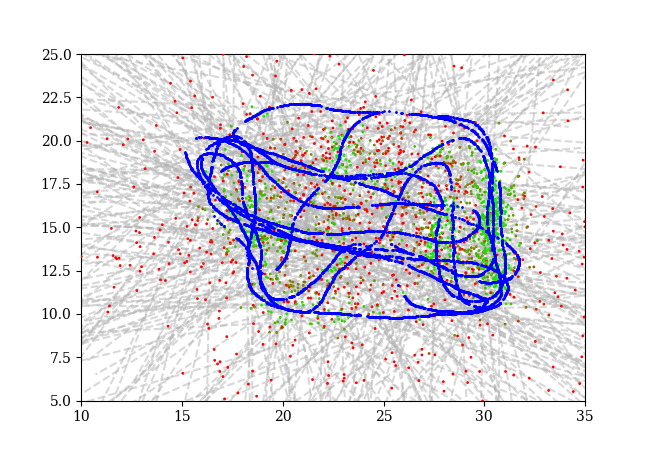}
    	\subcaption{Test 8.}
    	\label{figure_2dtraj1_9}
    \end{minipage}
    \hfill
	\begin{minipage}[t]{0.195\linewidth}
        \centering
    	\includegraphics[trim=61 38 48 46, clip, width=1.0\linewidth]{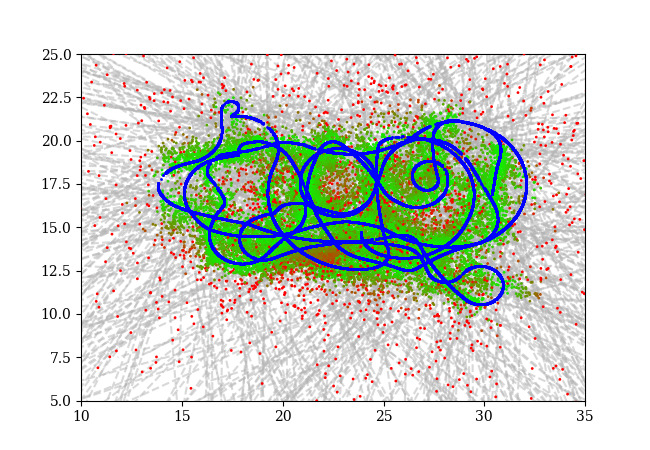}
    	\subcaption{Test 9.}
    	\label{figure_2dtraj1_10}
    \end{minipage}
    \hfill
	\begin{minipage}[t]{0.195\linewidth}
        \centering
    	\includegraphics[trim=61 38 48 46, clip, width=1.0\linewidth]{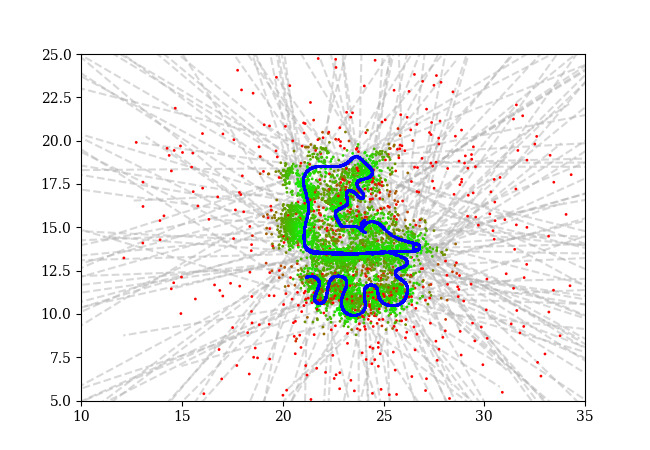}
    	\subcaption{Test 10.}
    	\label{figure_2dtraj1_11}
    \end{minipage}
    \hfill
	\begin{minipage}[t]{0.195\linewidth}
        \centering
    	\includegraphics[trim=61 38 48 46, clip, width=1.0\linewidth]{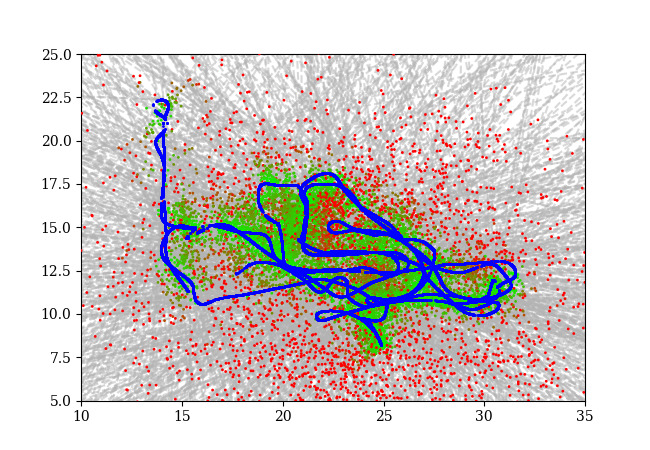}
    	\subcaption{Test 11.}
    	\label{figure_2dtraj1_12}
    \end{minipage}
    \caption{Evaluation of the predicted positions (green, red) against the ground truth trajectories (blue) for SfM for the train 1 dataset.}
    \label{figure_2dtraj1}
\end{figure*}

\begin{figure*}[!t]
    \centering
	\begin{minipage}[t]{0.195\linewidth}
        \centering
    	\includegraphics[trim=61 38 48 46, clip, width=1.0\linewidth]{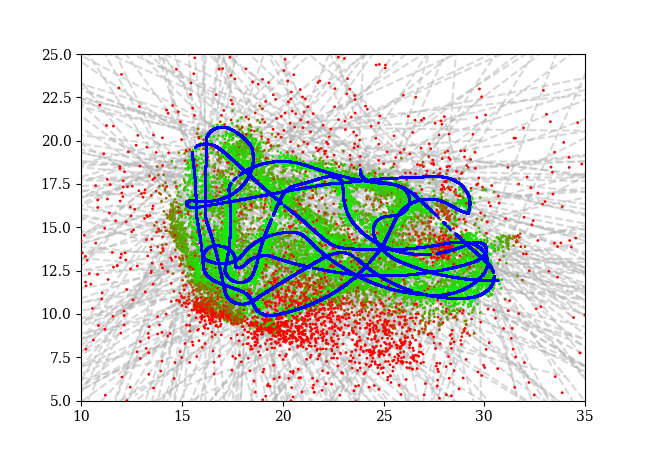}
    	\subcaption{Test 1.}
    	\label{figure_2dtraj2_1}
    \end{minipage}
    \hfill
	\begin{minipage}[t]{0.195\linewidth}
        \centering
    	\includegraphics[trim=61 38 48 46, clip, width=1.0\linewidth]{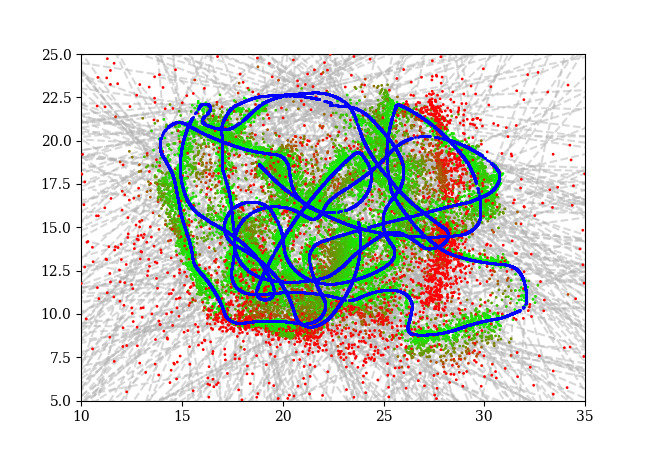}
    	\subcaption{Test 2.}
    	\label{figure_2dtraj2_2}
    \end{minipage}
    \hfill
	\begin{minipage}[t]{0.195\linewidth}
        \centering
    	\includegraphics[trim=61 38 48 46, clip, width=1.0\linewidth]{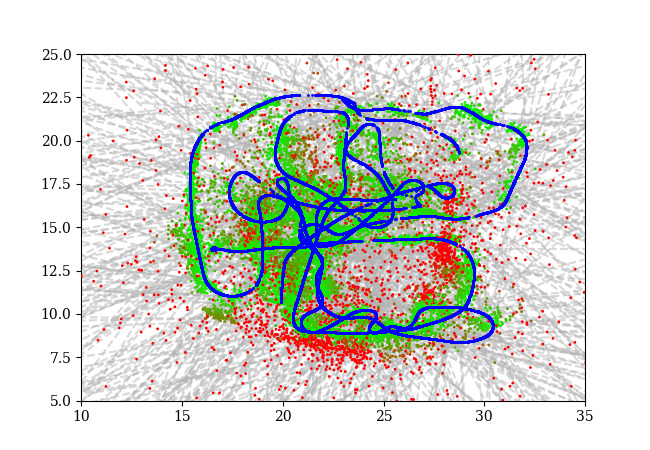}
    	\subcaption{Test 3.}
    	\label{figure_2dtraj2_3}
    \end{minipage}
    \hfill
	\begin{minipage}[t]{0.195\linewidth}
        \centering
    	\includegraphics[trim=61 38 48 46, clip, width=1.0\linewidth]{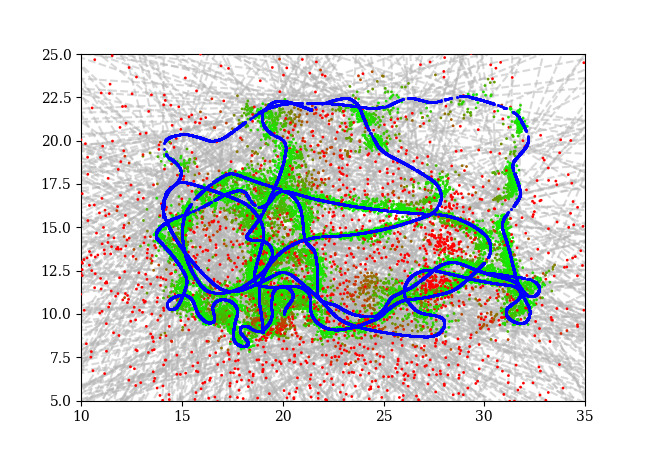}
    	\subcaption{Test 4.}
    	\label{figure_2dtraj2_4}
    \end{minipage}
    \hfill
	\begin{minipage}[t]{0.195\linewidth}
        \centering
    	\includegraphics[trim=61 38 48 46, clip, width=1.0\linewidth]{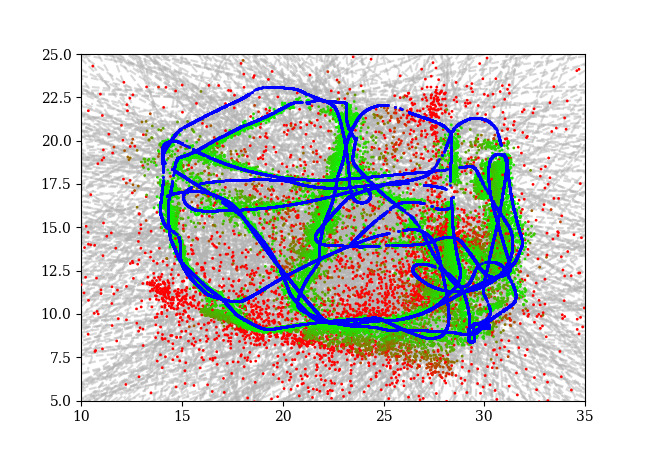}
    	\subcaption{Test 6.}
    	\label{figure_2dtraj2_6}
    \end{minipage}
	\begin{minipage}[t]{0.195\linewidth}
        \centering
    	\includegraphics[trim=61 38 48 46, clip, width=1.0\linewidth]{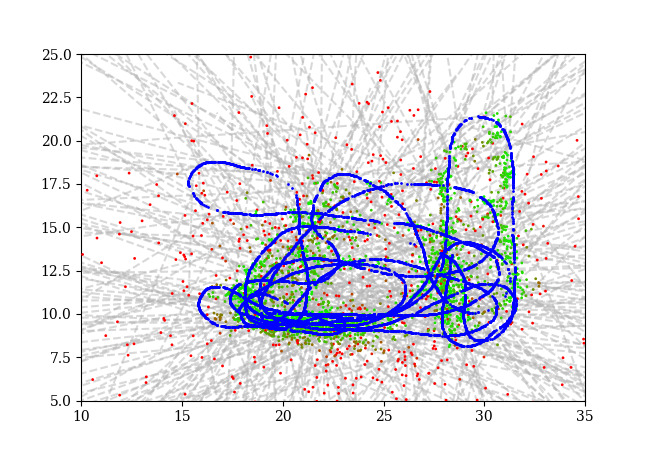}
    	\subcaption{Test 7.}
    	\label{figure_2dtraj2_7}
    \end{minipage}
    \hfill
	\begin{minipage}[t]{0.195\linewidth}
        \centering
    	\includegraphics[trim=61 38 48 46, clip, width=1.0\linewidth]{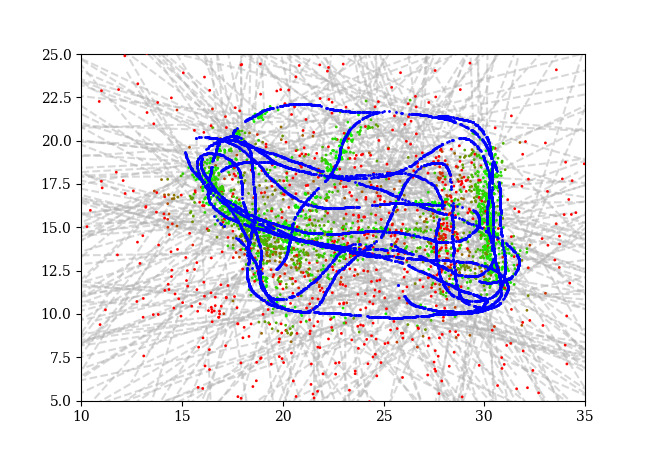}
    	\subcaption{Test 8.}
    	\label{figure_2dtraj2_9}
    \end{minipage}
    \hfill
	\begin{minipage}[t]{0.195\linewidth}
        \centering
    	\includegraphics[trim=61 38 48 46, clip, width=1.0\linewidth]{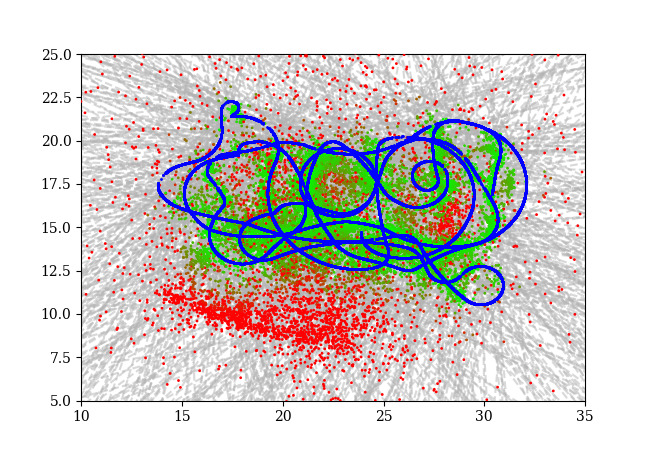}
    	\subcaption{Test 9.}
    	\label{figure_2dtraj2_10}
    \end{minipage}
    \hfill
	\begin{minipage}[t]{0.195\linewidth}
        \centering
    	\includegraphics[trim=61 38 48 46, clip, width=1.0\linewidth]{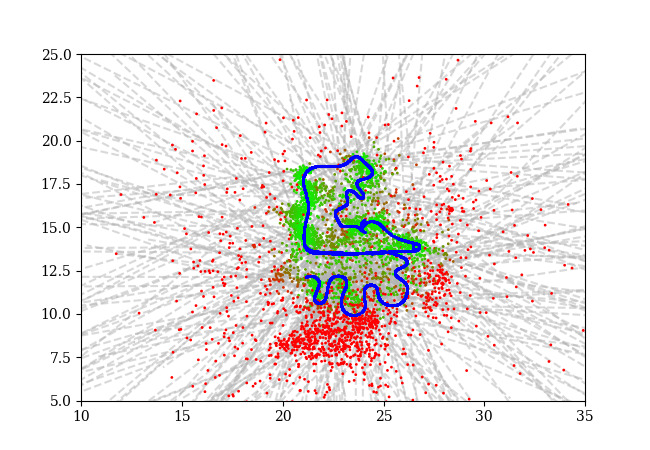}
    	\subcaption{Test 10.}
    	\label{figure_2dtraj2_11}
    \end{minipage}
    \hfill
	\begin{minipage}[t]{0.195\linewidth}
        \centering
    	\includegraphics[trim=61 38 48 46, clip, width=1.0\linewidth]{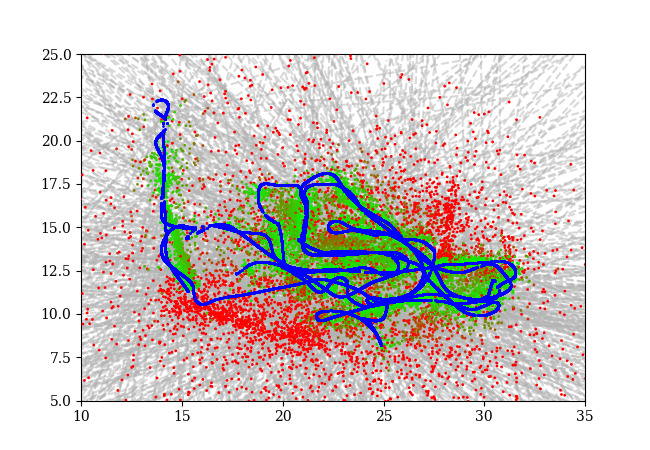}
    	\subcaption{Test 11.}
    	\label{figure_2dtraj2_12}
    \end{minipage}
    \caption{Evaluation of the predicted positions (green, red) against the ground truth trajectories (blue) for SfM for the train 2 dataset.}
    \label{figure_2dtraj2}
\end{figure*}

\begin{figure*}[!t]
    \centering
	\begin{minipage}[t]{0.195\linewidth}
        \centering
    	\includegraphics[trim=61 38 48 46, clip, width=1.0\linewidth]{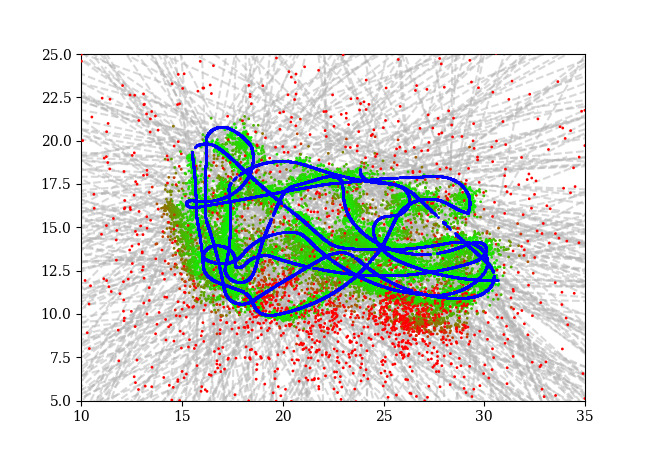}
    	\subcaption{Test 1.}
    	\label{figure_2dtraj3_1}
    \end{minipage}
    \hfill
	\begin{minipage}[t]{0.195\linewidth}
        \centering
    	\includegraphics[trim=61 38 48 46, clip, width=1.0\linewidth]{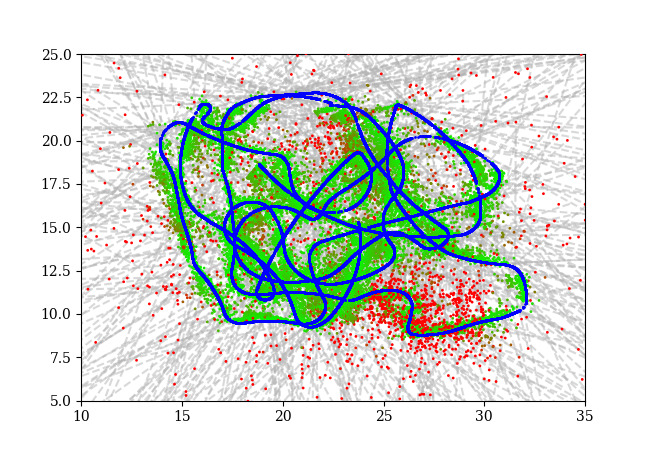}
    	\subcaption{Test 2.}
    	\label{figure_2dtraj3_2}
    \end{minipage}
    \hfill
	\begin{minipage}[t]{0.195\linewidth}
        \centering
    	\includegraphics[trim=61 38 48 46, clip, width=1.0\linewidth]{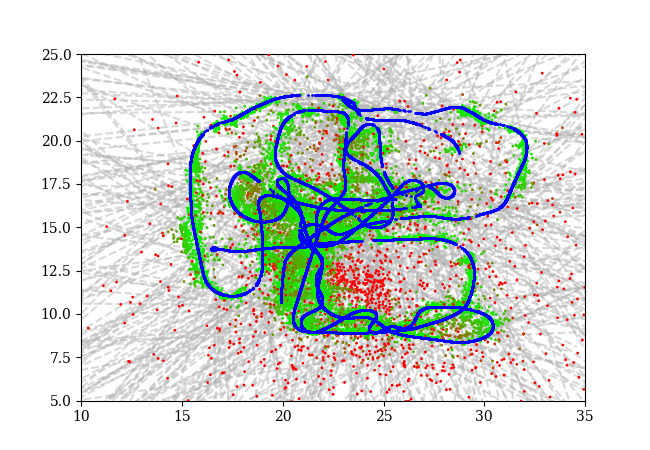}
    	\subcaption{Test 3.}
    	\label{figure_2dtraj3_3}
    \end{minipage}
    \hfill
	\begin{minipage}[t]{0.195\linewidth}
        \centering
    	\includegraphics[trim=61 38 48 46, clip, width=1.0\linewidth]{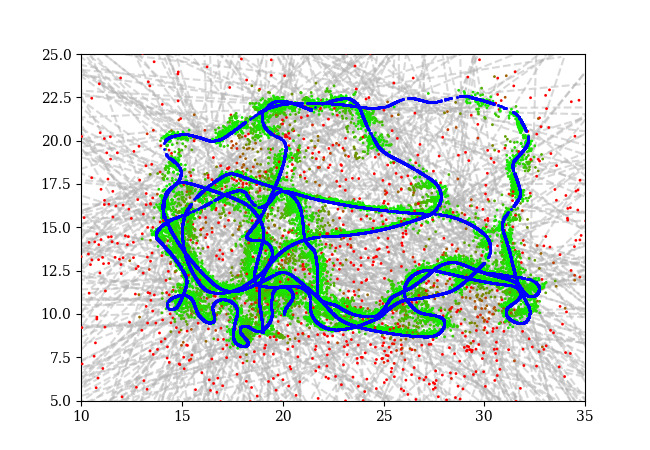}
    	\subcaption{Test 4.}
    	\label{figure_2dtraj3_4}
    \end{minipage}
    \hfill
	\begin{minipage}[t]{0.195\linewidth}
        \centering
    	\includegraphics[trim=61 38 48 46, clip, width=1.0\linewidth]{test_6_3_0.3.jpg}
    	\subcaption{Test 6.}
    	\label{figure_2dtraj3_6}
    \end{minipage}
	\begin{minipage}[t]{0.195\linewidth}
        \centering
    	\includegraphics[trim=61 38 48 46, clip, width=1.0\linewidth]{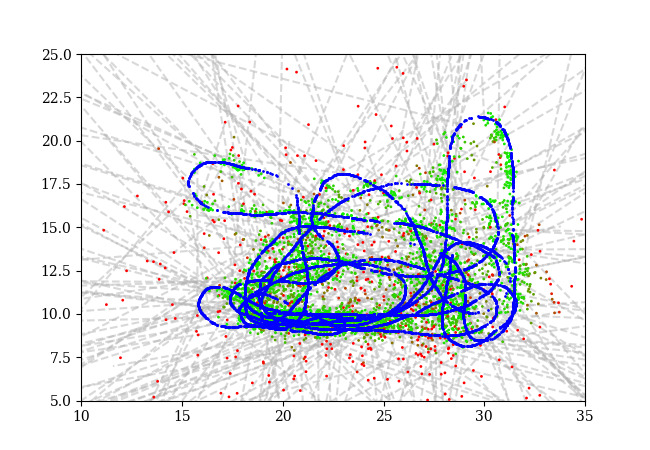}
    	\subcaption{Test 7.}
    	\label{figure_2dtraj3_7}
    \end{minipage}
    \hfill
	\begin{minipage}[t]{0.195\linewidth}
        \centering
    	\includegraphics[trim=61 38 48 46, clip, width=1.0\linewidth]{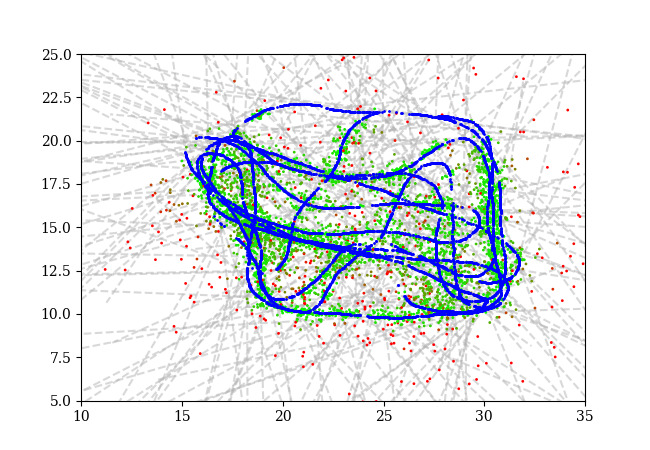}
    	\subcaption{Test 8.}
    	\label{figure_2dtraj3_9}
    \end{minipage}
    \hfill
	\begin{minipage}[t]{0.195\linewidth}
        \centering
    	\includegraphics[trim=61 38 48 46, clip, width=1.0\linewidth]{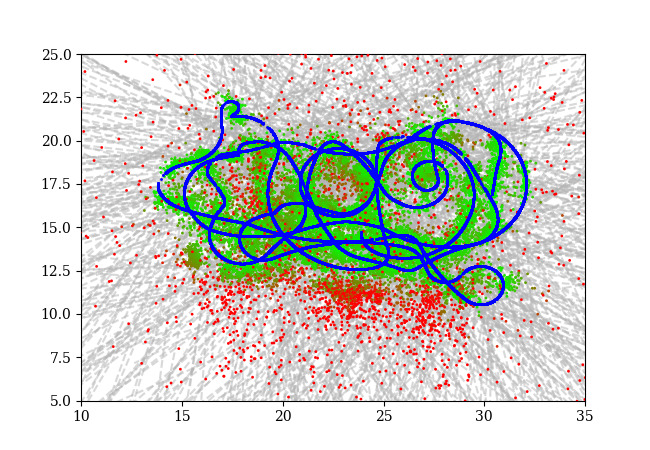}
    	\subcaption{Test 9.}
    	\label{figure_2dtraj3_10}
    \end{minipage}
    \hfill
	\begin{minipage}[t]{0.195\linewidth}
        \centering
    	\includegraphics[trim=61 38 48 46, clip, width=1.0\linewidth]{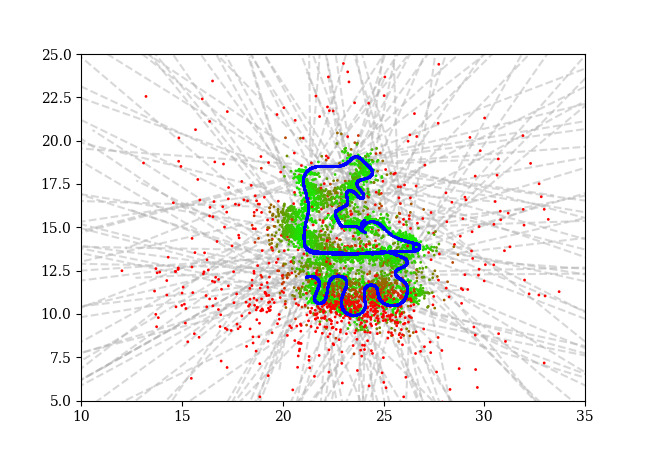}
    	\subcaption{Test 10.}
    	\label{figure_2dtraj3_11}
    \end{minipage}
    \hfill
	\begin{minipage}[t]{0.195\linewidth}
        \centering
    	\includegraphics[trim=61 38 48 46, clip, width=1.0\linewidth]{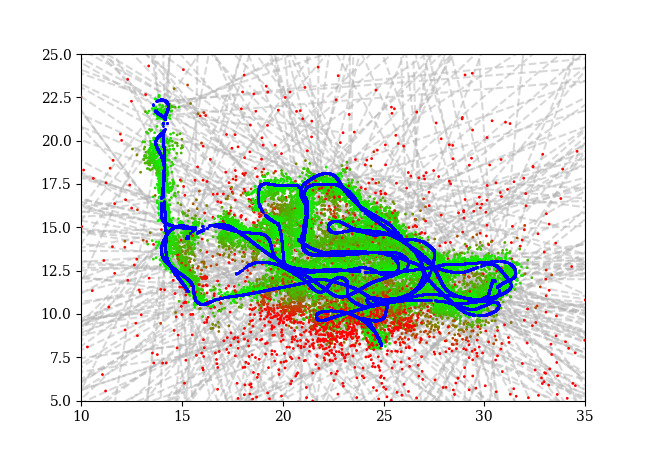}
    	\subcaption{Test 11.}
    	\label{figure_2dtraj3_12}
    \end{minipage}
    \caption{Evaluation of the predicted positions (green, red) against the ground truth trajectories (blue) for SfM for the train 3 dataset.}
    \label{figure_2dtraj3}
\end{figure*}

\begin{figure*}[!t]
    \centering
	\begin{minipage}[t]{0.195\linewidth}
        \centering
    	\includegraphics[trim=61 38 48 46, clip, width=1.0\linewidth]{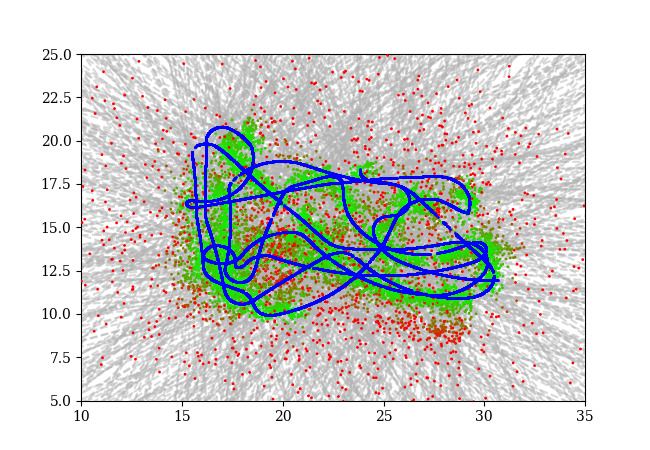}
    	\subcaption{Test 1.}
    	\label{figure_2dtraj4_1}
    \end{minipage}
    \hfill
	\begin{minipage}[t]{0.195\linewidth}
        \centering
    	\includegraphics[trim=61 38 48 46, clip, width=1.0\linewidth]{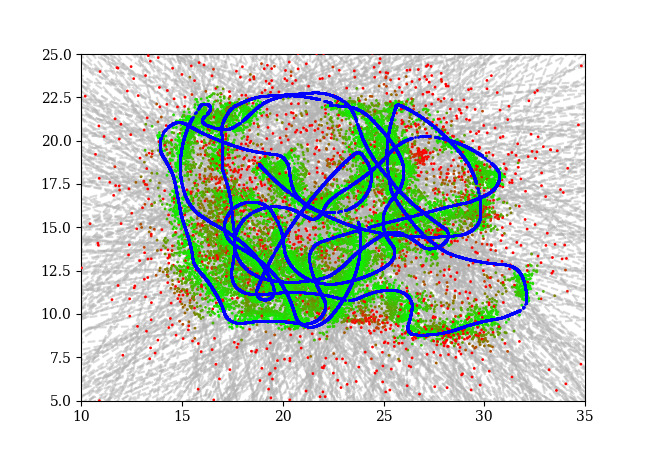}
    	\subcaption{Test 2.}
    	\label{figure_2dtraj4_2}
    \end{minipage}
    \hfill
	\begin{minipage}[t]{0.195\linewidth}
        \centering
    	\includegraphics[trim=61 38 48 46, clip, width=1.0\linewidth]{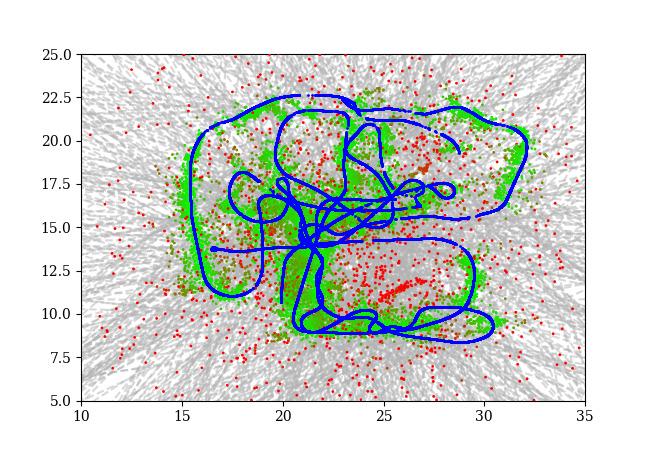}
    	\subcaption{Test 3.}
    	\label{figure_2dtraj4_3}
    \end{minipage}
    \hfill
	\begin{minipage}[t]{0.195\linewidth}
        \centering
    	\includegraphics[trim=61 38 48 46, clip, width=1.0\linewidth]{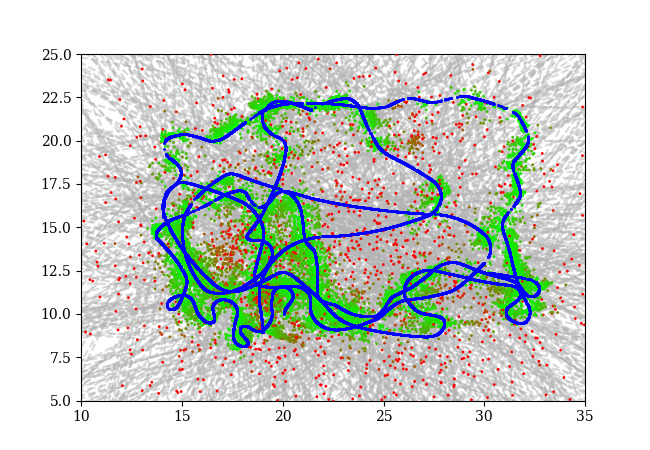}
    	\subcaption{Test 4.}
    	\label{figure_2dtraj4_4}
    \end{minipage}
    \hfill
	\begin{minipage}[t]{0.195\linewidth}
        \centering
    	\includegraphics[trim=61 38 48 46, clip, width=1.0\linewidth]{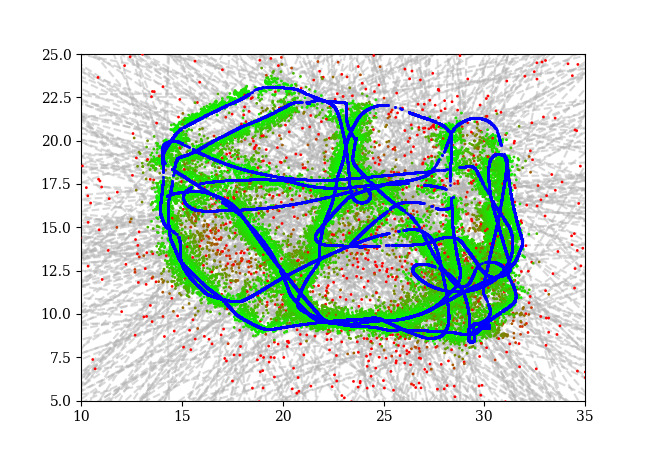}
    	\subcaption{Test 6.}
    	\label{figure_2dtraj4_6}
    \end{minipage}
	\begin{minipage}[t]{0.195\linewidth}
        \centering
    	\includegraphics[trim=61 38 48 46, clip, width=1.0\linewidth]{test_7_4_0.2.jpg}
    	\subcaption{Test 7.}
    	\label{figure_2dtraj4_7}
    \end{minipage}
    \hfill
	\begin{minipage}[t]{0.195\linewidth}
        \centering
    	\includegraphics[trim=61 38 48 46, clip, width=1.0\linewidth]{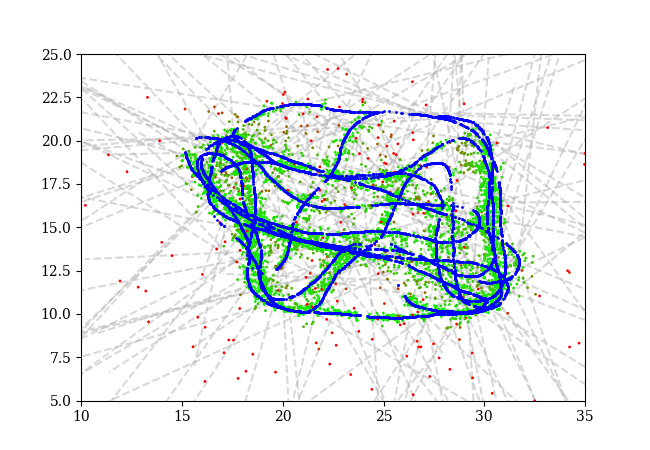}
    	\subcaption{Test 8.}
    	\label{figure_2dtraj4_9}
    \end{minipage}
    \hfill
	\begin{minipage}[t]{0.195\linewidth}
        \centering
    	\includegraphics[trim=61 38 48 46, clip, width=1.0\linewidth]{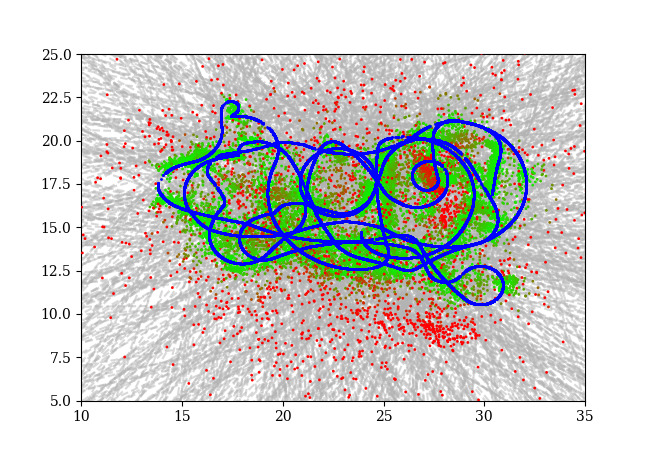}
    	\subcaption{Test 9.}
    	\label{figure_2dtraj4_10}
    \end{minipage}
    \hfill
	\begin{minipage}[t]{0.195\linewidth}
        \centering
    	\includegraphics[trim=61 38 48 46, clip, width=1.0\linewidth]{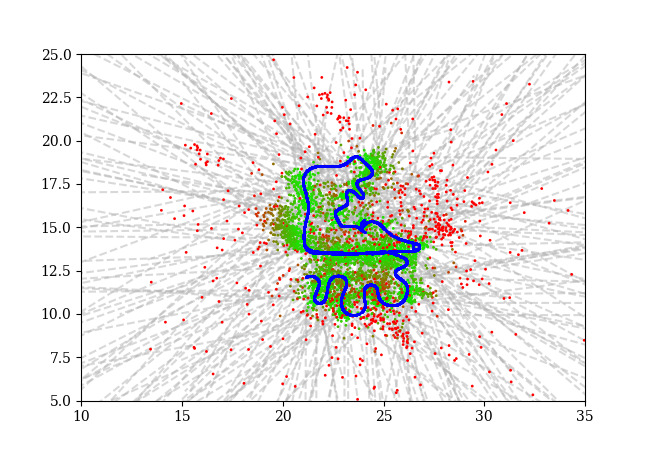}
    	\subcaption{Test 10.}
    	\label{figure_2dtraj4_11}
    \end{minipage}
    \hfill
	\begin{minipage}[t]{0.195\linewidth}
        \centering
    	\includegraphics[trim=61 38 48 46, clip, width=1.0\linewidth]{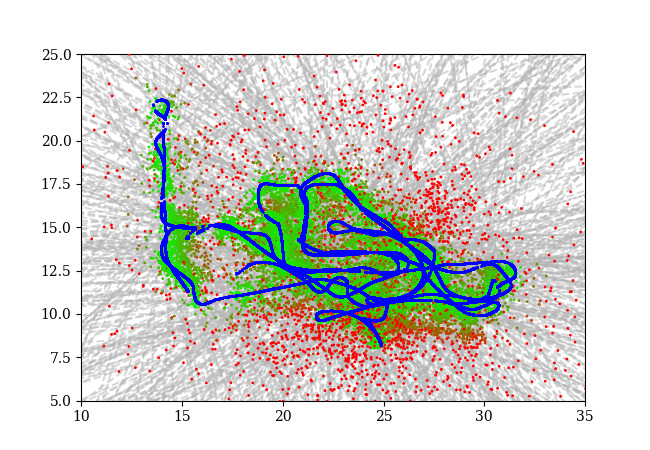}
    	\subcaption{Test 11.}
    	\label{figure_2dtraj4_12}
    \end{minipage}
    \caption{Evaluation of the predicted positions (green, red) against the ground truth trajectories (blue) for SfM for the train 4 dataset.}
    \label{figure_2dtraj4}
\end{figure*}

\begin{figure*}[!t]
    \centering
	\begin{minipage}[t]{0.195\linewidth}
        \centering
    	\includegraphics[trim=61 38 48 46, clip, width=1.0\linewidth]{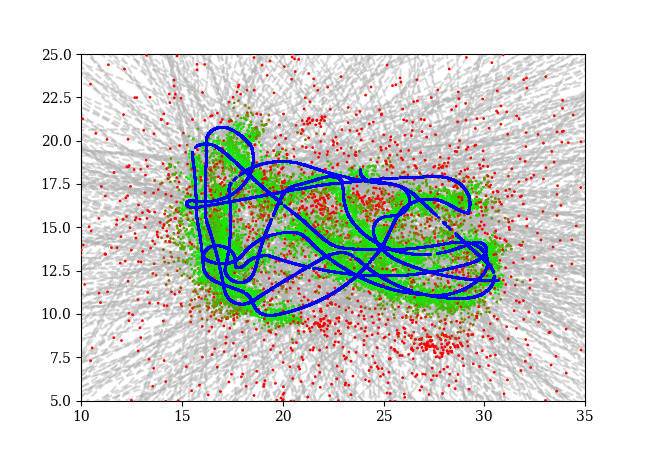}
    	\subcaption{Test 1.}
    	\label{figure_2dtraj5_1}
    \end{minipage}
    \hfill
	\begin{minipage}[t]{0.195\linewidth}
        \centering
    	\includegraphics[trim=61 38 48 46, clip, width=1.0\linewidth]{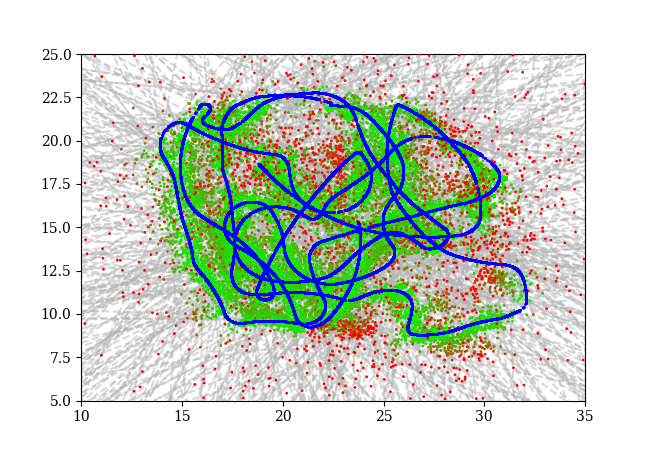}
    	\subcaption{Test 2.}
    	\label{figure_2dtraj5_2}
    \end{minipage}
    \hfill
	\begin{minipage}[t]{0.195\linewidth}
        \centering
    	\includegraphics[trim=61 38 48 46, clip, width=1.0\linewidth]{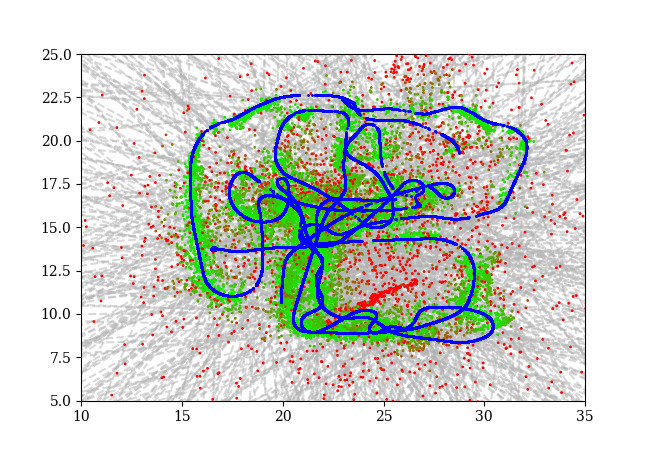}
    	\subcaption{Test 3.}
    	\label{figure_2dtraj5_3}
    \end{minipage}
    \hfill
	\begin{minipage}[t]{0.195\linewidth}
        \centering
    	\includegraphics[trim=61 38 48 46, clip, width=1.0\linewidth]{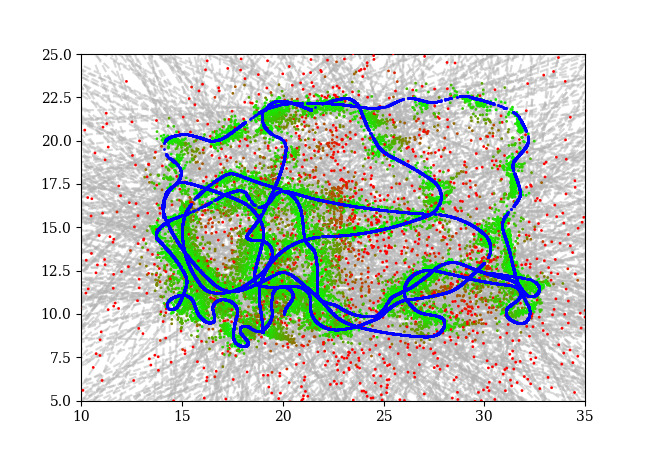}
    	\subcaption{Test 4.}
    	\label{figure_2dtraj5_4}
    \end{minipage}
    \hfill
	\begin{minipage}[t]{0.195\linewidth}
        \centering
    	\includegraphics[trim=61 38 48 46, clip, width=1.0\linewidth]{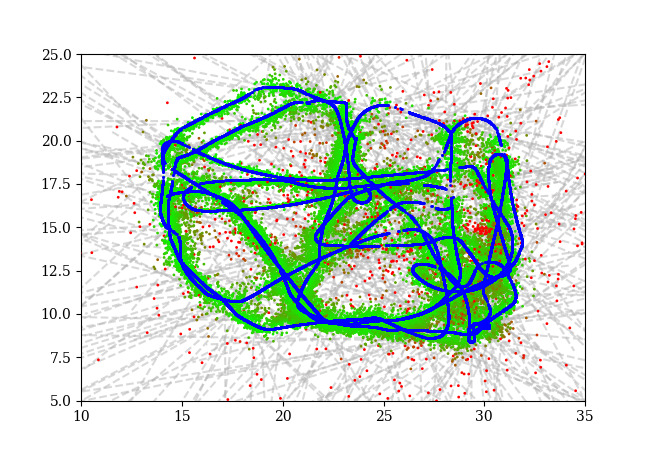}
    	\subcaption{Test 6.}
    	\label{figure_2dtraj5_6}
    \end{minipage}
	\begin{minipage}[t]{0.195\linewidth}
        \centering
    	\includegraphics[trim=61 38 48 46, clip, width=1.0\linewidth]{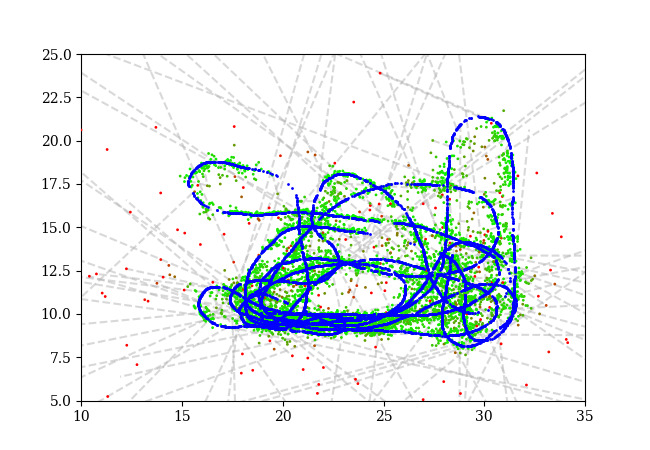}
    	\subcaption{Test 7.}
    	\label{figure_2dtraj5_7}
    \end{minipage}
    \hfill
	\begin{minipage}[t]{0.195\linewidth}
        \centering
    	\includegraphics[trim=61 38 48 46, clip, width=1.0\linewidth]{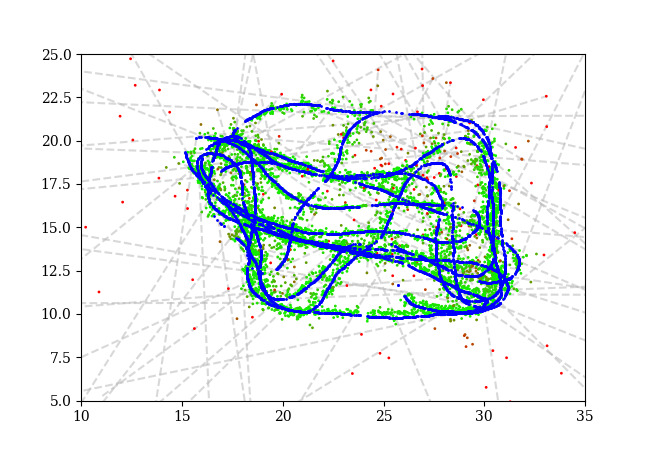}
    	\subcaption{Test 8.}
    	\label{figure_2dtraj5_9}
    \end{minipage}
    \hfill
	\begin{minipage}[t]{0.195\linewidth}
        \centering
    	\includegraphics[trim=61 38 48 46, clip, width=1.0\linewidth]{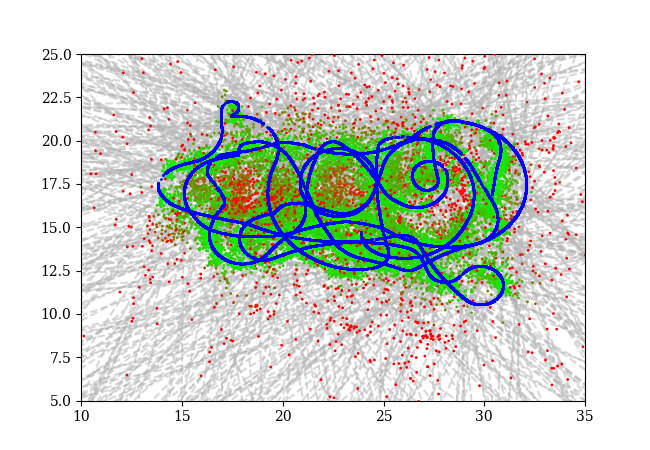}
    	\subcaption{Test 9.}
    	\label{figure_2dtraj5_10}
    \end{minipage}
    \hfill
	\begin{minipage}[t]{0.195\linewidth}
        \centering
    	\includegraphics[trim=61 38 48 46, clip, width=1.0\linewidth]{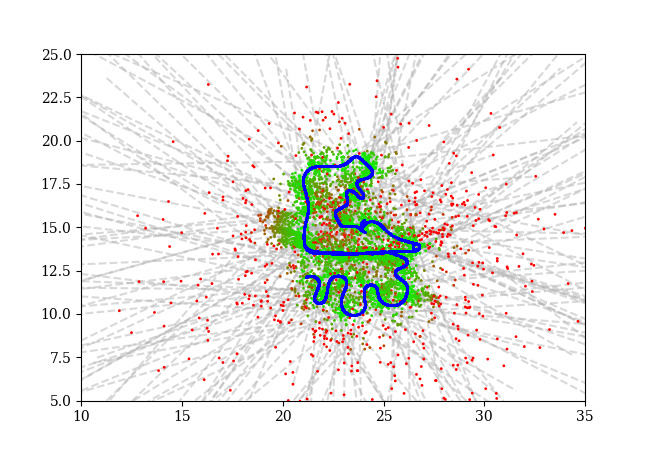}
    	\subcaption{Test 10.}
    	\label{figure_2dtraj5_11}
    \end{minipage}
    \hfill
	\begin{minipage}[t]{0.195\linewidth}
        \centering
    	\includegraphics[trim=61 38 48 46, clip, width=1.0\linewidth]{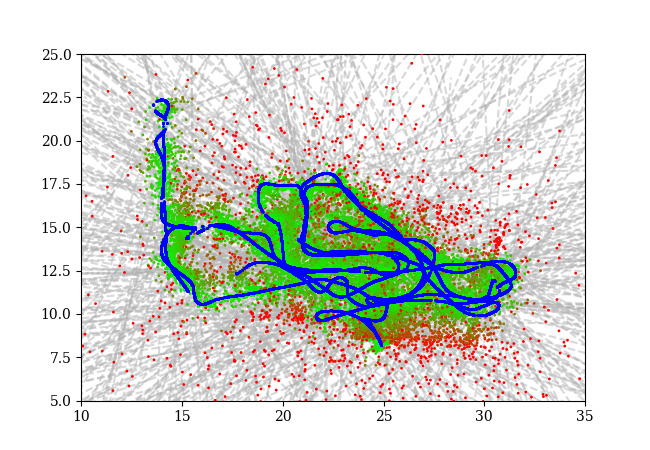}
    	\subcaption{Test 11.}
    	\label{figure_2dtraj5_12}
    \end{minipage}
    \caption{Evaluation of the predicted positions (green, red) against the ground truth trajectories (blue) for SfM for the train 5 dataset.}
    \label{figure_2dtraj5}
\end{figure*}

\begin{figure*}[!t]
    \centering
	\begin{minipage}[t]{0.195\linewidth}
        \centering
    	\includegraphics[trim=61 38 48 46, clip, width=1.0\linewidth]{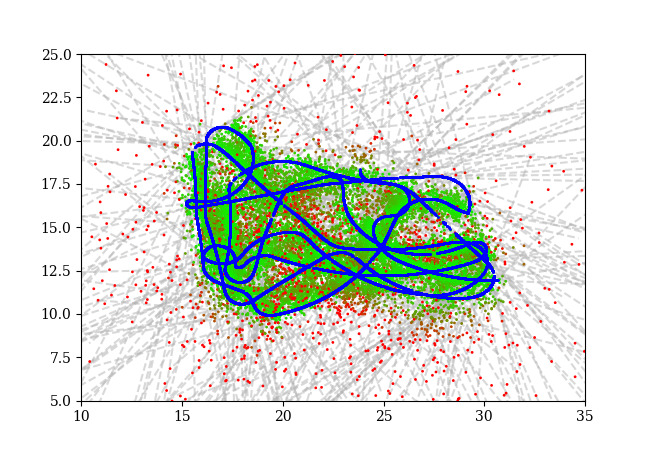}
    	\subcaption{Test 1.}
    	\label{figure_2dtraj6_1}
    \end{minipage}
    \hfill
	\begin{minipage}[t]{0.195\linewidth}
        \centering
    	\includegraphics[trim=61 38 48 46, clip, width=1.0\linewidth]{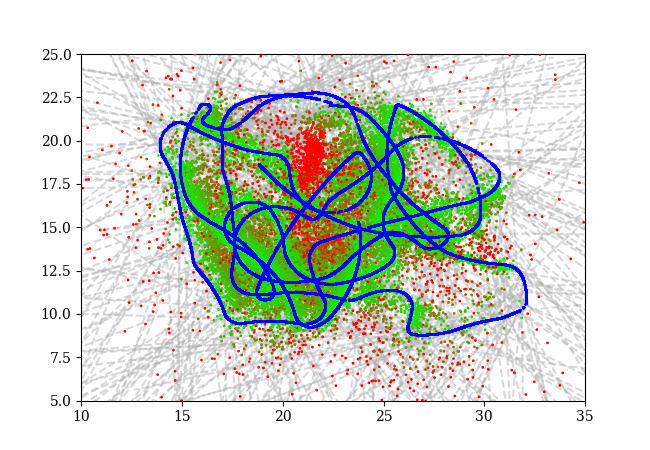}
    	\subcaption{Test 2.}
    	\label{figure_2dtraj6_2}
    \end{minipage}
    \hfill
	\begin{minipage}[t]{0.195\linewidth}
        \centering
    	\includegraphics[trim=61 38 48 46, clip, width=1.0\linewidth]{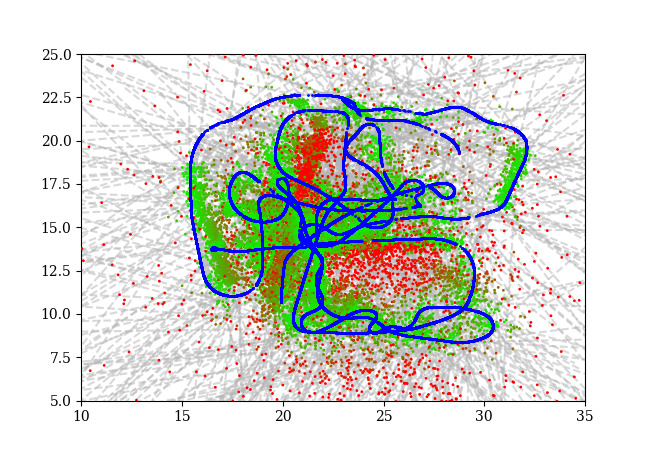}
    	\subcaption{Test 3.}
    	\label{figure_2dtraj6_3}
    \end{minipage}
    \hfill
	\begin{minipage}[t]{0.195\linewidth}
        \centering
    	\includegraphics[trim=61 38 48 46, clip, width=1.0\linewidth]{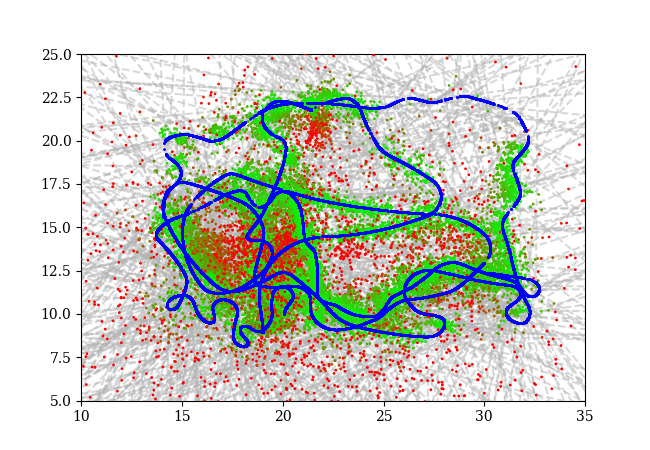}
    	\subcaption{Test 4.}
    	\label{figure_2dtraj6_4}
    \end{minipage}
    \hfill
	\begin{minipage}[t]{0.195\linewidth}
        \centering
    	\includegraphics[trim=61 38 48 46, clip, width=1.0\linewidth]{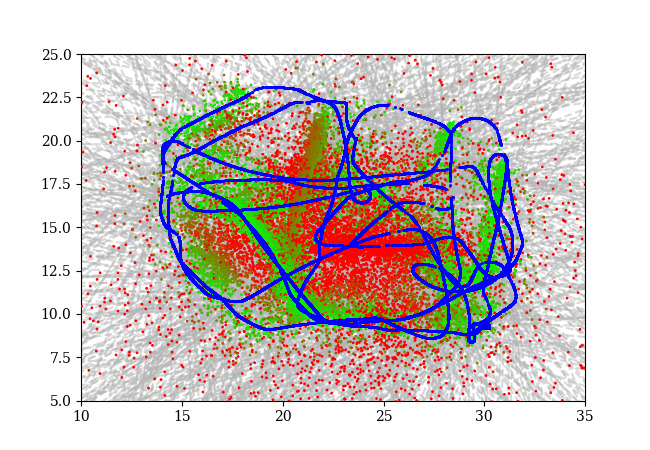}
    	\subcaption{Test 6.}
    	\label{figure_2dtraj6_6}
    \end{minipage}
	\begin{minipage}[t]{0.195\linewidth}
        \centering
    	\includegraphics[trim=61 38 48 46, clip, width=1.0\linewidth]{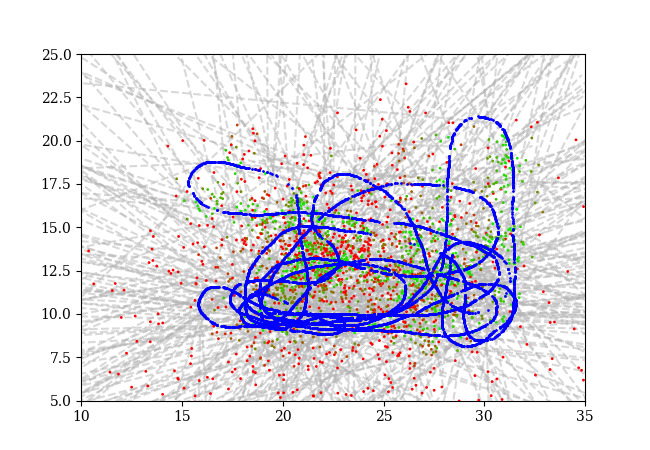}
    	\subcaption{Test 7.}
    	\label{figure_2dtraj6_7}
    \end{minipage}
    \hfill
	\begin{minipage}[t]{0.195\linewidth}
        \centering
    	\includegraphics[trim=61 38 48 46, clip, width=1.0\linewidth]{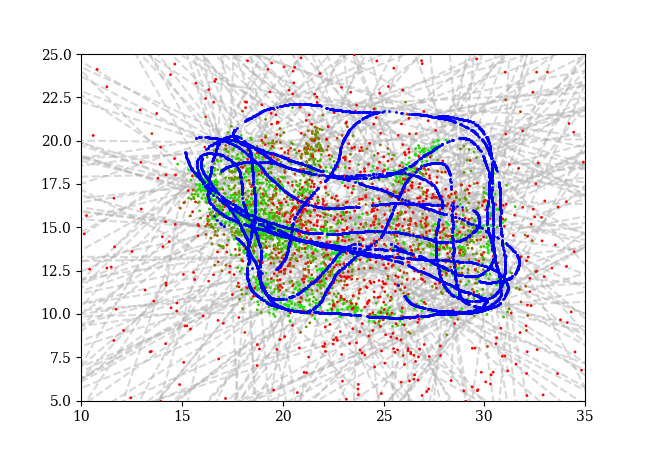}
    	\subcaption{Test 8.}
    	\label{figure_2dtraj6_9}
    \end{minipage}
    \hfill
	\begin{minipage}[t]{0.195\linewidth}
        \centering
    	\includegraphics[trim=61 38 48 46, clip, width=1.0\linewidth]{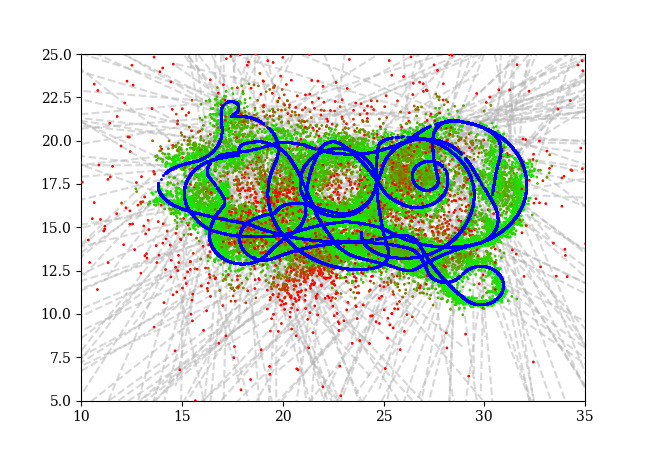}
    	\subcaption{Test 9.}
    	\label{figure_2dtraj6_10}
    \end{minipage}
    \hfill
	\begin{minipage}[t]{0.195\linewidth}
        \centering
    	\includegraphics[trim=61 38 48 46, clip, width=1.0\linewidth]{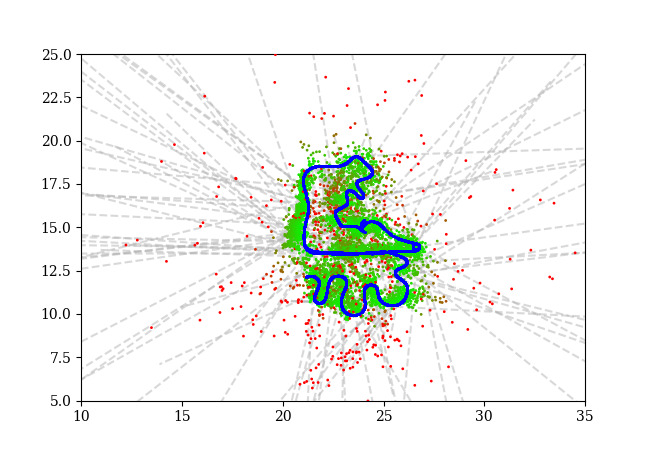}
    	\subcaption{Test 10.}
    	\label{figure_2dtraj6_11}
    \end{minipage}
    \hfill
	\begin{minipage}[t]{0.195\linewidth}
        \centering
    	\includegraphics[trim=61 38 48 46, clip, width=1.0\linewidth]{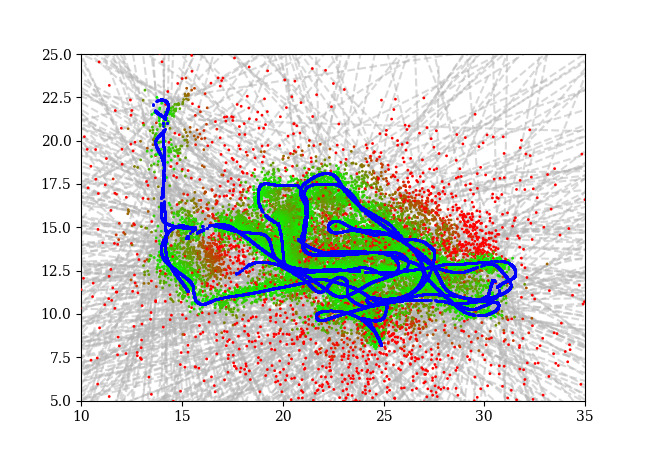}
    	\subcaption{Test 11.}
    	\label{figure_2dtraj6_12}
    \end{minipage}
    \caption{Evaluation of the predicted positions (green, red) against the ground truth trajectories (blue) for SfM for the train 6 dataset.}
    \label{figure_2dtraj6}
\end{figure*}

\begin{figure*}[!t]
    \centering
	\begin{minipage}[t]{0.195\linewidth}
        \centering
    	\includegraphics[trim=61 38 48 46, clip, width=1.0\linewidth]{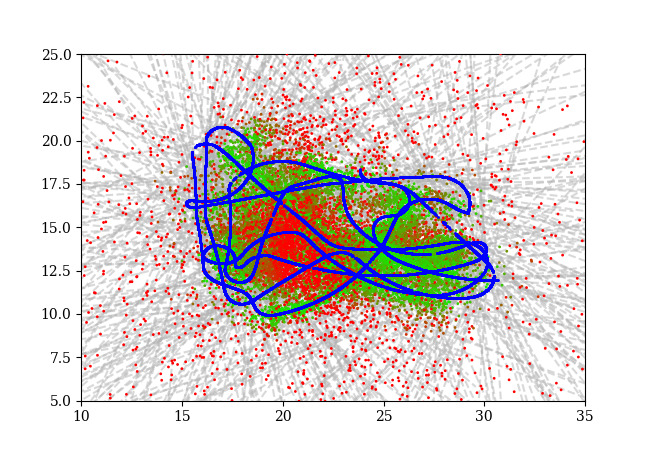}
    	\subcaption{Test 1.}
    	\label{figure_2dtraj7_1}
    \end{minipage}
    \hfill
	\begin{minipage}[t]{0.195\linewidth}
        \centering
    	\includegraphics[trim=61 38 48 46, clip, width=1.0\linewidth]{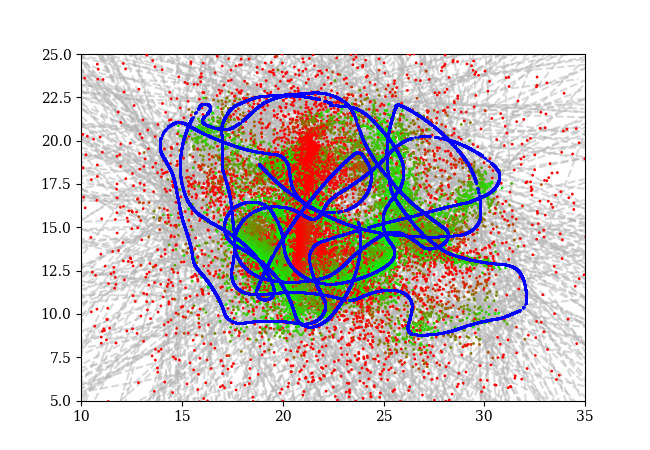}
    	\subcaption{Test 2.}
    	\label{figure_2dtraj7_2}
    \end{minipage}
    \hfill
	\begin{minipage}[t]{0.195\linewidth}
        \centering
    	\includegraphics[trim=61 38 48 46, clip, width=1.0\linewidth]{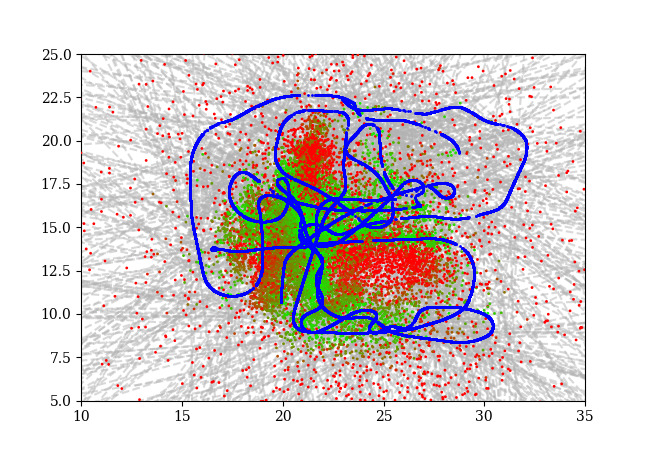}
    	\subcaption{Test 3.}
    	\label{figure_2dtraj7_3}
    \end{minipage}
    \hfill
	\begin{minipage}[t]{0.195\linewidth}
        \centering
    	\includegraphics[trim=61 38 48 46, clip, width=1.0\linewidth]{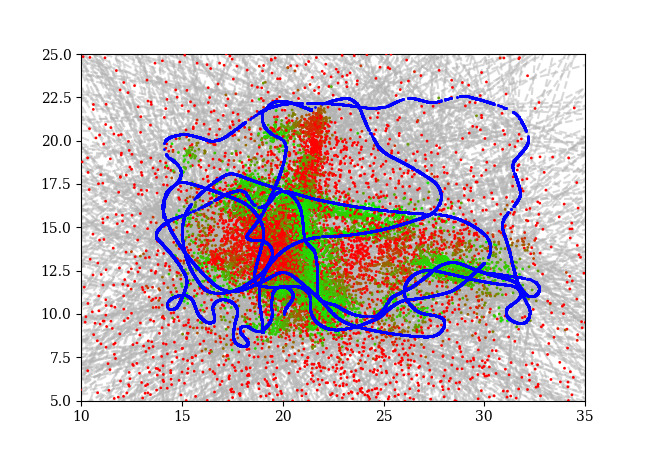}
    	\subcaption{Test 4.}
    	\label{figure_2dtraj7_4}
    \end{minipage}
    \hfill
	\begin{minipage}[t]{0.195\linewidth}
        \centering
    	\includegraphics[trim=61 38 48 46, clip, width=1.0\linewidth]{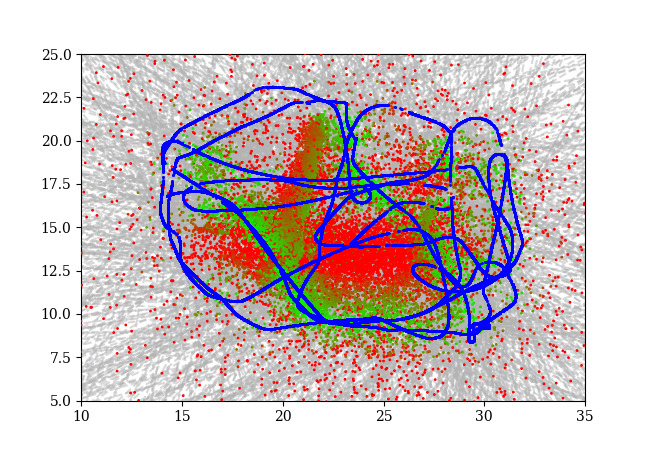}
    	\subcaption{Test 6.}
    	\label{figure_2dtraj7_6}
    \end{minipage}
	\begin{minipage}[t]{0.195\linewidth}
        \centering
    	\includegraphics[trim=61 38 48 46, clip, width=1.0\linewidth]{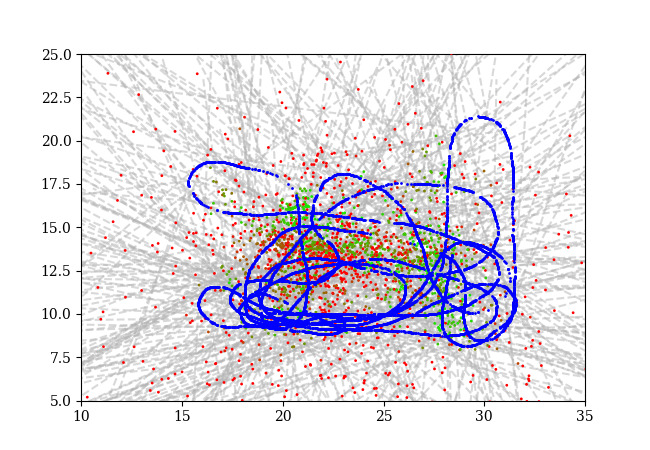}
    	\subcaption{Test 7.}
    	\label{figure_2dtraj7_7}
    \end{minipage}
    \hfill
	\begin{minipage}[t]{0.195\linewidth}
        \centering
    	\includegraphics[trim=61 38 48 46, clip, width=1.0\linewidth]{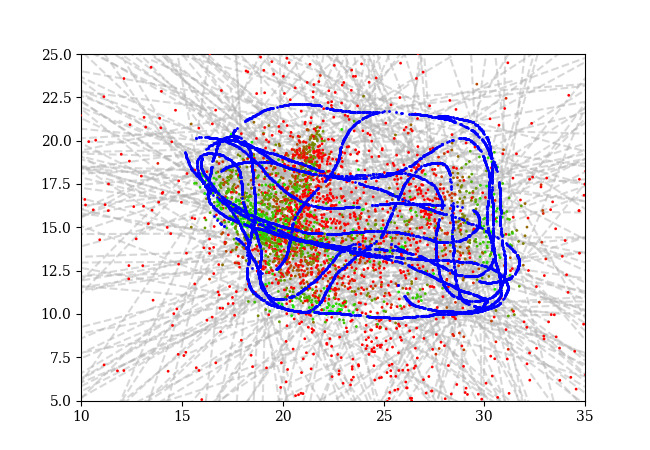}
    	\subcaption{Test 8.}
    	\label{figure_2dtraj7_9}
    \end{minipage}
    \hfill
	\begin{minipage}[t]{0.195\linewidth}
        \centering
    	\includegraphics[trim=61 38 48 46, clip, width=1.0\linewidth]{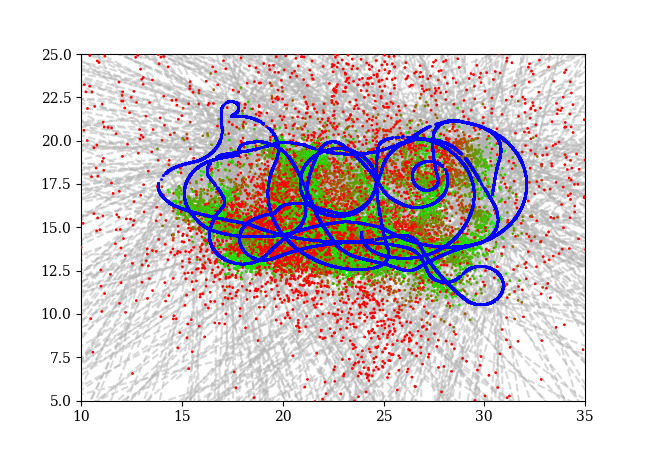}
    	\subcaption{Test 9.}
    	\label{figure_2dtraj7_10}
    \end{minipage}
    \hfill
	\begin{minipage}[t]{0.195\linewidth}
        \centering
    	\includegraphics[trim=61 38 48 46, clip, width=1.0\linewidth]{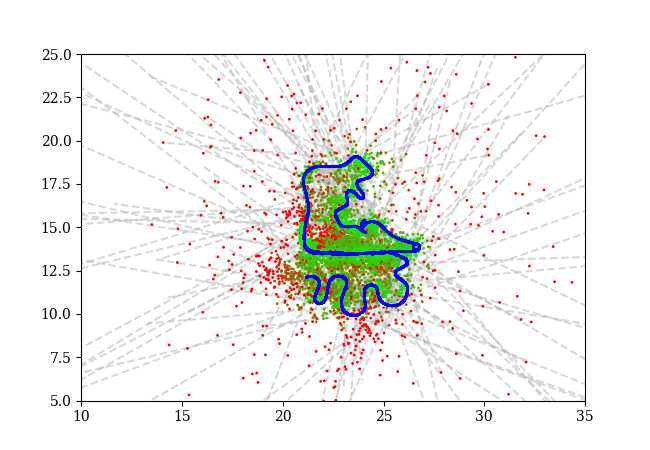}
    	\subcaption{Test 10.}
    	\label{figure_2dtraj7_11}
    \end{minipage}
    \hfill
	\begin{minipage}[t]{0.195\linewidth}
        \centering
    	\includegraphics[trim=61 38 48 46, clip, width=1.0\linewidth]{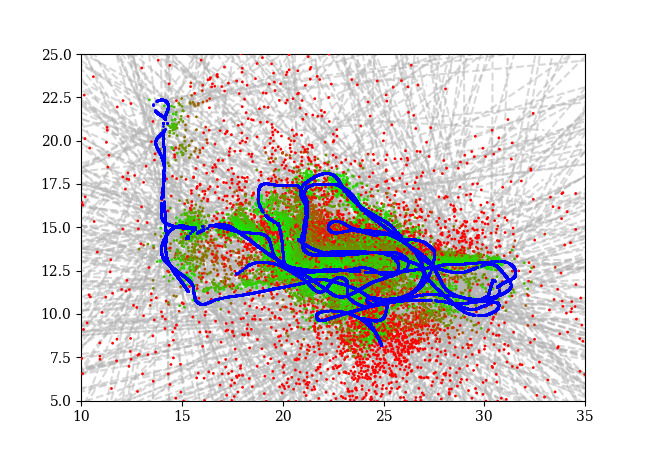}
    	\subcaption{Test 11.}
    	\label{figure_2dtraj7_12}
    \end{minipage}
    \caption{Evaluation of the predicted positions (green, red) against the ground truth trajectories (blue) for SfM for the train 7 dataset.}
    \label{figure_2dtraj7}
\end{figure*}

\begin{figure*}[!t]
    \centering
	\begin{minipage}[t]{0.195\linewidth}
        \centering
    	\includegraphics[trim=61 38 48 46, clip, width=1.0\linewidth]{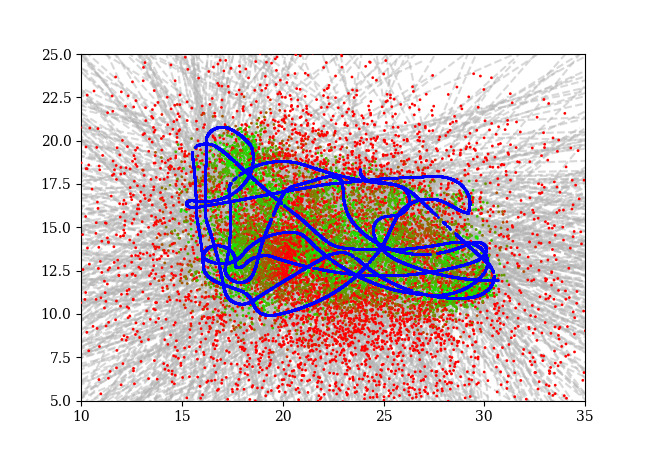}
    	\subcaption{Test 1.}
    	\label{figure_2dtraj8_1}
    \end{minipage}
    \hfill
	\begin{minipage}[t]{0.195\linewidth}
        \centering
    	\includegraphics[trim=61 38 48 46, clip, width=1.0\linewidth]{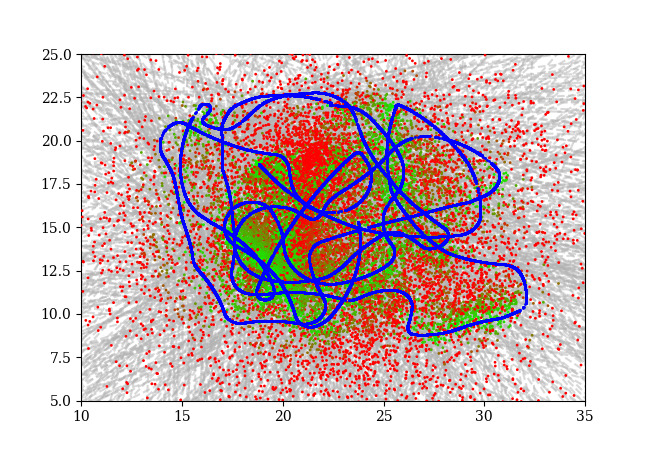}
    	\subcaption{Test 2.}
    	\label{figure_2dtraj8_2}
    \end{minipage}
    \hfill
	\begin{minipage}[t]{0.195\linewidth}
        \centering
    	\includegraphics[trim=61 38 48 46, clip, width=1.0\linewidth]{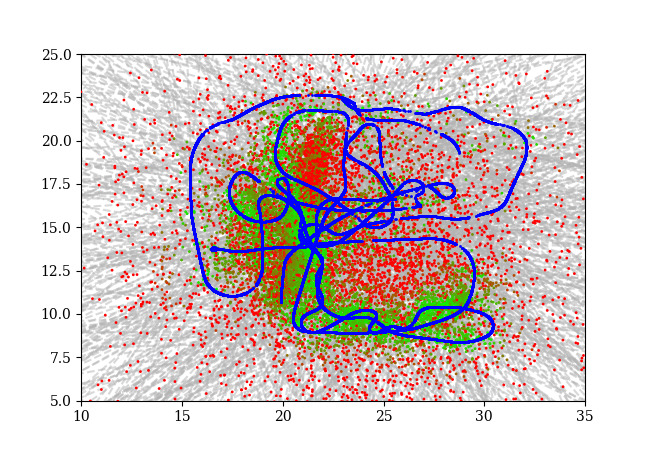}
    	\subcaption{Test 3.}
    	\label{figure_2dtraj8_3}
    \end{minipage}
    \hfill
	\begin{minipage}[t]{0.195\linewidth}
        \centering
    	\includegraphics[trim=61 38 48 46, clip, width=1.0\linewidth]{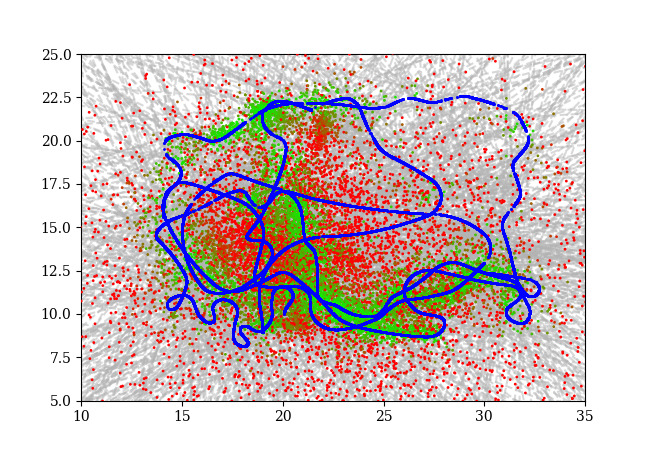}
    	\subcaption{Test 4.}
    	\label{figure_2dtraj8_4}
    \end{minipage}
    \hfill
	\begin{minipage}[t]{0.195\linewidth}
        \centering
    	\includegraphics[trim=61 38 48 46, clip, width=1.0\linewidth]{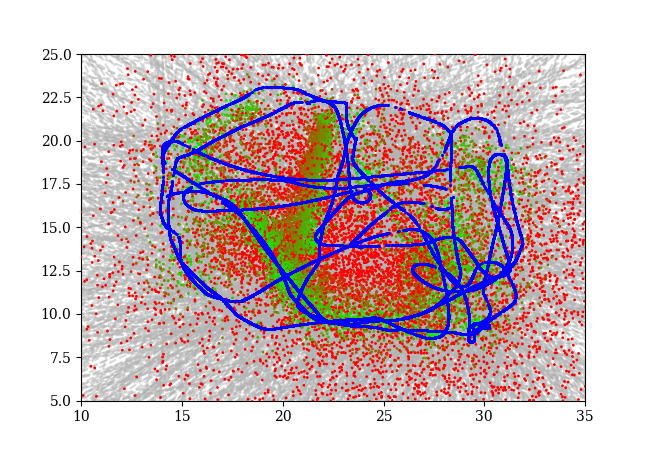}
    	\subcaption{Test 6.}
    	\label{figure_2dtraj8_6}
    \end{minipage}
	\begin{minipage}[t]{0.195\linewidth}
        \centering
    	\includegraphics[trim=61 38 48 46, clip, width=1.0\linewidth]{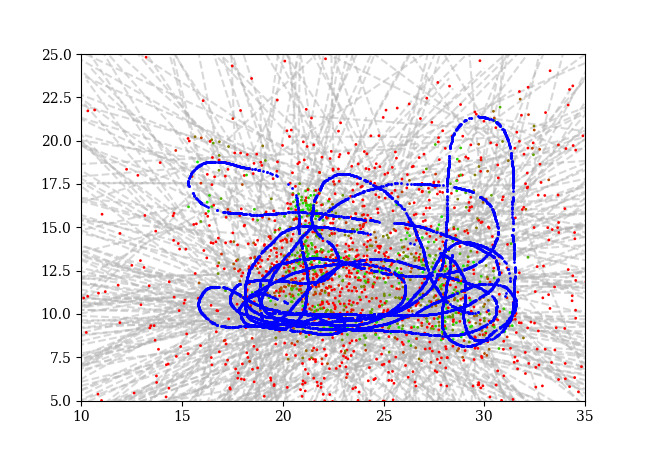}
    	\subcaption{Test 7.}
    	\label{figure_2dtraj8_7}
    \end{minipage}
    \hfill
	\begin{minipage}[t]{0.195\linewidth}
        \centering
    	\includegraphics[trim=61 38 48 46, clip, width=1.0\linewidth]{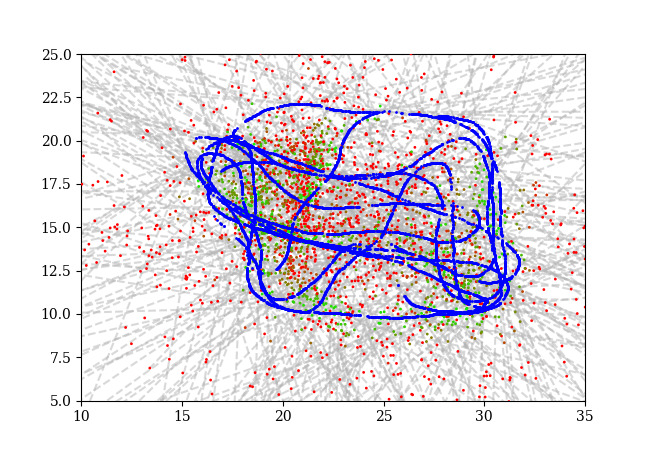}
    	\subcaption{Test 8.}
    	\label{figure_2dtraj8_9}
    \end{minipage}
    \hfill
	\begin{minipage}[t]{0.195\linewidth}
        \centering
    	\includegraphics[trim=61 38 48 46, clip, width=1.0\linewidth]{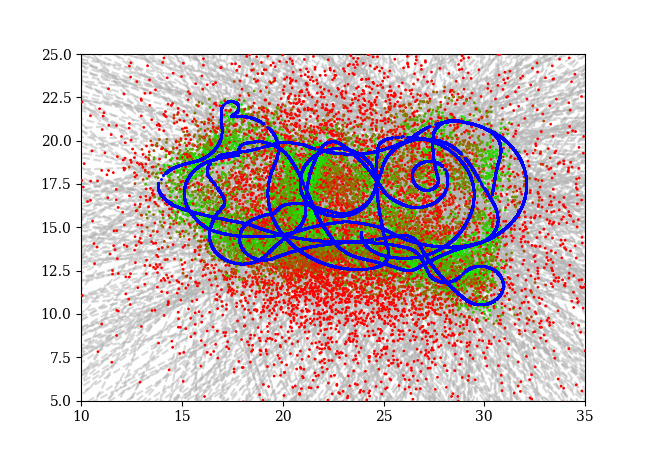}
    	\subcaption{Test 9.}
    	\label{figure_2dtraj8_10}
    \end{minipage}
    \hfill
	\begin{minipage}[t]{0.195\linewidth}
        \centering
    	\includegraphics[trim=61 38 48 46, clip, width=1.0\linewidth]{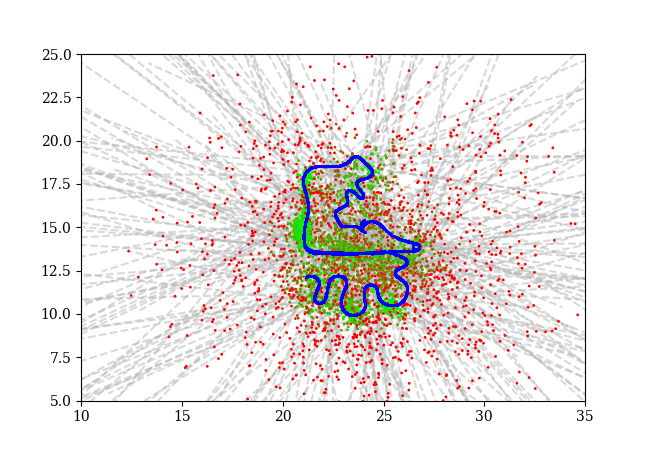}
    	\subcaption{Test 10.}
    	\label{figure_2dtraj8_11}
    \end{minipage}
    \hfill
	\begin{minipage}[t]{0.195\linewidth}
        \centering
    	\includegraphics[trim=61 38 48 46, clip, width=1.0\linewidth]{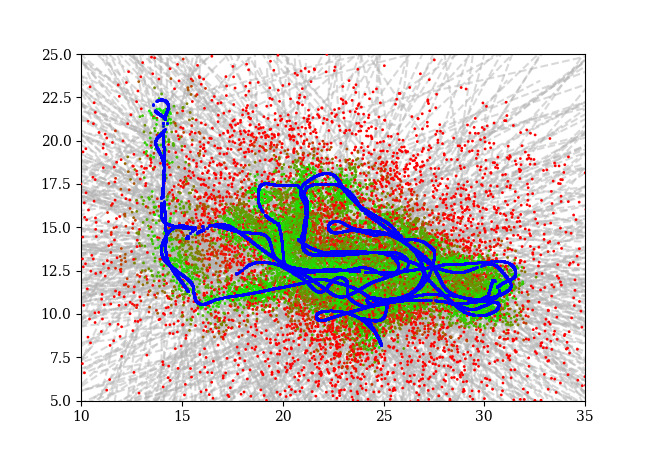}
    	\subcaption{Test 11.}
    	\label{figure_2dtraj8_12}
    \end{minipage}
    \caption{Evaluation of the predicted positions (green, red) against the ground truth trajectories (blue) for SfM for the train 8 dataset.}
    \label{figure_2dtraj8}
\end{figure*}

Figure~\ref{figure_2dtraj1} to Figure~\ref{figure_2dtraj8} contain visualizations of the predicted absolute poses for SfM, encompassing all eight training datasets and ten test datasets.

\clearpage

\section*{Biography}

\bio{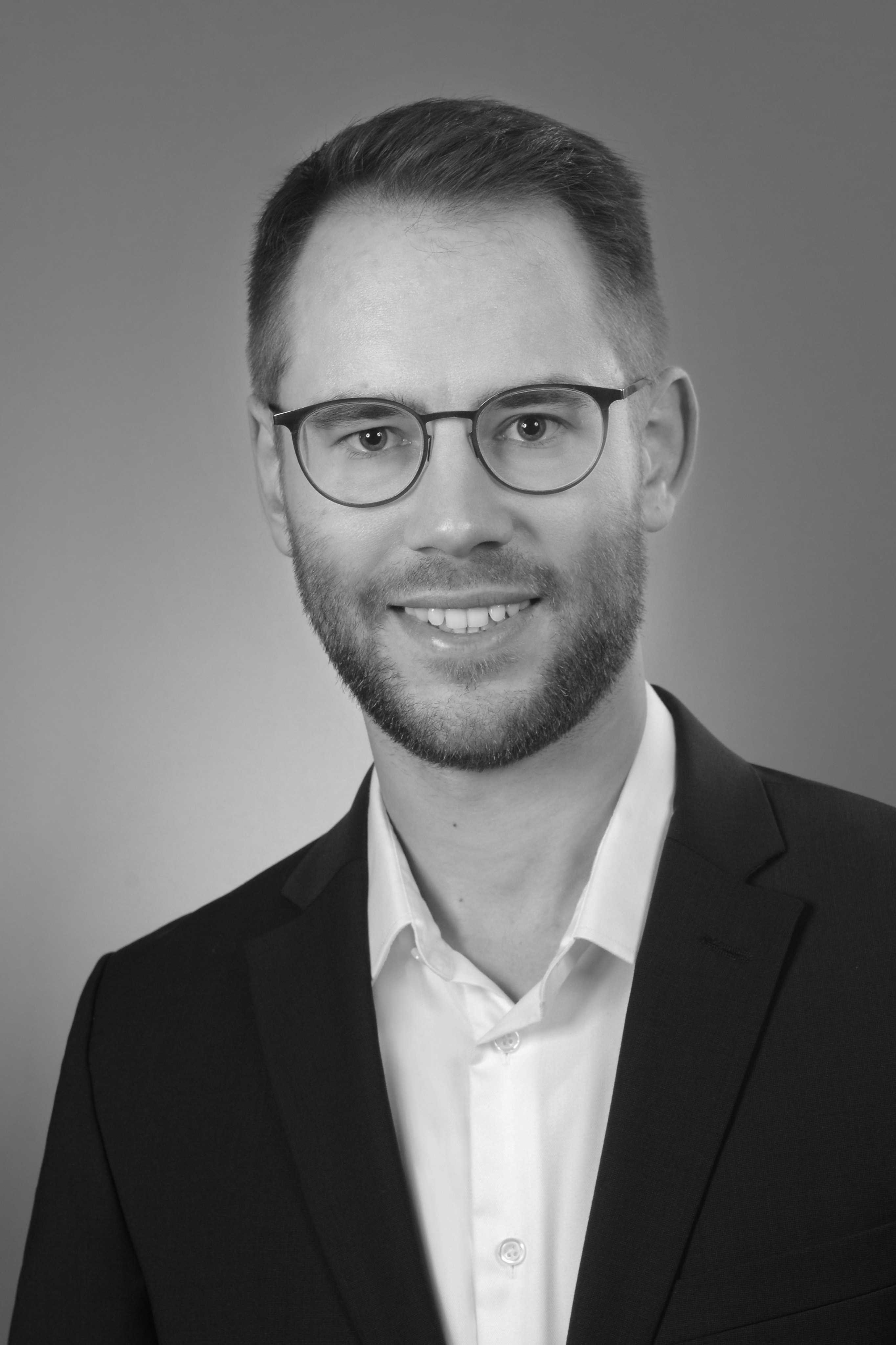}
\textbf{Felix Ott} received his MSc. degree in Computational Engineering at the FAU Erlangen-N{\"u}rnberg in 2019. He joined the Hybrid Positioning \& Information Fusion group in the Locating and Communication Systems department at Fraunhofer IIS. In 2023 he received his Ph.D. at the Ludwig-Maximilians University in Munich working in the Probabilistic Machine and Deep Learning group. His research covers multimodal information fusion for self-localization.
\endbio
\vspace{-0.1cm}
\bio{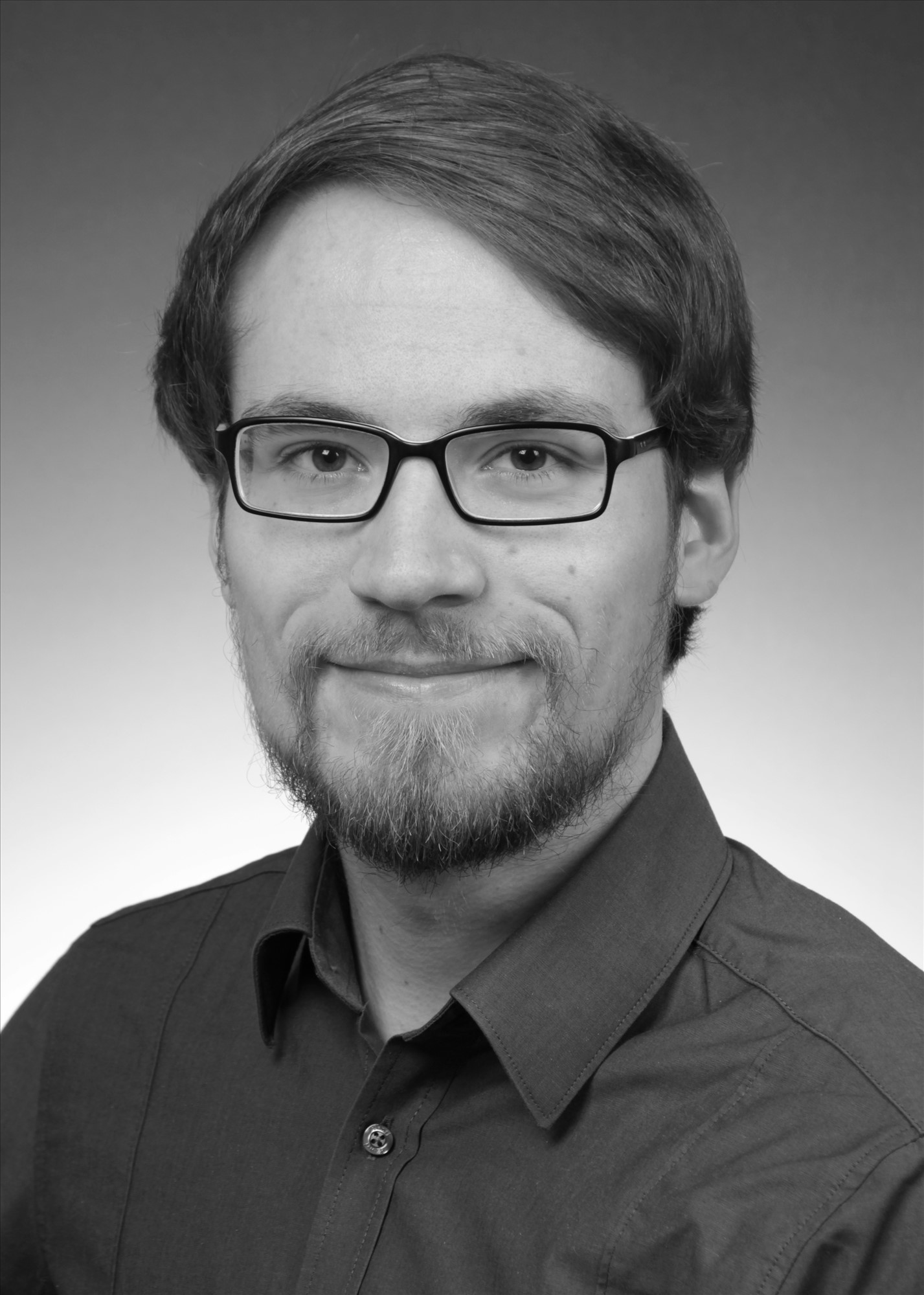}
\textbf{Lucas Heublein} received his M.Sc. degree in Integrated Life Science at the FAU Erlangen-Nürnberg. In 2020, he started his Computer Science degree at the FAU. He joined the Hybrid Positioning \& Information Fusion group at the Fraunhofer IIS in 2020 as a student assistant.
\endbio
\vspace{1.0cm}
\bio{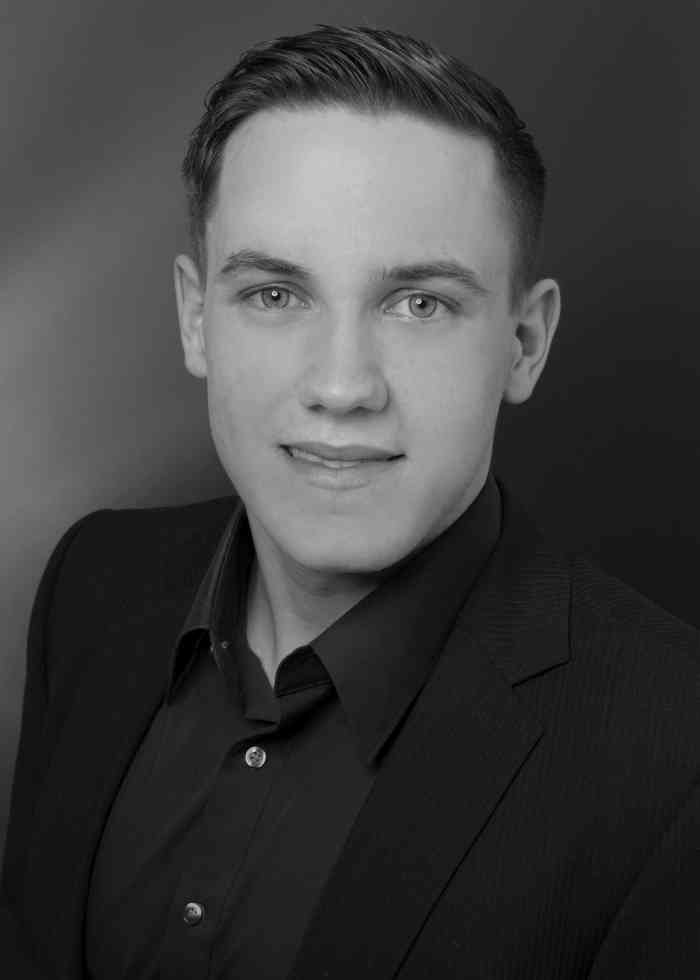}
\textbf{David Rügamer} is an interim professor for Computational Statistics at the RWTH Aachen. Before he was research associate, lecturer and interim professor for Data Science at the LMU Munich, where he also received his Ph.D. in 2018. His research is concerned with scalability of statistical modeling as well as machine learning for functional and multimodal data.
\endbio
\vspace{0.3cm}
\bio{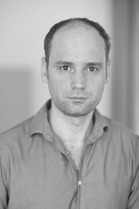}
\textbf{Bernd Bischl} is a full professor for statistical learning and data science and a director of the Munich Center of Machine Learning. His research focuses amongst other things on AutoML, interpretable machine learning and ML benchmarking.
\endbio
\vspace{1.6cm}
\bio{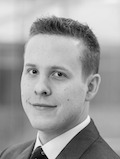}
\textbf{Christopher Mutschler} leads the precise positioning and analytics department at Fraunhofer IIS. Prior to that, Christopher headed the Machine Learning \& Information Fusion group. He gives lectures on machine learning at the FAU Erlangen-N{\"u}rnberg (FAU), from which he also received both his Diploma and PhD in 2010 and 2014 respectively. Christopher’s research combines machine learning with radio-based localization.
\endbio

\end{document}